# Effect-driven interpretation

## Functors for natural language composition

**Elements in Semantics**

---

Dylan Bumford
*University of California, Los Angeles*

Simon Charlow
*Yale University*

**Abstract:** Computer programs are often factored into pure components — simple, total functions from inputs to outputs — and components that may have side effects — errors, changes to memory, parallel threads, abortion of the current loop, etc. In this Element, we make the case that human languages are similarly organized around the give and pull of pure values and impure processes, and we'll aim to show how denotational techniques from computer science can be leveraged to support elegant and illuminating analyses of natural language composition.

# Contents

## Reader's Guide

This Element introduces and surveys an approach to natural language composition in the presence of enriched meanings. We borrow ideas, techniques, and vocabulary directly from a large body of research and practice in computer science centered around the concept of a functor. We'll get into all this in Chapter 1, but because the source material that we draw on spans so many linguistic domains and theoretical disciplines, we offer here a few preparatory remarks.

The compositional challenges that we address in this book arise when apparently independent semantic phenomena are brought into empirical contact. In short, how do you put two meanings together if those meanings have completely different mathematical structures? We offer a number of case studies in the text related to matters of anaphora and binding, indeterminacy and questions, presupposition, quantification, supplementation, and association with focus. On the one hand, readers do not need any particular domain expertise in any of these linguistic subjects to see the issues that crop up or to follow the solutions we propose. On the other hand, we will not spend any time motivating or discussing the individual domain-specific theories that we take as our starting points.

This means that the only advice we can offer readers hoping to develop a thorough understanding of the semantics of 'wh'-words or pronouns or focus-operators, say, is to consult the references that we cite in the relevant sections. Or better yet, don't consult them and just roll with it. Our chief aim is to abstract out what these sorts of theories have in common in order to create a plug-and-play compositional architecture. So the specifics of this or that phenomenon are mostly beside the point. Consequently, there is not going to be much in the way of novel empirical predictions or comparisons with other theories. The objective, after all, is to build a framework in which those other, more targeted theories can slot in. The predictions are only as good as those theories are.

As far as formal pre-requisites go, we hope that the entire Element is accessible to anyone familiar with the common aims and methods of compositional natural language semantics, as presented in any standard graduate textbook (e.g., Chierchia and McConnell-Ginet 1990, Heim and Kratzer 1998). As usual, denotations and grammatical operations are expressed using lambda terms and set comprehensions. Readers wishing to brush up on how these devices are used in semantics are referred to Winter's (2016) recent textbook for a particularly close notational match to the present work.

In addition, to make this all a little more concrete and a little more fun, we include snippets of code throughout the text implementing key ideas. The code is written in the programming language Haskell, whose construction and development have been heavily influenced by the algebraic concepts we discuss. This makes the language almost eerily well-suited to linguistic analysis, as we will see especially in Chapter 4. Unfortunately, space precludes us from providing any proper introduction to programming in Haskell, but fortunately, such introductions are easy to come by. Readers with no programming experience but an education in semantics, might start with van Eijck and Unger's (2010) book-length tutorial on how to use Haskell as a metalanguage for semantic theorizing. More experienced readers who just need a Haskell primer in order to play around with the code here could probably get by with the information at `https://learnxinyminutes.com/docs/haskell`. There are of course many complete books on Haskell (e.g., Hutton 2016), but we really only use a tiny sliver of the language, mostly just pattern matching and list comprehensions. And in any case, the syntax was designed to mimic standard mathematical notation, so we hope the code will be quite readable even to those unfamiliar with the language.

Finally, readers who don't want to bother with the code, but do want to see the ideas here in action, or maybe double check their work, are invited to visit `https://dylanbumford.com/effects.html`. The online demo there parses sentences and displays their semantic derivations in the style of this Element. Users may select different fragments of natural language and toggle various modes of combination on and off to see how these choices drive interpretations.

# 1 Introducing effects

## 1.1 Type-driven compositional semantics

Contemporary natural language semantic theories are generally drafted around three theoretically load-bearing components:

- a **lexicon** spelling out the meanings of individual expressions
- a **syntax** detailing the grammatical arrangement of these expressions
- a theory of **composition** describing how the meanings of complex expressions are built from their parts

Perhaps the simplest commonly-practiced architecture holds that expressions are constituted into binary-branching trees with lexical items at the leaves. Composition is governed by a simple theory of **types**, in the sense of Church (1940). A simple type is either a base type, corresponding to some fundamental ontological category (an entity $\mathtt{e}$, an event $\mathtt{v}$, a truth value $\mathtt{t}$, etc.), or is constructed from two other types with an arrow $\tau_1 \to \tau_2$, corresponding to a function from $\tau_1$ to $\tau_2$.

These types regulate an inventory of combinators, or **modes of combination**, which determine the range of ways that two meaningful expressions may combine. Given the simple type theory, by far the most common choices are forward and backward function application. For small, extensional fragments of language, this is often all that is required. But one may find appeals to other modes, including for example, various flavors of function composition (Ades and Steedman 1982, Dowty 1988), relation restriction (Kratzer 1996, Chung and Ladusaw 2003), and set intersection (Kamp 1975, Siegel 1976).

With a grammar and a bank of combinatory modes in hand, composition is then said to be **type-driven** (Klein and Sag 1985). The types of two daughter nodes are matched against the possible modes of combination. If the left type instantiates a combinator's first argument, and the right type its second, then the denotations of the daughters are composed as the combinator dictates. This basic setup is illustrated in Figure 1.

For this compositional regimen to make any sense, there must be an airtight correspondence between an expression's type and its denotation. An expression of type $\mathtt{e} \to \mathtt{t}$ must in fact denote a function whose domain is the set of ordinary entities and whose co-domain is the set of truth values. And an expression which denotes such a property of entities must in fact have type $\mathtt{e} \to \mathtt{t}$. In this sense, the semantics is said to be **strongly typed**, or type-safe.

**Types:**

$$\begin{aligned} \tau ::= &\ \mathsf{e} \mid \mathsf{t} \mid \cdots && \text{Base types} \\ \mid &\ \tau \to \tau && \text{Function types} \end{aligned}$$

**Modes of combination:**

$$(>) :: (\alpha \to \beta) \to \alpha \to \beta \qquad \text{Forward Application}$$
$$f > x := f\,x$$

$$(<) :: \alpha \to (\alpha \to \beta) \to \beta \qquad \text{Backward Application}$$
$$x < f := f\,x$$

$$(\circ) :: (\beta \to \varsigma) \to (\alpha \to \beta) \to \alpha \to \varsigma \qquad \text{Function Composition}$$
$$f \circ g := \lambda x.\, f\,(g\,x)$$

$$(\sqcap) :: (\mathsf{e} \to \mathsf{t}) \to (\mathsf{e} \to \mathsf{t}) \to \mathsf{e} \to \mathsf{t} \qquad \text{Predicate Modification}$$
$$f \sqcap g := \lambda x.\, f\,x \wedge g\,x$$

$$(\upharpoonleft) :: (\alpha \to \beta \to \mathsf{t}) \to (\beta \to \mathsf{t}) \to \alpha \to \beta \to \mathsf{t} \qquad \text{Relation Restriction}$$
$$r \upharpoonleft p := \lambda x \lambda y.\, p\,y \wedge r\,x\,y$$

**Figure 1** A simple type-driven grammar

Throughout this Element, we will display type-driven derivations as trees recording the types of the constituents derived. Below each branching node in a derivation tree, we will identify the mode of combination used to combine that node's daughters. An example, using some of the combinators in Figure 1, is given in (1.1). Sometimes, as in (1.1), we will also annotate constituents with their meanings, where lexical denotations are printed in **bold**.[1] However, denotations are always recoverable by recursively applying the combinators below nodes to the denotations of their daughters, so they are generally omitted except where we think they may be clarifying.

[1] In lambda terms, we follow Church's (1940) convention that the scope of a variable-binding operator extends as far to the right as possible, and applications are written as juxtapositions that associate to the left. So $\lambda f.\ f\,x\,y$ abbreviates $\lambda f.\ ((f\,x)\,y)$, not $((\lambda f.\ f)\,x)\,y$. Correspondingly, types associate to the right, so that $\alpha \to \beta \to \varsigma$ abbreviates $\alpha \to (\beta \to \varsigma)$, not $(\alpha \to \beta) \to \varsigma$.

(1.1)

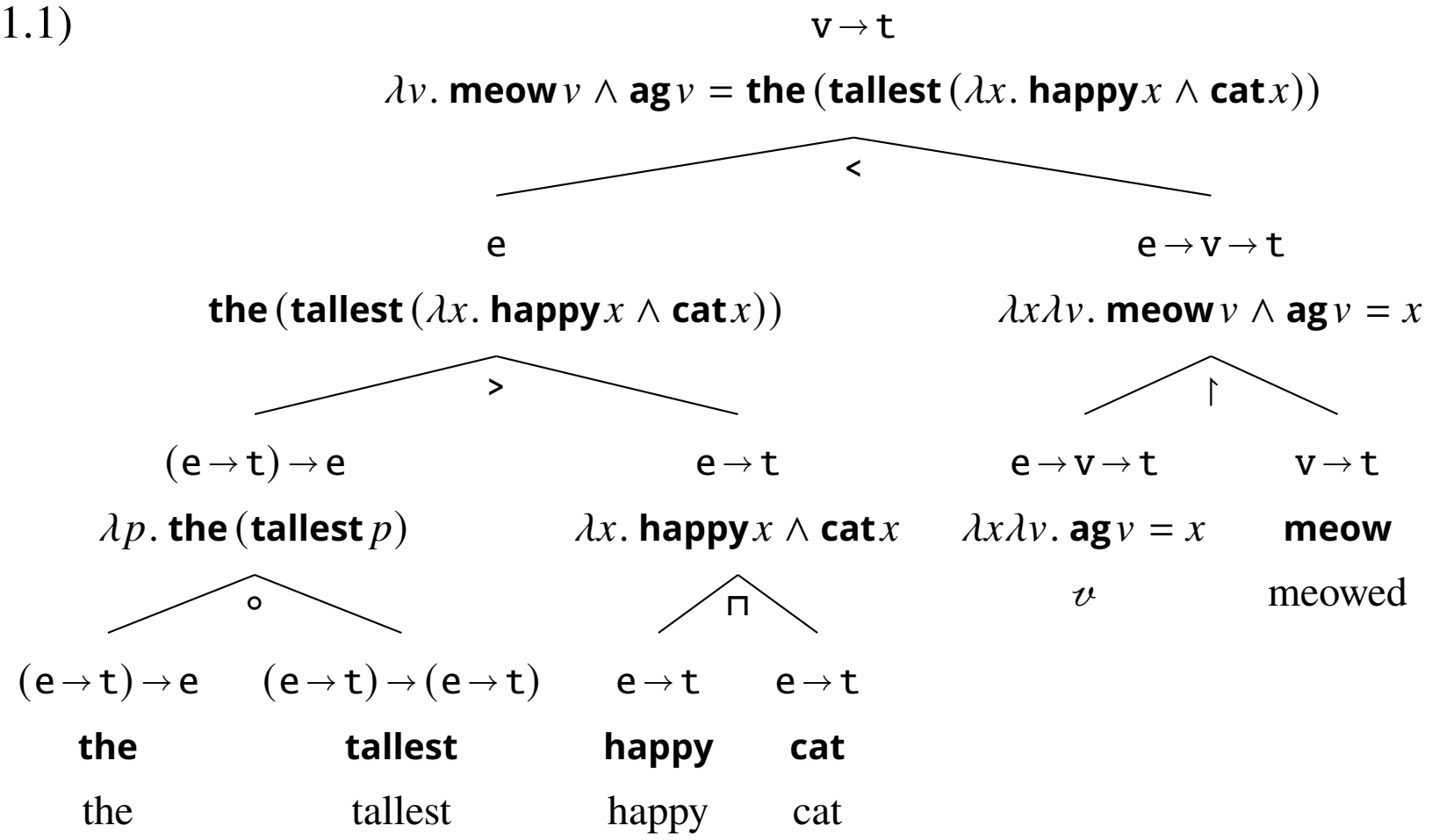

As it happens, the more or less canonical combinators in Figure 1 have the property that at most one of them will apply to any given pair of daughters. This guarantees that interpretation is **deterministic**: for any specific syntactic structure, there will be at most one interpretation (though there may of course be more than one well-typed structure for a given string). In principle, however, with a larger or different inventory of modes of combination, there could be a node at which more than one combinatory rule is applicable. In this case, the types would under-determine the meaning of the complex expression, predicting only *a set* of possible interpretations. Indeed, we will at many places in this Element make use of this compositional indeterminacy in the prediction of various systematic ambiguities.

## 1.2 More than a name

One consequence of the picture in Figure 1 is that, generally speaking, when two expressions have the same syntactic distribution, they must also have the same type. As every student of semantics knows, this leads immediately to trouble. Famously, quantificational noun phrases occur in the same syntactic positions as proper names. For instance, like names, they are just as felicitous in object positions as they are in subject positions:

(1.2a) {Jupiter, every planet} followed the moon.

(1.2b) The moon followed {Jupiter, every planet}.

But given the grammar in Figure 1, the only type that can be combined in the gaps of both (1.3a) and (1.3b) is e.

(1.3a)

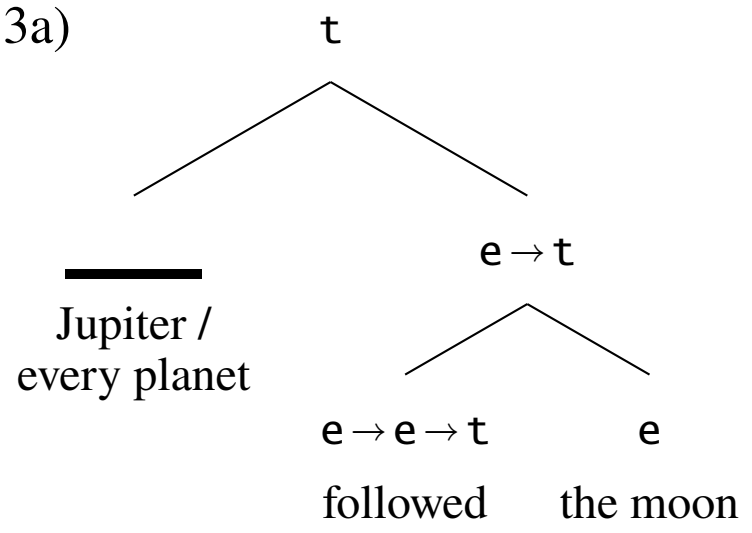

(1.3b)

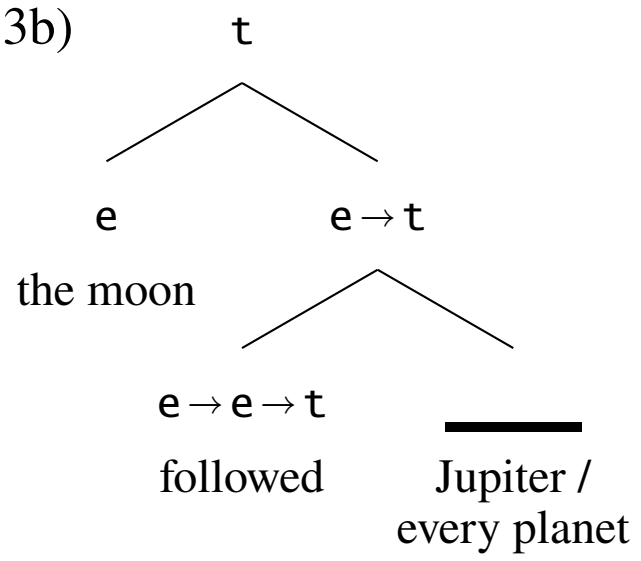

For proper names this is sensible, of course. But for quantificational phrases it is absurd. Well-rehearsed entailment patterns show that there can be no singular entity that is the referent of ‘every planet’, or ‘no planet’, or ‘exactly one planet’ (e.g., Geach 1972: Ch. 1.5). And so, given the required type-safety of the semantics, they cannot be expressions of type e.

Solutions to this problem break in every conceivable direction: reject that these are the trees that are interpreted, reject that these are the correct types of the lexical items, reject that these are the only modes of combination, reject the type system altogether, and so on. But one perhaps underappreciated aspect of the compositional conundrum laid bare in (1.3) is that it is not at all specific to matters of quantification. The basic nature of the problem is that an expression, say ‘every planet’, appears to play exactly the same argument-structural role as a name, yet clearly contributes something other than a simple referent to the meaning of the sentence. One might say the same is true of ‘wh’-expressions or disjunctions, as in (1.4). These are not (obviously) quantifiers, but they are (obviously) not names either. Where a name would provide a single, determinate referent to the predicate it saturates, these **indeterminate** expressions offer only a slate of candidates, a set of potential values conjured up in parallel.

(1.4a) Which planet is next to the moon?

(1.4b) The moon is next to Jupiter or Mars.

Definites too do not have the same semantic profile as names, despite saturating the same argument positions. What they refer to, or whether they refer at all, depends on what things are contextually salient. But an expression that does not have a stable individual referent cannot, strictly speaking, have type e. Pronouns, demonstratives, and indexicals even more clearly depend on context in a way that the type e does not represent. That is, none of the sentences in (1.5) can be said to denote a single, fixed proposition. Exactly what they say depends on when and where they are said.

(1.5a) The planet in the West is next to the moon.

(1.5b) It is next to the moon.

(1.5c) This planet that I'm looking at is next to the moon.

In fact, with a bit of prosodic focus, any noun phrase can be made to contribute more to the meaning of the sentence than just its referent. The sentences in (1.6) do not have the same truth conditions. The first is true when nothing relevant is visible except for Jupiter's moon; the second true when no other relevant moon is visible except for Jupiter's. The only difference between them is which phrase is focused, so the meanings of the focused phrases must somehow encode the necessary information.

(1.6a) Only [JUPITER'S MOON] is visible.

(1.6b) Only [JUPITER]'s moon is visible.

Likewise, even more transparently, a noun phrase may always be supplemented by various sorts of parenthetical and appositive constructions. For instance, the supplemented subject of (1.7a) and object of (1.7b) clearly saturate the same argument positions as the plain name that anchors them. Yet the supplemented phrases obviously add propositional information to the sentence, information that cannot be found in any individual entity.

(1.7a) Mars, which is now next to the moon, is setting.

(1.7b) Jupiter has passed Mars, which is now next to the moon.

Again, all of the phrases in (1.4)–(1.7) behave for compositional purposes as if they denoted ordinary entities. They are, by and large, grammatical and sensible wherever something of type e would be grammatical and sensible. But they cannot, or at least can't merely, denote entities. Their semantics is necessarily more complicated than this.

And these kinds of discrepancies are by no means limited to noun phrases. There are adjunct 'wh'-expressions that modify the same sorts of properties as ordinary adjuncts, and disjunctions of every category. There are verbs that differ only in their presuppositions. There are pro-forms and gaps and traces over arbitrarily complex types, parentheticals that can attach to almost anything, and of course quantifiers galore. All over the place we see expressions with interesting and complicated semantic properties that are nevertheless squeezed into positions where semantically boring expressions are expected, and this does not seem to disrupt composition in the slightest.

In this Element we adopt the view that these sorts of enriched expressions ought to be analogized to **impure** components of programming languages. This view has many precedents. Chief among them are analyses in the dynamic bent of Heim (1982), especially as described by Groenendijk and Stokhof (1991), Muskens (1990, 1996), van Eijck (2001), and colleagues, as well as analyses making heavy use of continuation-passing, including de Groote (2001), Barker (2002), Barker and Shan (2014), and Kiselyov and Shan (2014). The perspective we present here seeks to unify some of this work and much independent work on composition in various empirical domains.

In so doing, we largely follow the program introduced to linguistics by Shan (2001a, 2005), and further developed by Charlow (2014), Asudeh and Giorgolo (2020), and others. This program borrows directly from a tradition in computer science incorporating concepts from Category Theory that isolate repeating algebraic patterns that arise when working with various mathematical structures. These patterns have been used to explicate the semantics of common programming language constructs, and also to streamline the design of type-safe programming languages themselves.

As linguists, we benefit from this work on both ends. Natural languages are much like *weakly-typed* programs. The types of such programs generally reveal something about the sort of value that a program aims to compute, but may say nothing about what that computation consists in, or what sorts of things might happen along the way as the computation unfolds. As we've already sketched, in natural languages too all sorts of complex semantic activity can be packed into a phrase that the grammatical system — the compiler if you will — is effectively blind to. In that respect, the mathematical concepts that prop up the denotational semantics of such semantically devious programming languages can be imported, often wholesale, to linguistics. In contrast, *strongly-typed* programming languages are engineered to be as explicit as possible about whatever computational shenanigans a program engages in. Because they are so well-behaved by design, such languages can serve as a concrete, executable formalism for describing the behavior of less transparent, more promiscuous languages, including the ones we use every day.

## 1.3 Effects in programming languages

The constructs of so-called **imperative** programming languages are often divided into two categories. On the one hand there are **pure** expressions that carry out the basic business of determining a concrete result, or **value**. These include **literal** expressions like characters, strings, numbers, and booleans, whose values are fixed and stable, what we might call the names of the language. Other pure expressions include the (total) functions that convert one value to another. Usually a language provides a library of such functions, including straightforward boolean and arithmetic operations, like negation and addition and such. These we might think of as the ordinary nouns and verbs and adjectives of the language. Semantically, they are no more interesting than their graphs, the pairings between their inputs and outputs.

On the other hand there are **statements**. These are the bits of program text that determine the evaluation order and control flow of a computation; that is, what happens when the program is **executed**. Where a pure expression is intended to refer to some or other plain value, a statement is intended to *do something*, like change an address in memory, allocate and/or assign a value to a variable, throw an error, spawn multiple threads, start a loop, or print something to the screen. These sorts of non-referential processes are loosely referred to as the **(side) effects** of a computation.

As in natural language, program snippets containing commands like these may appear in the same places that pure program snippets would. For instance, the function `plus` on the left of Figure 2 is pure. It simply returns the sum of its two inputs. The function `showplus` on the right is impure. It also returns the sum of its two inputs, but additionally prints some words to the screen. In any given calling context, like `5 * ___`, the two functions will yield the same value. But when the one on the right is executed, the additional words are printed, as displayed in Figure 2.

How then do programming language theorists think about the meanings of programs like `showplus`? The answers to that question are certainly no less varied and creative than the answers that linguists have given to the issues of composing the phrases introduced above. But, excitingly, they are often different! In this Element, we do not pretend to offer a survey of the semantic methods deployed by computer scientists for reasoning about effects. Instead we concentrate on a few of the algebraic and combinatorial techniques that have proved useful already in linguistic theorizing.

The first order of business is rectifying the situation with the types. Type-safety requires that any semantically-relevant behavior of a program be reflected

```
function plus(x,y) = {

  return x + y
}

> 5 * plus(3, 7)

50
```

```
function showplus(x,y) = {
  console.log("doing (+)");
  return x + y
}

> 5 * showplus(3, 7)
"doing (+)"
50
```

**Figure 2** Pure and impure javascript programs

in its type. The same is true in language. To model the variety of behaviors on display in (1.4)–(1.7), it will be helpful to first expand the type system.

## 1.4 Algebraic Data Types

Some of the exhibited linguistic effects seem to call for denotations with *multiple dimensions* of meaning. The natural mathematical setting for modeling multi-dimensionality is a tuple, with different semantic dimensions in different coordinates. We thus introduce **product types** $\alpha \times \beta$ to model meanings that carry both type $\alpha$ and type $\beta$ content. Denotationally, an expression of type $\alpha \times \beta$ takes its meaning from the Cartesian product of $\alpha$ and $\beta$.

(1.8) Mars, a planet :: $\mathtt{e} \times \mathtt{t}$

$[\![\text{Mars, a planet}]\!] = \langle \mathbf{m}, \mathbf{planet}\ \mathbf{m} \rangle$

Other effects seem to call for denotations with *multiple variants* of meaning. For instance, a definite description will either refer to an object (type $\mathtt{e}$), or it will fail to refer, returning a computational dead end. We might model a failure of reference with a type $\perp$ whose only value is #. The description therefore denotes a computation that will return a value of one of two distinct types, either type $\mathtt{e}$ or type $\perp$. The type of such a computation is naturally modeled by a **sum type** $\alpha + \beta$. Denotationally, an expression of type $\alpha + \beta$ takes its meaning from the disjoint union of $\alpha$ and $\beta$.

(1.9) the planet :: $\mathtt{e} + \perp$

$[\![\text{the planet}]\!] = x \text{ if } \mathbf{planet} = \{x\} \text{ else } \#$

Types built from product, sum, and arrow constructors are called **Algebraic Data Types**. Finally, we will want to be able to define the types of values that are drawn from *powersets* of specific domains. For this we will slightly abuse the notation $\{\alpha\}$ to represent the type whose members are sets of type-$\alpha$ things. For instance, the canonical Hamblin denotation for a 'wh'-argument is a set of entities (Hamblin 1973), and its type therefore is $\{e\}$.

(1.10) which planet :: $\{e\}$
$[\![\text{which planet}]\!] = \{x \mid \textbf{planet}\,x\}$

With these algebraic data types as scaffolding for structured values, the entries in Table 1 spell out some more or less standard semantic characterizations of the constructions in (1.4)–(1.7). Our goal is certainly not to advocate for these or any other specific analyses of the expressions, but rather to explore an approach to compositionality in the face of such semantic complexities. To that end, these lexical entries are intended primarily as illustrative test cases for the techniques outlined in what follows, so we will not pause to seriously motivate any of these lexical entries on their raw linguistic merits.

The first four rows of Table 1 are the examples we've seen, deploying the complex types that represent the different kinds of structured values we will make use of. Row 1 construes the denotation of a pronoun as a function that selects a referent from some sort of linguistic context $i$. Commonly in semantics, the type $i$ is identified with (infinite) sequences of individuals, also known as **assignments**, and antecedent selection with a projection function $\pi$ targeting a particular coordinate of that sequence (Tarski 1956, Heim and Kratzer 1998). This is certainly not the only choice, as we'll see in Chapter 2, but as far as we are aware, every semantics for anaphoric expressions has this basic functional shape. The definite description in Row 2 denotes a "partial object". If its restrictor describes a single entity in its context of utterance, then it refers to that entity; otherwise, it fails to refer (Strawson 1950, Cooper 1975). Row 3 analyzes the supplemented name as semantically bidimensional, denoting both an ordinary referent and a fact about that referent (Potts 2005, McCready 2010, Asudeh and Giorgolo 2020). And in Row 4, we have a 'wh'-expression referring indeterminately to the set of values that might answer whatever sentence contains it (Hamblin 1973, Hagstrom 1998). We will see plenty of examples of this sort of example in Chapters 2 and 3.

The next four rows of Table 1 are slightly more complex, involving nested type constructors. Quantificational phrases like 'no planet' denote Generalized Quantifiers, properties of properties (Montague 1973, Barwise and Cooper 1981). Focused phrases like 'JUPITER' are both bidimensional and indeterminate;

| Expression | Type | Denotation |
|---|---|---|
| it | $\mathtt{i} \to \mathtt{e}$ | $\lambda i.\, \pi i$ |
| the planet | $\mathtt{e} + \bot$ | $x$ if $\mathbf{planet} = \{x\}$ else # |
| Jupiter, a planet | $\mathtt{e} \times \mathtt{t}$ | $\langle \mathbf{j}, \mathbf{planet}\,\mathbf{j} \rangle$ |
| which planet | $\{\mathtt{e}\}$ | $\{x \mid \mathbf{planet}\, x\}$ |
| no planet | $(\mathtt{e} \to \mathtt{t}) \to \mathtt{t}$ | $\lambda Q.\, \neg \exists x.\, \mathbf{planet}\, x \wedge Q\, x$ |
| JUPITER | $\mathtt{e} \times \{\mathtt{e}\}$ | $\langle \mathbf{j}, \{x \mid x \sim \mathbf{j}\} \rangle$ |
| as for Jupiter | $\mathtt{s} \to (\mathtt{e} \times \mathtt{s})$ | $\lambda s.\, \langle \mathbf{j}, \mathbf{j} +\!\!+ s \rangle$ |
| a planet | $\mathtt{s} \to \{\mathtt{e} \times \mathtt{s}\}$ | $\lambda s.\, \{\langle x, x +\!\!+ s \rangle \mid \mathbf{planet}\, x\}$ |

**Table 1** Example noun phrases, their types and denotations

a certain entity $\mathbf{j}$ is named, while the alternatives to that entity $\{x \mid x \sim \mathbf{j}\}$ are evoked (Rooth 1985). In Rows 7 and 8, we list a topic-marked name, 'as for Jupiter', and an indefinite, 'a planet', in the style of dynamic semantics. Topics, almost by definition, not only refer but also put their referent front and center on some sort of evolving discourse stage. This is naturally modeled by the deterministic update in the table, which both holds out $\mathbf{j}$ for composition and also rotates $\mathbf{j}$ to the front of a context $s$. As with pronouns, linguistic theories will differ in exactly how they model topic lists and topic list re-centering, which we represent abstractly as $+\!\!+$ (Grosz, Joshi, and Weinstein 1995, Bittner 2001). Typical dynamic analyses of indefinites are similar, in that they change the conversational state so that their witnesses are available for anaphora. But like 'wh'-expressions, indefinites do not necessarily single out unique referents, and so in general correspond to updates that are **nondeterministic**. Again, theories may differ in how they model discourse contexts, but all dynamic semantics for indefinites in the wake of Heim (1982) have the basic relational update procedure from inputs $s$ to modified outputs $x +\!\!+ s$ (Groenendijk and Stokhof 1991, Muskens 1990, Dekker 1994).

The perspective we would like to encourage here views these expressions as denoting particular kinds of **computations**, specifically computations that yield entities. A pronoun reads a referent from an environment. An appositive writes a fact to the common ground. A 'wh'-expression forks the composition into several parallel paths, one for each of its potential answers. A definite description is a program that might crash if executed in the wrong situation. An indefinite modifies what referents are in memory, and possibly what addresses they're stored in. And so on.

Remember, all of these expressions appear in positions where ordinary entities are expected. And naturally their denotations all, one way or another, contain, return, manipulate, abstract and/or quantify over entities. Following the programming literature, we will sometimes talk about the entities in these types as being situated in a particular **computational context**, and use the term **effect** to refer to whatever a computation does with them.

To make this formal, let us introduce *ad-hoc* type constructors for each of the effects in Table 1. Here we use the term (unary) **type constructor** to mean a function from types to types. We find the constructor for each effect by abstracting over $\mathtt{e}$ with a type variable $\alpha$. For instance, the $\mathtt{R}$ constructor encodes the effect of reading from an environment of type $\mathtt{i}$. The $\mathtt{W}$ constructor encodes the effect of logging a message of type $\mathtt{t}$. The $\mathtt{M}$ constructor the effect of possibly failing. And so on.

(1.11)
$$\mathtt{R}\,\alpha ::= \mathtt{i} \to \alpha \qquad \mathtt{C}\,\alpha ::= (\alpha \to \mathtt{t}) \to \mathtt{t}$$
$$\mathtt{W}\,\alpha ::= \alpha \times \mathtt{t} \qquad \mathtt{F}\,\alpha ::= \alpha \times \{\alpha\}$$
$$\mathtt{M}\,\alpha ::= \alpha + \bot \qquad \mathtt{S}\,\alpha ::= \{\alpha\}$$
$$\mathtt{T}\,\alpha ::= \mathtt{s} \to (\alpha \times \mathtt{s}) \qquad \mathtt{D}\,\alpha ::= \mathtt{s} \to \{\alpha \times \mathtt{s}\}$$

Then we can express our dictionary of noun phrases as in Table 2. This makes clear that all of these expressions are entity-directed computations. The particular nature of each computation is encoded in the structure of the effect defined by its type constructor in (1.11).

| Expression | Type | Denotation |
|---|---|---|
| it | $\mathtt{R\,e}$ | $\lambda i.\, \pi\, i$ |
| the planet | $\mathtt{M\,e}$ | $x$ if $\mathbf{planet} = \{x\}$ else $\#$ |
| Jupiter, a planet | $\mathtt{W\,e}$ | $\langle \mathbf{j}, \mathbf{planet}\,\mathbf{j} \rangle$ |
| which planet | $\mathtt{S\,e}$ | $\{x \mid \mathbf{planet}\, x\}$ |
| no planet | $\mathtt{C\,e}$ | $\lambda Q.\, \neg \exists x.\, \mathbf{planet}\, x \wedge Q\, x$ |
| JUPITER | $\mathtt{F\,e}$ | $\langle \mathbf{j}, \{x \mid x \sim \mathbf{j}\} \rangle$ |
| as for Jupiter | $\mathtt{T\,e}$ | $\lambda s.\, \langle \mathbf{j}, \mathbf{j} +\!\!+\, s \rangle$ |
| a planet | $\mathtt{D\,e}$ | $\lambda s.\, \{\langle x, x +\!\!+\, s \rangle \mid \mathbf{planet}\, x\}$ |

**Table 2** Example noun phrases, with types encoded by effect constructors

So far all we have done is relabel the types of different kinds of values. None of this provides any immediate relief to the problem at hand, which is to fit these expressions into syntactic contexts that only know how to process an ordinary entity. On the contrary, hiding all the mathematical structure of the types would seem to preclude rather than facilitate type-driven composition.

But it turns out that all of these constructors — `R`, . . . , `D` — share a few very important algebraic properties. And these properties allow us to ignore whatever is specific to the kind of computation represented by the type, and get on with the work of passing the underlying entity into the predicate that it saturates. This not only paves the way for composition, it also reveals a certain uniformity that is completely lost in all of the independent theorizing about the various linguistic phenomena. At the same time, it frees researchers working on independent semantic problems to concentrate on their effects of interest without inventing idiosyncratic mechanisms of scope, type-shifting, and combination.

In the coming chapters, we spell out some of the algebraic properties of these effects. But first, let us lay some computational groundwork for the discussion. As mentioned in Section 1.1, the compositional framework we will present is categorematic and type-driven. That is, the ways of interpreting a constituent will depend only on the types of its daughters. Moreover, the rules determining which pairs of types lend themselves to which modes of combination are going to be entirely formal and decidable. You just have to look at the types and see if they match the rules.

In other words, it should all be stupid enough that a computer can do it for us. Let us then put our machine where our mouth is, and implement a very simple type-driven interpreter.

## 1.5 Implementing a type-driven interpreter

To get the ball rolling, let's start with the type system in Figure 1, setting aside effects for now. Throughout this Element, we will build on this interpreter, folding in the combinatoric operations that we introduce as we go.

Our goal here is to have the computer figure out every way that a sentence can be interpreted, knowing only its constituency structure and the types of its lexical items. As such, we focus here on the process of type-driven combination, rather than any aspect of parsing. We will assume then that the sentences to be interpreted are pre-assembled into binary-branching trees. Here is a Haskell data type representing such parsed structures:

```haskell
data Syn
  = Leaf String
  | Branch Syn Syn
```

A piece of syntax `Syn` is either a `Leaf` containing a `String` (the name of the lexical item), or a `Branch`ing node containing two subtrees.

The goal is to determine the possible modes of combination for each node and what the resulting type would be. It is, essentially, to deduce the semantic derivation in (1.12b) — which completely determines the meanings in (1.1) — from the tree in (1.12a), where only constituency structure and lexical types are known.

(1.12a)

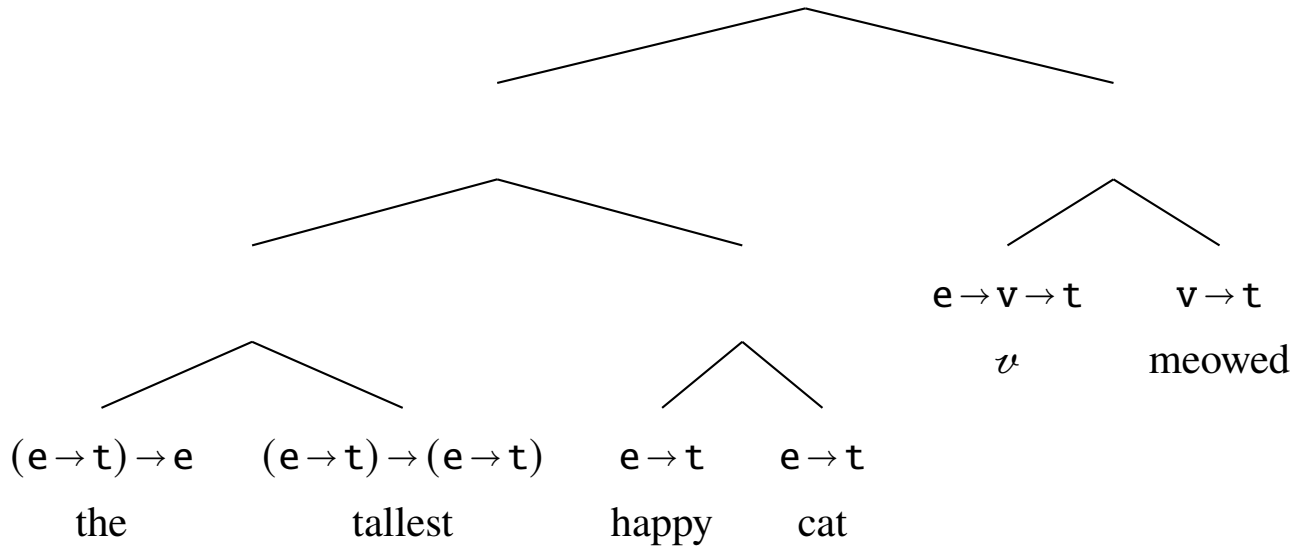

(1.12b)

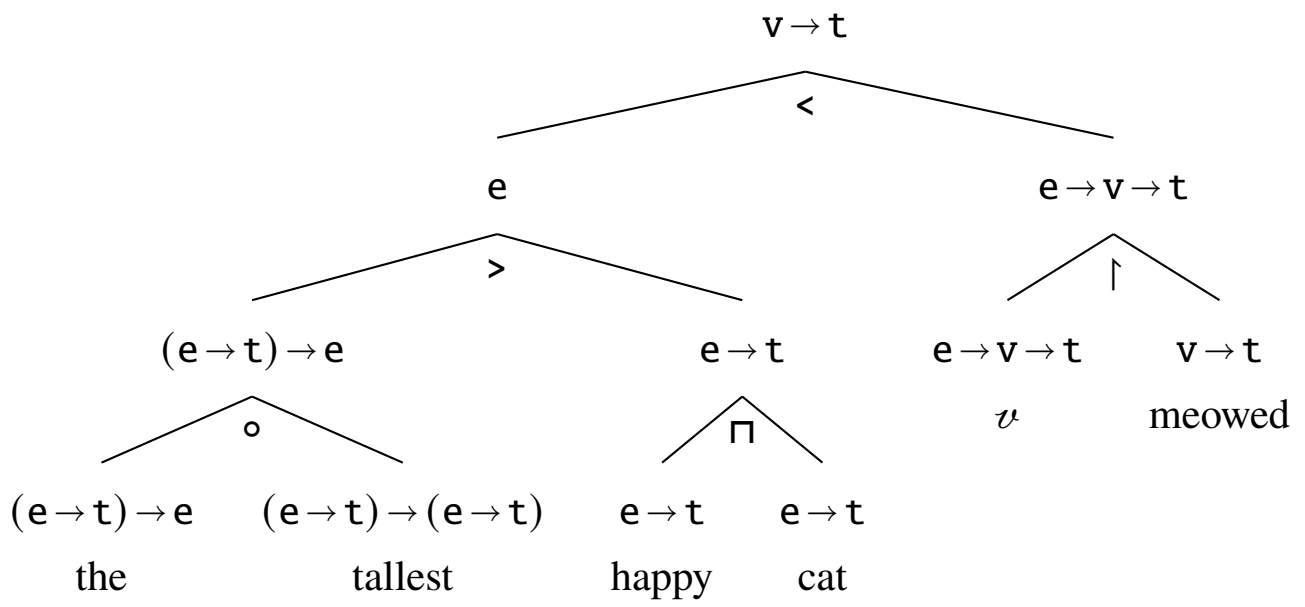

To do this, we define a data structure for type- and mode-annotated trees, which we call `Sem` trees, analogous to the `Syn` trees defined above. A `Sem` tree is either a typed lexical item, or a combination of two `Sem` objects (themselves type- and mode-annotated trees), annoted with a result type and a mode of combination. A type `Ty` is either atomic — `E`, `T`, or `V` — or complex — an arrow between two other types. And a mode of combination `Mode` is, for now, just a tag indicating which of the modes from Figure 1 applies. They are just Haskell versions of the symbols in (1.12b).

```
data Sem
  = Lex Ty String
  | Comb Ty Mode Sem Sem
  deriving (Show)

data Ty
  = E | T | V          -- base types
  | Ty :-> Ty          -- function types
  deriving (Eq, Show) -- Ty's can be compared for equality, printed

data Mode
  = FA -- forward application
  | BA -- backward application
  | PM -- predicate modification
  | FC -- forward composition
       -- other basic modes of combination, as desired
  deriving (Show)
```

The engine of the type-driven logic is implemented by the function `modes`. All it does is pattern-match on the two types that it is handed. If the left type is an arrow type `a :-> b` and the right type is `a`, then `FA` is an applicable mode of combination, and the result will be of type `b`. Vice versa for `BA`. If both inputs are arrow types with co-domain `T` and identical domain types, then `PM` applies. And so on.

```
modes :: Ty -> Ty -> [(Mode, Ty)]
modes l r = case (l, r) of
  (a :-> b, _      ) | r == a -> [(FA, b)]
  (_      , a :-> b) | l == a -> [(BA, b)]
  (a :-> T, b :-> T) | a == b -> [(PM, a :-> T)]
  (c :-> d, a :-> b) | b == c -> [(FC, a :-> d)]
  --        ...          ==      ...
  _                          -> []
```

Notice that the result of applying `modes` to two types `l` and `r` is a *list* of possible results. If any of the substantive cases match, the result is a singleton list containing just the mode appropriate to that case. If none of them match, the result is an empty list. So the result of any call of `modes l r` is at most a singleton list. Thus the interpreter is deterministic. In later chapters, this will not be the case. The result of combination may contain 0, 1, or more modes and the corresponding result types.

Finally, we define the interpreter `synsem`. Given a lookup table of lexical types `Lexicon` and a syntactic object `Syn`, it returns a list of possible type- and mode-annotations: `[Sem]`. It does this by a very straightforward recursion, bottoming out at the leaves with an appeal to the lexicon. Given a branching node `Branch lsyn rsyn`, it interprets the left branch, interprets the right branch, and then combines the types of the results with `modes`. Because interpretation is relational rather than functional, each of these actions yields not one, but a list of intermediate results. The final collection of `Sem` annotations is determined by taking each possible interpretation `lsem` of the left branch, each possible interpretation `rsem` of the right branch, and each possible way of putting such interpretations together to yield a new `Comb`ined constituent.

```
type Lexicon = String -> [Ty]

synsem :: Lexicon -> Syn -> [Sem]
synsem lex syn = case syn of
  (Leaf w)            -> [Lex t w | t <- lex w]
  (Branch lsyn rsyn) ->
    [ Comb ty op lsem rsem
      | lsem     <- synsem lex lsyn
      , rsem     <- synsem lex rsyn
      , (op, ty) <- modes (getType lsem) (getType rsem) ]
  where
    getType (Lex ty _)       = ty
    getType (Comb ty _ _ _) = ty
```

# 2 Functors

## 2.1 Maps and mapping

The computational contexts encoded in the types of Table 2 are quite varied, but they are all known to category theorists and computer scientists as **functors**. Intuitively speaking, a constructor $\Sigma$ is functorial if its parameter $\alpha$ remains accessible to manipulation despite being embedded in the $\Sigma$ structure. It should moreover not make any difference what kind of thing $\alpha$ is. Just knowing how it is situated inside the $\Sigma$ structure should be enough to know how it could be adjusted.

For instance, consider the constructor $\boxed{\mathsf{S}\,\alpha} = \{\alpha\}$. A value of type $\boxed{\mathsf{S}\,\mathbb{N}}$ is a set of natural numbers $X$. Given a function $k : \mathbb{N} \to \mathbb{N}$, we can **map** $k$ over $X$ by applying it pointwise to the elements of $X$. We could do the same if $k$ instead converted numbers to strings, or if $X$ were full of entities and $k$ a function from entities to truth values. In every case, the way $k : \alpha \to \beta$ is used to update a set $X$ of $\alpha$ values is the same: $X' = \{k\,x \mid x \in X\}$, as seen in (2.1).

(2.1)

| $X$ | $k$ | $X'$ |
|---|---|---|
| $\{1, 2, 3\}$ | $(\lambda x.\, x + 1)$ | $\{k\,x \mid x \in X\} = \{2, 3, 4\}$ |
| $\{1, 2, 3\}$ | $(\lambda x.\, \textbf{repeat}\ x\ \textbf{"c"})$ | $\{k\,x \mid x \in X\} = \{\textbf{"c"}, \textbf{"cc"}, \textbf{"ccc"}\}$ |
| $\{\textbf{j}, \textbf{m}, \textbf{s}\}$ | $(\lambda x.\, x = \textbf{j})$ | $\{k\,x \mid x \in X\} = \{\textbf{true}, \textbf{false}\}$ |

Similarly, a value in the domain of $\boxed{\mathsf{R}\,\alpha}$ is a function that assigns an $\alpha$ to every input $\mathtt{i}$. We might think of such a function as an $\mathtt{i}$-indexed family of $\alpha$ values. Given a function $X$ in the domain of $\boxed{\mathsf{R}\,\alpha}$ and a function $k : \alpha \to \beta$, we can map $k$ over $X$ by applying it to the value $X$ takes at each index. That is, given an index $i$, we first retrieve the $\alpha$ located at $X\,i$, and then apply $k$ to the result. Again, it's clear we can do this no matter the return type of $k$, or indeed the types of the values held in $X$, as illustrated in (2.2), where the index domain $\mathtt{i}$ is $\{\triangle, \square\}$.

(2.2)

| $X$ | $k$ | $X'$ |
|---|---|---|
| $\begin{bmatrix} \triangle \mapsto 3 \\ \square \mapsto 7 \end{bmatrix}$ | $(\lambda x.\, x + 1)$ | $(\lambda i.\, k\,(X\,i)) = \begin{bmatrix} \triangle \mapsto 4 \\ \square \mapsto 8 \end{bmatrix}$ |
| $\begin{bmatrix} \triangle \mapsto 1 \\ \square \mapsto 3 \end{bmatrix}$ | $(\lambda x.\, \textbf{repeat}\ x\ \textbf{"c"})$ | $(\lambda i.\, k\,(X\,i)) = \begin{bmatrix} \triangle \mapsto \textbf{"c"} \\ \square \mapsto \textbf{"ccc"} \end{bmatrix}$ |
| $\begin{bmatrix} \triangle \mapsto \textbf{s} \\ \square \mapsto \textbf{j} \end{bmatrix}$ | $(\lambda x.\, x = \textbf{j})$ | $(\lambda i.\, k\,(X\,i)) = \begin{bmatrix} \triangle \mapsto \textbf{false} \\ \square \mapsto \textbf{true} \end{bmatrix}$ |

Formally, then, a type constructor $\Sigma$ is a functor if there is a **map** operation ($\bullet$) with the type indicated in (2.3) respecting the laws in (2.4).

(2.3) $(\bullet) :: (\alpha \to \beta) \to \boxed{\Sigma\,\alpha} \to \boxed{\Sigma\,\beta}$

(2.4) **Identity:** $\mathbf{id} \bullet X = X$

**Composition:** $f \bullet (g \bullet X) = (f \circ g) \bullet X$

Note that ($\bullet$) here is **polymorphic** in its input. It must be defined for *any* types $\alpha$ and $\beta$, which means it must be capable of lifting any arbitrary function to a corresponding function on computations $\boxed{\Sigma\,\alpha}$ and $\boxed{\Sigma\,\beta}$. If ($\bullet$) lifts all functions "in the same way", then it is said to be **parametric**. A precise definition of parametricity is beyond the scope of this Element (see Reynolds 1983), but intuitively, a function is parametric if it never asks any questions about the types of its arguments.

For instance, the mapping operations for $\mathtt{R}$ and $\mathtt{S}$ defined above are parametric, in that they perform exactly the same action no matter the type of the function $k$ they are handed. Moreover, since these operations satisfy the laws in (2.4), they establish the functoriality of $\mathtt{R}$ and $\mathtt{S}$, respectively.

(2.5) $k \bullet_{\mathtt{R}} X := \lambda i.\, k\,(X\,i)$ $\qquad$ $k \bullet_{\mathtt{S}} X := \{k\,x \mid x \in X\}$

Readers are invited to take a second to convince themselves the laws are satisfied in both cases. Actually, conveniently, whenever ($\bullet$) is parametric, the first law entails the second. This is sometimes called the **free theorem** for functors (Wadler 1989). Even better, for any given constructor $\Sigma$, there is at most one parametric function ($\bullet$) that satisfies **Identity**. This means that if there is a uniform mapping operation for an effect $\Sigma$, then it is unique.

Intuitively, what those functor laws say is that for an operation ($\bullet$) to count as a map, it should not interact in any way with the effect structure of $\Sigma$, concentrating all its energy on applying $k$ to the embedded $\alpha$. So mapping the identity function $\mathbf{id} := \lambda x.\, x$ over a structure $X$ should not change anything at all about $X$. This is because $\mathbf{id}$ does not change the $\alpha$ element(s) in $X$, and ($\bullet$) does not change the non-$\alpha$ elements in $X$ or the structure of $X$ itself. Additionally, a map should be a homomorphism with respect to function composition; it shouldn't matter whether you map a composite operation $f \circ g := \lambda x.\, f\,(g\,x)$ over $X$, or first map $g$ over $X$ and then $f$ over the result.

Not every notion of a computational context is like this. For instance, the type $\boxed{\mathtt{E}\,\alpha} ::= \alpha \to \alpha$ is not functorial. A value in the domain of $\boxed{\mathtt{E}\,\alpha}$ is a function from $\alpha$ to $\alpha$. There is no obvious sense in which such a function can be construed as a structure enveloping some $\alpha$ values. Concretely, there's no principled way to map a function $k : \alpha \to \beta$ over an $\boxed{\mathtt{E}\,\alpha}$ computation to get an $\boxed{\mathtt{E}\,\beta}$ computation. This would amount to a way of composing an $\alpha \to \beta$ function with an $\alpha \to \alpha$ function to get a $\beta \to \beta$ function.

In Haskell, the $(\bullet)$ operation is known as `fmap`.

```
class Functor f where
  fmap :: (a -> b) -> f a -> f b
```

For many type constructors `Σ`, finding the relevant mapping operation is a straightforward matter of following the types. Take `M`, for instance. Handed a function $k : \alpha \to \beta$ and a computation of type $\boxed{\mathtt{M}\,\alpha}$, we must produce a computation of type $\boxed{\mathtt{M}\,\beta}$. Well, a computation of type $\boxed{\mathtt{M}\,\alpha}$ is either an $\alpha$ or an error #, and the computation we need to build is either a $\beta$ or an error. So, two options: if we got lucky and got an $\alpha$, we should apply $k$ to it to create our $\beta$; if we got unlucky, we remain unlucky. In other words:

(2.6) $\quad k \bullet_{\mathtt{M}} X := \#\ \textbf{if}\ X = \#,\ \textbf{otherwise}\ (k\,X)$

What else could we possibly write down that would have the required type? Nothing, in fact. This is the only parametric function of type $(\alpha \to \beta) \to \boxed{\mathtt{M}\,\alpha} \to \boxed{\mathtt{M}\,\beta}$ that there is.

Indeed, any constructor whose parameter instances $\alpha$ are all in positive positions is guaranteed to be a functor, and its $(\bullet)$ mechanically derivable.[2] (This actually allows `Functor` instances to be automatically inferred in Haskell via `deriving Functor`!) As it happens, this includes all of the examples in Table 2. Many of these effects are so essential to the day to day organization of programs that they have canonical names in Haskell, spelled out in Table 3. The `fmap`s of these constructors are defined in the standard prelude or in standard libraries. We give simplified versions of these definitions in Appendix A3.

Notice that some of the constructors in the table have additional type parameters. For instance, where we write $\boxed{\mathtt{R}_\iota\,\alpha} ::= \iota \to \alpha$, the corresponding Haskell constructor is declared as `data Reader i a = Reader (i -> a)`. The `i` here identifies the type of the environment that anaphoric computations read from. As discussed in Section 1.4, this environment can take a great variety of forms, depending on the linguistic theory and phenomena of interest, so any particular computation will need to specify which form is assumed. Likewise, the W/`Writer` effect is parameterized by what kind of values are stored in the supplemental dimension, and the T/`State` effect by how discourse contexts are encoded, and so on. Since these choices mostly do not matter for our purposes, we will generally leave these parameters implicit to avoid typographic clutter, writing, for instance, $R\ e$. When we do make particular choices, we will discuss the relevant types in surrounding prose.

[2]If a type is viewed as a tree with the constructors as node labels and their parameter types as daughters, then (i) the root of every type is considered positive, and (ii) every node preserves the polarity of its parent except for the left-hand side of the arrow, which reverses it. So, for instance, in the type $\alpha \times (((\alpha \to \beta) \to \alpha) \times (\beta \to \alpha))$, every $\alpha$ is in a $\boxed{\text{positive}}$ position and every $\beta$ a $\boxed{\text{negative}}$ one, as seen by inspecting the tree:

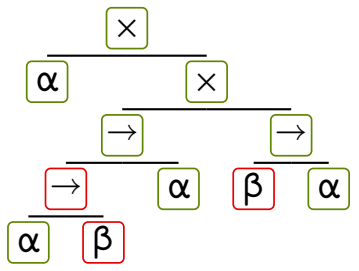

| Formal Type | Haskell Analog |
|---|---|
| $R_\iota\ \alpha ::= \iota \to \alpha$ | `data Reader i a = Reader (i -> a)` |
| $W_o\ \alpha ::= \alpha \times o$ | `data Writer t a = Writer (a, t)` |
| $M\ \alpha ::= \alpha + \bot$ | `data Maybe a = Just a \| Nothing` |
| $S\ \alpha ::= \{\alpha\}$ | `[a] -- the list constructor [] is built-in syntax` |
| $T_\sigma\ \alpha ::= \sigma \to (\alpha \times \sigma)$ | `data State s a = State (s -> (a, s))` |
| $C_\rho\ \alpha ::= (\alpha \to \rho) \to \rho$ | `data Cont r a = Cont ((a -> r) -> r)` |

**Table 3** Haskell conventions for common effect types

### *2.1.1 Mapping as a mode of combination*

How does any of this help with problems of composition stressed in Chapter 1? No doubt the simplest thing to do would be to add (•) to the inventory of combinatory modes, perhaps with a forward version and a backward version in analogy with ordinary forward and backward Function Application.

Such combinators, defined in Figure 3, provide for derivations like those in (2.7). Both of the derived constituents in (2.7) combine a pronoun with an ordinary predicate of entities. To keep things quite simple at the start, let us adopt a **variable-free** view of anaphora, wherein the computations denoted by pronouns are mere requests for antecedents, nothing more (see, e.g., Jacobson 1999, 2014, et seq.). In this conception, anaphora resolution is the process of choosing how such requests should be fulfilled (what values to pass in for the open arguments), but the pronouns themselves do no selectional work as part of their semantics. That is, a pronoun's job is just to make the request, and then to hand over the antecedent it receives for further composition. They are, therefore, identity functions on antecedents.

**Combinators**

$$(>) :: (\alpha \to \beta) \to \alpha \to \beta$$ Forward Application
$$f > a \coloneqq f\,a$$

$$(<) :: \alpha \to (\alpha \to \beta) \to \beta$$ Backward Application
$$a < f \coloneqq f\,a$$

$$\vdots$$ Other Basic Combinators

$$(\bullet_{>}) :: (\alpha \to \beta) \to \boxed{\Sigma\,\alpha} \to \boxed{\Sigma\,\beta}$$ Forward Map
$$f \bullet_{>} A \coloneqq f \bullet A$$

$$(\bullet_{<}) :: \boxed{\Sigma\,\alpha} \to (\alpha \to \beta) \to \boxed{\Sigma\,\beta}$$ Backward Map
$$A \bullet_{<} f \coloneqq f \bullet A$$

**Figure 3** Adding basic (•) combinators to the grammar

(2.7)

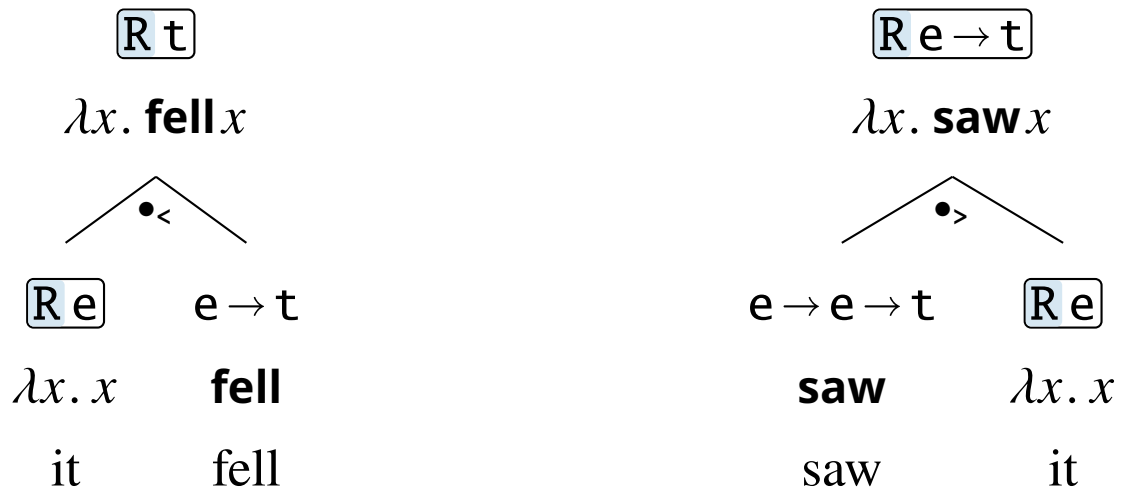

We should highlight here that $\boxed{\mathtt{R}\,\mathtt{e} \to \mathtt{t}}$ stands for $\mathtt{R}(\mathtt{e} \to \mathtt{t})$. In general, the scope of a constructor prefixed to a bounding box $\boxed{\phantom{x}}$ is the entire box. This saves parentheses and results in more readable complex types.

One nice aspect of the grammar in Figure 3 is that there is no substantive difference between the way effect-typed expressions compose in subject vs. object positions. As usual with natural language, a function may occur on either the left or right side of its argument, but the semantics is the same in both cases. However, with just the combinatory inventory of Figure 3, it is not possible to combine a computational predicate — type $\boxed{\mathtt{R}\,\mathtt{e} \to \mathtt{t}}$ — with an ordinary subject — type $\mathtt{e}$. This is because the mapping operation (•) always lifts an ordinary *function* over an effectful *argument*, though what is needed in this case is the application of an effectful function to an ordinary argument. We could add further modes of combination that do this using (•), or we could allow ordinary arguments to be **lifted** into ordinary functions à la Partee (1986).

(2.8) $(\cdot)^{\text{LIFT}} :: \alpha \to (\alpha \to \tau) \to \tau$

$x^{\text{LIFT}} := \lambda k.\, k\, x$

This latter technique is shown in (2.9a), which certainly solves the immediate problem of composing an ordinary value with a computation that yields a function. But the real trouble for this simple-minded approach to incorporating functoriality begins when an expression and its sister *both* denote computations. This situation is illustrated in (2.9b). Even with free lifting, there is no way to put the subject and predicate of (2.9b) together.[3]

(2.9a)

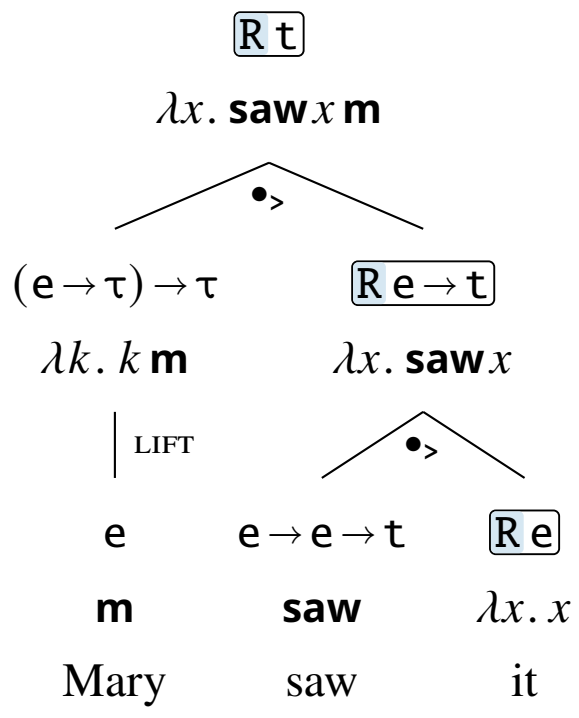

(2.9b)

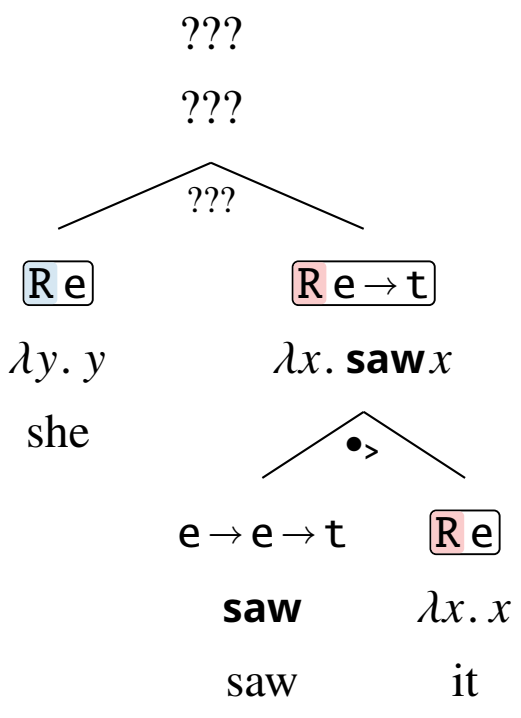

## 2.2 Higher-order effects

The first step in redressing this unfortunate incomposability is deciding what sort of thing (2.9b) ought to denote. The type of the subject, R e, indicates that it needs an antecedent. The type of the predicate, R e → t, indicates that it also, independently, needs an antecedent. The denotation of the full sentence should honor both of these requests, so its type should indicate that it needs *two* antecedents; after one request has been fulfilled, say by passing a value in for the object, the other request will remain outstanding. It is, in this sense, a computation whose result is another computation. We will refer to such nested computational structures as exhibiting **higher-order effects**.

[3] The various colors used to highlight type constructors have no semantic significance; they are just there to help see at a glance how effects are distributed throughout a denotation.

(2.10) she saw it :: $\mathtt{R}\,(\mathtt{R}\,\mathtt{t})$

$[\![\text{she saw it}]\!] = \lambda x \lambda y.\, \textbf{saw}\, x\, y$

In the next two sections, we show two ways to use the functoriality of $\mathtt{R}$ to arrive at the denotation in (2.10). Both take advantage of the following idea. Even though $\mathtt{R}\,\mathtt{e}$ and $\mathtt{R}\,(\mathtt{e} \to \mathtt{t})$ cannot be combined (because neither can be mapped over the other), the *underlying* type of the predicate, $\mathtt{e} \to \mathtt{t}$, could be combined with the *full* type of the subject $\mathtt{R}\,\mathtt{e}$, as in (2.7). And because $\mathtt{R}$ is a functor, we should be able to map this underlying mode of combination, type $(\mathtt{e} \to \mathtt{t}) \to \mathtt{R}\,\mathtt{e} \to \mathtt{R}\,\mathtt{t}$, over the $\mathtt{R}\,(\mathtt{e} \to \mathtt{t})$ predicate. Since the mode of combination that combines the subject, $\mathtt{R}\,\mathtt{e}$, with the underlying type of the predicate, $\mathtt{e} \to \mathtt{t}$, is $(\bullet)$, what we need then is some way to map $(\bullet)$ itself over one of the daughters.

### 2.2.1 Mapping in the language

The most direct route to this sort of higher-order mapping is to add $(\bullet)$ to the object language. Then $(\bullet)$ might be mapped over a computation just the same as any other function. This is effectively the strategy that Jacobson (1999) adopts, though with combinators specific to $\mathtt{R}$ and none of our effect-oriented conceit.[4]

[4]The reader may wish to compare (2.12) to Jacobson's (1999: p. 139) Example (31), given in (2.11a), and transliterated to our notation in (2.11b) below. The translation proceeds by rewriting $\mathbf{g}_0$ as $(\bullet)$ and recognizing that Jacobson's $\mathbf{g}_n$ is equivalent to $\mathbf{g}_0\,\mathbf{g}_{n-1}$. Then in (2.11c), we write as many $(\bullet)$'s as possible as infix operators, yielding a logical form isomorphic to the tree in (2.12).

(2.11a) $\mathbf{g}_0\; (\textsc{lift}\, [\![\text{his mother}]\!])\; (\mathbf{g}_1\; (\mathbf{g}_0\; [\![\text{loves}]\!]\; [\![\text{his dog}]\!]))$

(2.11b) $(\bullet)\; (\textsc{lift}\, [\![\text{his mother}]\!])\; ((\bullet)\; (\bullet)\; ((\bullet)\; [\![\text{loves}]\!]\; [\![\text{his dog}]\!]))$

(2.11c) $\textsc{lift}\, [\![\text{his mother}]\!] \bullet ((\bullet) \bullet ([\![\text{loves}]\!] \bullet [\![\text{his dog}]\!]))$

(2.12)

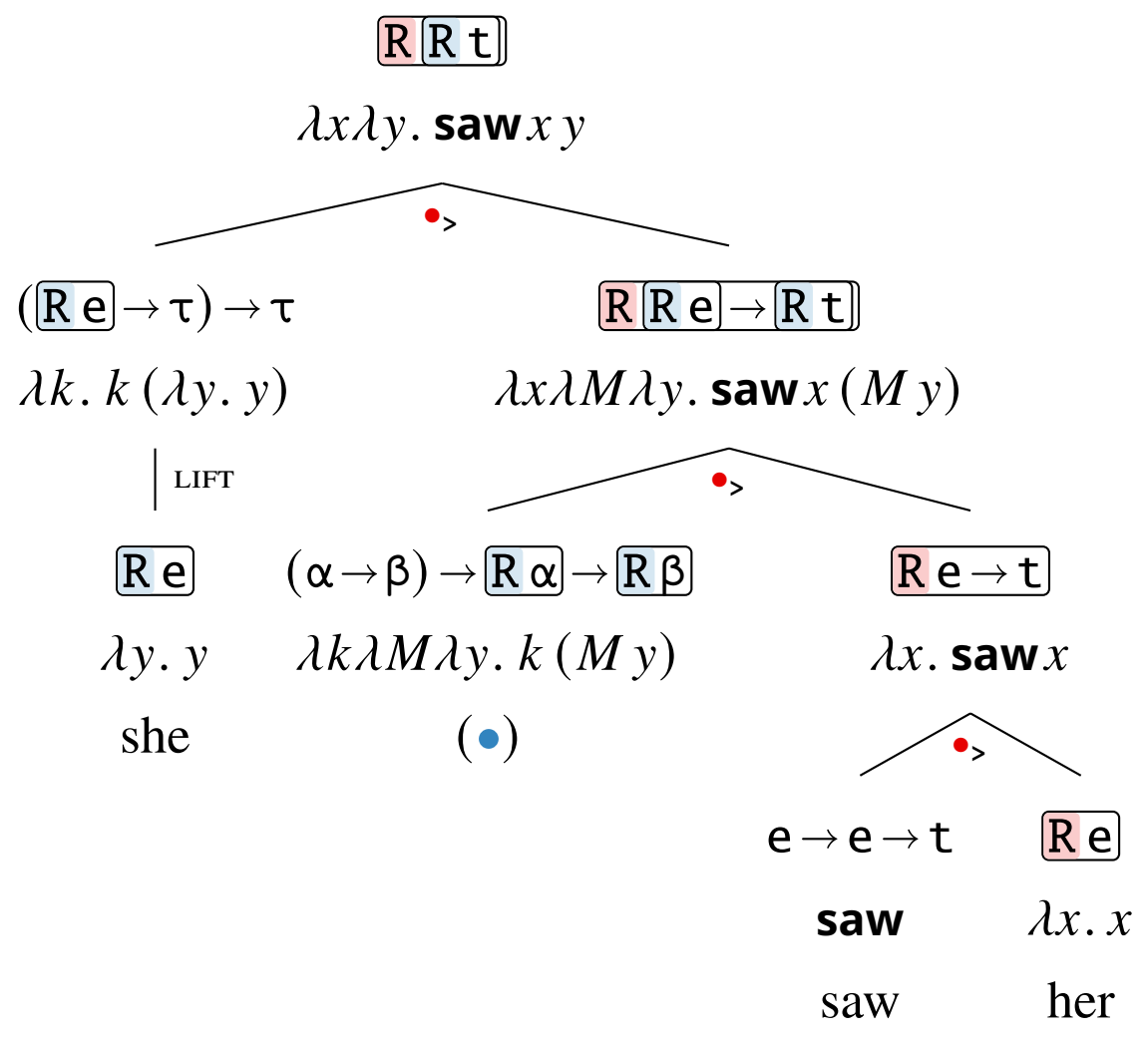

Because all of the effects we've introduced are functorial, and the only interesting aspects of the derivation in (2.12) are the (•)s, the tree is a template for composition with any of the enriched meanings of Table 2. For instance, switching R for S, immediately derives a multiple-'wh' question as in (2.13).

(2.13)

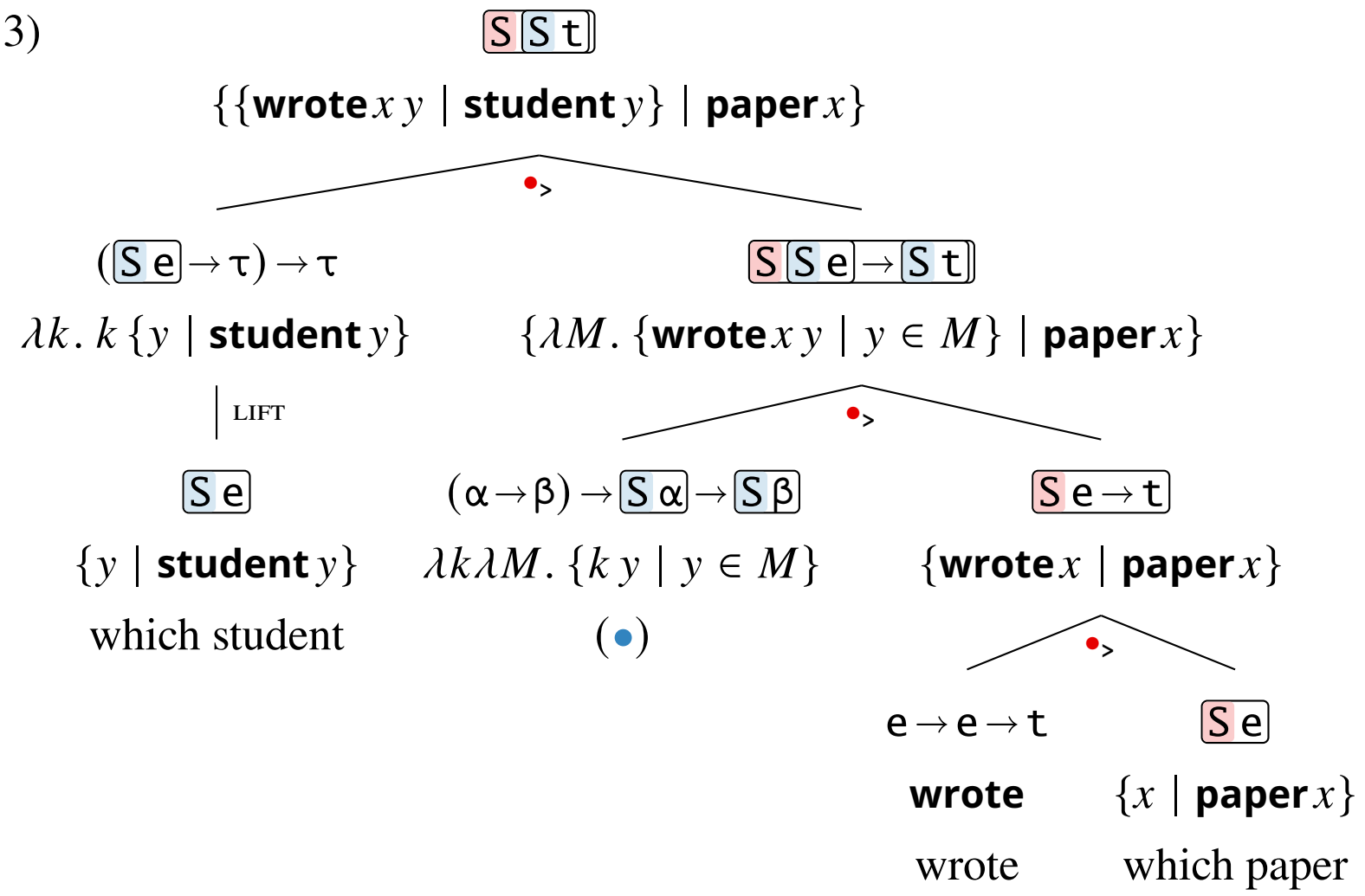

Moreover, the various instances of (•) are completely independent. All that matters is that the effects of the subject and object are both functorial; but they needn't be *the same* functor. So the template works just as well for sentences containing different kinds of effects, rather than multiple instances of an effect, as in (2.14).

(2.14)

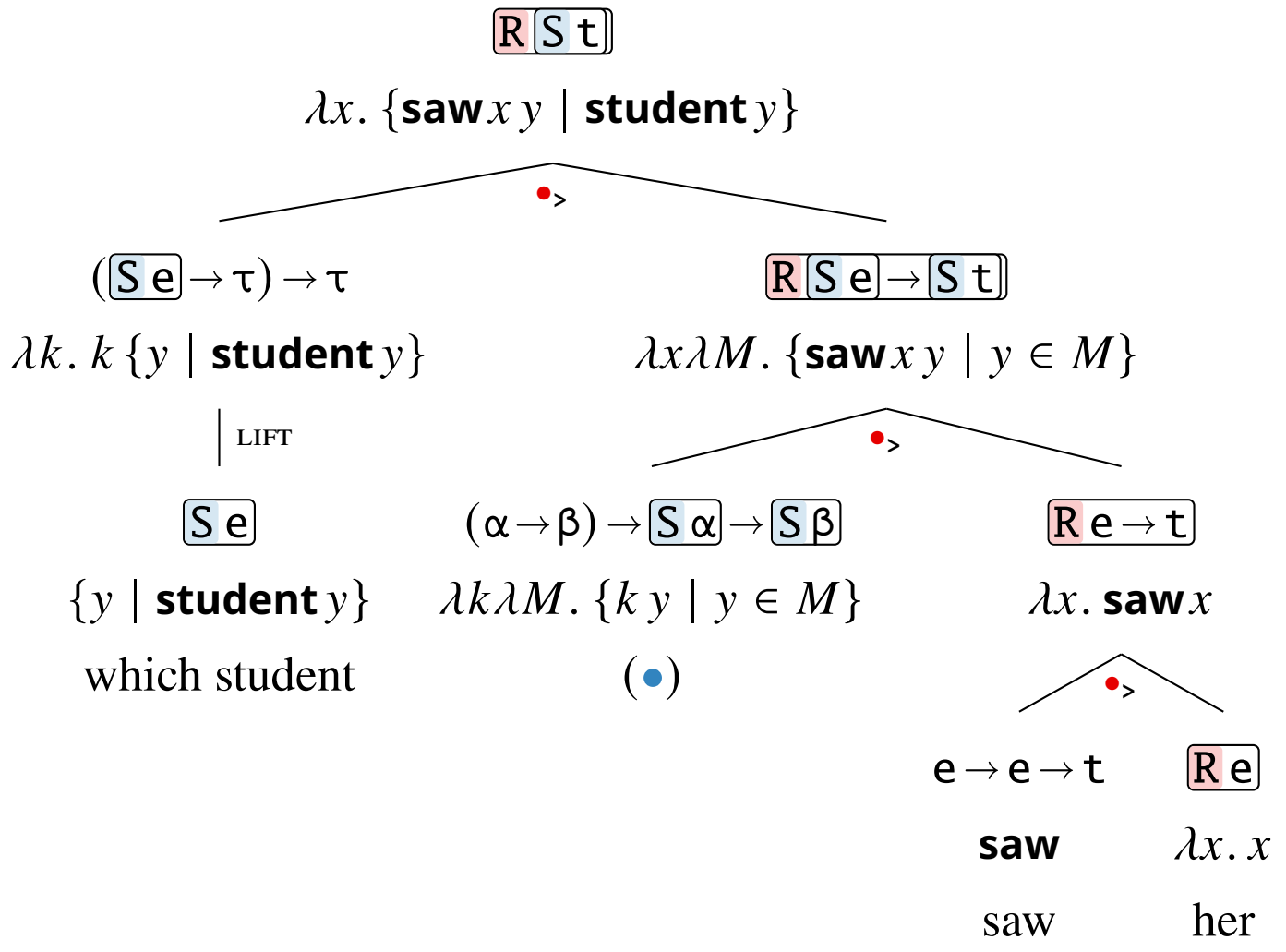

As straightforward as this object-language reification of (•) is, there are a few reasons for discontent. On the empirical side, effects can pop up just about anywhere, including in the daughters of constituents that would otherwise combine via arbitrary modes of combination. As a result, (•) is not the only mode of combination that will need to be mapped. For instance, as things stand there is no way to combine an ordinary property with a computational one.

(2.15)

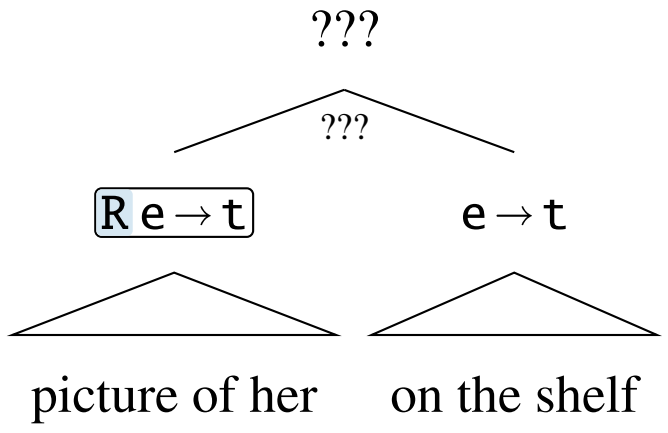

Following (2.12), we would have to treat the Predicate Modification combinator as a lexical item, or at least as a unary type-shifter like LIFT so that it can be partially applied. Doing this would make possible the derivation in (2.16).

(2.16)

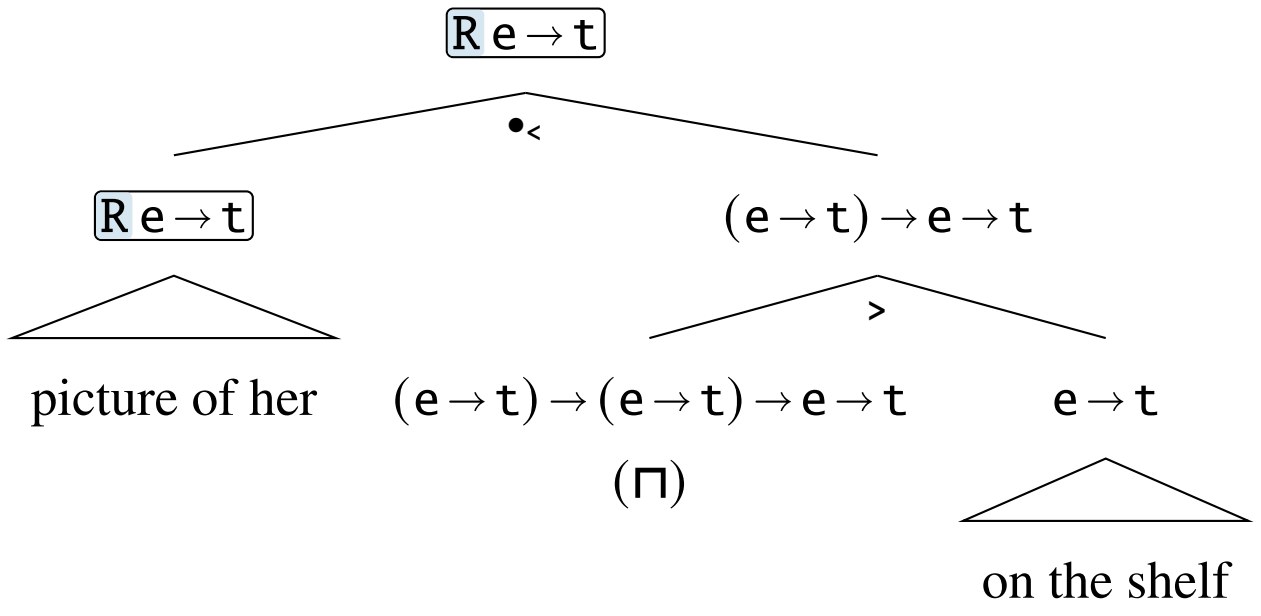

In this manner, eventually all modes of combination will need to be realized lexically. Whether this is syntactically justifiable is open to debate, but it certainly increases the distance between the forms that are uttered and the forms that are interpreted. And as a practical matter, the resulting combinatorial system is admittedly unwieldy. Even with practice, derivations are hard to find. Sentences with multiple effects often require a great deal of creativity to compose, mapping and lifting partially applied combinators over constituents.

Anyone who needs convincing of this should try deriving the sentence in (2.12) so that the subject's antecedent-request outscopes the object's. That is, the blue R should precede the red R in the final type signature R R t, and the final denotation should be $\lambda y \lambda x.\, \textbf{saw}\, x\, y$. (It is possible.) Worse, given that combinators can apply to one another iteratively and without bound, it can be exceedingly difficult to rule readings out. You never quite know when some cleverer insertion of maps and lifts would do the trick.

### *2.2.2 Mapping as a higher-order mode of combination*

For these reasons, we offer an alternative to the Jacobsonian vision. Once more, the key to putting together sentences with multiple non-interacting effects is the ability to map ($\bullet$) itself. But as seen in (2.16), this is not enough. We will want to be able to map an arbitrary mode of combination ($*$) over one of the daughters. In the previous section, this was accomplished by embedding combinators in the object language and using $\bullet_>$ as a binary mode of combination to partially apply them to the relevant daughters one at a time.

Instead, we might just as well provision the grammar with a means of constructing complex combinators from simpler ones, just as complex types are constructed from simpler types. In general, if there is a mode ($*$) that can combine constituents $M :: \sigma$ and $N :: \tau$, then there should also be a mode to combine constituents $M :: \sigma$ and $N' :: \boxed{\Sigma\,\tau}$, provided that $\Sigma$ is a functor. Intuitively, there is a $\tau$ thing sitting inside $N'$ just waiting to be combined with $M$ via $(*)$. So the enriched mode should map $(\lambda b.\, M * b)$ over $N'$. And vice versa, if $M'$ is of type $\boxed{\Sigma\,\sigma}$ and $N :: \tau$. These **mode-transforming** operations are defined in (2.17), and the grammar incorporating these higher-order combinators is given in Figure 4.

(2.17a) $\overleftarrow{\mathsf{F}}\,(*)\, E_1\, E_2 \coloneqq (\lambda a.\, a * E_2) \bullet E_1$

(2.17b) $\overrightarrow{\mathsf{F}}\,(*)\, E_1\, E_2 \coloneqq (\lambda b.\, E_1 * b) \bullet E_2$

From these, the $(\bullet_{>})$ mode emerges as the special case of $\overrightarrow{\mathsf{F}}$ applied to $(>)$ and $(\bullet_{<})$ as $\overleftarrow{\mathsf{F}}$ applied to $(<)$, as seen in (2.18).

(2.18)

$$\begin{aligned} \overrightarrow{\mathsf{F}} > f\, E_2 &= (\lambda b.\, f > b) \bullet E_2 \\ &= (\lambda b.\, f\, b) \bullet E_2 \\ &= f \bullet E_2 \\ &= f \bullet_{>} E_2 \end{aligned} \qquad \begin{aligned} \overleftarrow{\mathsf{F}} < E_1\, f &= (\lambda a.\, a < f) \bullet E_1 \\ &= (\lambda a.\, f\, a) \bullet E_1 \\ &= f \bullet E_1 \\ &= E_1 \bullet_{<} f \end{aligned}$$

Example derivations using these higher-order modes of combination are given in (2.19). Notice that it does not matter which daughter the effect is in or which daughter takes the other as argument (if either). Because the higher-order map and lower-order mode are independent, our combinatory bases are covered however the functions and effects are oriented.

(2.19a)

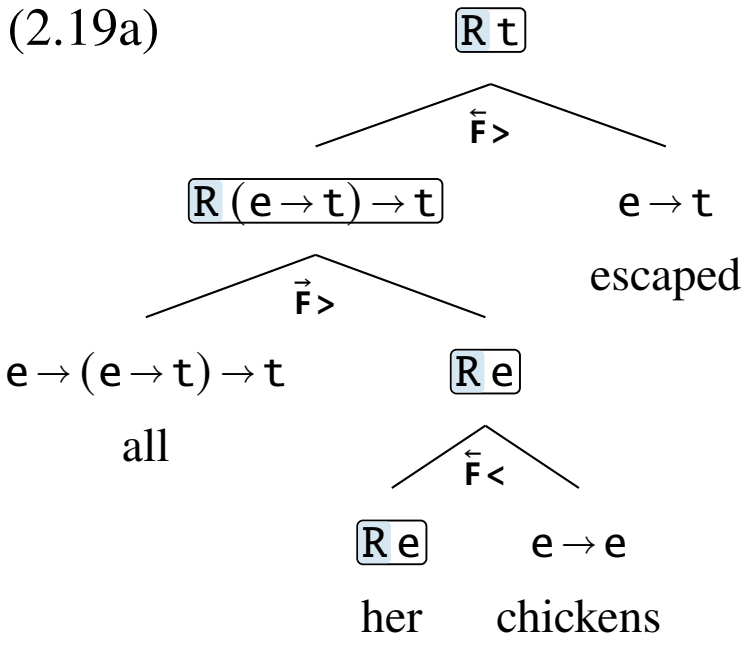

(2.19b)

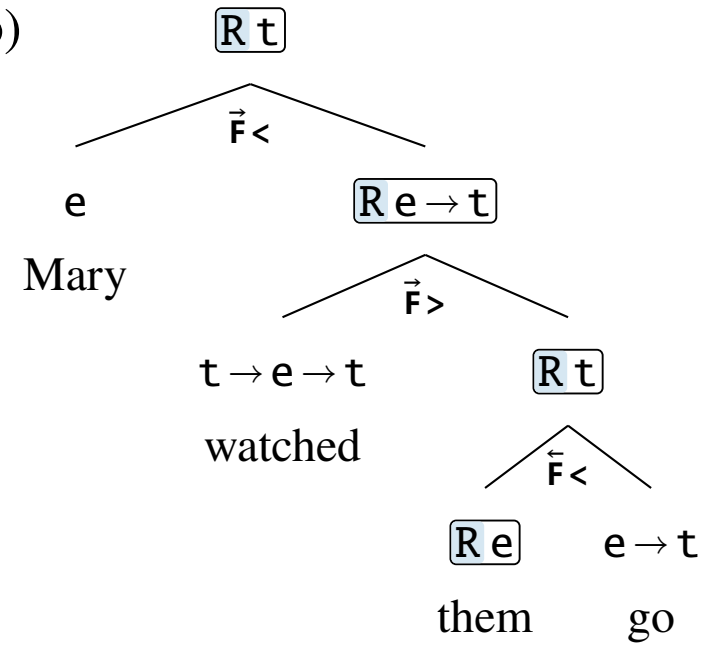

(2.19c)

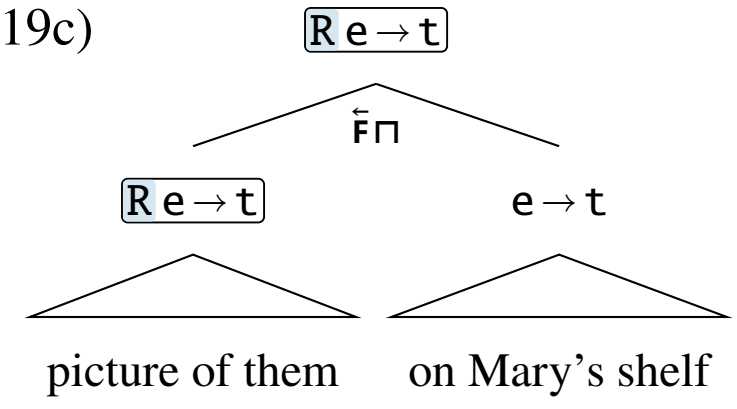

(2.19d)

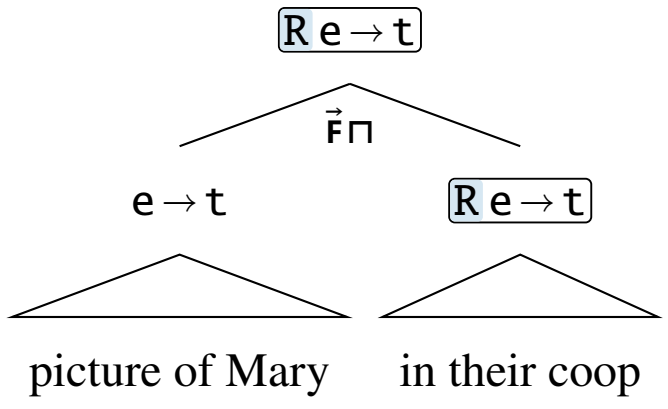

**Types:**

| | |
|---|---|
| $\tau ::= \mathtt{e} \mid \mathtt{t} \mid \cdots$ | Base types |
| $\mid\ \tau \to \tau$ | Function types |
| $\mid\ \tau \times \tau$ | Product types |
| $\mid\ \tau + \tau$ | Sum types |
| $\mid\ \{\tau\}$ | Set types |
| $\mid\ \boxed{\Sigma\,\tau}$ | Computation types |

**Effects:**

| | |
|---|---|
| $\Sigma ::= \mathtt{R}$ | Input |
| $\mid\ \mathtt{W}$ | Output |
| $\mid\ \mathtt{S}$ | Indeterminacy |
| $\mid\ \ldots$ | $\ldots$ |

**Basic Combinators:**

$$(\mathbin{>}) :: (\alpha \to \beta) \to \alpha \to \beta \qquad \text{Forward Application}$$
$$f \mathbin{>} x \coloneqq f\,x$$

$$(\mathbin{<}) :: \alpha \to (\alpha \to \beta) \to \beta \qquad \text{Backward Application}$$
$$x \mathbin{<} f \coloneqq f\,x$$

$$(\sqcap) :: (\mathtt{e} \to \mathtt{t}) \to (\mathtt{e} \to \mathtt{t}) \to \mathtt{e} \to \mathtt{t} \qquad \text{Predicate Modification}$$
$$f \sqcap g \coloneqq \lambda x.\, f\,x \wedge g\,x$$

$\ldots$ $\ldots$

**Meta-combinators:**

$$\overleftarrow{\mathsf{F}} :: (\sigma \to \tau \to \omega) \to \boxed{\Sigma\,\sigma} \to \tau \to \boxed{\Sigma\,\omega} \qquad \text{Map Left}$$
$$\overleftarrow{\mathsf{F}}\,(*)\,E_1\,E_2 \coloneqq (\lambda a.\, a * E_2) \bullet E_1$$

$$\overrightarrow{\mathsf{F}} :: (\sigma \to \tau \to \omega) \to \sigma \to \boxed{\Sigma\,\tau} \to \boxed{\Sigma\,\omega} \qquad \text{Map Right}$$
$$\overrightarrow{\mathsf{F}}\,(*)\,E_1\,E_2 \coloneqq (\lambda b.\, E_1 * b) \bullet E_2$$

**Figure 4** A type-driven grammar with functorial effects

Obviously the lexical types and constituencies here should be taken with a grain of salt. The point is just to illustrate that the effects themselves no longer necessitate any type-shifting or covert object-language combinators. The maps are in the mergers, so derivations have exactly the same shape as they would if they were effect-free.

On occasion, authors have suggested that combination might alternate as needed between variants of $(>)$, $(\bullet_>)$, $(<)$, and $(\bullet_<)$ for some specific effect or other (e.g., Krifka (1992: 25) for focus, Hagstrom (1998: 142) for questions). But the clearest antecedent for isolating an implicit map and parameterizing it to arbitrary underlying modes of combination, as we have done here, comes from Barker and Shan (2014: 118ff). There, versions of $\overrightarrow{\mathsf{F}}$ and $\overleftarrow{\mathsf{F}}$ (and also $\mathbf{A}$, which we introduce in Chapter 3) are defined for the continuation functor $\mathsf{C}$. The technique is introduced as a means of tractably approximating a grammar with free-floating type-shifters of the sort sketched in Section 2.2.1, rather than as providing an alternative to it, but the authors recognized that the recursion in the higher-order rules gives rise to a kind of derivational scope ambiguity, as we discuss in the next section. Throughout this Element, we build on this insight and attempt to show the virtues of generalizing this method to arbitrary functorial effects, and later extend the higher-order combinators to other sorts of compositionally fruitful algebraic structures.

## 2.3 Effect layering

Importantly, the operators in (2.17) are iterative in the sense that they take a binary mode $(*)$ and return a new binary mode, $\overleftarrow{\mathsf{F}}*$ or $\overrightarrow{\mathsf{F}}*$. This means they can in principle apply to their own output. For instance, since $\overleftarrow{\mathsf{F}}*$ is a binary mode, $\overleftarrow{\mathsf{F}}(\overleftarrow{\mathsf{F}}*)$ is yet another mode, as is $\overrightarrow{\mathsf{F}}(\overleftarrow{\mathsf{F}}*)$, and likewise for $\overleftarrow{\mathsf{F}}(\overrightarrow{\mathsf{F}}*)$ and $\overrightarrow{\mathsf{F}}(\overrightarrow{\mathsf{F}}*)$.

What do these higher-order combinators amount to? Particularly illuminating are the cases where both daughters are computations: $\overleftarrow{\mathsf{F}}(\overrightarrow{\mathsf{F}}*)$ and $\overrightarrow{\mathsf{F}}(\overleftarrow{\mathsf{F}}*)$. Cranking through the definitions gives:

(2.20)
$$\overleftarrow{\mathsf{F}}(\overrightarrow{\mathsf{F}}*) = \lambda E_1 \lambda E_2.\, (\lambda a.\, (\lambda b.\, a * b) \bullet E_2) \bullet E_1$$
$$\overrightarrow{\mathsf{F}}(\overleftarrow{\mathsf{F}}*) = \lambda E_1 \lambda E_2.\, (\lambda b.\, (\lambda a.\, a * b) \bullet E_1) \bullet E_2$$

For illustrative purposes, let us rewrite these equations with the order of $(\bullet)$'s arguments flipped, so that the computation comes first and the to-be-mapped function second. That is, let's swap out $(\bullet)$ for $(\bullet_<)$, making the relevant adjustments. This gives the equations in (2.21).

(2.21)
$$\overleftarrow{\mathsf{F}}(\overrightarrow{\mathsf{F}}*) = \lambda E_1 \lambda E_2.\, E_1 \bullet_< (\lambda a.\, E_2 \bullet_< (\lambda b.\, a * b))$$
$$\overrightarrow{\mathsf{F}}(\overleftarrow{\mathsf{F}}*) = \lambda E_1 \lambda E_2.\, E_2 \bullet_< (\lambda b.\, E_1 \bullet_< (\lambda a.\, a * b))$$

In this form, the derived higher-order modes reveal a striking resemblance to **Quantifier Raising** (though most functors have nothing to do with quantification and there is certainly no syntactic raising here), as illustrated below.

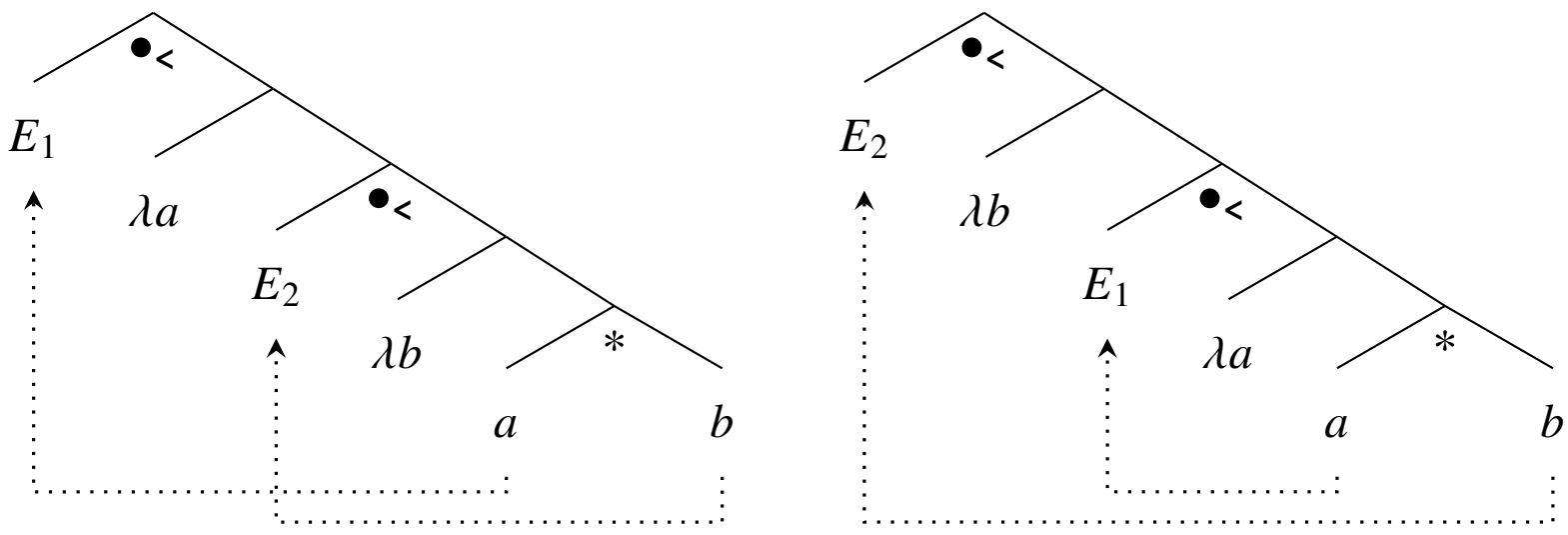

We discuss the relationship between scope and effects in Chapter 4, but it is worth noting that the diagrams above already suggest a sense in which computation-denoting constituents **take scope** over their compositional contexts. Choosing to map anything over the left daughter gives the left daughter's effect priority over whatever is mapped. For instance, if both daughters request antecedents, then the left daughter's request will come first. Choosing instead to map something over the right daughter gives the right daughter's effect priority. The difference can be seen in (2.22).[5]

(2.22)

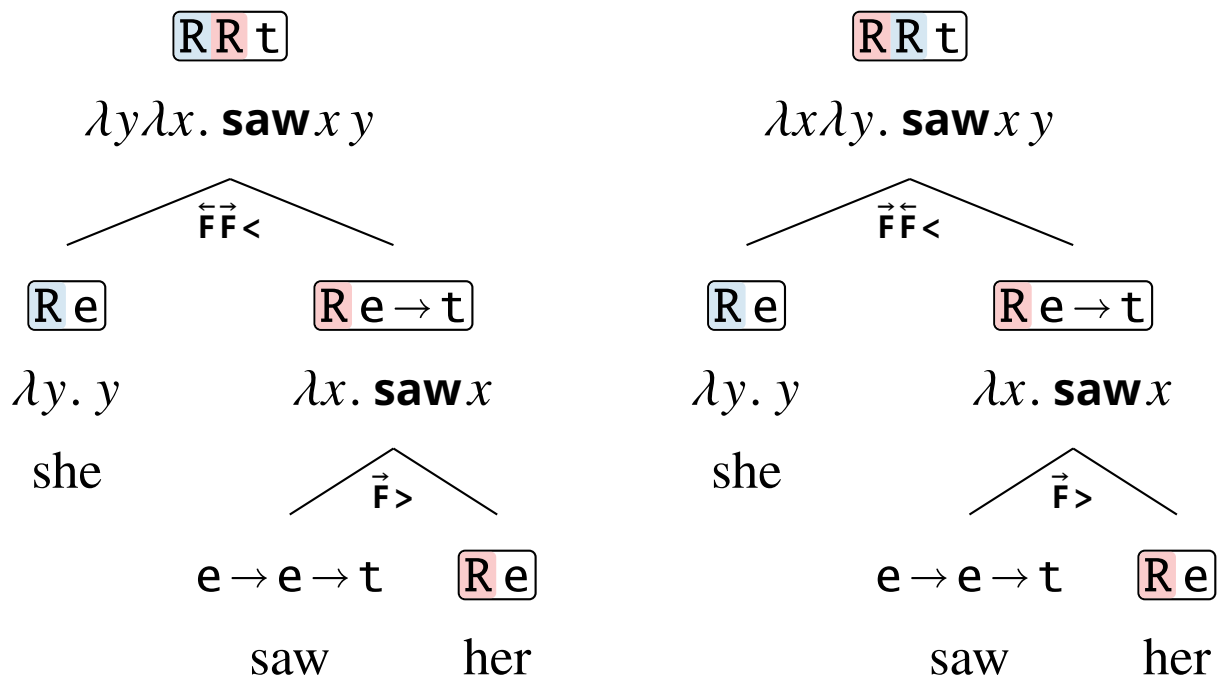

[5]To cut out some notational clutter, we will often display both higher-order effect signatures and higher-order modes of combination as lists, rather than right-nested embeddings:

$$\boxed{\mathtt{I}\ \Sigma\ \mathtt{K}\ \Lambda\ \alpha} \equiv \boxed{\mathtt{I}\ \boxed{\Sigma\ \boxed{\mathtt{K}\ \boxed{\Lambda\ \alpha}}}} \qquad\qquad \overleftarrow{\mathsf{F}}\,(\overrightarrow{\mathsf{F}}\,(\overrightarrow{\mathsf{F}}>)) \equiv \overleftarrow{\mathsf{F}}\overrightarrow{\mathsf{F}}\overrightarrow{\mathsf{F}}>$$

What this priority amounts to depends on the nature of the effect. But because the operations involved here work for any functorial constructor, we can immediately combine constituents with different kinds of effects, often in multiple ways. For instance, a context-sensitivite predicate and an indeterminate subject can be combined in the two ways shown in (2.23).

(2.23a)

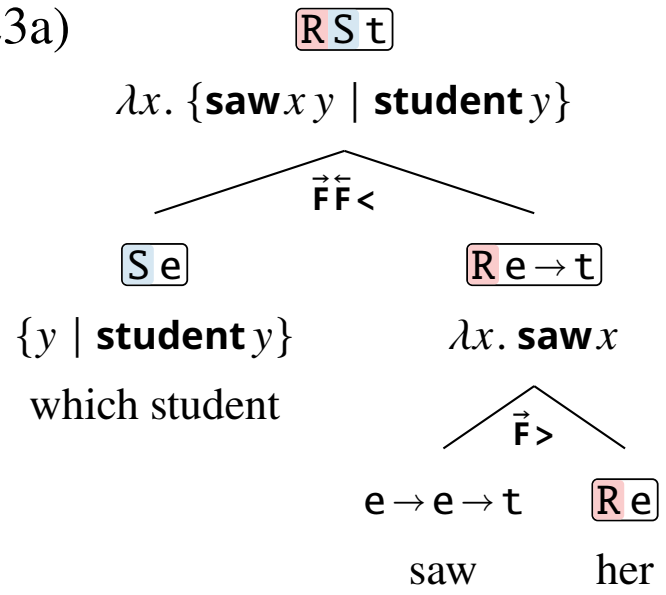

(2.23b)

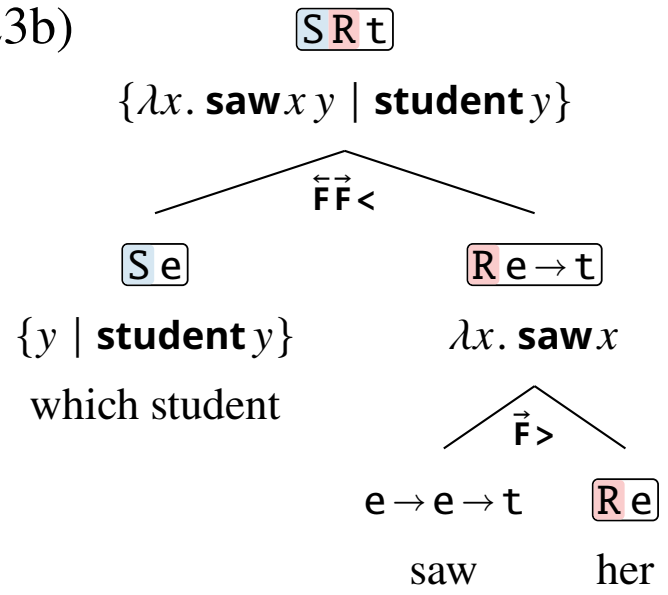

In fact, denotations along both of these lines have been proposed in the literature on questions and indeterminacy. The denotation at the root of (2.23a) is a variable-free version of what would standardly be assigned in a "Hamblin Semantics" (Hamblin 1973, Hagstrom 1998, and Kratzer and Shimoyama 2002, among others). Here, the sentence demands to know who the referent of 'her' is, and only then determines a set of propositions corresponding to the question 'which student saw her'.

The denotation at the root of (2.23b) is a variable-free version of the sort of meaning assigned to focus structures in Rooth 1985, ambiguous expressions in Poesio 1996, and questions in Romero and Novel 2013. In this style, the sentence immediately evaluates to a set of values, one per student, but each one of those values is semantically incomplete, still waiting to find out who 'her' refers to.

All of those authors felt compelled to *choose* one or the other form, and indeed to set *all* of the nodes in the tree in that type. This is partly because they took it for granted that one should commit to a particular fixed shape for all denotations with a single application-like mode of combination. The principal issues motivating one or the other form have to do with binding and quantification, which we will see more of in Chapters 3 and 4. For now, suffice it to say that the functoriality of R and S, together with higher-order modes of combination, means that there is no immediate pressure to settle on one of the meanings in (2.23). We needn't generalize to either potential "worst case".

To close this section, let us look at the other two cases of higher-order combination, in which a single daughter harbors two effects:

(2.24) $\overleftarrow{\mathsf{F}}\,(\overleftarrow{\mathsf{F}}\,*) = \lambda E_1 \lambda E_2.\, (\lambda X.\, (\lambda a.\, a * E_2) \bullet X) \bullet E_1$

$\overrightarrow{\mathsf{F}}\,(\overrightarrow{\mathsf{F}}\,*) = \lambda E_1 \lambda E_2.\, (\lambda X.\, (\lambda a.\, E_1 * a) \bullet X) \bullet E_2$

In particular, if the left daughter, $f$, has type $\sigma \to \tau$, and the right daughter $\mathcal{J}$ has type $\boxed{\Sigma\,\boxed{K\,\sigma}}$, then these may be combined as

$$\boxed{\Sigma\,\boxed{K\,\tau}}$$
$$(\lambda K.\, f \bullet_{K} K) \bullet_{\Sigma} \mathcal{J}$$
$$\overrightarrow{\mathsf{F}}\overrightarrow{\mathsf{F}}\,>$$
$$\sigma \to \tau \qquad \boxed{\Sigma\,\boxed{K\,\sigma}}$$
$$f \qquad \mathcal{J}$$

If $f$ is the identity function, then the respective **Identity** laws for $\Sigma$ and $K$ guarantee that the result, $(\lambda K.\, f \bullet_{K} K) \bullet_{\Sigma} \mathcal{J}$, is equal to $\mathcal{J}$. What this shows is that the $\overrightarrow{\mathsf{F}}\,(\overrightarrow{\mathsf{F}}\,>)$ combinator is a map for the composite effect $\boxed{\Sigma K\,\alpha} ::= \boxed{\Sigma\,\boxed{K\,\alpha}}$. (This is in fact what underlies our notational conflation.) In other words, we have discovered a fundamental property of functors: they are **closed under composition**. If $\Sigma$ and $K$ are both functors, then $\Sigma K$ is also a functor, with $(\bullet_{\Sigma K}) = \overrightarrow{\mathsf{F}}\,(\overrightarrow{\mathsf{F}}\,>)$.

## 2.4 Functors and pseudoscope

Absent any sort of closure operators, which we discuss in Chapter 3, functorial effects **percolate** up the tree in which they're composed. This is plainly evident from the types. In (2.25), for example, the genitive pronoun embedded in the object introduces an anaphoric dependency to the denotation, and that dependency is inherited by every node dominating it in the tree.

(2.25)

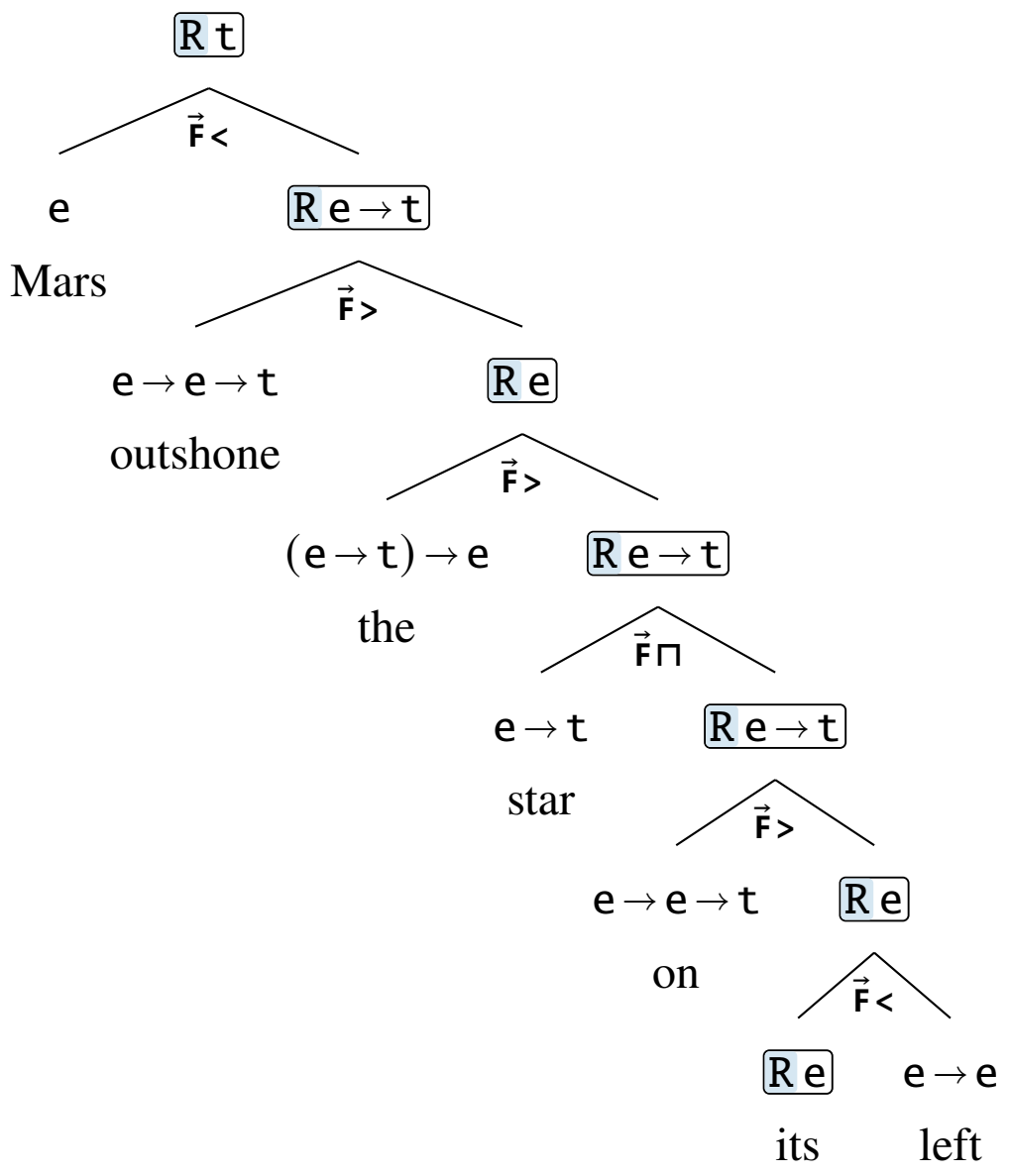

Once composed, the entire sentence becomes context-dependent; it requests an antecedent in order to compute a truth value. In a sense, the pronoun's computational effect — requesting an antecedent — is displaced from the location of the pronoun itself. In fact, (2.25) is equivalent to what we'd get if we "Quantifier"-Raised the pronoun and mapped its syntactic context, as in (2.26).

(2.26)

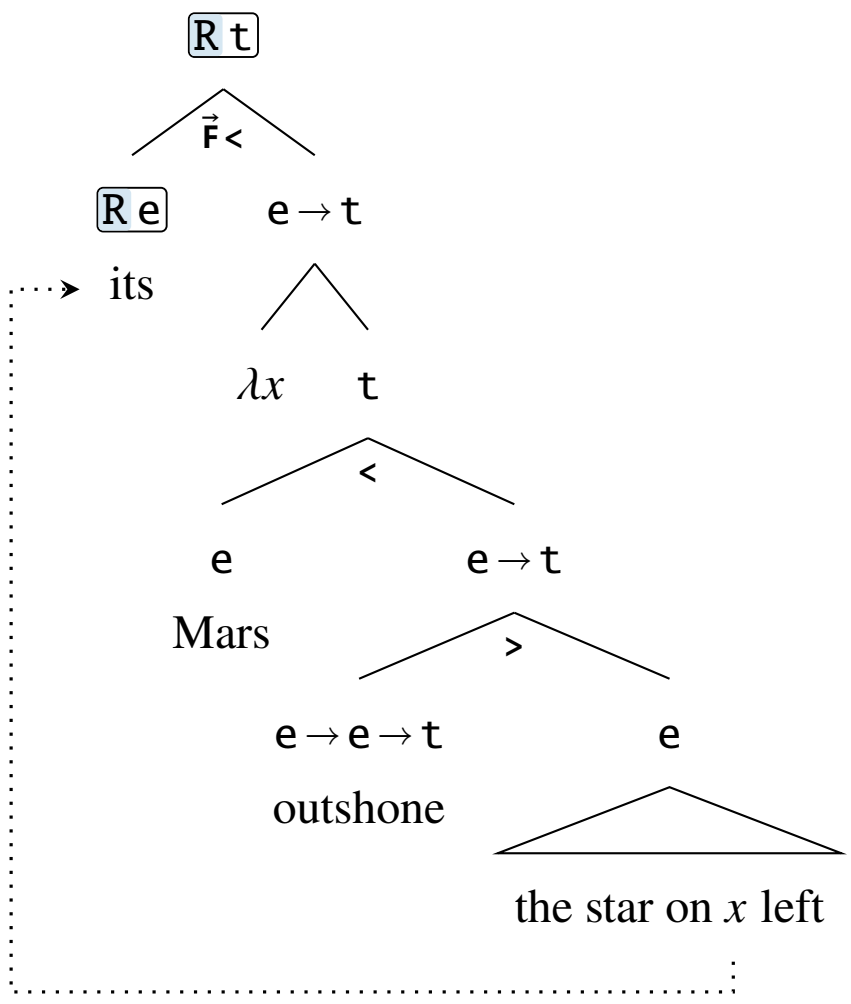

With a little effort, this equivalence can be seen to follow from the Law of Composition in (2.4). Just rewriting the law as in (2.27) using the combinators introduced in this chapter, together with a suggestive but meaningless arrow, gets most of the way there.

(2.27) $f \bullet_{>} (g \bullet_{>} A) = A \bullet_{<} (\lambda x.\, f > (g > x))$

Since the equivalence is algebraic rather than due to some quirk of context-dependence, it guarantees that for *any* functorial effect, mapping can be seen as a means of giving the effect scope over its compositional context. That is, repeatedly mapping over an embedded computation is equivalent to scoping the computation out of the way and mapping once where it lands.

But importantly, the scope provided by ($\bullet$) does not depend on any assumptions about syntactic transformations. In particular, there is no reason to expect effect-percolation to exhibit sensitivity to the sorts of **islands** that govern movement. And indeed, all of the effects in Table 2 — with the exception of quantifiers, which we return to in Chapter 5 — are island *in*sensitive.

For instance, consider the examples in (2.28).

(2.28a) Who remembers when who left?

(2.28b) Mary only gets mad when JOHN leaves the lights on

(2.28c) Mary hopes that because her cat, named Sassy, is home, John is too

(2.28d) Mary's being out of town means that if you don't see John's car, you can be sure nobody's home

These sentences all contain an effect-denoting expression within an island, either an embedded question or embedded adjunct. Yet they all also have readings in which the semantic force of that embedded expression is felt outside of the island. For example, the question in (2.28a) can be understood as asking which pairs of people $\langle a, b\rangle$ are such that $a$ remembers when $b$ left. The sentence in (2.28b) declares John to be the only person such that Mary gets mad when he leaves the lights on. Example (2.28c) commits the speaker to Mary's cat being named Sassy, regardless of what Mary knows or whether her cat is home. Likewise, (2.28d), as a whole, presupposes that John has a car, even though the presupposition trigger 'John's car' is in an embedded hypothetical.

In these construals, the effect-generating expressions are sometimes said to take **exceptional scope** over the islands that embed them. The capacity for exceptional scope is a hallmark of a functorial effect, about which we will have more to say in Chapter 3.

## 2.5 Implementing functorial effects in the type-driven interpreter

Notice that all of the derivations in Section 2.3 are syntactically spare. Nothing is inserted into the plain constituency tree, and no types are shifted. Expressions are simply combined according to their types, using the higher-order modes of combination to navigate effects, *as needed*. As discussed in Chapter 1, one of the main benefits of this type-driven approach is that it takes the creativity out of composition. In this section, we demonstrate that there is an effective procedure for determining all of the possible combinations of any two types, just as there was for the basic grammar in Figure 1. We do this by extending the Haskell interpreter of Section 1.5 to cover the grammar of Figure 4.

First, we need to expand our representation of types to incorporate type constructors modeling effects. Following Figure 4, we say that a type `Ty` can be atomic or functional, as before, but also now computational. A computation type is parameterized by an effect `EffX`, which we discuss below.

```
data Ty
  = E | T | V      -- primitive types
  | Ty :-> Ty      -- function types
  | Comp EffX Ty   -- computation types
  deriving (Eq, Show)
```

Next, we expand our inventory of combinatory modes. The new modes $\vec{\mathsf{F}}$ and $\overleftarrow{\mathsf{F}}$ are meta-combinators; they take modes as arguments and return modes.

(2.29)
$$\vec{\mathsf{F}}\,(*)\,E_1\,E_2 \coloneqq (\lambda b.\,E_1 * b) \bullet E_2$$
$$\overleftarrow{\mathsf{F}}\,(*)\,E_1\,E_2 \coloneqq (\lambda a.\,a * E_2) \bullet E_1$$

Our representations `MR` and `ML` of these meta-combinators are thus parameterized by this underlying mode $(*)$ :: `Mode`.

```
data Mode
  = FA | BA | PM        -- basic modes of combination
  | MR Mode | ML Mode   -- map right and map left
  deriving (Show)
```

As far as type-driven combination is concerned, the only thing we need to know about an effect is its label, and whether or not it is a functor. None of the grammatical, combinatoric operations inspect the internal structure of an effect. Indeed, this is the whole point of the algebraic abstractions. Knowing that an effect `Σ` is functorial is enough to know that it can be combined using $\vec{\mathsf{F}}/\overleftarrow{\mathsf{F}}$. Of course the actual semantics will depend on how the effect is encoded (whether it is a product, a set, a function into sets, etc.) and how (•) is defined for that encoding, but the logic of type-driven composition needn't bother with such matters.

Consequently, the representation of effects `EffX` includes just enough information to drive the combinatorics, namely a label indicating what kind of effect it is and parameters for whatever incidental data the computation is specialized to (the type of the environment it reads from, or the type of the data that it stores, or the type of context it quantifies over, etc.). Note that all of the effects we consider here are functors, so the `functor` predicate happens to be vacuous. But we include it for good measure, and to set the stage for future chapters.

```
data EffX
  = SX    -- computations with indeterminate results
  | RX Ty -- computations that query an environment of type Ty
  | WX Ty -- computations that store information of type Ty
  | CX Ty -- computations that quantify over Ty contexts
  -- and so on for other effects, as desired
  deriving (Eq, Show)

functor :: EffX -> Bool
functor _ = True
```

With these representations fixed, we define the logic of combination in the function `combine` below. This is again a simple matter of pattern-matching on the types of the daughters. For starters, if the daughters `l` and `r` can be combined via any of the basic modes of combination from Chapter 1, then go for it. Recall that the function `modes` returns a list containing whatever basic `Mode`s are applicable to combining `l` and `r`, together with the `Ty`pe that would result from so combining them.

In addition, we check for two other possibilities. The function `addMR` returns an empty list — adding no new modes of combination to what the basic `modes` was able to find — unless the right daughter's type is a computation type `Comp f t` with a functorial effect `f`. If it is, then we try to `combine` the left daughter `l` with the right daughter's underlying type `t`. That recursive call will produce a list of possible `Mode`s and resulting `Ty`pes. If there is no way to combine `l` and `t`, then the new list will again be empty, adding nothing to the basic `modes`. But if it is possible the combine `l` and `t`, then for each way of doing so `(op, u)`, we build a new higher-order mode `MR op`, signaling that `op` can be mapped over the right daughter, resulting in a combined type `Comp f u`.

The case for checking that the left daughter `l` is functorial is exactly symmetric to the right one. Importantly, these two investigations `addMR` and `addML` are not exclusive. If both daughters are functorial, and the underlying types can be combined, they will both return new substantive modes of combination.

```
combine :: Ty -> Ty -> [(Mode, Ty)]
combine l r =
  -- see if any basic modes of combination work
  modes l r
  -- if the right daughter is functorial, try to map over it
  ++ addMR l r
  -- if the left daughter is functorial, try to map over it
  ++ addML l r

addMR l r = case r of
  Comp f t | functor f
    -> [ (MR op, Comp f u) | (op, u) <- combine l t ]
  _ -> [                                            ]

addML l r = case l of
  Comp f s | functor f
    -> [ (ML op, Comp f u) | (op, u) <- combine s r ]
  _ -> [                                            ]
```

The behavior of `combine` can be appreciated by loading these definitions into `ghci` (the interactive Haskell interpreter), and then querying the results of some of the key type combinations considered in this chapter. The following examples correspond, respectively, to combinations of R e and e→t, e and R e→t, R e and R e→t, and S e and R e→t.

```
ghci> combine (Comp (RX E) E) (E :-> T)
[(ML BA, Comp (RX E) T)]
```

```
ghci> combine E (Comp (RX E) (E :-> T))
[(MR BA, Comp (RX E) T)]
```

```
ghci> combine (Comp (RX E) E) (Comp (RX E) (E :-> T))
[(MR (ML BA), Comp (RX E) (Comp (RX E) T)),
 (ML (MR BA), Comp (RX E) (Comp (RX E) T))]
```

```
ghci> combine (Comp SX E) (Comp (RX E) (E :-> T))
[(MR (ML BA), Comp (RX E) (Comp SX T)),
 (ML (MR BA), Comp SX (Comp (RX E) T))]
```

Finally, the top-level interpreter that annotates trees is exactly as it was in Chapter 1, except that the basic `modes` function is upgraded to the recursive `combine` function.

```
synsem :: Lexicon Syn -> [Sem]
synsem lex syn = case syn of
  (Leaf w)           -> [Lex t w | t <- lex w]
  (Branch lsyn rsyn) ->
    [ Comb ty op lsem rsem
      | lsem     <- synsem lex lsyn
      , rsem     <- synsem lex rsyn
      , (op, ty) <- combine (getType lsem) (getType rsem) ]
  where
    getType (Lex ty _)       = ty
    getType (Comb ty _ _ _) = ty
```

# 3 Applicative Functors

## 3.1 Merging effects

As seen in Chapter 2, one of the benefits of using (•) or $\vec{\mathsf{F}}$ / $\overleftarrow{\mathsf{F}}$ for composition is that it accommodates any number and variety of computations. Whatever their shape, the side effects of these computations are passed over in order to get at the underlying argument-structural values and compose them as appropriate.

However, this means that when two constituents with independent effects are combined, the result is necessarily **higher-order**, a computation that returns another computation. For instance, with two 'wh'-expressions, we end up with a set of sets of propositions. With two pronouns, we end up with a function from an antecedent to a function from an antecedent to a proposition. And so on.

(3.1)

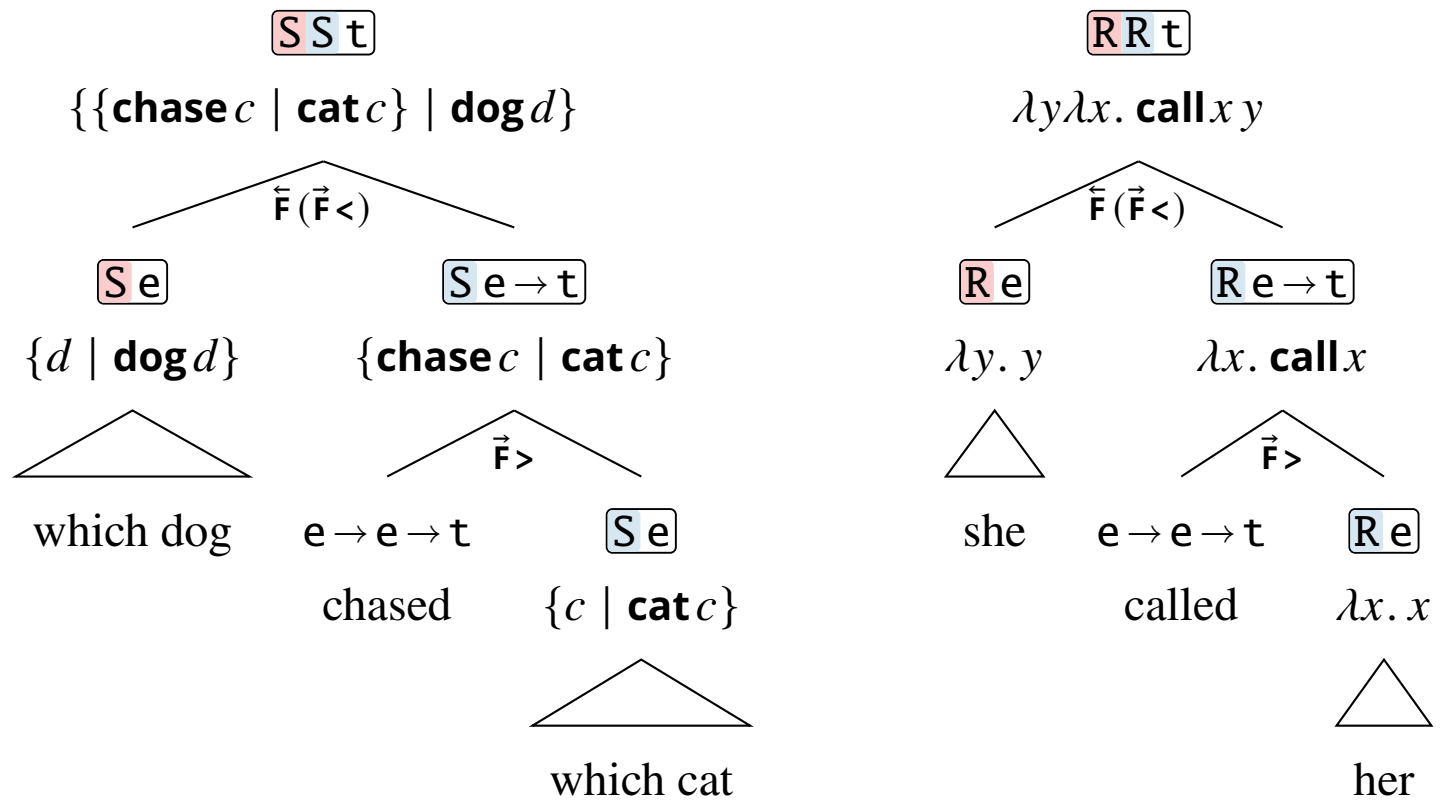

Even among theories that countenance these sorts of higher-order meanings, they are not generally taken to represent the default interpretations of such sentences, much less their only interpretations. Consider, for instance, the typical **variable-full** format for managing pronouns (as in, e.g., Heim and Kratzer 1998). Every constituent is evaluated relative to a sequence/variable assignment, whether it contains any pronouns or not. If it does, those pronouns select values from particular coordinates of the assignment. If it does not, the assignment is ignored. And importantly, when two sisters are both evaluated, they are evaluated at *the same* assignment.

A fragment with this shape is outlined in Figure 5. Compared to the grammar in Figure 4, several things stand out. The environment-sensitivity effect is always outermost in a type, and every expression's type is of the form $\boxed{\mathtt{R}\,\sigma}$ for some ordinary type $\sigma$. Accordingly, all of the modes of combination expect their daughters to be environment-sensitive, and return an environment-sensitive result. This means that lexical items themselves must all be coerced into environment-reading computations, whether they pay any real attention to the environment or not.

Exactly the same pattern emerges in a standard Hamblin grammar for questions (Hamblin 1973), as sketched in Figure 6. Every constituent denotes an indeterminate computation — modeled by the set of values it might return — whether that constituent contains a ‘wh’-expression or not. If it does, then the ‘wh’-expressions generate genuine alternatives. If it does not, then the denotation is a singleton value. Importantly, when two sisters are both evaluated, they are evaluated *pointwise*, so that the alternatives generated in the two daughters are amalgamated into a single large set.

Again, as seen in Figure 6, indeterminacy is pervasive and top-level; every expression’s type is of the form $\boxed{\mathtt{S}\,\sigma}$. All the modes of combination expect indeterminate daughters, and return indeterminate results. And all lexical items are coerced into indeterminate computations, whether they generate any alternatives or not.

These grammars then deliver derivations as in (3.2), where the various computational components are continuously merged into a single effect layer whenever two constituents are combined.

(3.2)

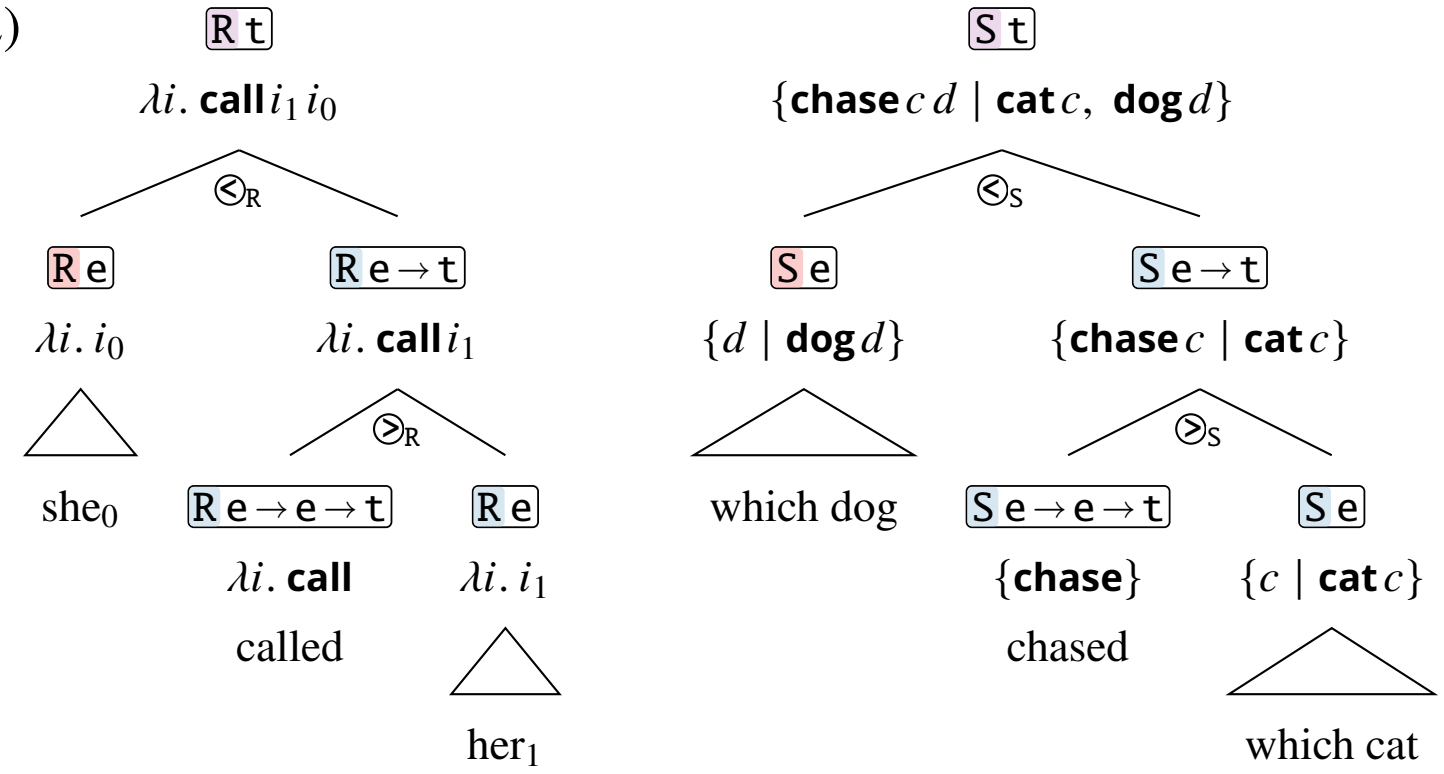

However, both of these grammars are examples of generalization to the worst case, forcing *every constituent* to match the complexity of the few constituents of semantic interest. Unfortunately, neither case is as bad as it can get. Simply mixing the two kinds of phenomena is beyond the reach of either grammar. The rigid, pervasive replacement of ordinary combinatory modes with effect-specific

**Types:**

$\sigma ::= \mathsf{e} \mid \mathsf{t} \mid \ldots$ Base pre-types

$\quad \mid \ \sigma \to \sigma$ Function pre-types

$\tau ::= \boxed{\mathsf{R}\,\sigma}$ Expression types

**Combinators:**

$(\circledR_{\mathrm{R}}) :: \boxed{\mathsf{R}\,\alpha\to\beta} \to \boxed{\mathsf{R}\,\alpha} \to \boxed{\mathsf{R}\,\beta}$

$F \circledR_{\mathrm{R}} A \coloneqq \lambda i.\, F\,i\,(A\,i)$

$(\circledL_{\mathrm{R}}) :: \boxed{\mathsf{R}\,\alpha} \to \boxed{\mathsf{R}\,\alpha\to\beta} \to \boxed{\mathsf{R}\,\beta}$

$A \circledL_{\mathrm{R}} F \coloneqq \lambda i.\, F\,i\,(A\,i)$

**Lexicon:**

$\text{it}_n :: \boxed{\mathsf{R}\,\mathsf{e}}$
$\coloneqq \lambda i.\, i_n$

$\text{Mars} :: \boxed{\mathsf{R}\,\mathsf{e}}$
$\coloneqq \lambda i.\, \mathbf{m}$

$\text{cat} :: \boxed{\mathsf{R}\,\mathsf{e}\to\mathsf{t}}$
$\coloneqq \lambda i.\, \mathbf{cat}$

$\text{nobody} :: \boxed{\mathsf{R}\,(\mathsf{e}\to\mathsf{t})\to\mathsf{t}}$
$\coloneqq \lambda i.\, \mathbf{nobody}$

...

**Figure 5** Env.-sensitive grammar

**Types:**

$\sigma ::= \mathsf{e} \mid \mathsf{t} \mid \ldots$ Base pre-types

$\quad \mid \ \sigma \to \sigma$ Function pre-types

$\tau ::= \boxed{\mathsf{S}\,\sigma}$ Expession types

**Combinators:**

$(\circledR_{\mathrm{S}}) :: \boxed{\mathsf{S}\,\alpha\to\beta} \to \boxed{\mathsf{S}\,\alpha} \to \boxed{\mathsf{S}\,\beta}$

$F \circledR_{\mathrm{S}} A \coloneqq \{f\,a \mid f \in F,\ a \in A\}$

$(\circledL_{\mathrm{S}}) :: \boxed{\mathsf{S}\,\alpha} \to \boxed{\mathsf{S}\,\alpha\to\beta} \to \boxed{\mathsf{S}\,\beta}$

$A \circledL_{\mathrm{S}} F \coloneqq \{f\,a \mid f \in F,\ a \in A\}$

**Lexicon:**

$\text{who} :: \boxed{\mathsf{S}\,\mathsf{e}}$
$\coloneqq \{x \mid \mathbf{person}\,x\}$

$\text{Mars} :: \boxed{\mathsf{S}\,\mathsf{e}}$
$\coloneqq \{\mathbf{m}\}$

$\text{cat} :: \boxed{\mathsf{S}\,\mathsf{e}\to\mathsf{t}}$
$\coloneqq \{\mathbf{cat}\}$

$\text{nobody} :: \boxed{\mathsf{S}\,(\mathsf{e}\to\mathsf{t})\to\mathsf{t}}$
$\coloneqq \{\mathbf{nobody}\}$

...

**Figure 6** Indeterminate grammar

variants limits the applicability of the fragment to just the specific effects described. What is gained in uniformity and simplicity is sacrificed in generality and extensibility.

## 3.2 Applicative Functors

Fortunately, the underlying strategy of both the Heim and Kratzer grammar for environment-sensitivity and the Hamblin grammar for interrogativity can be made modular and algebraic, in line with the generic mapping operation ($\bullet$) of Chapter 2. Both strategies draw on a pair of essential operations: foremost, a means of composing a computation that yields a function $\boxed{\Sigma\, \alpha \to \beta}$ with one that yields an argument $\boxed{\Sigma\, \alpha}$ to form a computation yielding a result $\boxed{\Sigma\, \beta}$; and secondarily, a means of injecting an ordinary value $\alpha$ into a **unitary**, or "trivial", computation.

Intuitively, a computation is trivial if it adds no effect of any consequence. It is a computation that does nothing except return a value. A trivial environment-sensitive computation is a constant function; it requests an environment but makes no use of it, so that it doesn't *really* read from the environment at all. A trivial indeterminate computation is a singleton set; it computes exactly one thread, so that there isn't *really* any parallelism to speak of. It is a singleton set.

Technically, what it means for a computation to be trivial depends on how effect-generating functions and effect-generating arguments are combined, and in particular on how the effects that they generate are combined. This is the only way to know whether the potentially trivial effect really does not change anything. One natural abstraction for this sort of relationship is known as an **applicative functor** (McBride and Paterson 2008, Kiselyov 2015), which we'll call an applicative for short.

A type constructor $\Sigma$ is applicative if there are operations $\eta$ and ($\circledast$) with the types indicated in (3.3) respecting the laws in (3.4).

(3.3)
$$\eta :: \alpha \to \boxed{\Sigma\, \alpha}$$
$$(\circledast) :: \boxed{\Sigma\, \alpha \to \beta} \to \boxed{\Sigma\, \alpha} \to \boxed{\Sigma\, \beta}$$

(3.4)

**Homomorphism**
$\eta\, f \circledast \eta\, x = \eta\,(f\, x)$

**Identity**
$\eta\, \mathbf{id} \circledast X = X$

**Interchange**
$\eta\,(\lambda k.\, k\, x) \circledast F = F \circledast \eta\, x$

**Composition**
$(\eta\,(\circ) \circledast F \circledast G) \circledast X = F \circledast (G \circledast X)$

In Haskell, these operations are known as `pure` and `(<*>)`.

```
class Functor f => Applicative f where
  pure  :: a -> f a
  (<*>) :: f (a -> b) -> f a -> f b
```

Alongside types for the `Applicative` operations `pure` and `(<*>)`, this definition declares `Applicative` as a **subclass** of `Functor`. That is, if `f` is `Applicative`, `f` is also necessarily a `Functor` with an associated `fmap :: (a -> b) -> f a -> f b` (this point is elaborated on below). So the Haskell subclass notation is somewhat 'backwards': `Applicative f` implies `Functor f`, rather than the converse.

It's easy to see that the combinators and implicit lexical coercion mechanisms of Figures 5 and 6 emerge from the following `R` and `S` applicative functors. Again, readers are encouraged to check that these respective pairs of operations satisfy the applicative laws in (3.4).

(3.5)
$$\eta_{\text{R}}\, x := \lambda i.\, x \qquad\qquad \eta_{\text{S}}\, x := \{x\}$$
$$F \circledast_{\text{R}} X := \lambda i.\, F\, i\, (X\, i) \qquad\qquad F \circledast_{\text{S}} X := \{f\, x \mid f \in F,\ x \in X\}$$

In fact, all of the effects in Table 2 are applicative functors. We provide the canonical parametric instances of $\eta$ and $\circledast$ for each of these type constructors in Appendix A4, though it's worth flagging that for some of the them, there are slight variants on the definitions that also satisfy the laws (see Section 3.5). And as might be expected from the terminology, every applicative functor is a functor, in that whenever $\eta$ and $(\circledast)$ satisfy the laws in (3.4), the $(\bullet)$ operation defined in (3.6) will satisfy the laws in (2.4). Indeed, using the recipe in (3.6), the **Identity** applicative law in (3.4) immediately entails the **Identity** functor law in (2.4). It follows then from the free theorem for functors that for any `Σ`, the $(\bullet)$ defined by (3.6) is the unique parameteric map for its type.

(3.6)
$$k \bullet m := \eta\, k \circledast m$$

Putting these operations to work in deriving natural language meanings looks much the same as it did in Chapter 2. We might, for instance, add directionally-oriented versions of $(\circledast)$ as new modes of combination, mirroring the naive functorial grammar of Figure 3. This is exactly the strategy underlying the Heim and Kratzer and Hamblin grammars in Figures 5 and 6, where:

(3.7a)
$$(\circledast_{>}) :: \boxed{\Sigma\,\alpha\to\beta}\to\boxed{\Sigma\,\alpha}\to\boxed{\Sigma\,\beta}$$
$$F \circledast_{>} X := F \circledast X$$

(3.7b)
$$(\circledast_{<}) :: \boxed{\Sigma\,\alpha}\to\boxed{\Sigma\,\alpha\to\beta}\to\boxed{\Sigma\,\beta}$$
$$X \circledast_{<} X := F \circledast X$$

If $\eta$ were additionally added as a unary **type-shifter**, like LIFT, then derivations would look exactly as they do in (3.1), except that the transitive verbs would be shifted by $\eta$ just before combination.

The trouble with this straightforward approach is also the same as in Chapter 2. There is no general way to combine constituents with multiple effects. That is, while ($\circledast$) guarantees a way to put together two type-`S` constituents, or two type-`R` constituents, it does not by itself suffice as a means to combine two `R S` constituents, or two `S R` constituents. And such constituents are by no means exotic. All it takes is a left branch with an alternative-generator and a pronoun, and a right branch with an alternative-generator and a pronoun, as in (3.8).

(3.8)

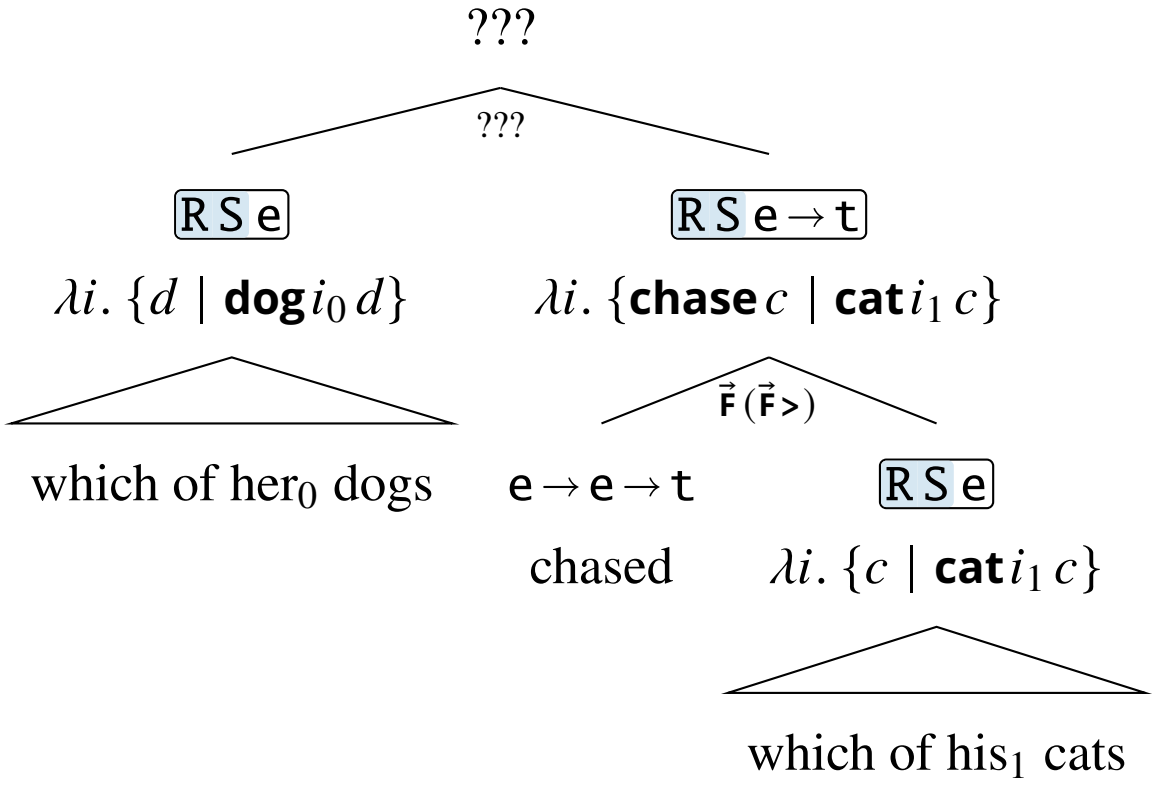

The solutions to higher-order applicative combination are again the same as in Chapter 2. The problem is that `R S e` and `R S e → t` cannot be combined via $⧀_{\mathtt{R}}$ because neither `R`-computation delivers a function. They both deliver further computations: `S e` and `S e → t`. But of course, *those* computations can be combined via $⧀_{\mathtt{S}}$. And since `R` is an applicative functor, we should be able to map the $⧀_{\mathtt{S}}$ mode of combination over the outer `R` layer, merging the two `R` effects in the process.

### *3.2.1 Applicatives in the language*

Just as in Section 2.2.1, higher-order effect management can be coaxed out of these basic applicative grammars by adding $\eta$ and ($\circledast$) directly to the lexicon (Charlow 2018, Charlow 2022). This immediately opens the door to derivations like (3.9), paralleling the derivations in (2.12)–(2.14).

(3.9)

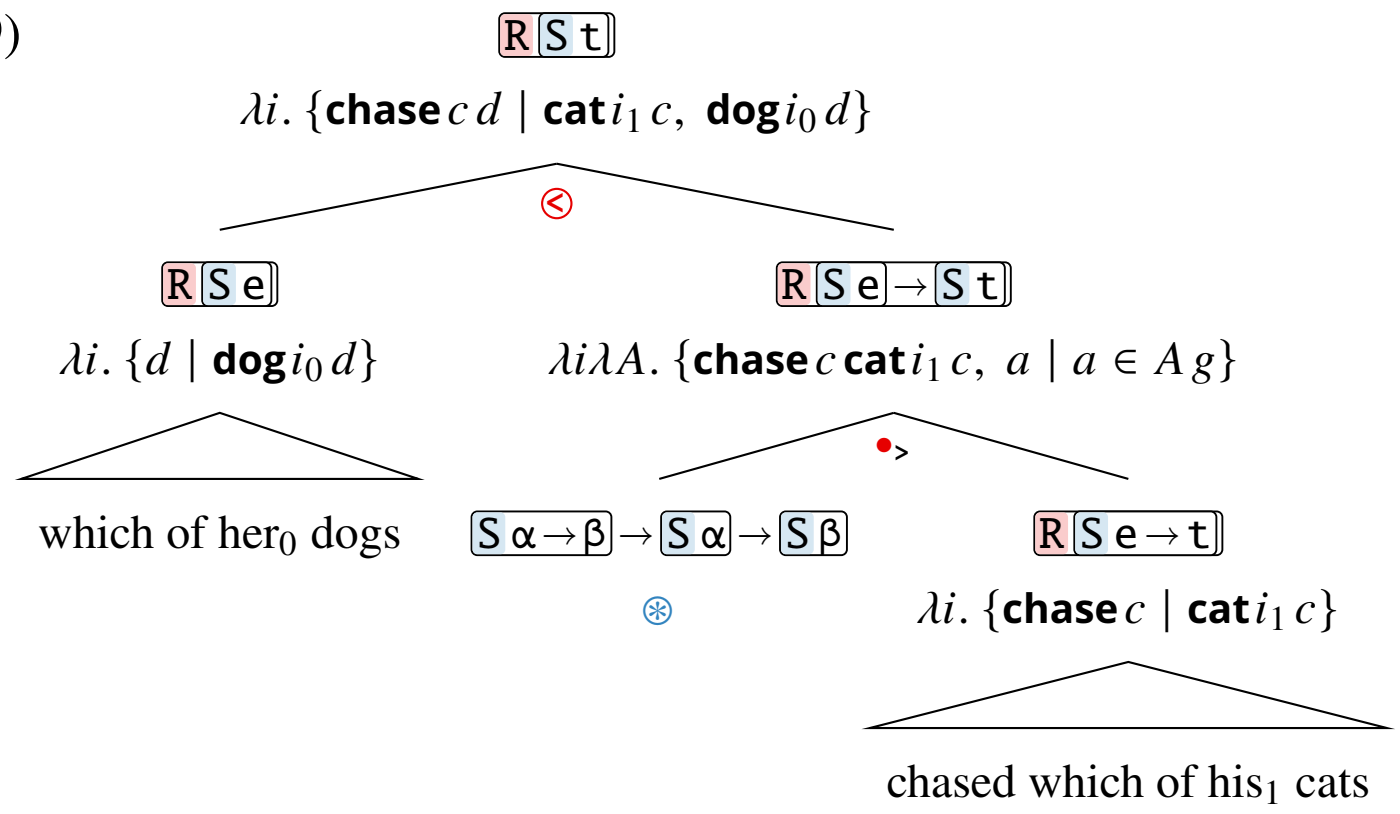

But this approach faces all the same challenges as adding (•) to the object language did in Chapter 2. Namely, it takes a great deal of ingenuity to find derivations when they exist, and even more ingenuity to find out when they don't. And what derivations you do discover are likely to be full of syntactically suspect combinatory operators in the leaves.

### 3.2.2 Structured application as a higher-order mode of combination

So instead, we follow the strategy of Section 2.2.2, provisioning the grammar with a means of (i) applying an arbitrary combinator to the underlying values of two computations, and (ii) merging their effects together in a singly-layered structure. That is, whenever there is a mode $(*)$ that might combine constituents $M :: \sigma$ and $N :: \tau$, then there should also be a mode to combine constituents $M' :: \boxed{\Sigma\,\sigma}$ and $N' :: \boxed{\Sigma\,\tau}$, so long as $\Sigma$ is an applicative functor. Intuitively, there is a $\sigma$ thing sitting inside $M'$, and a $\tau$ thing sitting inside $N'$, both ready to be combined via $(*)$. And since $\Sigma$ is applicative, there is a principled way of zipping up the computational contexts of $M'$ and $N'$. The relevant higher-order combinator is defined in (3.10).

(3.10) $\mathbf{A} :: (\sigma \to \tau \to \omega) \to \boxed{\Sigma\,\sigma} \to \boxed{\Sigma\,\tau} \to \boxed{\Sigma\,\omega}$

$\mathbf{A}\,(*)\,E_1\,E_2 \coloneqq \eta\,(*) \circledast E_1 \circledast E_2$

With these higher-order modes of combination, the earlier single-effect derivations in (3.2) are expressed as in (3.11).

(3.11)

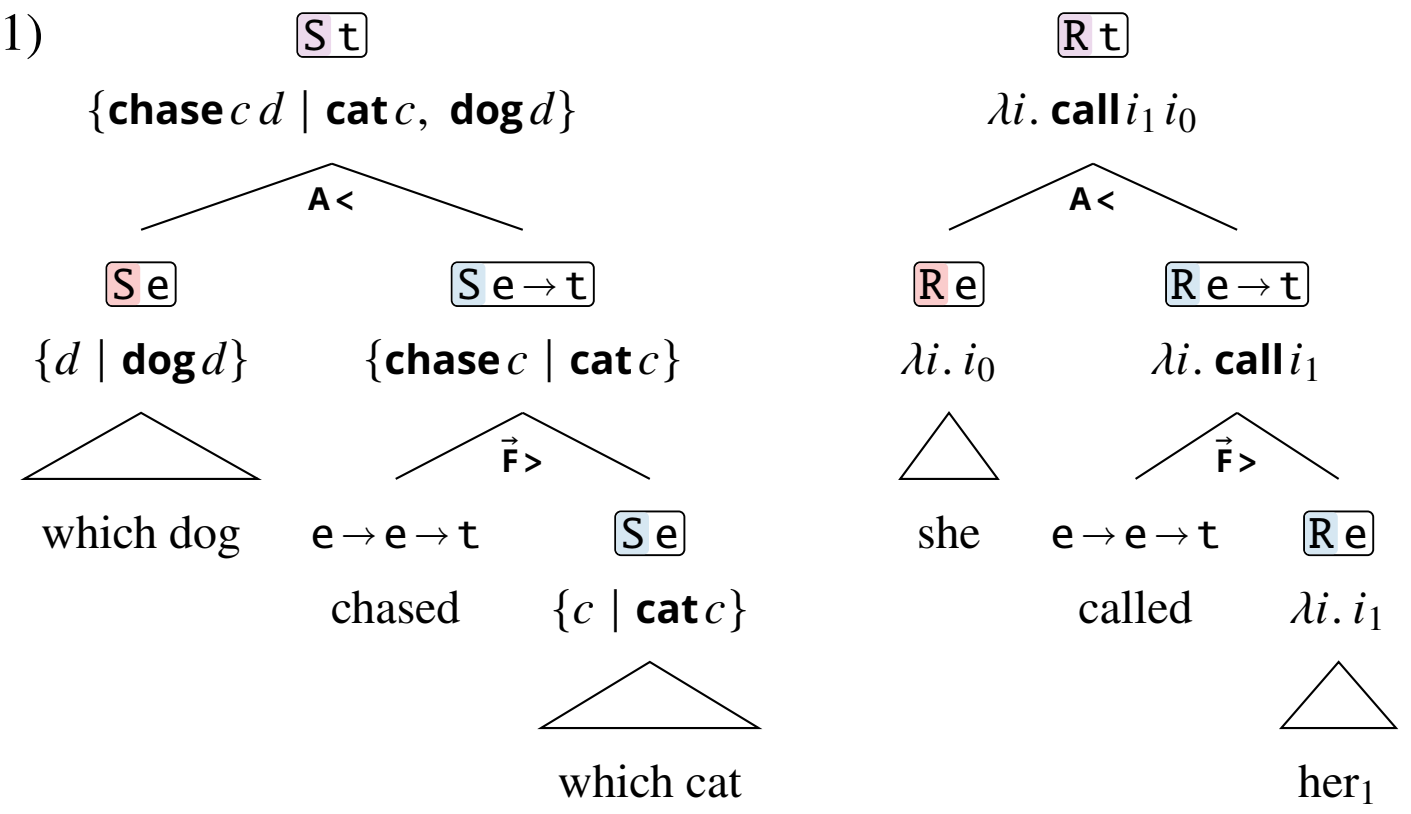

The troubling, composite-effect sentence of (3.8) is now derived in (3.12). Crucially, just as in Section 2.2.2, because the new **A** is higher-order, it can be iterated. That is, (**>**) is a way of putting together two meanings, and therefore (**A >**) is too, and so is **A** (**A >**), etc.

(3.12)

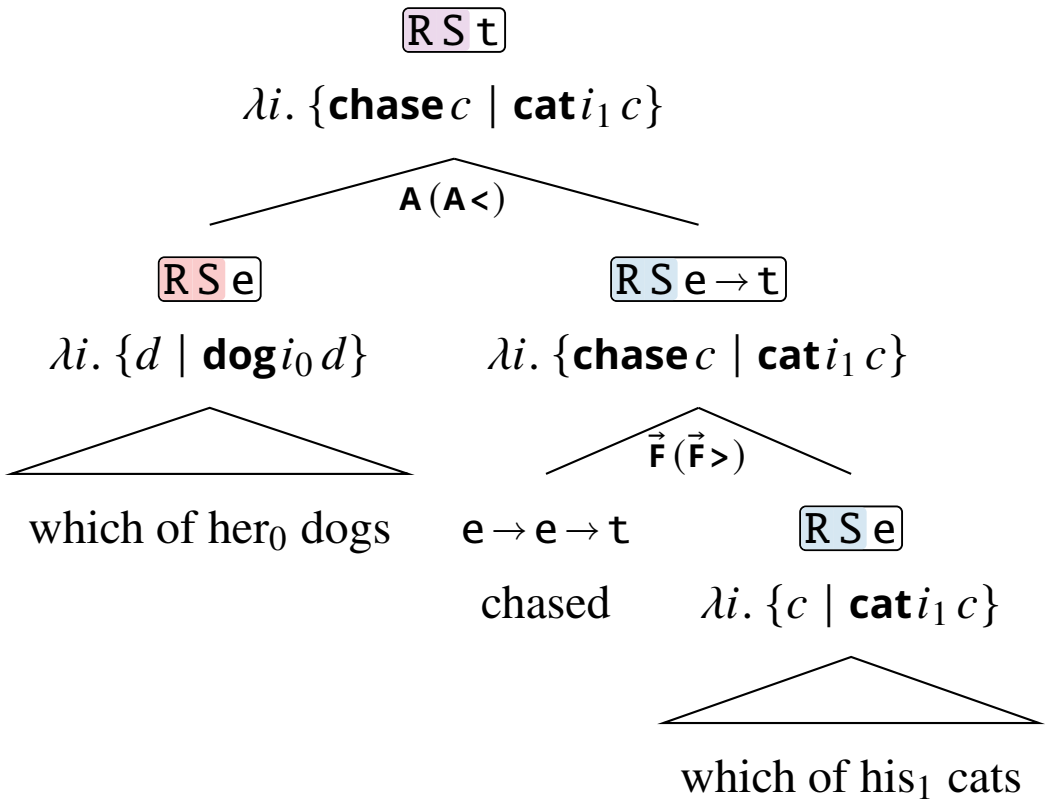

Like functors, applicatives are closed under composition. Given applicative $\Sigma$ and $K$, $\Sigma K$ is applicative, with lawful $\eta$ and ($\circledast$) operations built on those of $\Sigma$ and $K$. As with functors, higher-order applicative combination is an immediate consequence — a compositional theorem, if you like — of our effect-driven interpreter: $\mathcal{F}$ :: $\Sigma K\,\sigma \to \tau$ and $\mathcal{X}$ :: $\Sigma K\,\sigma$ may combine via **A** (**A >**) or **A** (**A <**), depending on whether $\mathcal{F}$ or $\mathcal{X}$ is on the left. Indeed, this is exactly what transpires in (3.12).

The combination of $\vec{\mathsf{F}}/\overleftarrow{\mathsf{F}}$ and **A** provides a lot of flexibility. For instance, the effects needn't be balanced the way they are in (3.12). Examples of computationally imbalanced daughters are shown in (3.13) and (3.14). Outer effects may be merged with **A** while inner ones are mapped over, or vice versa.

(3.13)

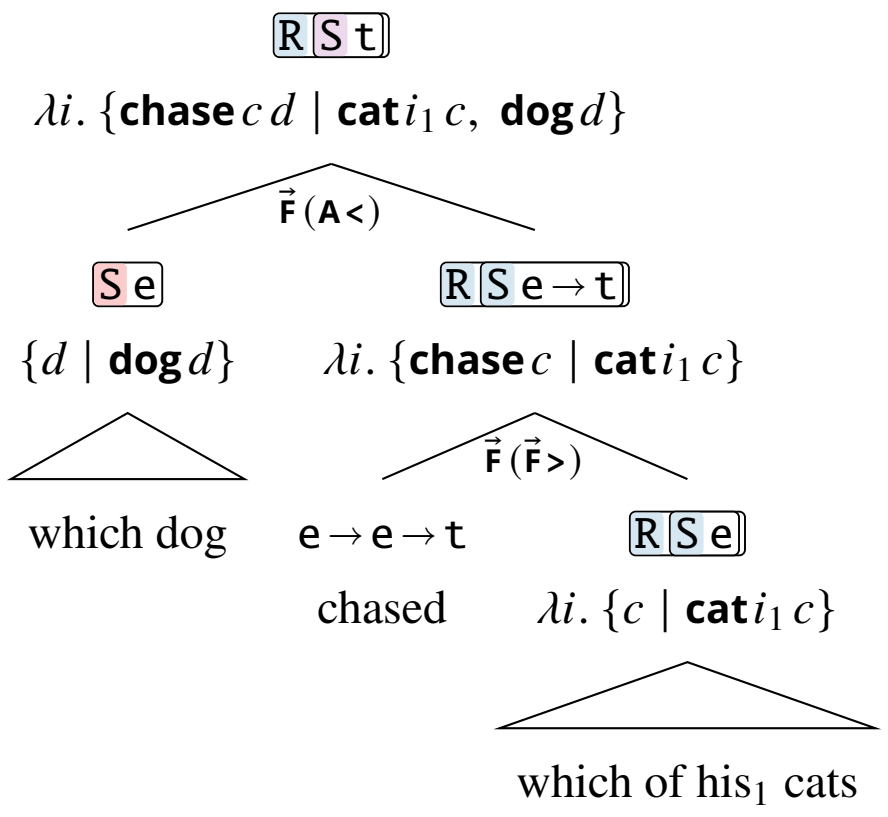

(3.14)

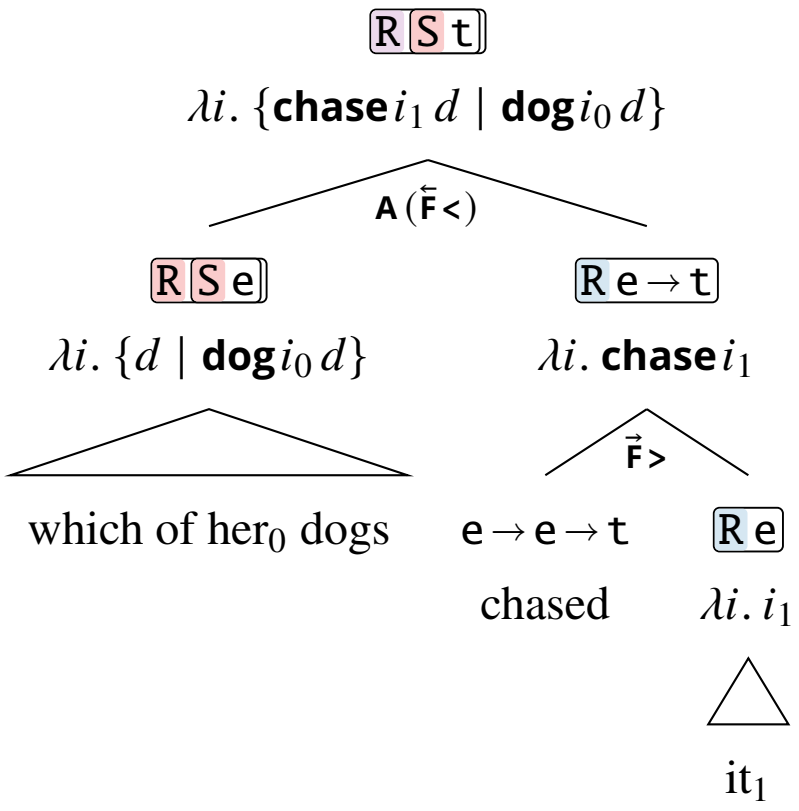

## 3.3 Selectivity and unselectivity

As discussed in Section 2.4, the functoriality of an effect gives rise to a kind of exceptional scope-taking. Barriers to movement obviously exert no direct influence on the scope of an effect, since the generators of effects do not move (or certainly do not need to move for their effect to spread upward). And barriers to quantifier scope, if they are separate from those for movement, likewise hold no particular sway over the percolation of effects since effects are not generally quantificational. In other words, island boundaries will simply get mapped over effectful computations, just like ordinary predicates.

But this is not to say it is impossible to operate on a computation itself, as opposed to the value it computes. Operators like this, which we will refer to broadly as **closure operators**, come in two flavors, distinguished more by their linguistic role than their formal properties. We'll define here a few such operators pulled from various semantic literatures, but keep in mind that the particular denotations and background domain-specific theories do not matter for any of the points we make about composition. Because the semantics is type-driven, any predictions about scope and effect order follow from the types.

First, there are linguistic items thought to **associate with an effect** in an essential capacity. The way we will use the term, an expression associates with an effect $\Sigma$ if it takes an argument of type $\boxed{\Sigma\,\alpha}$, for some type $\alpha$. That is, the expression's semantics *expects* a particular type of computation. The canonical examples in this category are **focus-sensitive items** like 'only' and 'also' (Rooth 1985). For instance, when handed a proposition $q$ together with a set $\mathcal{A}$ of alternatives to $q$, the lexical entry in (3.15a) checks whether $q$ is the only alternative in $\mathcal{A}$ that is true.

We would also include in this category alternative-sensitive expressions like the indeterminate quantifiers 'mo' and 'ka' in Japanese (Shimoyama 2006), free-choice modals (Aloni 2007), and even ordinary question-embedding attitudes like 'wonder' (Groenendijk and Stokhof 1984). Likewise, any dynamic semantic connective or determiner would count as associating with the referent-introducing effects of its arguments (e.g., Muskens 1990, Dekker 1994). The same goes for static variable-binding operators like the abstraction index of, e.g., Heim and Kratzer 1998, about which we will have more to say in Chapter 4 (as we'll discuss then, $\mathtt{V}$ is a synonym for $\mathtt{R}$).

(3.15a) only :: $\boxed{\mathtt{F}\,\mathtt{t}} \to \mathtt{t}$

$[\![\text{only}]\!] := \lambda\langle q, \mathcal{A}\rangle.\ \{p \in \mathcal{A} \mid p\} = \{q\}$

(3.15b) mo :: $\boxed{\mathtt{S}\,\mathtt{t}} \to \mathtt{t}$

$[\![\text{mo}]\!] := \lambda m.\ \bigwedge m$

(3.15c) may :: $\boxed{\mathtt{S}\,\mathtt{t}} \to \mathtt{t}$

$[\![\text{may}]\!] := \lambda m.\ \bigwedge\{\Diamond p \mid p \in m\}$

(3.15d) wonder :: $\boxed{\mathtt{S}\,\mathtt{t}} \to \mathtt{e} \to \mathtt{t}$

$[\![\text{wonder}]\!] := \lambda m \lambda x.\ \textbf{wonder}\, m\, x$

(3.15e) and :: $\boxed{\mathtt{D}\,\mathtt{t}} \to \boxed{\mathtt{D}\,\mathtt{t}} \to \boxed{\mathtt{D}\,\mathtt{t}}$

$[\![\text{and}]\!] := \lambda R \lambda L \lambda i.\ \{\langle p \wedge q, k\rangle \mid \langle p, j\rangle \in L\,i,\ \langle q, k\rangle \in R\,j\}$

(3.15f) $n$ :: $\boxed{\mathtt{V}\,\beta} \to \boxed{\mathtt{V}\,\mathtt{e} \to \beta}$

$[\![n]\!] := \lambda b \lambda g \lambda x.\ b\, g^{n \mapsto x}$

Second, we see a smattering of covert operators usually associated with some sort of complete **evaluation domain**, often a clause. Semantically, these sorts of operators might be thought of as executing the computations that their prejacents denote, and possibly evaluating the results. The most well-known examples are the many varieties of **existential closure** used in alternative- and dynamic-semantic settings (e.g., Heim 1982, Kratzer and Shimoyama 2002), which we'll take an in-depth look at imminently. But also potentially in this category are various **exhaustivity** operators that deny all stronger alternatives (e.g., Krifka's (1995) "ScalAssert"), the lowering operation of continuation semantics that bounds the scope of quantifiers (Barker 2002, Barker and Shan 2014) (to be discussed in Section 5.4), and any mechanism for locally accommodating a presupposition (e.g., Heim 1983, Beaver and Krahmer 2001).

(3.16a) ∃-clo :: $\boxed{\mathtt{S}\,\mathtt{t}} \to \mathtt{t}$

$[\![\exists\text{-clo}]\!] := \lambda m.\ \bigvee m$

(3.16b) ScalAssert :: $\boxed{\mathtt{F}\,\mathtt{t}} \to \mathtt{t}$

$[\![\text{ScalAssert}]\!] := \lambda \langle q, \mathcal{A}\rangle.\ \{p \in \mathcal{A} \mid p \Rightarrow q\} = \{q\}$

(3.16c) $\Downarrow$ :: $\boxed{\mathtt{C}\,\mathtt{t}} \to \mathtt{t}$

$[\![\Downarrow]\!] := \lambda m.\ m\,(\lambda p.\ p)$

(3.16d) Accom :: $\boxed{\mathtt{M}\,\mathtt{t}} \to \mathtt{t}$

$[\![\text{Accom}]\!] := \lambda m.$ **false** if $m = \#$ else $m$

When a constructor $\Sigma$ is applicative, any operator $\lightning$ :: $\boxed{\Sigma\,\sigma} \to \tau$ will in principle close over every effect in its scope simultaneously. Operators like this are sometimes said to be **unselective**, on analogy with the unselective binding of indefinites found in Lewis 1975 and Heim 1982. For instance, consider the conditional in (3.17). Thanks to the applicativity of $\mathtt{S}$, a single existential closure operator manages to capture the scope of all the indefinites in the antecedent (above ∃-clo, we suppress the restrictions on $\exists x, u, y$ to save space).

(3.17)

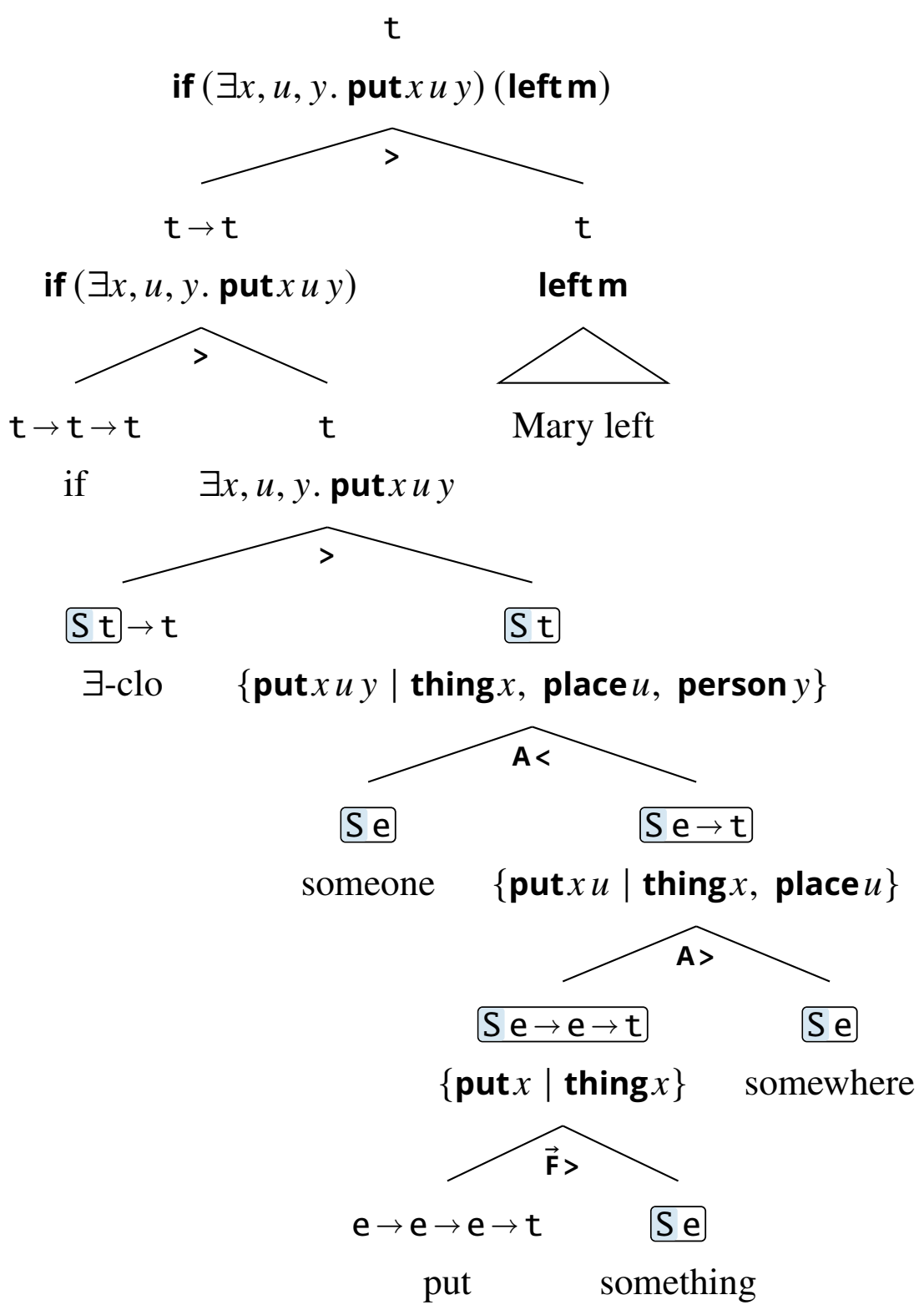

At the same time, every applicative functor is still a functor, which suffices to establish higher-order derivations of the antecedent. Unless something precludes such derivations, this predicts the availability of various exceptional-scope readings, in which the existential closure operator associates **selectively** with a subset of the indefinites in its scope. For instance, (3.18) mimics the derivation in (3.17) except that the alternatives generated by the direct object are consistently mapped-over at every node. The alternatives generated by the other two indefinites are merged as above and jointly closed over in the antecedent.

(3.18)

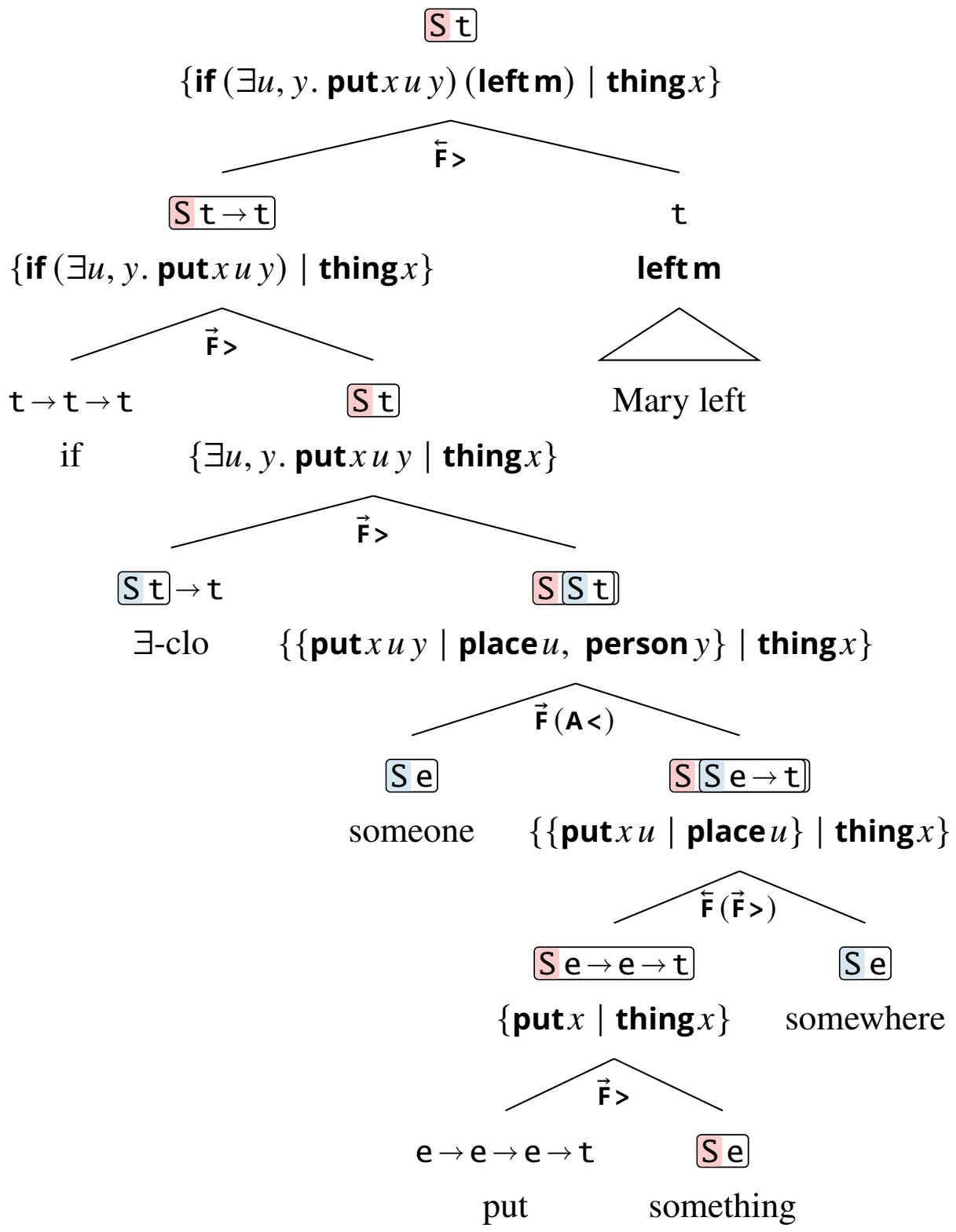

Even a single effect may in principle escape the scope of a closure operator, with the help of $\eta$. For demonstrative purposes, imagine the closure operator of (3.18) is obligatory, or perhaps even part of the semantics of the conditional.[6] Then the only way to combine a conditional with an ordinary, effect-free

[6]Conditionals are indeed occasionally thought to associate with static alternatives of the S variety (e.g., Alonso-Ovalle 2009), as in (3.19), and almost always thought to associate with dynamic alternatives of the D variety (e.g., Muskens 1996), as in (3.20).

(3.19) if :: $\boxed{\mathtt{S}\ \mathtt{t}} \to \boxed{\mathtt{S}\ \mathtt{t}} \to \mathtt{t}$

$$[\![\text{if}]\!] \coloneqq \lambda m \lambda n.\ \bigwedge\{p \Rightarrow \bigvee n \mid p \in m\}$$

(3.20) if :: $\boxed{\mathtt{D}\ \mathtt{t}} \to \boxed{\mathtt{D}\ \mathtt{t}} \to \boxed{\mathtt{D}\ \mathtt{t}}$

$$[\![\text{if}]\!] \coloneqq \lambda m \lambda n \lambda i.\ \{\langle \forall j.\ \langle \textbf{true}, j\rangle \in m\,i \Rightarrow \exists k.\ \langle \textbf{true}, k\rangle \in n\,j,\ i\rangle\}$$

antecedent would be to coerce the antecedent into a trivially indeterminate computation, one whose only thread is the ordinary proposition.

(3.21)

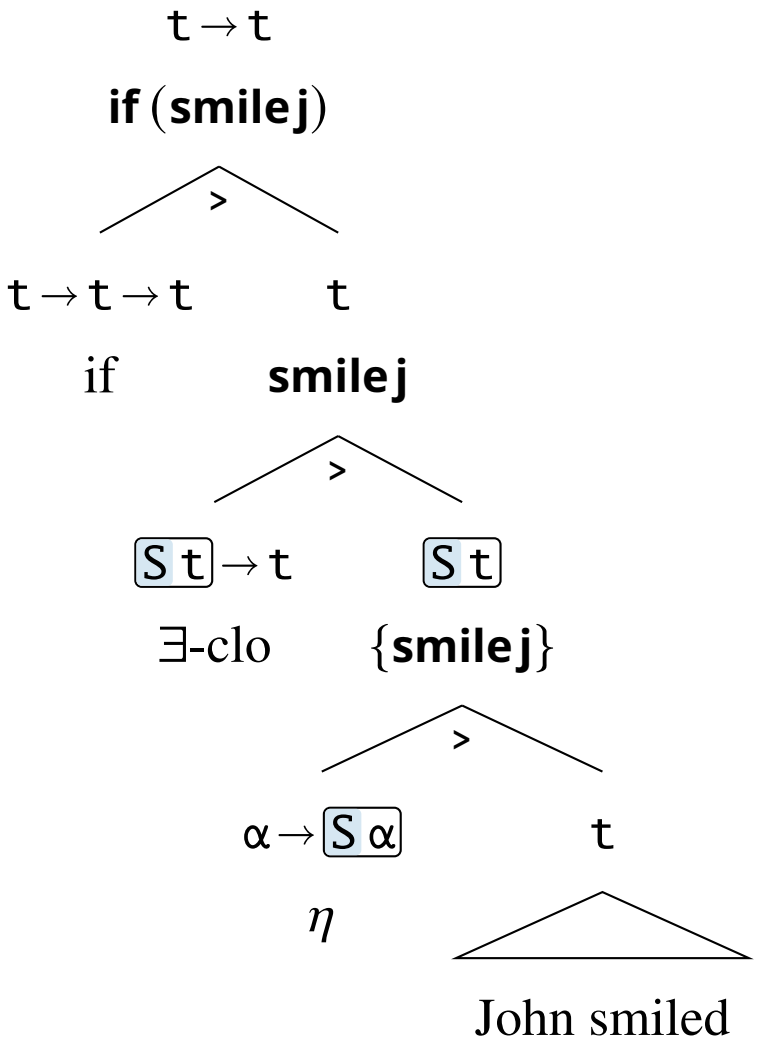

If the antecedent has an effect, but not an effect that ∃-clo knows what to do with, then the $\eta$-coercion must apply to the underlying, pure value of the antecedent.

(3.22)

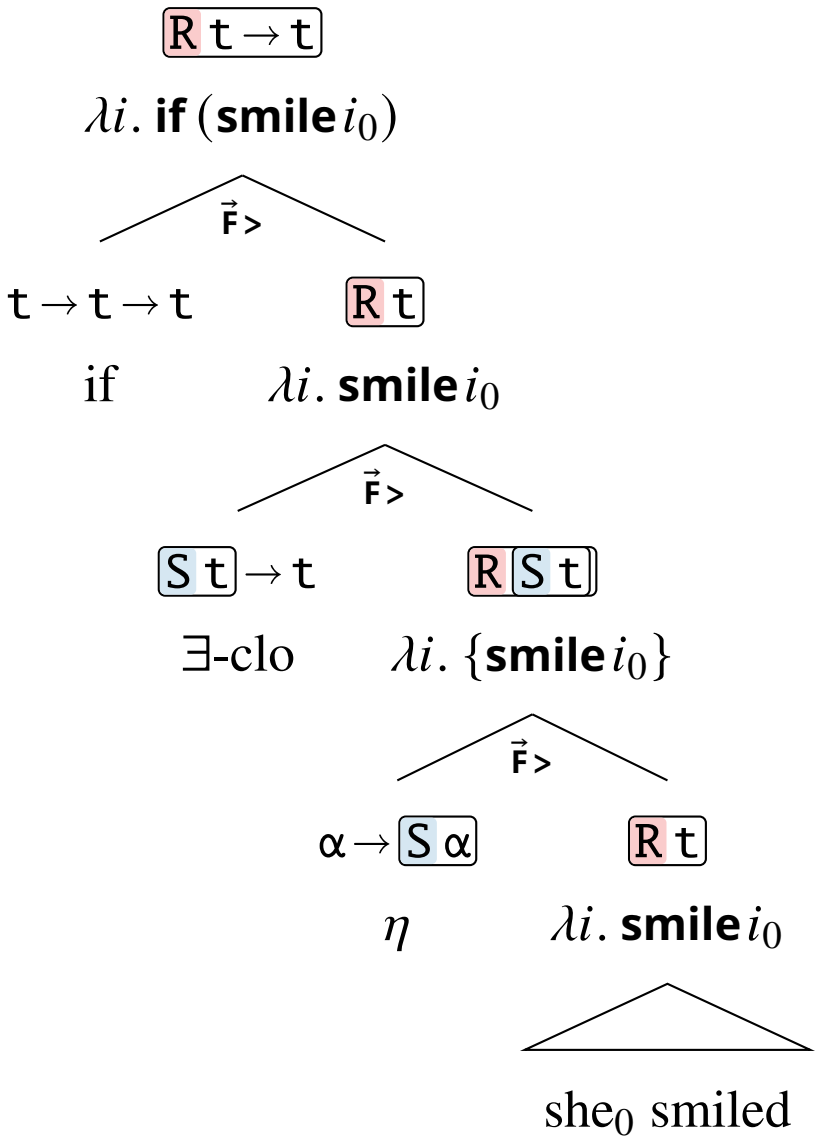

This same strategy, of applying $\eta$ to *the result* of a computation — corresponding to its underlying type — might just as well be used to coerce the propositions underlying an S-node, as in (3.23). In so doing, the existential force of the indefinite, carried by the alternatives it generates, escapes the closure operator and outscopes the conditional in which it appears.

(3.23)

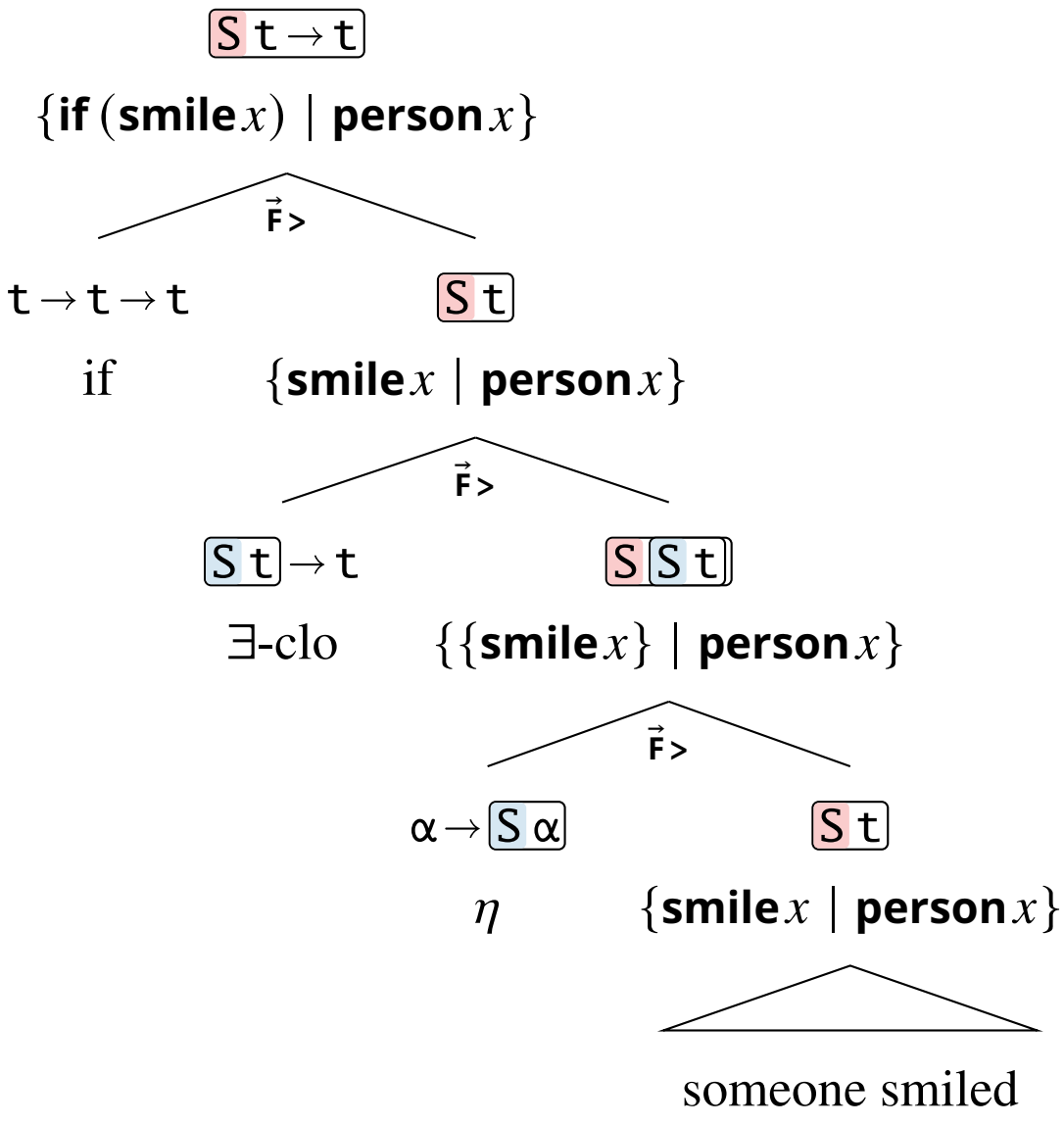

In this manner, cleverly allocated $\eta$ operators can play two apparently contrary roles in the grammar. They can inject pure values into computations that *feed* closure operators, creating the denotational structure that the operators expect. But in exactly the same way, they can render the underlying values of already-effectful denotations as dummy computations, providing decoy targets for closure operators, and in the process *shield* effects from the operators that threaten to consume them.

However, as in other places in this Element, we should like very much to eliminate cleverness from the picture. Once a theorist has fixed a grammar and a set of lexical items, it should not require an act of inspiration to figure out the range of meanings an expression may take. This motivated the re-framing of (•) and (⊛) in terms of $\vec{\mathbf{F}}/\overleftarrow{\mathbf{F}}$ and **A** above.

It is less obvious how to recast occurrences of $\eta$ in these terms because $\eta$ *creates* an effect, where $(\bullet)$ and $(\circledast)$ *handle* them. That is, because the latter combinators take computations as arguments, they are only applicable when at least one of the daughters denotes a computation. But $\eta$ is unary and completely parametric in its input, which means that it can in principle apply to any constituent of any type at any time, iteratively. However, $\eta$ifying a constituent is almost always pointless. It just injects a bit of harmless, lingering computational cruft. In fact, we suggest that in the presence of $\vec{\mathbf{F}}/\overleftarrow{\mathbf{F}}$, the only semantic use for $\eta$ is to feed a closure operator of some sort or another, as in the derivations of (3.18)–(3.23). We thus propose the higher-order modes of combination in (3.24), which differ only in whether the expectant operator is the left daughter (3.24a) or the right daughter (3.24b).

(3.24a) $\vec{\mathbf{U}} :: ((\sigma \to \sigma') \to \tau \to \omega) \to (\boxed{\Sigma\,\sigma} \to \sigma') \to \tau \to \omega$

$\vec{\mathbf{U}}\,(*)\,E_1\,E_2 := (\lambda a.\, E_1\,(\eta\, a)) * E_2$

(3.24b) $\overleftarrow{\mathbf{U}} :: (\tau \to (\sigma \to \sigma') \to \omega) \to \tau \to (\boxed{\Sigma\,\sigma} \to \sigma') \to \omega$

$\overleftarrow{\mathbf{U}}\,(*)\,E_1\,E_2 := E_1 * (\lambda b.\, E_2\,(\eta\, b))$

These rules say that a closure operator of type $\boxed{\Sigma\,\sigma} \to \sigma'$ may be combined with a prejacent of type $\tau$ whenever the function type $\sigma \to \sigma'$ could be combined with $\tau$. Here's how: convert the closure operator into an ordinary function of type $\sigma \to \sigma'$ by composing it with $\eta$. That is, create the function that will take in an input of type $\sigma$, inject it into $\Sigma$ with $\eta$, and then pass that newly created computation of type $\boxed{\Sigma\,\sigma}$ into the closure operator. Then combine this function of type $\sigma \to \sigma'$ with the other daughter of type $\tau$ in whatever way(s) make sense.

Derivations of (3.21) and (3.23) using this new meta-combinator are shown in (3.25). The complete applicative and functor rules are summarized in Figure 7.

(3.25)

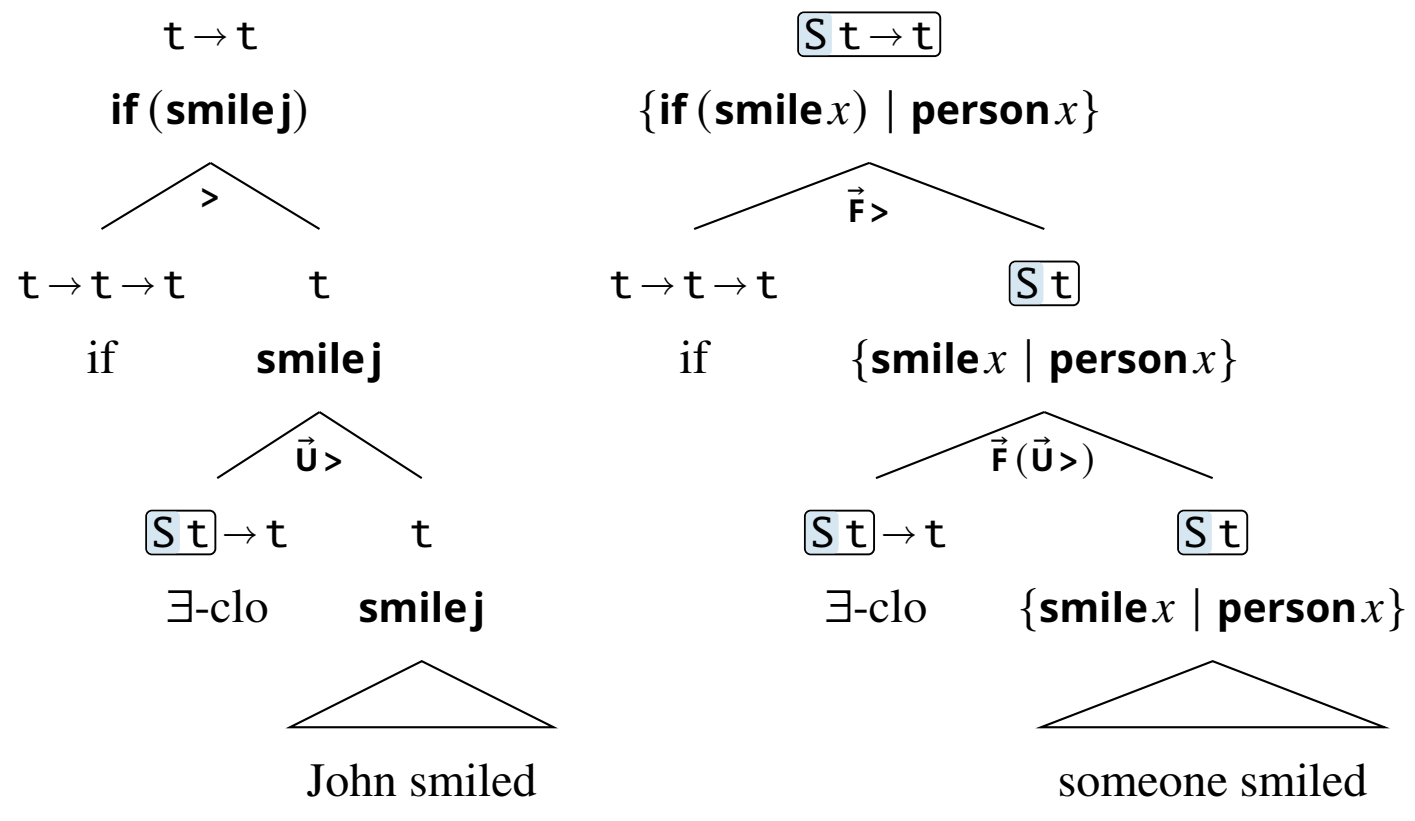

## 3.4 Applicative typology

The grammar of Figure 7 is in some sense maximally expressive relative to the applicative structure of an effect. It permits every possible agglomeration or stratification of structure, which in turn means that closure operators may capture anywhere from none to all of the effects in their prejacents.

It is perhaps worth taking stock of how the different ingredients of Figure 7 regulate this expressivity. One way to carry out this thought experiment is to investigate the capturing profile of closure operators under various ablations of the meta-combinator inventory.

For instance, with a purely functorial grammar that includes only $(\bullet)$, we predict that any operator $\lightning :: \boxed{\Sigma\,\sigma} \to \tau$ will necessarily associate with exactly one effect. The reason is that while $(\bullet)$ suffices to generate higher-order effects, the operator $\lightning$ only knows how to process a single layer. All the others will have to be mapped over it. And if there aren't any effects, then $\lightning$ is out of luck because with only $(\bullet)$, there is no way to create a computation where one did not exist before.

A grammar with only $(\circledast)$ and no mapping or unit combinators is essentially the Heim and Kratzer and Hamblin grammars of Figures 5 and 6. For starters, in order for composition to be possible at all, all lexical items will need to be coerced into the computation type $\Sigma$ in the lexicon. That done, the grammar will predict that all closure operators $\lightning$ necessarily associate with at least one effect, and also necessarily capture all of their prejacent's effects. Without $(\bullet)$, there is no way to scope an effect over $\lightning$, and without $\eta$, there is no way to fake an effect where one is not expressed.

Excluding only $\eta$ yields a grammar where all closure operators associate with at least one effect, but do so selectively ($(\bullet)$ allows all but one effect to optionally pass over the closure). On the flipside, excluding only $(\circledast)$ means all closure operators associate with at most one effect ($\eta$ can create computations at will, simulating effects where none are expressed, but no $(\circledast)$ means that each effect creates a distinct computational layer, of which $\lightning$ can target just one). Finally, excluding only $(\bullet)$ does not reduce the expressivity of the full grammar in Figure 7, provided that free insertions of $\eta$ are permitted. This is thanks to the equivalence in (3.6), which ensures the mapping operations can always be simulated by a composition of $\eta$ and $(\circledast)$.

All this is to say, just because a type constructor is mathematically applicative doesn't mean that the particular instance(s) of its associated $(\circledast)$ and $\eta$

**Types:**

$$\begin{aligned} \tau ::= {} & \mathtt{e} \mid \mathtt{t} \mid \cdots && \text{Base types} \\ \mid {} & \tau \to \tau && \text{Function types} \\ \mid {} & \ldots && \ldots \\ \mid {} & \boxed{\Sigma\, \tau} && \text{Computation types} \end{aligned}$$

**Effects:**

$$\begin{aligned} \Sigma ::= {} & \mathtt{R} && \text{Input} \\ \mid {} & \mathtt{W} && \text{Output} \\ \mid {} & \mathtt{S} && \text{Indeterminacy} \\ \mid {} & \ldots && \ldots \end{aligned}$$

**Basic Combinators:**

$$(>) :: (\alpha \to \beta) \to \alpha \to \beta \qquad \text{Forward Application}$$
$$f > x := f\, x$$

$$(<) :: \alpha \to (\alpha \to \beta) \to \beta \qquad \text{Backward Application}$$
$$x < f := f\, x$$

$$\ldots \qquad \ldots$$

**Meta-combinators:**

$$\overleftarrow{\mathsf{F}} :: (\sigma \to \tau \to \omega) \to \boxed{\Sigma\, \sigma} \to \tau \to \boxed{\Sigma\, \omega} \qquad \text{Map Left}$$
$$\overleftarrow{\mathsf{F}}\,(*)\, E_1\, E_2 := (\lambda a.\, a * E_2) \bullet E_1$$

$$\overrightarrow{\mathsf{F}} :: (\sigma \to \tau \to \omega) \to \sigma \to \boxed{\Sigma\, \tau} \to \boxed{\Sigma\, \omega} \qquad \text{Map Right}$$
$$\overrightarrow{\mathsf{F}}\,(*)\, E_1\, E_2 := (\lambda b.\, E_1 * b) \bullet E_2$$

$$\mathsf{A} :: (\sigma \to \tau \to \omega) \to \boxed{\Sigma\, \sigma} \to \boxed{\Sigma\, \tau} \to \boxed{\Sigma\, \omega} \qquad \text{Structured App}$$
$$\mathsf{A}\,(*)\, E_1\, E_2 := (\lambda a \lambda b.\, a * b) \bullet E_1 \circledast E_2$$

$$\overleftarrow{\mathsf{U}} :: (\sigma \to (\tau \to \tau') \to \omega) \to \sigma \to (\boxed{\Sigma\, \tau} \to \tau') \to \omega \qquad \text{Unit Left}$$
$$\overleftarrow{\mathsf{U}}\,(*)\, E_1\, E_2 := E_1 * (\lambda b.\, E_2\, (\eta\, b))$$

$$\overrightarrow{\mathsf{U}} :: ((\sigma \to \sigma') \to \tau \to \omega) \to (\boxed{\Sigma\, \sigma} \to \sigma') \to \tau \to \omega \qquad \text{Unit Right}$$
$$\overrightarrow{\mathsf{U}}\,(*)\, E_1\, E_2 := (\lambda a.\, E_1\, (\eta\, a)) * E_2$$

**Figure 7** A type-driven grammar with functors and applicatives

combinators need be included in a grammar or fragment. It is highly likely that different natural language operators evince different empirical capturing profiles (cf. S. Beck 2006 on intervention effects), so it is useful to know what happens when these algebraic toggles are set in various combinations.

## 3.5 Commutative and non-commutative effects

In this chapter we've concentrated on R and S effects as canonical examples of applicativity in linguistics. These two effects have in common that they are **commutative**: when using **A** to combine $F :: \boxed{\Sigma\, \sigma \to \tau}$ and $X :: \boxed{\Sigma\, \sigma}$, it doesn't matter which daughter is on the left and which is on the right. For any $F$ and $X$ with these types, we have the following equivalence.

(3.26) $\mathbf{A}{>}\, F\, X = \mathbf{A}{<}\, X\, F$

This is because the ($\circledast$) instance for R simply passes an incoming environment down to both computations $F$ and $X$ before applying the one to the other. And the ($\circledast$) instance for S takes the full cross-product of its daughters, selecting every element from the one and every element from the other independently.

But not all effects are like this. As a preview of Chapters 4 and 5, we draw attention to two such non-commutative effects. The first, T, is defined in (3.27).

(3.27)
$$\begin{aligned} \boxed{\mathtt{T}\,\alpha} &::= \mathtt{s} \to (\alpha \times \mathtt{s}) \\ \eta x &:= \lambda i.\, \langle x, i \rangle \\ F \circledast X &:= \lambda i.\, \langle f\, x, k \rangle, \textbf{ where } \langle f, j \rangle = F\, i \\ &\qquad\qquad\qquad\qquad\quad\; \langle x, k \rangle = X\, j \end{aligned}$$

The constructor T models a transition from one context s to another, computing an $\alpha$ along the way. This is a simplified version of the sort of state-updating denotations often seen in dynamic semantics. For instance, if we identify the context type s with a list of mentioned entities (Vermeulen 1993, van Eijck 2001), then we might imagine that the denotations of names add referents to the outgoing context, while the denotations of pronouns read antecedents from the incoming context.[7] Sequencing an expression containing a name and an expression containing a pronoun, *in that order*, results in a kind of dynamic binding (no scope or c-command required), as seen in (3.28).

[7]Note that this encoding conceives of referent-storage and referent-retrieval as two different operations that perform the same *kind* of computation (i.e., doing stuff with the discourse context). Thus the pronoun and the name have the same type, but with dual denotations. This is in contrast to the treatment we've drawn on extensively above, utilizing a more finely-grained R constructor reserved specifically for computations that *read* from a context (whether they write to it or not). We will have more to say about the choice between these in Chapters 4 and 5, after we've introduced dynamic semantics.

(3.28)

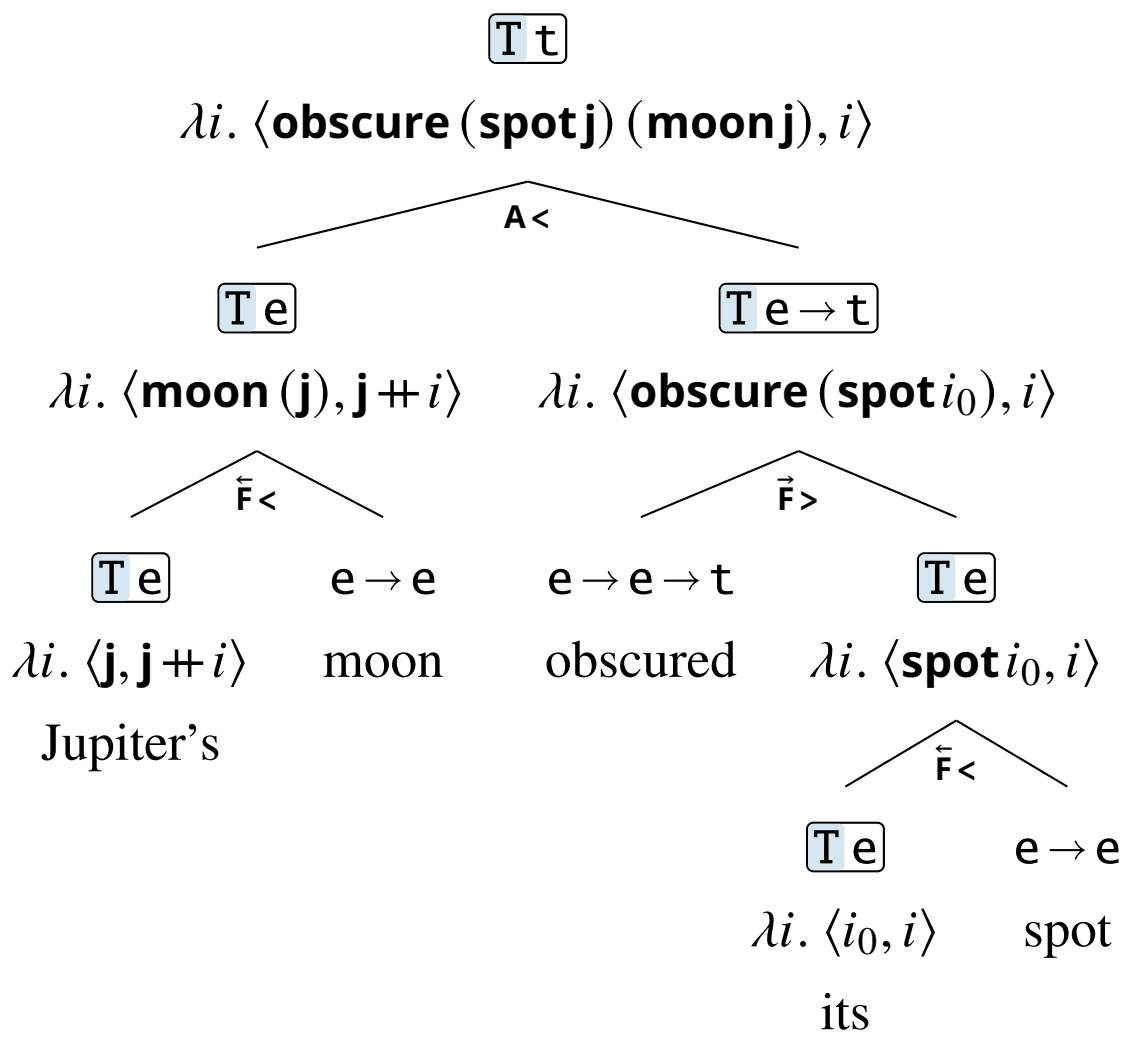

Importantly, reversing the order of the daughters would reverse which computation passes its output to the other. Compare $\mathsf{A>}\,F\,X$ and $\mathsf{A<}\,X\,F$ below:

(3.29a) $$\mathsf{A>}\,F\,X \coloneqq \lambda i.\,\langle f\,x, k\rangle, \;\textbf{where}\; \langle f, j\rangle = F\,i \qquad \langle x, k\rangle = X\,j$$

(3.29b) $$\mathsf{A<}\,X\,F \coloneqq \lambda i.\,\langle f\,x, k\rangle, \;\textbf{where}\; \langle x, j\rangle = X\,i \qquad \langle f, k\rangle = F\,j$$

In the first case, $F$ is evaluated at the input state $i$, and its output $j$ is passed in as input to $X$. In the second case, $X$ is evaluated at the input $i$, and its output $j$ is passed in as input to $F$. The only way for $F$ to influence the state that $X$ is evaluated against is for $F$ to come first (and vice versa).

The second non-commutative effect, defined by $\mathtt{C}$ in (3.30), models Generalized Quantifiers over the domain of α.

(3.30) $$\boxed{\mathtt{C}\,\alpha} ::= (\alpha \to \mathtt{t}) \to \mathtt{t}$$
$$\eta x \coloneqq \lambda c.\, c\,x$$
$$F \circledast X \coloneqq \lambda c.\, F\,(\lambda f.\, X\,(\lambda x.\, c\,(f\,x)))$$

Here, “applying” one quantifier $F :: \boxed{\mathtt{C}\ \sigma \to \tau}$ to another $X :: \boxed{\mathtt{C}\ \sigma}$ means passing the latter in as the part of the scope of the former (cf. Barker and Shan 2014; Shan and Barker 2006). Analogously to $\mathtt{T}$, our compositional system systematically uses **A** to grant the left daughter scope over the right, as seen in (3.31).

(3.31a) $\mathbf{A}{>}\, F\, X \coloneqq \lambda c.\, F\, (\lambda f.\, X\, (\lambda x.\, c\, (f\, x)))$

(3.31b) $\mathbf{A}{<}\, X\, F \coloneqq \lambda c.\, X\, (\lambda x.\, F\, (\lambda f.\, c\, (f\, x)))$

For $\mathtt{C}$ this left-over-right compositional default corresponds, naturally enough, to *surface scope* in sentences with multiple sources of quantification:

(3.32)

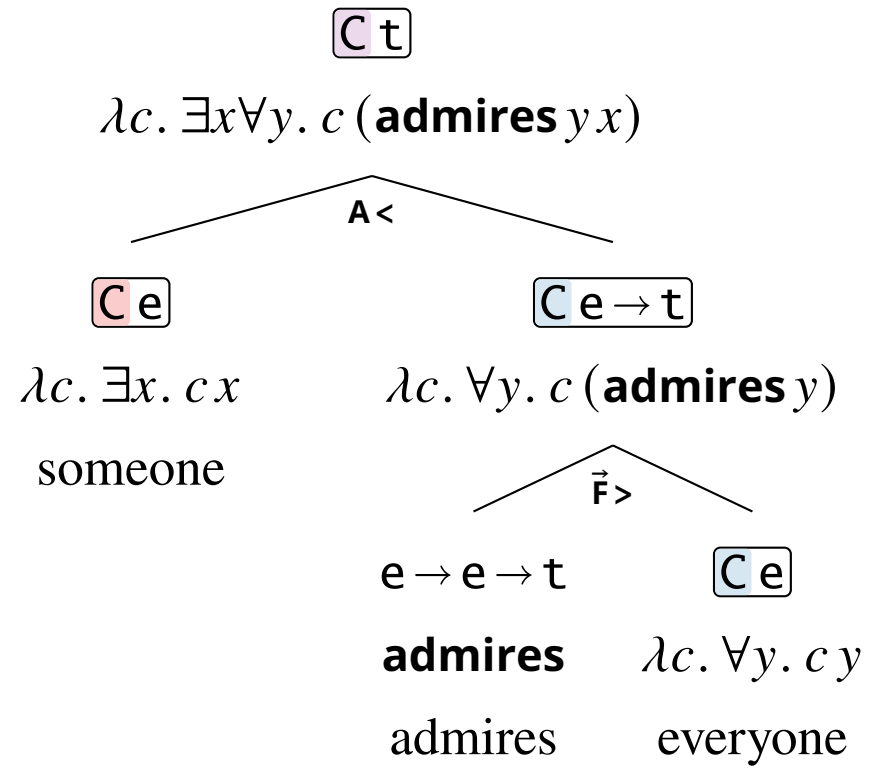

While derivations like this are nothing if not routine at this point, it is worth recalling that such constructions have played an outsize role in the history of semantics, and linguistics. The familiar type mismatch between the type $\mathtt{e} \to \mathtt{e} \to \mathtt{t}$ verb and type $\boxed{\mathtt{C}\ \mathtt{e}}$ object has been taken to motivate Quantifier Raising, LF, and the Y-model of syntax (Heim and Kratzer 1998; May 1985). In the effect-oriented interpreter, such innovations are unnecessary: the applicative and functorial nature of the $\mathtt{C}$ effect drive interpretation, and the linear bias of **A** corresponds to a grammatical default for surface scope. (We will see how inverse scope arises in Section 4.3.1.)

Some linguistic theories see the non-commutativity of $\mathtt{T}$- and $\mathtt{C}$-like computations as a fundamental starting point in explaining empirical patterns of scope and binding. Pronouns nearly always follow their antecedents, and “dynamic semantics” is to some extent just a cover term for theories in which discourse contexts are updated from left to right, largely in reaction to this bias in anaphora (Groenendijk and Stokhof 1988, Rothschild and Yalcin 2016). Similarly, people’s proclivity to interpret quantifiers with scopes that respect their surface order is well-attested, or at least often appealed to in the literature (Kroch 1974, Fodor 1982, Reinhart 1983, Shan and Barker 2006).

Of course there are huge classes of examples that run counter to these defaults. Fortunately, as we will see in the Chapter 4, modeling scope and binding phenomena with asymmetric modes of combination like $\mathbf{A}_{\text{C}}$ and $\mathbf{A}_{\text{T}}$ does not preclude grammatical mechanisms that invert this effect order. It just makes such additional mechanisms necessary, often resulting in more complex derivations. In this sense, the non-commutativity is felt to encapsulate the underlying or fundamental linearity of the processes, which emerges, all else equal, in interpretations of least resistance.

Alternatively, for any of these non-commutative effects where $\mathbf{A}{>}\, F\, X \neq \mathbf{A}{<}\, X\, F$, the combinator on the right-hand side of the inequality yields a second applicative functor for the same type.

(3.33) $$F \circledast'_{\Sigma} X := \mathbf{A}{<}\, X\, F$$
$$= (\cdot)^{\text{LIFT}} \bullet X \circledast_{\Sigma} F$$

This reversed applicative intentionally inverts the function-argument relationship of its daughters before calling ($\circledast$). In this way, the arguments are evaluated before the functions that take them. Correspondingly, the $\mathbf{A}$ meta-combinator that arises from this applicative will pass discourse referents from right to left and scope later constituents over earlier ones by default. This sort of pervasive anti-chronology presumably holds little interest on its own as far as natural languages go, in contrast to the left-to-right applicatives above. But researchers interested in preserving symmetry may opt to include *both* $\circledast_{\Sigma}$ and its twin $\circledast'_{\Sigma}$ in the combinatory inventory.

For example, working in a first-order applicative system analogous to the pointwise grammars of Figures 5 and 6, Barker (2002) proposed to replace (>) with both $\textcircled{>}_{\text{C}}$ and $\textcircled{>}'_{\text{C}}$ (and to replace (<) with $\textcircled{<}_{\text{C}}$ and $\textcircled{<}'_{\text{C}}$). Since $\textcircled{>}_{\text{C}}$ and $\textcircled{>}'_{\text{C}}$ have exactly the same type, but sequence their daughters in opposite orders, this has the effect of introducing systematically scope ambiguity at all ordinary branching nodes. For scope-flexible languages like English, this may hold some appeal. Though if there do turn out to be grammatically-correlated preferences for one reading or another, the explanation would have to come from somewhere other than the semantics, since $\circledast$ and $\circledast'$ are mathematically symmetric.

## 3.6 Implementing applicative effects in the type-driven interpreter

Adding applicativity to our interpreter will follow exactly the same format as adding functoriality did in Chapter 2. We start again by adding the structure-preserving application **A** and unit rules $\vec{\mathbf{U}}/\overleftarrow{\mathbf{U}}$. Again because these are meta-combinators, they are parameterized to the modes that would combine the relevant underlying types.

```
data Mode
  = FA | BA | PM        -- etc
  | MR Mode | ML Mode   -- map right and map left
  | AP Mode             -- structured application
  | UR Mode | UL Mode   -- unit right and unit left
```

For the type logic, we need a new predicate to characterize which effects have applicative instances. All of the effects presented in this Element except for `W` are applicative (see Appendix A4), so this predicate is almost as trivial as `functor`.

`W`-tagged computations are parameterized by the type of data that stored in the second dimension. In order to merge two `W` computations, there must be a way to fuse the secondary contents carried by the respective computations, and in order to lift an arbitrary value of type α into a computation of type `W` α, there must be some sort of trivial supplement that does not add any real information. And of course these supplemental operations and values should not interfere with the applicative laws. It turns out, this is guaranteed when the relevant notion of fusion is associative and has an identity element. Operations with this algebraic structure are known as **monoids**. So in the `applicative` clause for `W`, we add a predicate to check that the supplemental parameter type is a monoid. For this toy system, the only monoidal type is the type of truth values, where fusion is conjunction with truth as its identity element. In a more elaborate setup, the supplemental content might consist of a list of the individuals mentioned in the sentence, modeling a kind of discourse memory. Unremarkable values would be lifted into computations with memory by pairing them with the empty list, and two such lists might be fused by concatenation.

```
functor, applicative :: EffX -> Bool
functor     _      = True
applicative (WX s) = monoid s
applicative f      = functor f && True

monoid :: Ty -> Bool
monoid T = True
monoid _ = False
```

Extending the combination function `combine` follows the same recursive logic as in Chapter 2. When combining a left and right daughter, we start by including any of the earlier basic and functorial operations, and then we look to add applicative ones if possible. For instance, if both daughters are computation types `Comp f s` and `Comp g t` with the same applicative effect (`f == g, applicative f`), then try combining the underlying types `s` and `t`. For every way `(op, u)` that those underlying types can be combined, build a new composite mode of combination `AP op` with combined type `Comp f u`. The unit rules are similar, trying to combine underlying types, and building on the results.

```
combine :: Ty -> Ty -> [(Mode, Ty)]
combine l r =
  modes l r
  ++ addMR l r ++ addML l r
  -- if both daughters are applicative, try structured application
  ++ addAP l r
  -- if the left daughter closes an applicative effect,
  -- try to purify the right daughter
  ++ addUR l r
  -- if the right daughter closes an applicative effect,
  -- try to purify the left daughter
  ++ addUL l r

addAP l r = case (l, r) of
  (Comp f s, Comp g t) | f == g, applicative f
    -> [ (AP op, Comp f u) | (op, u) <- combine s t ]
  _ -> [                                            ]

addUR l r = case l of
  Comp f s :-> s' | applicative f
    -> [ (UR op, u) | (op, u) <- combine (s :-> s') r ]
  _ -> [                                              ]

addUL l r = case r of
  Comp f t :-> t' | applicative f
    -> [ (UL op, u) | (op, u) <- combine l (t :-> t') ]
  _ -> [                                              ]
```

As in the previous chapter, the extended functionality provided by the revised `combine` can be appreciated by firing up `ghci`. First, in addition to the higher-order functorial combinations of `S e` and `S e→t`, `combine` discovers a third, applicative combination using `AP` and resulting in `S t`. In the second example,

`S t` → `t` and `S t` combine to yield either `t` (in which the prejacent's effects are captured by the closure operator) or `S t` (in which the prejacent's effects escape closure; the two layerings of `MR` and `UR` turn out to be equivalent).

```
ghci> combine (Comp SX E) (Comp SX (E :-> T))
[(MR (ML BA), Comp SX (Comp SX T)),
 (ML (MR BA), Comp SX (Comp SX T)),
 (AP BA, Comp SX T)]
```

```
ghci> combine ((Comp SX T) :-> T) (Comp SX T)
[(FA, T),
 (MR (UR FA), Comp SX T),
 (UR (MR FA), Comp SX T)]
```

# 4 Monads

## 4.1 Motivating monads: effects inside effects

Up to this point we have ignored the internal compositional details of various noun phrases, writing things like:

| Expression | Type | Denotation |
|---|---|---|
| a cat | $\boxed{\mathtt{S}\,\mathtt{e}}$ | $\{x \mid \textbf{cat}\, x\}$ |
| the cat | $\boxed{\mathtt{M}\,\mathtt{e}}$ | $x$ if $\textbf{cat} = \{x\}$ else # |
| no cat | $\boxed{\mathtt{C}\,\mathtt{e}}$ | $\lambda Q.\, \neg\exists x.\, \textbf{cat}\, x \wedge Q\, x$ |
| ... | $\boxed{\Sigma\,\mathtt{e}}$ | ... |

Consider then the semantics of a determiner. Each determiner creates a specific sort of computation, the particulars of which depend on the property that restricts it. The indefinite article ‘a’, for example, creates an indeterminate computation by sifting each of its restrictor’s witnesses into an isolated compositional thread. The definite article ‘the’ introduces partiality by creating a computation that might crash, depending on the property it is handed. Given a property $P$, the quantifier ‘no’ determines a computational loop that walks through the $P$s one by one ensuring that none sit truthfully in their syntactic context $Q$. And so on for the others.

The natural denotations for these sorts of operations are functions from properties to computations.

| Expression | Type | Denotation |
|---|---|---|
| a | $(\mathtt{e} \to \mathtt{t}) \to \boxed{\mathtt{S}\,\mathtt{e}}$ | $\lambda P.\, \{x \mid P\, x\}$ |
| the | $(\mathtt{e} \to \mathtt{t}) \to \boxed{\mathtt{M}\,\mathtt{e}}$ | $\lambda P.\, x$ if $P = \{x\}$ else # |
| no | $(\mathtt{e} \to \mathtt{t}) \to \boxed{\mathtt{C}\,\mathtt{e}}$ | $\lambda P \lambda Q.\, \neg\exists x.\, P\, x \wedge Q\, x$ |
| ... | $(\mathtt{e} \to \mathtt{t}) \to \boxed{\Sigma\,\mathtt{e}}$ | ... |

Types like these are the converses of the closure operators in the previous section. Where a closure operator $\lightning :: \boxed{\Sigma\,\mathtt{t}} \to \mathtt{t}$ *reduces* an effect $\boxed{\Sigma\,\mathtt{t}}$ to a value $\mathtt{t}$, a determiner *generates* an effect $\boxed{\Sigma\,\mathtt{e}}$ from a value $(\mathtt{e} \to \mathtt{t})$. In category theoretic settings, types with this general shape, $\sigma \to \boxed{\Sigma\,\tau}$, are known as **Kleisli arrows**, and types with the shape of closure operators, $\boxed{\Sigma\,\sigma} \to \tau$, known as **co-Kleisli arrows**. Simple examples of composition with these types are given in (4.1).

(4.1a)

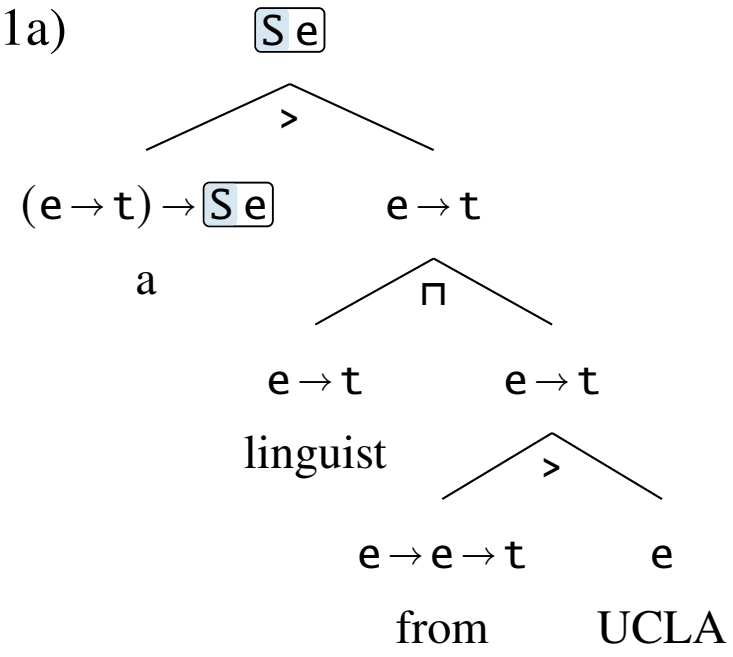

(4.1b)

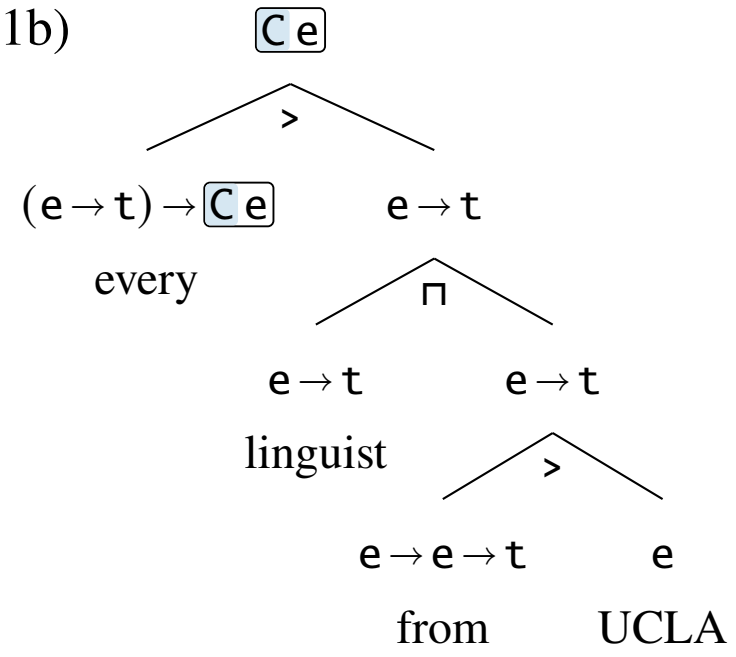

What happens when the restrictor itself denotes a computation, as in (4.2a)? Nothing particularly special. The modes of combination at the nodes of the noun phrase must be mapped over the new effect, but are otherwise unchanged.

(4.2a)

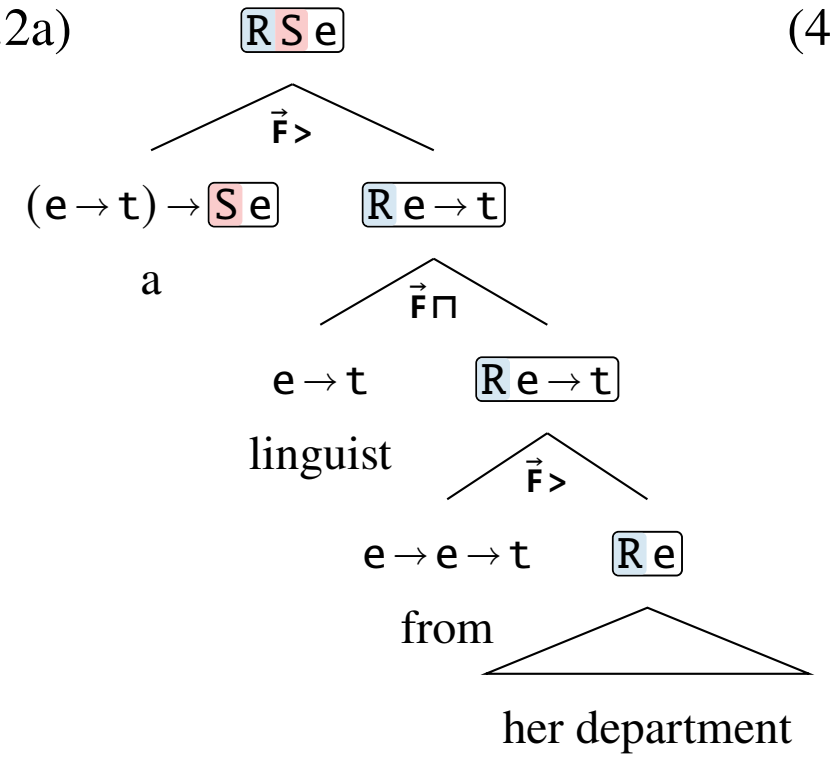

(4.2b)

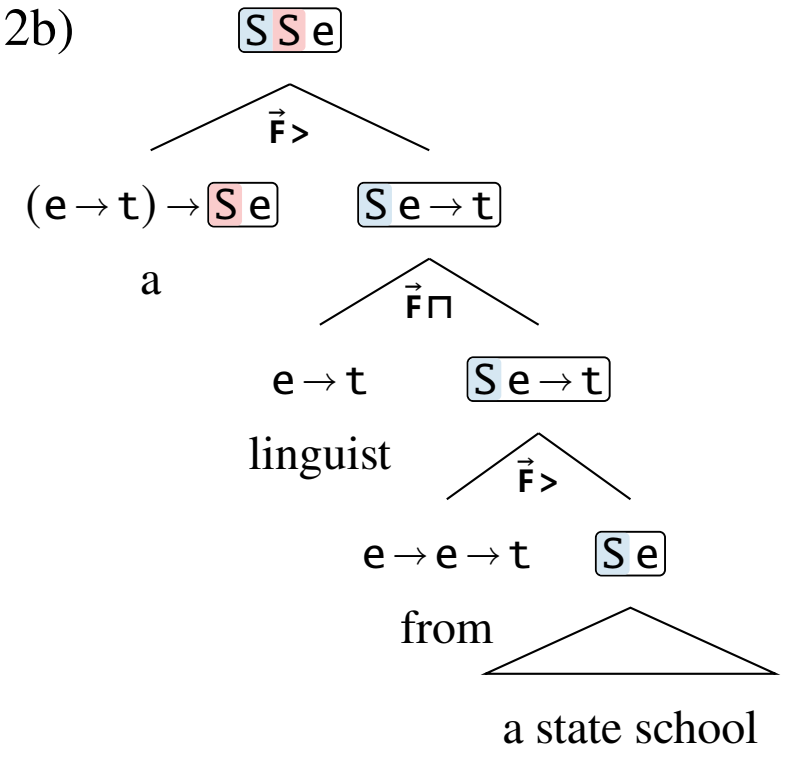

Of note, however, is that (4.2a) is *not ambiguous*. Only one layering of the two effects is possible. The restrictor's effect must take wider scope than the effect introduced by the determiner. There is simply no other way to combine the pieces.

This is true even when the computation in the restrictor is of the same ilk as the determiner's, as seen in (4.2b). The indefinite needs an ordinary property to use as a concrete basis for filtering the domain, but its sister here is indeterminate, holding simultaneously one property per state school. With the algebraic ingredients developed thus far, the way to feed those underlying singular properties to the determiner — the *only* way — is to map the determiner over the effects of the restrictor, as in (4.2b). This means that the resulting denotation is necessarily higher-order, *even though the* S *effect is applicative*.

Compare the derivations in (4.3a) and (4.3b). The un-nested configuration in (4.3a) has a now familiar applicative combination, merging the cat and box alternatives into one set of propositions. But the nested configuration in (4.3b) has no such combination. We are necessarily left with a set of sets of propositions. For each box, we compute a set of propositions containing as many claims as there are cats.[8]

(4.3a)

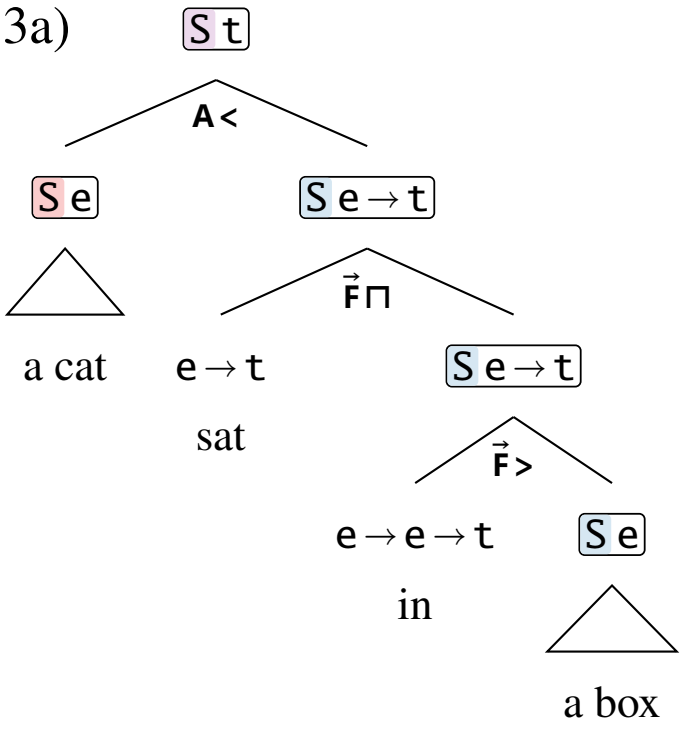

(4.3b)

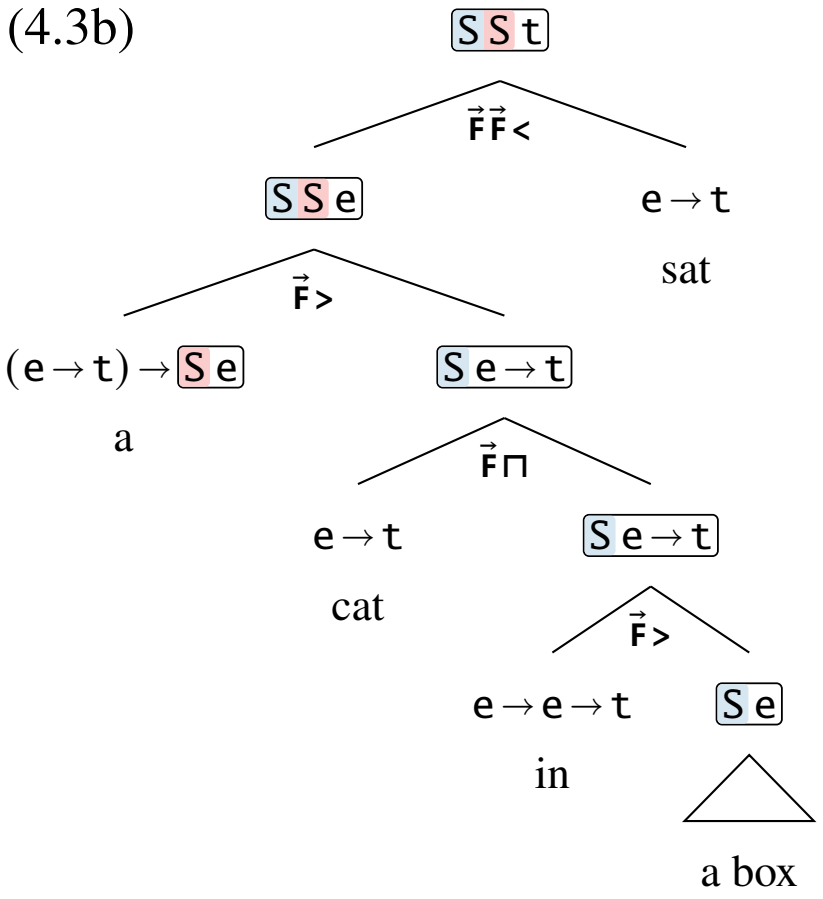

This predicts that an otherwise unselective closure operator will, extraordinarily, fail to capture effects that happen to be nested in the arguments of Kleisli arrows. For instance, assuming as in Chapter 3 that the antecedent of a conditional is associated with an existential closure operator, the current state of affairs predicts that a nested indefinite will *necessarily* take exceptional scope over the conditional.

[8] For illustration, we assume that 'in' is type e→e→t in both its adnominal and adverbial guises, but we stress that this is not intended to be a realistic type of meaning for the adverbial case.

(4.4)

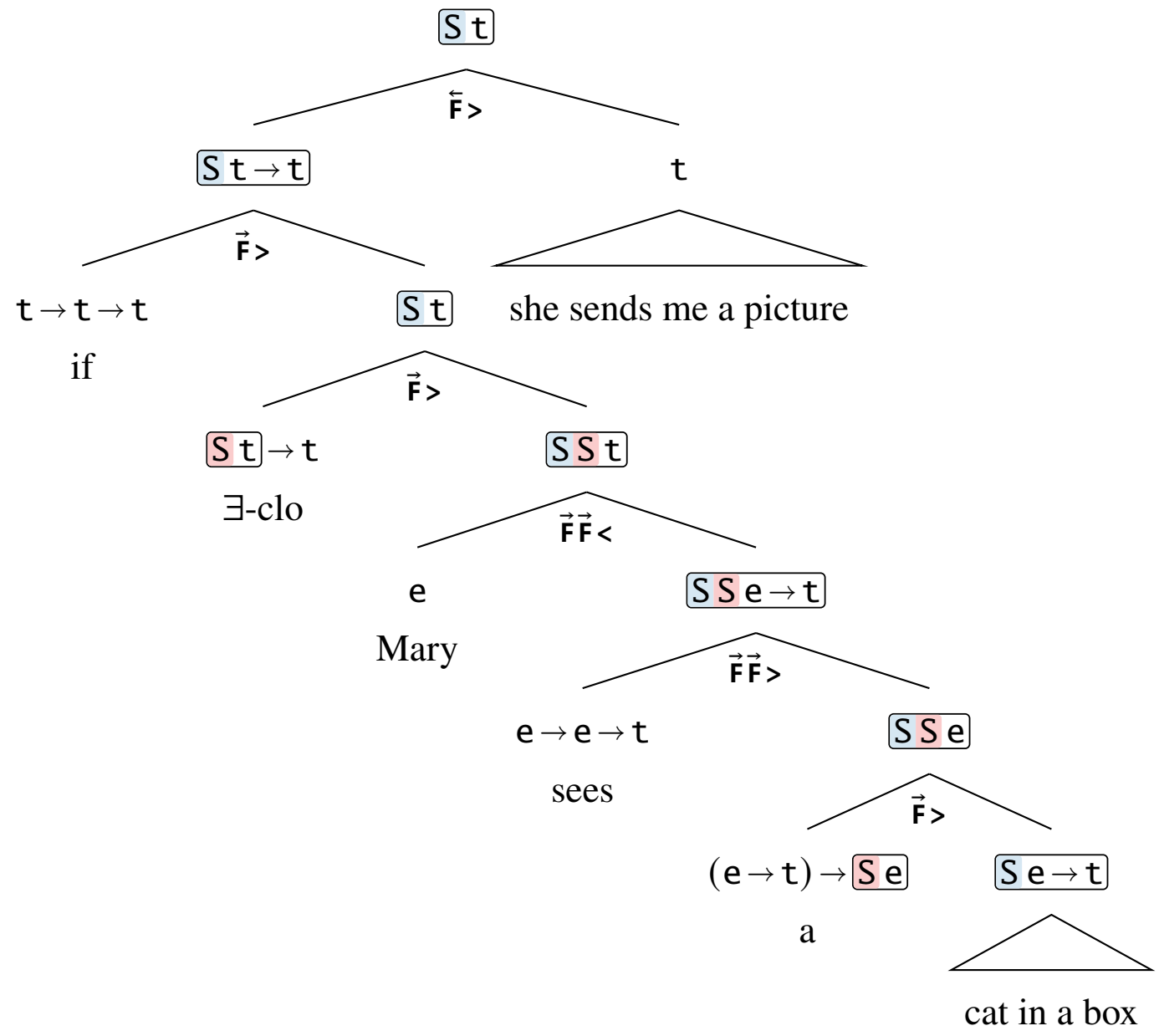

Certainly for English indefinites at least, this prediction is false. The particular problem might be solved by adding more ∃-clo operators in the antecedent, but that strategy is less sensible when the closure operator is a lexical item that associates unselectively with effects. For illustration, consider the entry in (4.5) defining 'can' as an alternative-sensitive modal operator (see, e.g., Aloni 2007; Goldstein 2019 for discussion of analyses along these lines). Given a set of options *m* stemming from the alternative-generators in its prejacent, ⟦can⟧ ensures that every proposition in *m* is a live possibility.

(4.5) can :: $\boxed{\texttt{S}\ \texttt{e} \to \texttt{t}} \to \texttt{e} \to \texttt{t}$

$$[\![\text{can}]\!] \coloneqq \lambda m \lambda x.\ \bigwedge\{\Diamond(Px) \mid P \in m\}$$

As things stand, we predict a difference between the possible readings of (4.6a) and (4.6b). The latter is predicted to confer total freedom to Mary in her choice of apples and blankets. The former only grants universal permission to the apples of a particular blanket. This is empirically disappointing, since in reality it doesn't matter whether it's the apples or Mary that's on the blanket. Both parses have unselective readings.

(4.6a)

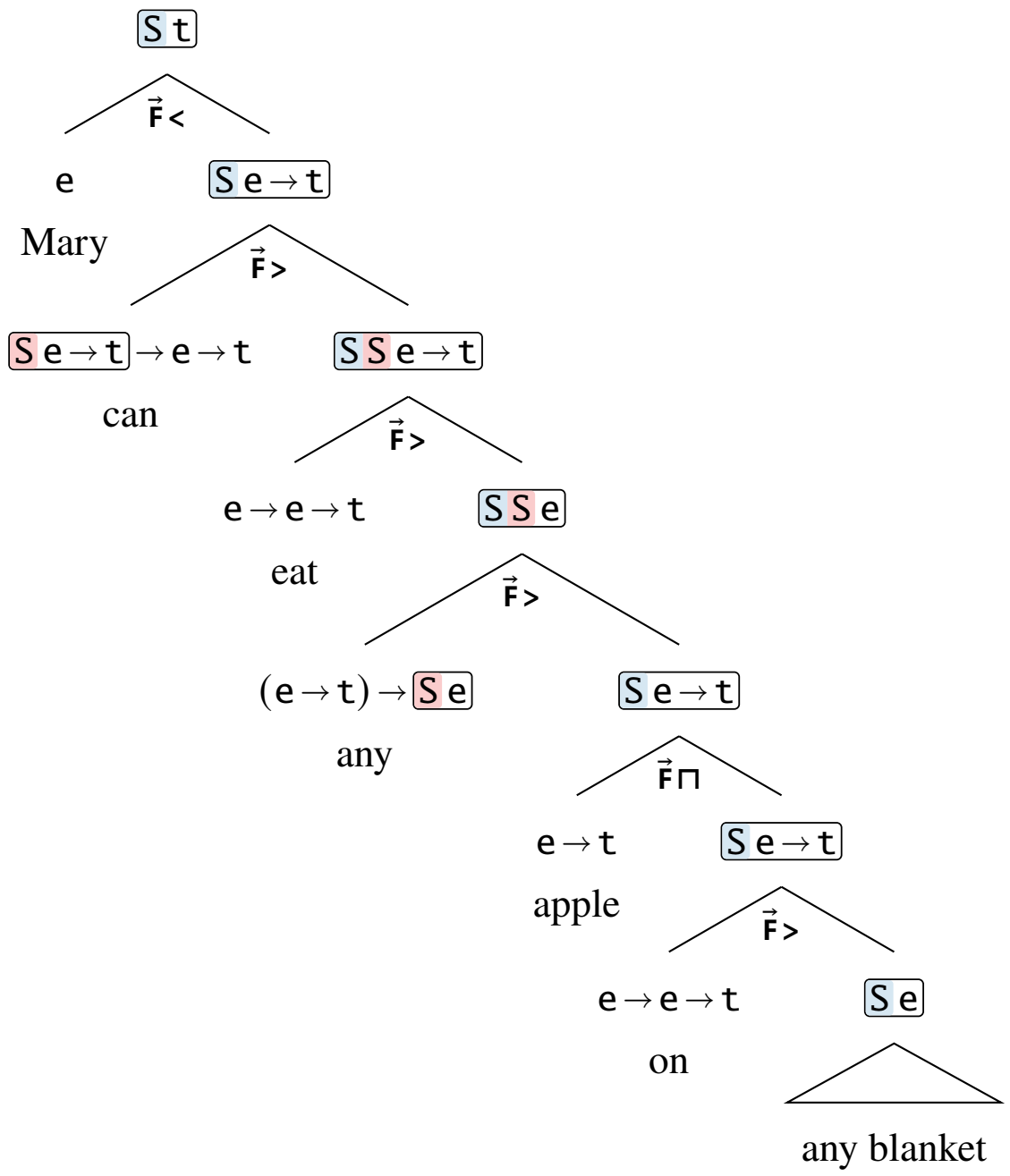

(4.6b)

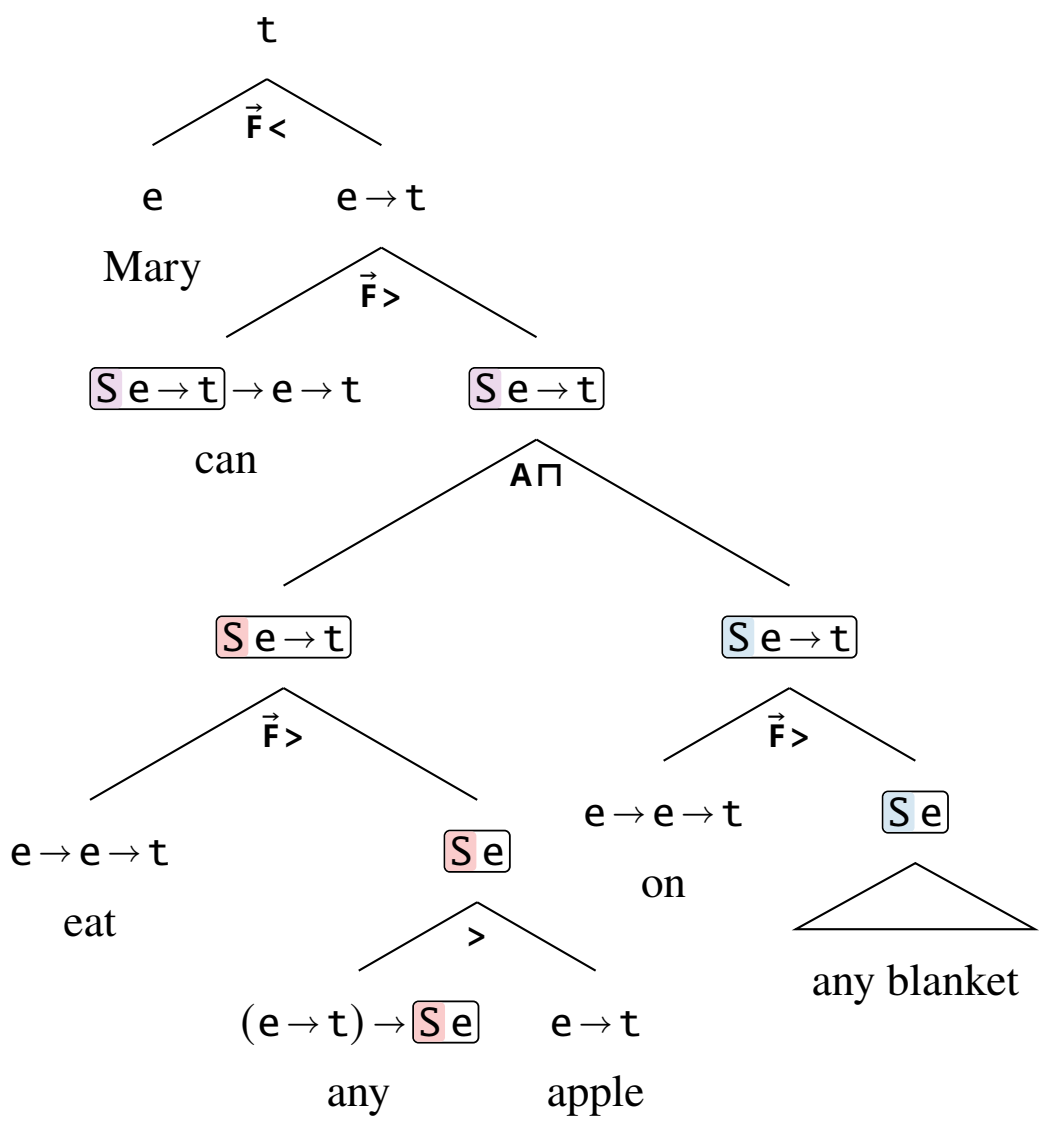

The point is not to argue for any particular analysis of English free choice semantics or indefinite scope delimitation. The point is that if there is *any* operator that can associate with nested effects just the same as it can with un-nested effects, that pattern is beyond the expressive reach of the current grammar.

And such nestings do not only occur in the arguments of determiners. They are liable to pop up any time a Kleisli arrow appears. For instance, consider the abstraction operator commonly used to connect displaced elements with their argument positions, defined with an effect-theoretic type in (4.7). The contructor $\mathtt{V}$ here is a synonym for $\mathtt{R}_{\vec{e}}$ (with $\vec{e}$ the type of sequences of individuals, i.e., assignments), but in anticipation of the discussion of binding in Chapter 5, we give it its own constructor (see, e.g., Büring 2005 for arguments that the assignments used to bind traces should be kept distinct from those that bind pronouns).

(4.7) $$n :: \boxed{\mathtt{V}\,\beta} \to \boxed{\mathtt{V}\,\mathtt{e} \to \beta}$$
$$[\![n]\!] := \lambda b \lambda g \lambda x.\, b\, g^{n \mapsto x}$$

Given an environment-dependent meaning $m :: \boxed{\mathtt{V}\,\beta}$, the abstraction $n$ binds an argument $x$ to the $n$th coordinate of the environment that $m$ is evaluated in. Any trace that accesses this coordinate will therefore resolve to $x$ when evaluated.

If the prejacent of $n$ contains an effect, so that $\beta = \boxed{\Sigma \cdots}$, then the return type of $n$ will contain a Kleisli arrow: $\mathtt{e} \to \boxed{\Sigma \cdots}$. What happens when the specifier of the $n$ contains the same effect, as in (4.8)? (Here, and in the next few examples, we assume overt movement of the subject out of *v*P.) For no good reason, we are forced into a denotation where these effects are nested rather than merged. Again, this seems empirically inadequate, since 'Mary knows who ate what' can mean that Mary knows every pair in the $[\![\text{ate}]\!]$ relation. That is, the attitude verb can quantify unselectively over the 'wh'-elements in its complement.

(4.8)

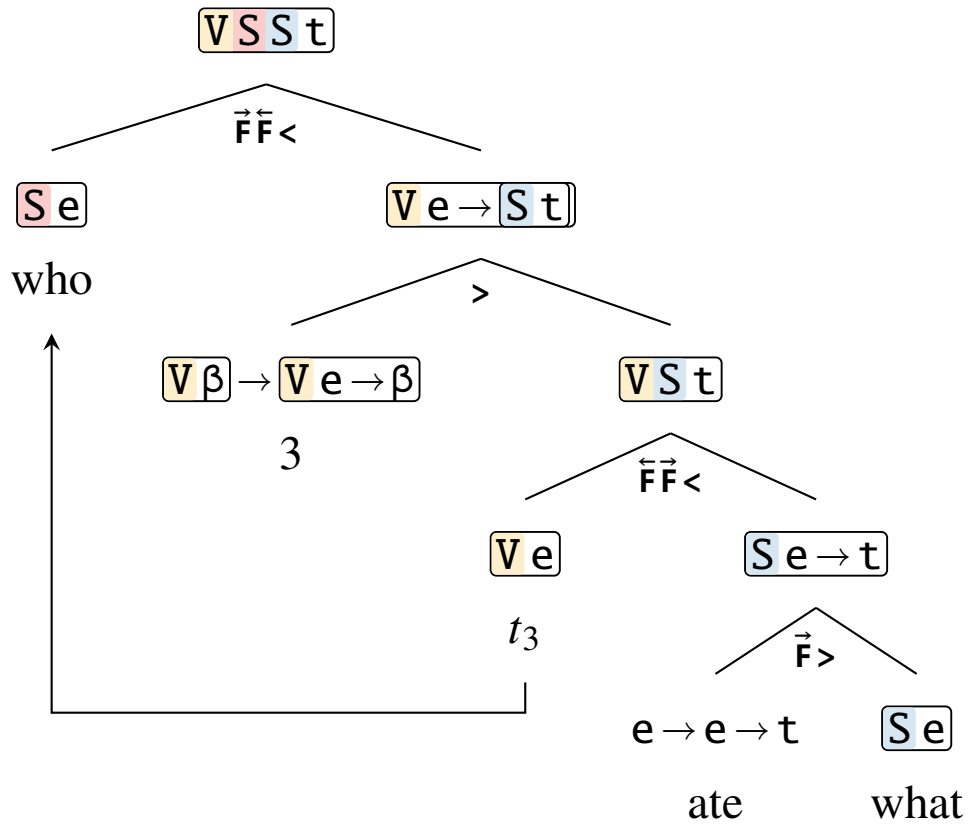

The problem arises regardless of the inner effect. The movement of any semantically rich constituent will result in a layering rather than merging of semantic structure:

(4.9a)

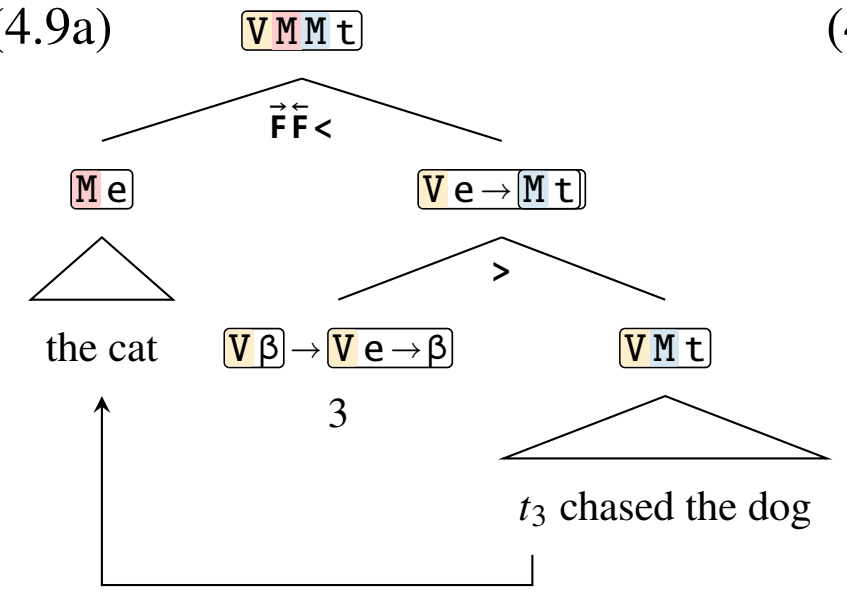

(4.9b)

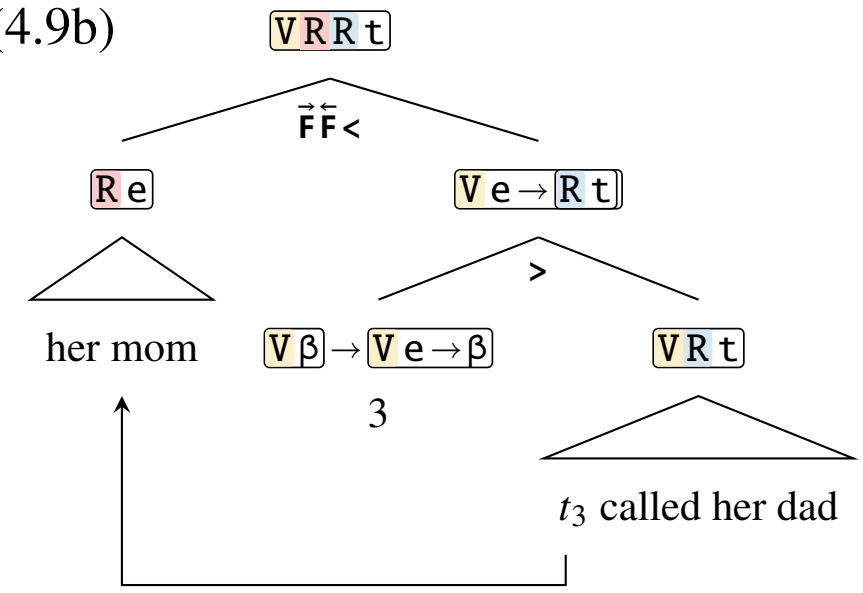

In (4.9b), for instance, the requests from the two pronouns cannot be unified, even though they likely refer to the same antecedent. Compare this with what happens when the two effects occur in nearly any other configuration. With 'her mom' just a little bit higher or lower than the abstraction, the requests of the two anaphoric expressions can be co-valued, as usual.

(4.10)

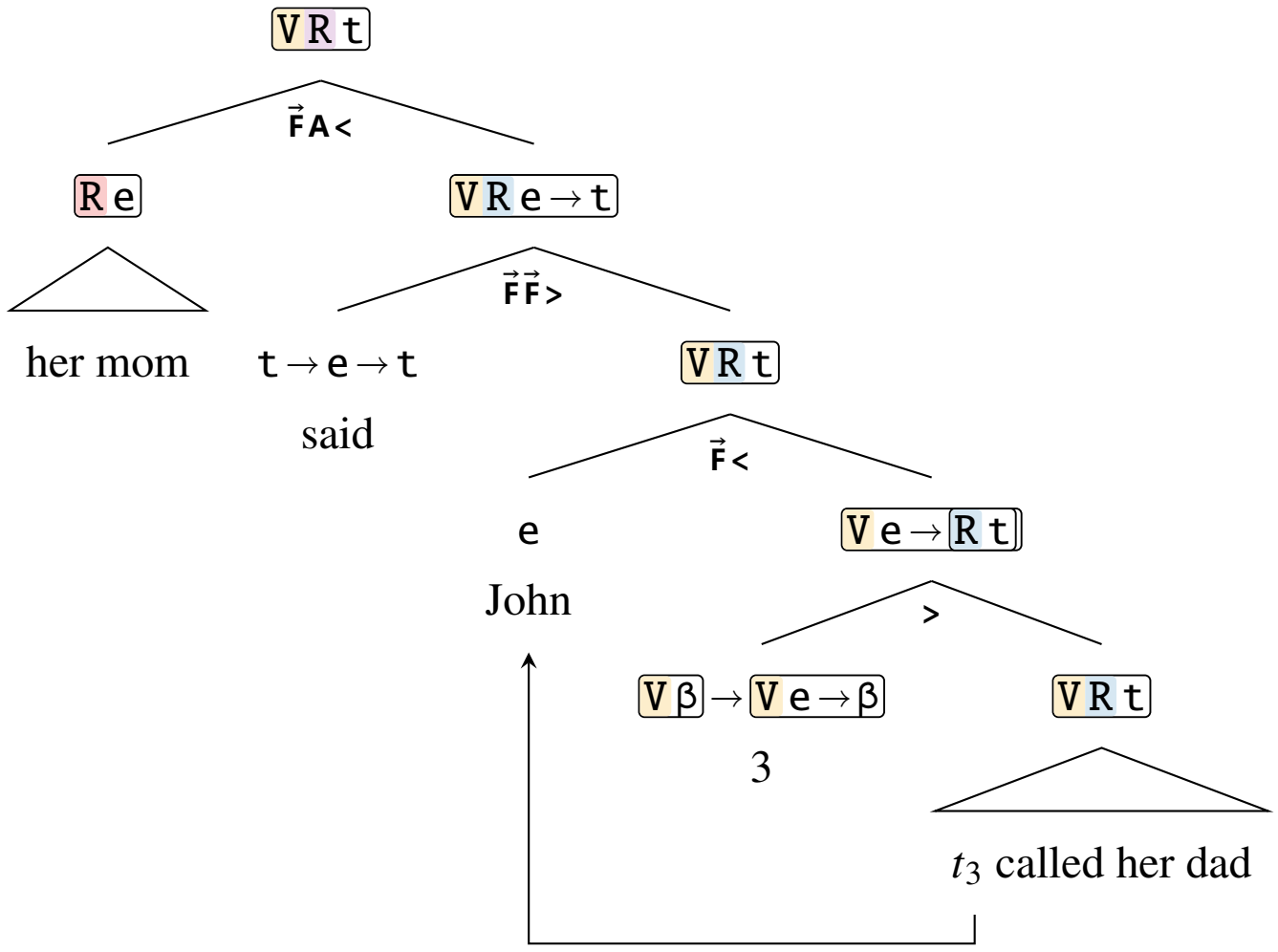

(4.11)

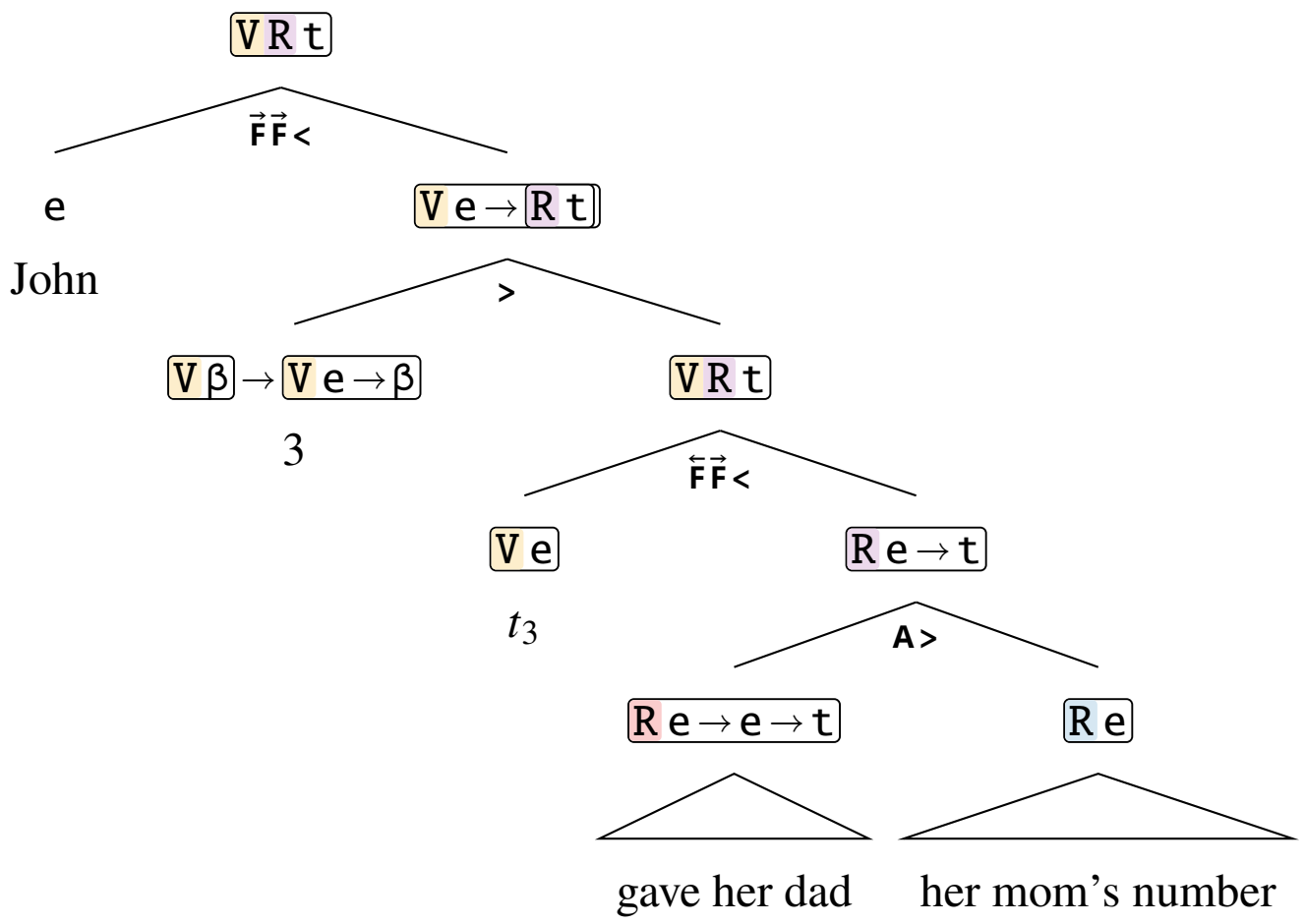

What is needed then is a way to combine a Kleisli arrow $k :: \sigma \to \boxed{\Sigma\,\tau}$ with a computation $m :: \boxed{\Sigma\,\sigma}$ so as to produce a composite computation $\boxed{\Sigma\,\tau}$ that amalgamates the effects of $m$ and $k$. It is not hard to see that such combinations are easy to define when $\Sigma$ is $\mathtt{R}$ or $\mathtt{S}$, which immediately irons out the wrinkles in the examples above. Analogous combinations are in fact definable for all of the computation types in Table 2. Moreover, the various combinators share important algebraic relationships with ($\bullet$), $\eta$, and ($\circledast$). We turn to these relationships in the next section.

## 4.2 Flattening effects

Let's start with $\mathtt{R}$. The task is to find a way of putting something of type $\sigma \to \boxed{\mathtt{R}\,\tau}$ together with something of type $\boxed{\mathtt{R}\,\sigma}$. The obvious candidate — indeed the only polymorphic function with this type — is given in (4.12).

(4.12) $(\mathrel{=\!\!\ll}_{\mathtt{R}}) :: (\sigma \to \boxed{\mathtt{R}\,\tau}) \to \boxed{\mathtt{R}\,\sigma} \to \boxed{\mathtt{R}\,\tau}$

$k \mathrel{=\!\!\ll}_{\mathtt{R}} m := \lambda i.\, k\,(m\,i)\,i$

Notice that this just permutes the arguments of the corresponding ($\circledast$) operation on $\mathtt{R}$, which is fitting since ($\mathrel{=\!\!\ll}$) and ($\circledast$) differ only in the type of their first argument: $\sigma \to \boxed{\mathtt{R}\,\tau}$ vs. $\boxed{\mathtt{R}\,\sigma \to \tau}$.[9] The only difference between these types is whether the environment argument corresponding to $\mathtt{R}$ comes before or after the ordinary argument corresponding to $\sigma$.

[9] Whereas ($\bullet_{\mathtt{R}}$) is Jacobson's (1999) **g**, ($\mathrel{=\!\!\ll}_{\mathtt{R}}$) is none other than Jacobson's **z**.

Let's try $\mathtt{S}$. We're now seeking an operation $(\lessdot_{\mathtt{S}})$ to combine a function $k :: \sigma \to \boxed{\mathtt{S}\,\tau}$ with an argument $m :: \boxed{\mathtt{S}\,\sigma}$. The definition in (4.13) is a natural choice.

(4.13) $(\lessdot_{\mathtt{S}}) :: (\sigma \to \boxed{\mathtt{S}\,\tau}) \to \boxed{\mathtt{S}\,\sigma} \to \boxed{\mathtt{S}\,\tau}$

$$k \lessdot_{\mathtt{S}} m := \bigcup\{k\,x \mid x \in m\}$$

This time there are other functions we could imagine doing the job. The big union, for instance, could just as well be swapped out for a big intersection, as far as the types are concerned. But there is an important sense in which the definition in (4.13) preserves all of the effect structure of $m$ and $k$. A grand intersection would likely throw out alternatives generated by at least one of $m$ and $k$, certainly not something we'd want from a mode of combination.

One way to formalize the sense in which $(\lessdot)$ does not add or lose any information about the effects generated by $m$ and $k$ is to note that for both $\mathtt{R}$ and $\mathtt{S}$, we have the following equivalences.

(4.14a)
$$\begin{aligned} &\lambda x.\, k \lessdot_{\mathtt{R}} (\eta_{\mathtt{R}}\, x) \\ &= \lambda x.\, k \lessdot_{\mathtt{R}} ((\lambda y \lambda_.\, y)\, x) \\ &= \lambda x.\, k \lessdot_{\mathtt{R}} (\lambda_.\, x) \\ &= \lambda x \lambda i.\, k\,((\lambda_.\, x)\, i)\, i \\ &= \lambda x \lambda i.\, k\, x\, i \\ &= k \end{aligned}$$

(4.15a)
$$\begin{aligned} &\lambda x.\, k \lessdot_{\mathtt{S}} (\eta_{\mathtt{S}}\, x) \\ &= \lambda x.\, k \lessdot_{\mathtt{S}} ((\lambda y.\, \{y\})\, x) \\ &= \lambda x.\, k \lessdot_{\mathtt{S}} \{x\} \\ &= \lambda x.\, \{z \mid a \in \{x\},\ z \in k\, a\} \\ &= \lambda x.\, \{z \mid z \in k\, x\} \\ &= k \end{aligned}$$

(4.14b)
$$\begin{aligned} &\eta_{\mathtt{R}} \lessdot_{\mathtt{R}} m \\ &= (\lambda x \lambda_.\, x) \lessdot_{\mathtt{R}} m \\ &= \lambda i.\, (\lambda x \lambda_.\, x)\,(m\, i)\, i \\ &= \lambda i.\, m\, i \\ &= m \end{aligned}$$

(4.15b)
$$\begin{aligned} &\eta_{\mathtt{S}} \lessdot_{\mathtt{S}} m \\ &= (\lambda x.\, \{x\}) \lessdot_{\mathtt{S}} m \\ &= \{z \mid a \in m,\ z \in (\lambda x.\, \{x\})\, a\} \\ &= \{a \mid a \in m\} \\ &= m \end{aligned}$$

The reductions in (4.14a) and (4.15a) guarantee that no information in $k$ is added or lost when it is combined via $(\lessdot)$; since $\eta$ creates a trivial computation, the only effects in $k \lessdot \eta\, x$ should come from $k$. The reductions in (4.14b) and (4.15b) guarantee that no information is added or lost when $m$ is combined via $(\lessdot)$; again the reason is that since $\eta$ doesn't do anything interesting, the only modification to $m$ would have to come from $(\lessdot)$.

An applicative functor for which there is such a well-behaved ($\lll$) operator is known as a **monad**. Formally, to count as well-behaved, the operator should respect the laws in (4.16). The first two equations are just generalizations of the facts observed above. We discuss the significance of **Associativity** in Section 4.3.

(4.16) **Left Identity:** $\eta \lll m = m$

**Right Identity:** $k \lll \eta x = k x$

**Associativity:** $k \lll (c \lll m) = (\lambda x.\, k \lll c x) \lll m$

In Haskell, the ($\lll$) operation is spelled `(=<<)`. In practice, it is often convenient to work with a version of ($\lll$) that takes its arguments in the opposite order:

(4.17) $m \ggg k := k \lll m$

Indeed, in Haskell, it is this flipped version that the standard `Monad` type class implements, where it is given the name `(>>=)`, pronounced “bind”. Obviously `(=<<)` and `(>>=)` are interdefinable. Also, for historical reasons, the `pure` operation guaranteed by the applicativity of the constructor is redundantly specified in the `Monad` class, where it is called `return`.

```
class Applicative f => Monad f where
  (>>=) :: f a -> (a -> f b) -> f b

  return :: a -> f a
  return = pure

(=<<) :: (a -> f b) -> f a -> f b
k =<< m = m >>= k
```

One thing to notice about the ($\lll$) operations defined in (4.12) and (4.13) is that they both implicitly incorporate the corresponding definitions of ($\bullet$) for their types. This is certainly easiest to see in (4.13), which is clearly a mapping of $k$ over $m$ — $k \bullet_{\mathrm{S}} m = \{k x \mid x \in m\}$ — followed by a flattening with $\bigcup$. We might just as well have written $k \lll_{\mathrm{S}} m := \bigcup(k \bullet_{\mathrm{S}} m)$.

Upon inspection, (4.12) can also be seen as a mapping of $k$ over $m$ — $k \bullet_{\mathrm{R}} m = \lambda i.\, k\,(m i)$ — followed by a sort of flattening, namely, the re-use of the argument $i$. The traditional name for this argument-duplicating operation is $\mathbf{W} := \lambda M \lambda i.\, M i i$ (Szabolcsi 1989). Using this, we might just as well have written (4.12) as $k \lll_{\mathrm{R}} m := \mathbf{W}(k \bullet_{\mathrm{R}} m)$.

And in general, every monadic (=≪) :: $(\alpha \to \Sigma\,\beta) \to \Sigma\,\alpha \to \Sigma\,\beta$ is a composition of an effect-mapping operation (•) :: $(\alpha \to \beta) \to \Sigma\,\alpha \to \Sigma\,\beta$ and an effect-flattening operation $\mu$ :: $\Sigma\,\Sigma\,\beta \to \Sigma\,\beta$. That is, for every monad $\Sigma$, there is a $\mu$ such that:

(4.18) $k \lll m = \mu\,(k \bullet m)$

Hence, it suffices to define an operation $\mu$ :: $\Sigma\,\Sigma\,\beta \to \Sigma\,\beta$ that renders the derived (=≪) lawful. Alternatively, we may take (=≪) as primitive, with $\mu\,M := \textbf{id} \lll M$.

Haskell's customary name for $\mu$ is `join`. For quite obscure technical and historical reasons, `join` is omitted from the standard `Monad` type class, but in principle it could have been defined as follows.

```
class Applicative f => Monad f where
  join :: f (f a) -> f a

  return :: a -> f a
  return = pure

m >>= k = join (fmap k m)
k =<< m = m >>= k
```

Finally, we should point out that every monad determines an applicative functor via the equation in (4.19a). That is, whenever (≫=) and $\eta$ satisfy the monad laws in (4.16), the operator defined in (4.19a) together with the same $\eta$ will satisfy the applicative laws in (3.4). And every monad determines a functor via the recipe in (4.19b). The **Left Identity** monad law in (4.16) transparently guarantees that the operation defined in (4.19b) satisfies the **Identity** functor law in (2.4). So just as in Chapter 3, the free theorem for functors guarantees that the (•) defined by (4.19b) is the unique parametric map for its type, in agreement with the maps deduced in the previous chapters.

(4.19a) $F \circledast X = F \ggg\!= \lambda f.\, X \ggg\!= \lambda x.\, \eta\,(f\,x)$

(4.19b) $k \bullet m = m \ggg\!= \lambda x.\, \eta\,(k\,x)$

As might be expected at this point, all of the effects introduced so far in this Element (see Table 2 for reference) are monadic.[10] We provide standard definitions of the monad operators for these constructors in Appendix A5.

[10]The caveat from Section 3.6 applies again in the case of `W`: the action that fuses together two computations' supplemental contents needs to be associative and needs to have an identity for a proper definition of $\eta$. In Table 2, these supplements are propositions, fused together by conjunction.

*4.2.1 Flattening as a higher-order mode of combination*

Incorporating (≪) into a grammar presents the same options as in the preceding chapters. First, since the nodes that caused compositional trouble in (4.3b)/(4.8) have exactly the types that (≪) and (≫) are suited to combine, those immediate problems would be resolved just by adding these two operators to the inventory of combinatory modes. This would be exactly analogous to the naive additions of $(\bullet_>)$ and $(\bullet_<)$ in Chapter 2, or ⧁ and ⧀ in Chapter 3.

With these, we might put together perfectly respectable derivations as in (4.20) and (4.21).

(4.20)

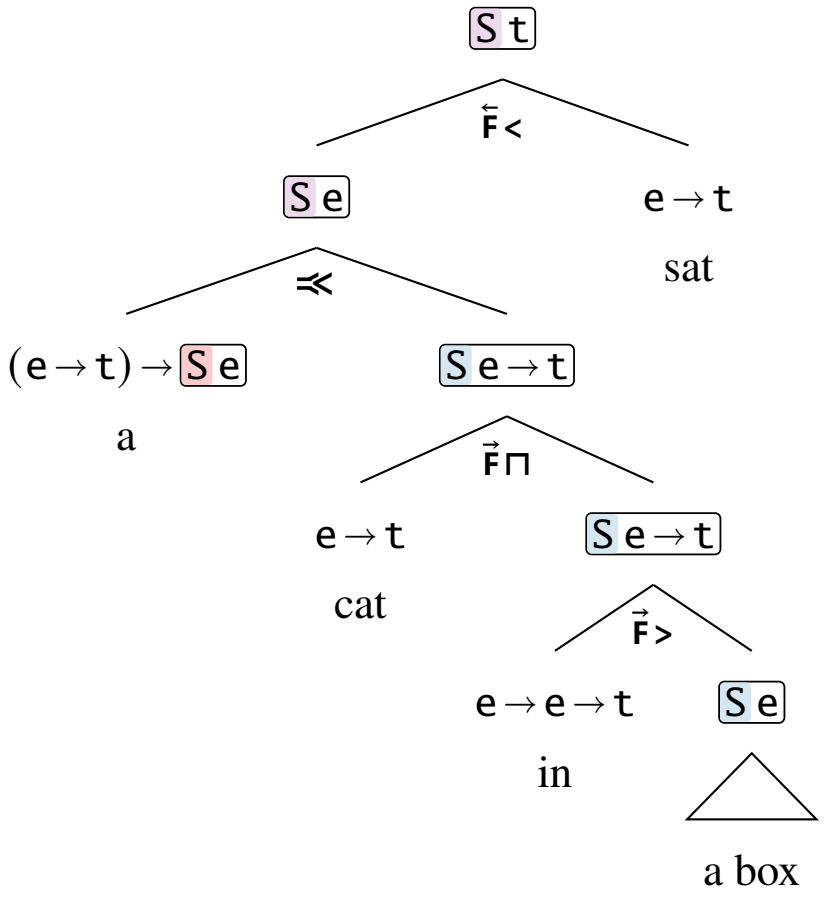

(4.21)

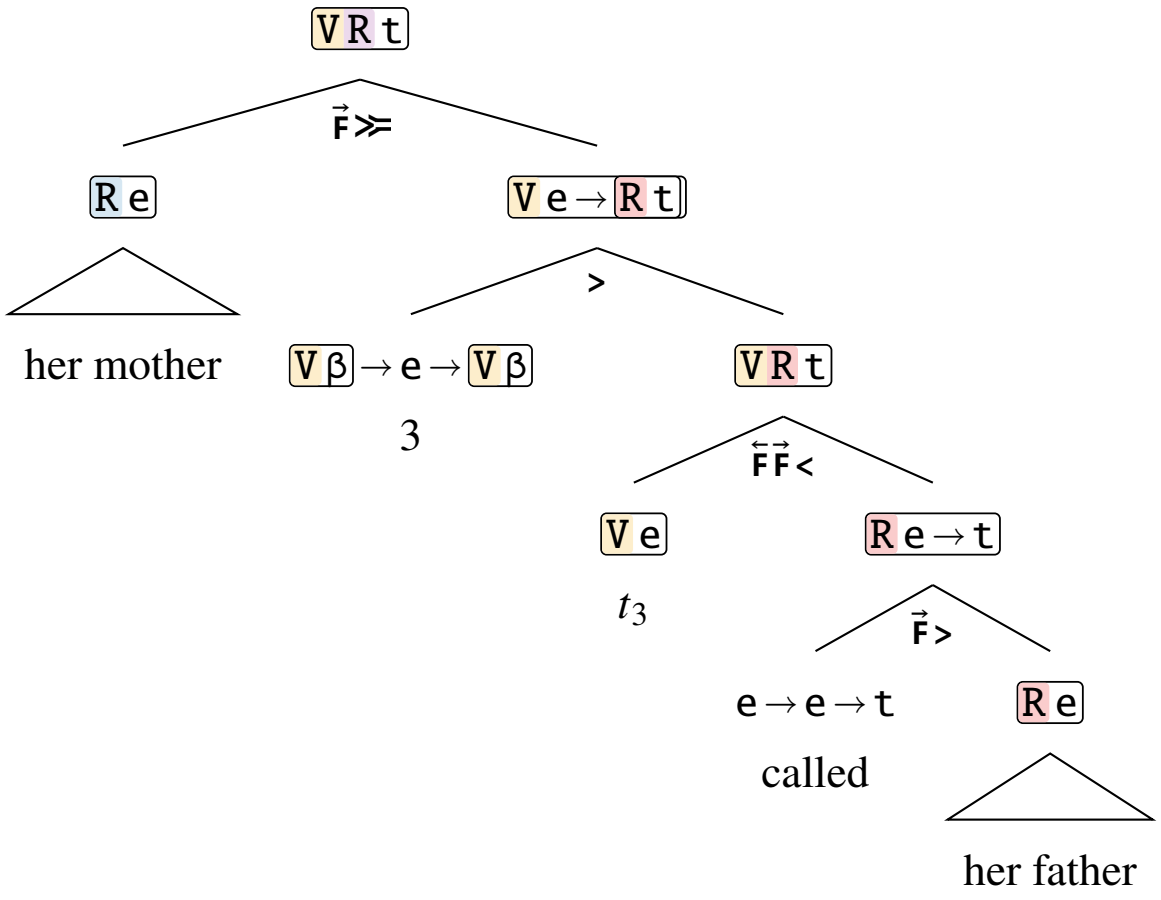

But since the grammar we have developed so far already has robust resources for mapping left and right with forward and backward application, we might as well make use of the equivalence in (4.18). This way we add only one meta-rule for exploiting the monadic nature of effects, defined in (4.22). The entire grammar, extended with **J**, is presented in Figure 8.

(4.22) $\mathbf{J}\,(*)\,E_1\,E_2 \coloneqq \mu\,(E_1 * E_2)$

When the parameter $(*)$ to this rule is $\vec{\mathbf{F}}\,\mathbf{>}$, the result is equivalent to $(\Lleftarrow)$, and when the parameter is $\overleftarrow{\mathbf{F}}\,\mathbf{<}$, the result is equivalent to $(\ggg)$. Thus we would derive (4.20), for instance, as in (4.23) instead.

(4.23)

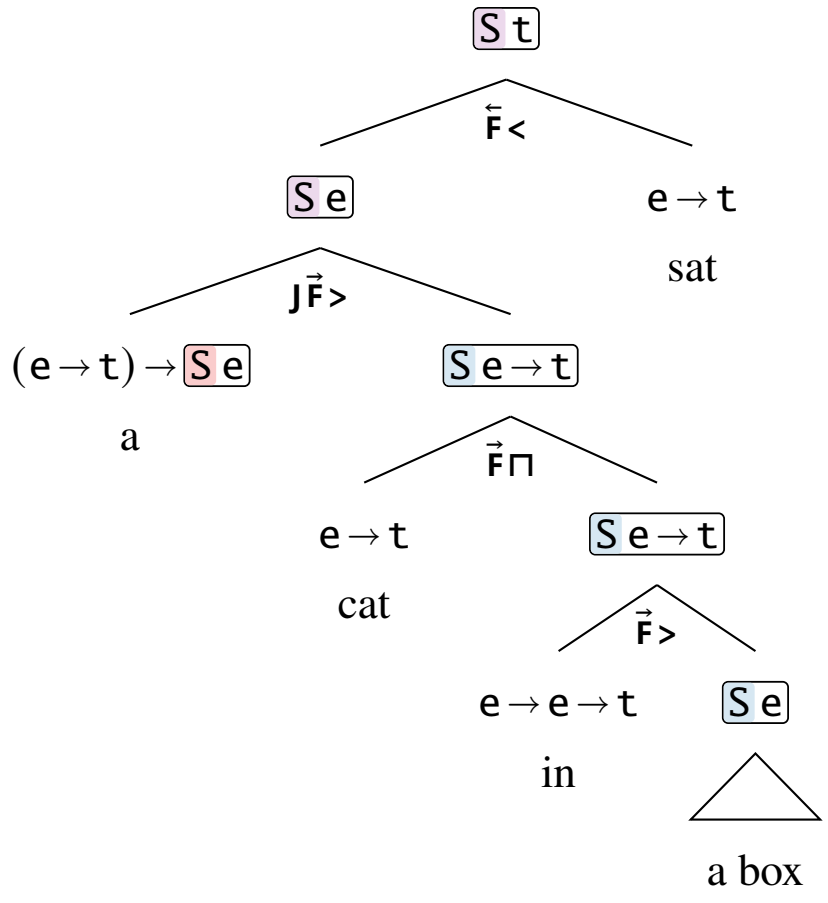

And being a mode of combination, the **J** rule may itself appear in the argument to other meta-combinators like $\vec{\mathbf{F}}$, $\overleftarrow{\mathbf{F}}$, and **A**. This guarantees that incidental occurrences of other kinds of effects can continue to bubble up even as monadic effects are ironed out below them. The derivation in (4.24) provides an example.

(4.24)

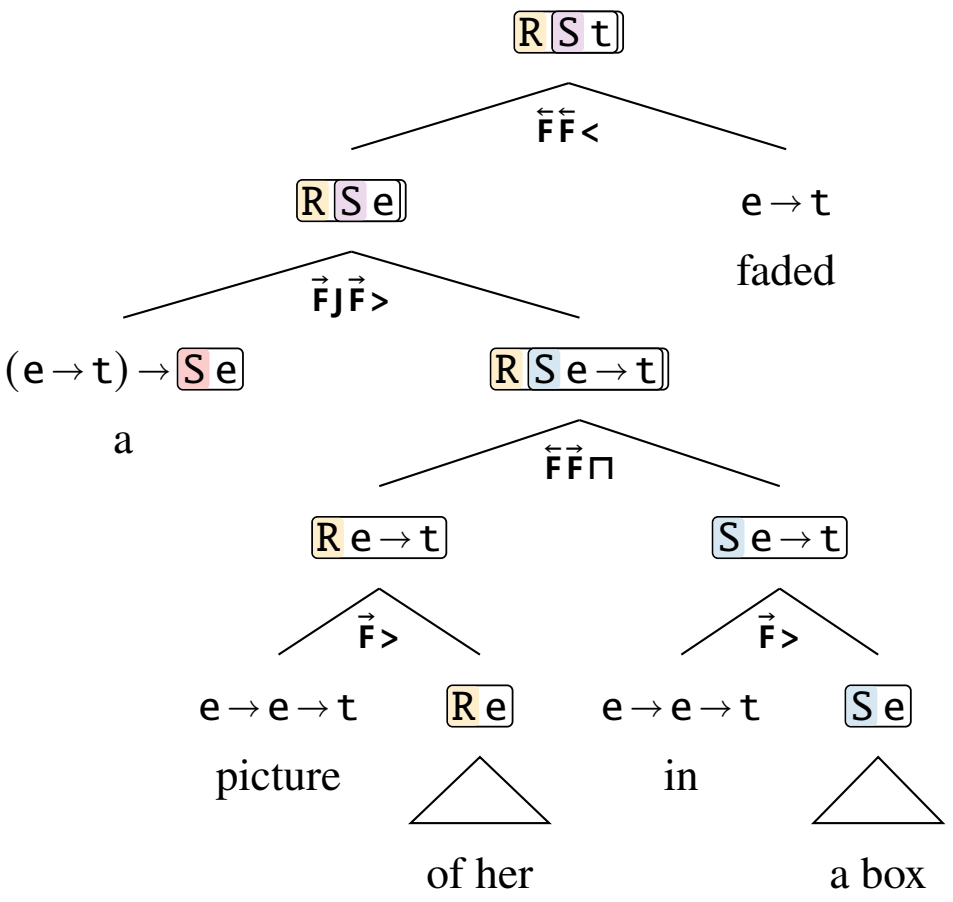

## 4.3 Scope and ≫=-ing

### 4.3.1 LFs and abstraction

Monads have played an important mathematical role in provisioning **imperative programming languages** with denotational semantics. These are languages that include commands for the sorts of actions described in Chapter 1: assigning values to variables, throwing errors, starting loops, etc. Such commands are described as denoting computations (as opposed to values) in exactly the way that the expressions in Table 2 have been here. And sequences of two such commands, one of which depends on the value computed by the other, are given meanings in terms of (≫=).

**Functional programming languages**, unlike imperative languages, and unlike natural languages, make these denotational mechanisms explicit in the syntax of expressions. In essence, everything that is packed into the modes of combination in Figure 8 (or left implicit in an imperative language), must be typed out as part of the program itself in a functional language like Haskell. This means there are a lot of explicit `>>=`s and `fmap`s gluing everything together. In fact, `>>=` has played such an outsized role in structuring Haskell programs that it has its own syntax called `do`-notation.

`do`-blocks represent sequences of monadic actions, chained together by `>>=`s. They are intended to resemble an imperatively organized program while

**Types:**

$$\tau ::= \mathtt{e} \mid \mathtt{t} \mid \cdots$$ Base types
$$\mid\ \tau \to \tau$$ Function types
$$\mid\ \ldots$$ ...
$$\mid\ \boxed{\Sigma\,\tau}$$ Computation types

**Effects:**

$$\Sigma ::= \mathtt{R}$$ Input
$$\mid\ \mathtt{W}$$ Output
$$\mid\ \mathtt{S}$$ Indeterminacy
$$\mid\ \ldots$$ ...

**Basic Combinators:**

$$(\mathbf{>}) :: (\alpha \to \beta) \to \alpha \to \beta$$ Forward Application
$$f \mathbf{>} x \coloneqq f\,x$$

$$(\mathbf{<}) :: \alpha \to (\alpha \to \beta) \to \beta$$ Backward Application
$$x \mathbf{<} f \coloneqq f\,x$$

... ...

**Meta-combinators:**

$$\overleftarrow{\mathsf{F}} :: (\sigma \to \tau \to \omega) \to \boxed{\Sigma\,\sigma} \to \tau \to \boxed{\Sigma\,\omega}$$ Map Left
$$\overleftarrow{\mathsf{F}}\,(*)\,E_1\,E_2 \coloneqq (\lambda a.\, a * E_2) \bullet E_1$$

$$\overrightarrow{\mathsf{F}} :: (\sigma \to \tau \to \omega) \to \sigma \to \boxed{\Sigma\,\tau} \to \boxed{\Sigma\,\omega}$$ Map Right
$$\overrightarrow{\mathsf{F}}\,(*)\,E_1\,E_2 \coloneqq (\lambda b.\, E_1 * b) \bullet E_2$$

$$\mathsf{A} :: (\sigma \to \tau \to \omega) \to \boxed{\Sigma\,\sigma} \to \boxed{\Sigma\,\tau} \to \boxed{\Sigma\,\omega}$$ Structured App
$$\mathsf{A}\,(*)\,E_1\,E_2 \coloneqq (\lambda a \lambda b.\, a * b) \bullet E_1 \circledast E_2$$

$$\overleftarrow{\mathsf{U}} :: (\sigma \to (\tau \to \tau') \to \omega) \to \sigma \to (\boxed{\Sigma\,\tau} \to \tau') \to \omega$$ Unit Left
$$\overleftarrow{\mathsf{U}}\,(*)\,E_1\,E_2 \coloneqq E_1 * (\lambda b.\, E_2\,(\eta\,b))$$

$$\overrightarrow{\mathsf{U}} :: ((\sigma \to \sigma') \to \tau \to \omega) \to (\boxed{\Sigma\,\sigma} \to \sigma') \to \tau \to \omega$$ Unit Right
$$\overrightarrow{\mathsf{U}}\,(*)\,E_1\,E_2 \coloneqq (\lambda a.\, E_1\,(\eta\,a)) * E_2$$

$$\mathsf{J} :: (\sigma \to \tau \to \boxed{\Sigma\,\Sigma\,\omega}) \to \sigma \to \tau \to \boxed{\Sigma\,\omega}$$ Join
$$\mathsf{J}\,(*)\,E_1\,E_2 \coloneqq \mu\,(E_1 * E_2)$$

**Figure 8** A type-driven grammar with monads

maintaining referential transparency and type safety. For instance, the following program block has the intuitive behavior enumerated to its right.

```
s = do x <- m
       y <- o x
       return (p y)
```

1. Compute `m` to get a value `x`
2. Pass `x` to `o` to compute a value `y`
3. Pass `y` to `p`; package the result as a computation

This block is mechanically "de-sugared" by the compiler into a right-nested sequence of binds:

```
s = m >>= (\x -> o x >>= (\y -> return (p y)))
```

Essentially, each `v <- m ...` is translated as `m >>= (\v -> ...)`. This means that in a **do**-block, there must be some monad $\Sigma$ such that:

- Every expression to the right of `<-` has type $\Sigma\,\alpha$
- The last line is an expression of type $\Sigma\,\zeta$ for some type $\zeta$
- The whole block then has type $\Sigma\,\zeta$

Like everything in Haskell, **do**-blocks are just expressions. They may appear anywhere that any other expression of the same type might appear. In particular, since a **do**-block denotes a computation of type $\Sigma\,\zeta$ for some monad $\Sigma$, it may itself occur on the right side of a `<-` in a larger **do**-block. For instance, taking advantage of the equivalences in (4.19b), we might write the **Composition Law** for functors — $f \bullet (g \bullet M) = (f \circ g) \bullet M$ — as an equivalence between programs, as in (4.25).

(4.25)

```
do y <- { do x <- m
             return (g x) }
   return (f y)
```

=

```
do x <- m
   return (f (g x))
```

With this in mind, let us revisit the basic derivation in (4.3a), repeated below as (4.26).

(4.26)

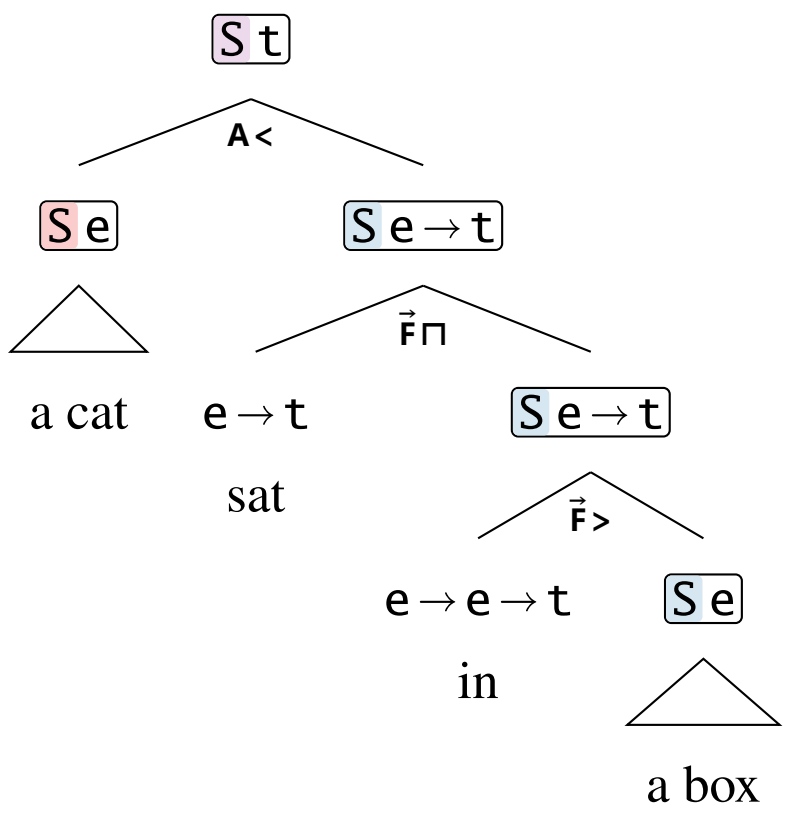

Starting with the prepositional phrase, the definition of **F⃗** tells us that this constituent's meaning is computed by mapping the preposition **in** over the indefinite **a box**. In Haskell, this is represented by the program `fmap in' (a box)` (`in` is a reserved keyword in Haskell, hence the `in'`). Given the equivalence in (4.19b), this could just as well be expressed as `a box >>= \x -> return (in' x)`. And written in `do`-notation, this is the program in (4.28).

(4.27)

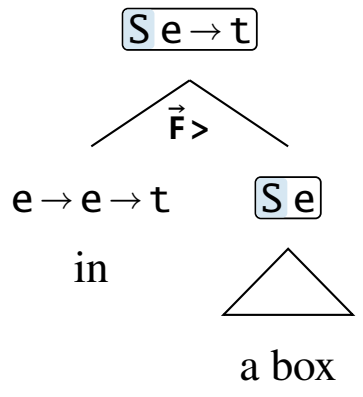

(4.28)

```
do x <- a box
   return (in' x)
```

Repeating this translation at the next constituent up delivers the program in (4.30), where `(&)` is (⊓). And given the functor law spelled out in (4.25), this is equivalent to (4.31).

(4.29)

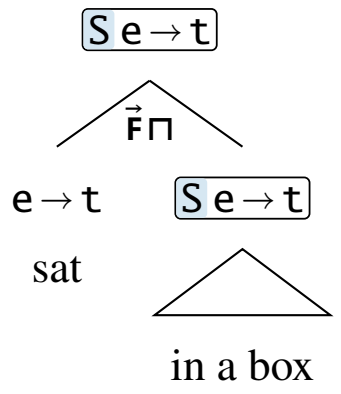

(4.30)

```
do p <- { do x <- a box
             return (in' x) }
   return (sat & p)
```

(4.31)

```
do x <- a box
   return (sat & in' x)
```

Finally, using the monadic encoding of (⊛) in (4.19a), the top level constituent is computed by the program in (4.33), which, again given the functor law in (4.25), is equivalent to (4.34).

(4.32)

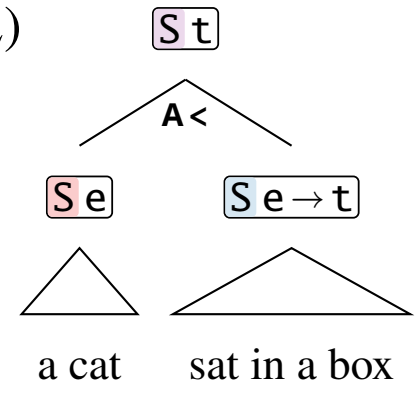

(4.33)

```
do z <- a cat
   p <- { do x <- a box
             return (sat & in' x) }
   return (p z)
```

(4.34)

```
do z <- a cat
   x <- a box
   return ((sat & in' x) z)
```

De-sugaring the `do`-notation into `>>=`s, and drawing this program as a tree yields the quasi-derivation in (4.35). Linguists accustomed to analyzing sentences in terms of their "Logical Forms" (LFs) may be struck by how much this looks like Quantifier Raising (May 1985, Heim and Kratzer 1998). All of the properly computational expressions are raised from their argument positions. Those positions are instead filled with variables that are abstracted over where the fancy constituent lands, forming a kind of "scope" for the computation. But these computations are not quantificational; they do not take scopes as arguments. Instead, the enriched content and its **continuation** — the compositional context in which the enriched content is situated — are combined via (≫=).[11]

(4.35)

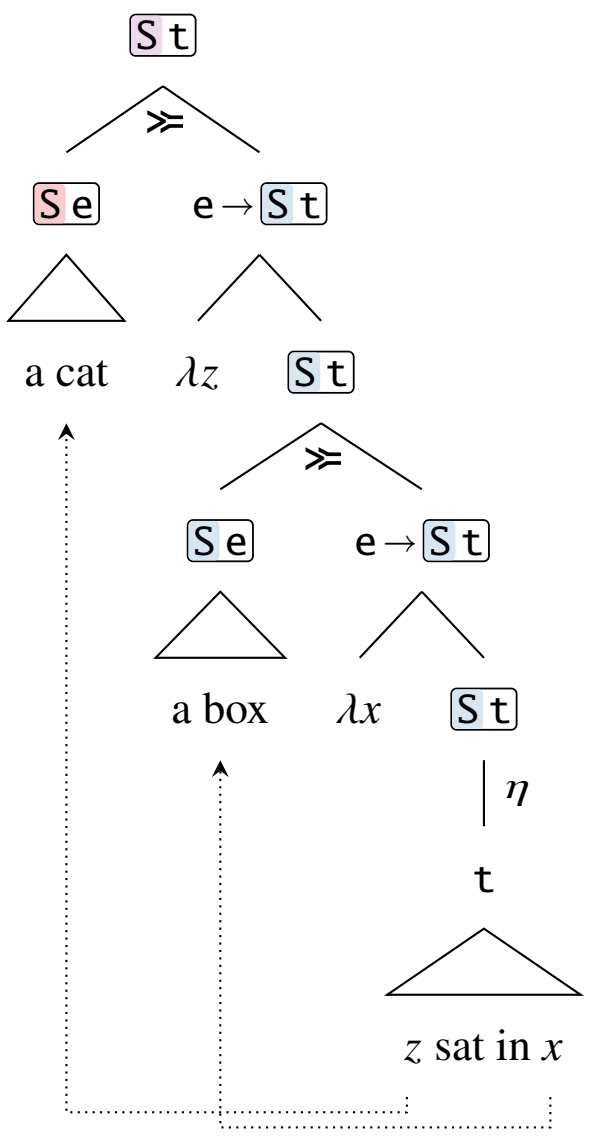

(4.36)

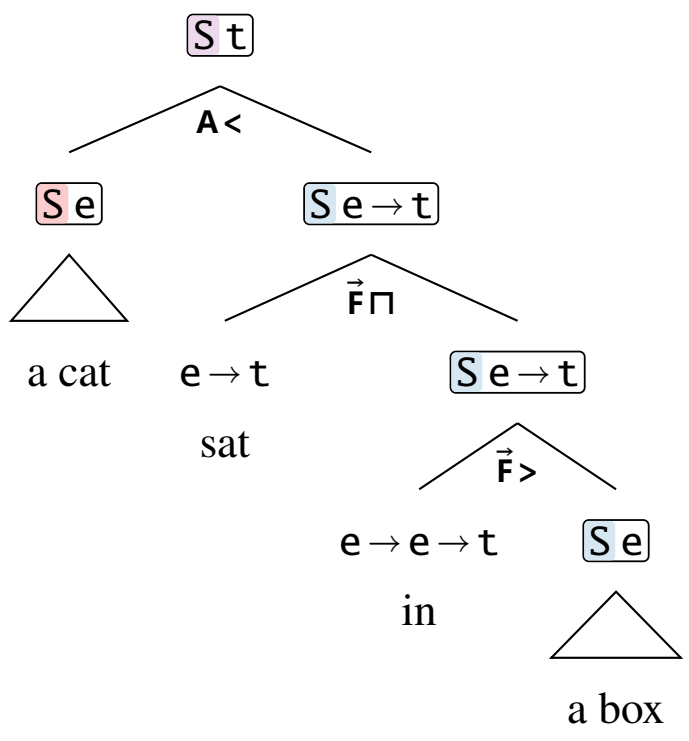

In this sense, Haskell's `do`-notation is very much like the linguist's LF. Content that cannot be interpreted *in situ* via the basic modes of combination is moved out of the way, leaving a named trace as a placeholder. All of the day-to-day argument-structural logic of the derivation is performed with this variable. At the top, the results of this ordinary calculation are folded over the computational structure of the extra-ordinary expression.

[11]The tree in (4.35), and several others in the remainder of this section, contain lambdas and variables. These should not be confused with the abstractions and traces introduced in Chapter 3 to deal with overt movement. The latter are ordinary object-language expressions, interpreted using the same sort of explicit, categorematic semantics as everything else. The lambdas here are really in the metalanguage, with the modes of combination. We will be content with this abuse of notation, as our goal in this section is just to bring out connections between the LF approach with covert scope-taking and the higher-order combinatorial approach developed in the Element.

Remarkably, the monad laws guarantee that this transformational derivation and the original, *in situ* derivation in (4.36) are equivalent, *for any monadic type* Σ. We don't even need to compute the denotations. But (4.35) is not the only conceivable LF for the sentence. Naturally, there is another well-typed derivation in which the object outscopes the subject, as shown in (4.37), corresponding to the `do`-block in (4.38).

(4.37)

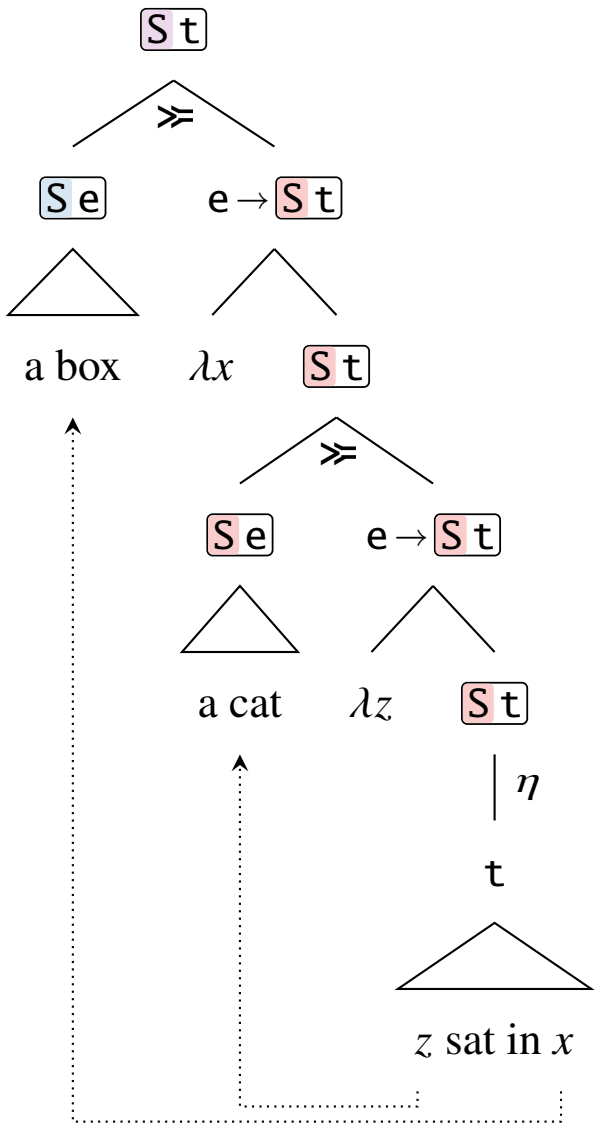

(4.38)

```
do x <- a box
   z <- a cat
   return ((sat & in' x) z)
```

If (4.36) is algebraically equivalent to (4.35), we might reasonably ask if there is an *in situ* derivation equivalent to (4.37). It turns out there is. The surface ordering of effects in (4.36) stems from the application of **A** at the root, which by definition sequences effects from left to right. But **A** is not the only mode of combination the type-driven interpreter will find for these types. Just as in Chapter 2, it is possible to put these nodes together in such a way that we end up with a higher-order meaning of type S S t in which the right daughter ends up outside the left, shown in (4.39a). And given this, there must be yet another mode of combination that follows this one up with **J**, shown in (4.39b).

(4.39a)

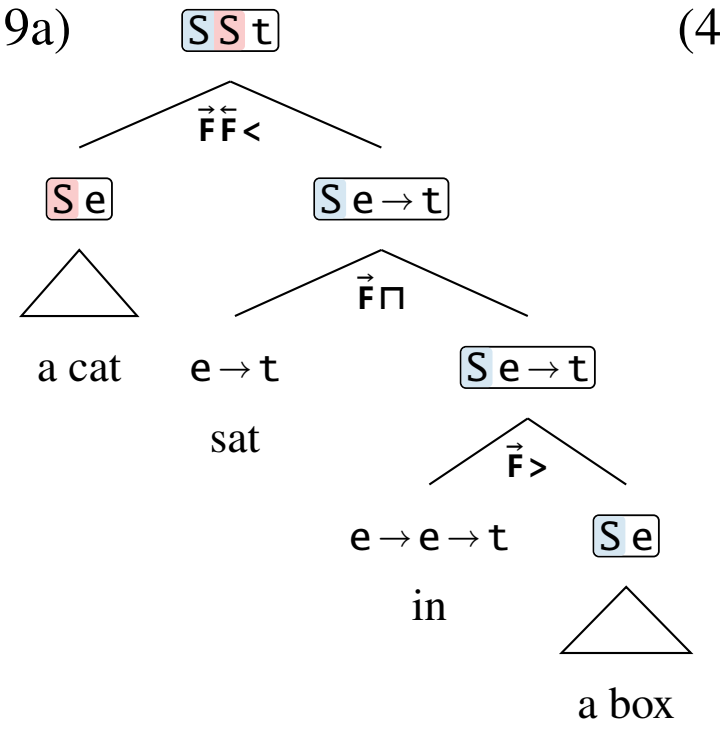

(4.39b)

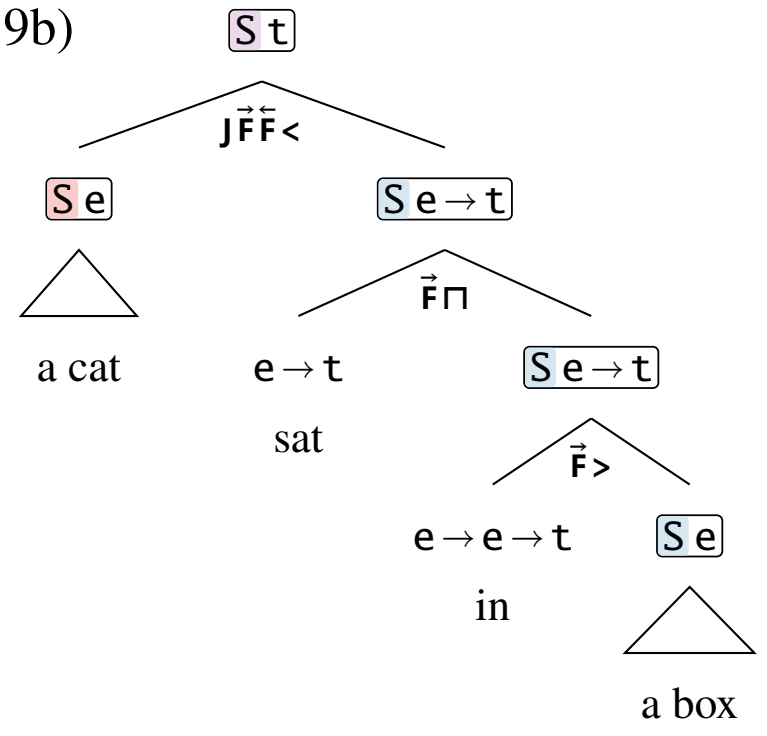

Let $F = [\![\text{sat in a box}]\!]$ and $A = [\![\text{a cat}]\!]$. Without unpacking the sets that these expressions denote, we can reason as follows about the meaning of (4.39b).

(4.40) $\mathbf{J}\,(\vec{\mathbf{F}}\,(\overleftarrow{\mathbf{F}}\,{<}))\,A\,F$

$= \mu\,((\lambda p.\,(\lambda z.\,p\,z) \bullet A) \bullet F)$ — def. of $\mathbf{J}$, $\vec{\mathbf{F}}$, and $\overleftarrow{\mathbf{F}}$

$= F \ggg \lambda p.\,(\lambda z.\,p\,z) \bullet A$ — (4.18): $\mu\,(k \bullet m) = m \ggg k$

$= F \ggg \lambda p.\,A \ggg \lambda z.\,\eta\,(p\,z)$ — (4.19b): $k \bullet m = m \ggg \lambda z.\,\eta\,(k\,z)$

$= ([\![\text{a box}]\!] \ggg \lambda x.\,\eta\,(\mathbf{sat} \sqcap \mathbf{in}\,x)) \ggg \lambda p.\,(A \ggg \lambda z.\,\eta\,(p\,z))$ — expand $F$, per (4.31)

$= [\![\text{a box}]\!] \ggg \lambda x.\,(\eta\,(\mathbf{sat} \sqcap \mathbf{in}\,x) \ggg \lambda p.\,(A \ggg \lambda z.\,\eta\,(p\,z)))$ — (4.16): assoc. law

$= [\![\text{a box}]\!] \ggg \lambda x.\,(A \ggg \lambda z.\,\eta\,((\mathbf{sat} \sqcap \mathbf{in}\,x)\,z))$ — (4.16): right id law

$= [\![\text{a box}]\!] \ggg \lambda x.\,([\![\text{a cat}]\!] \ggg \lambda z.\,\eta\,((\mathbf{sat} \sqcap \mathbf{in}\,x)\,z))$ — expand $A$

Since this last line is exactly the tree in (4.37), we have found the inverse-scope derivation. That is, the monad laws guarantee that the *in situ* combination in (4.39b) is equivalent to the transformational LF in (4.37).

At the level of meanings, this is all a bit extravagant, since (4.35) and (4.37) compute the same set of propositions: $\bigcup\{\{(\mathbf{sat} \sqcap \mathbf{in}\,x)\,z \mid \mathbf{cat}\,z\} \mid \mathbf{box}\,x\} = \bigcup\{\{(\mathbf{sat} \sqcap \mathbf{in}\,x)\,z \mid \mathbf{box}\,x\} \mid \mathbf{cat}\,z\} = \{(\mathbf{sat} \sqcap \mathbf{in}\,x)\,z \mid \mathbf{cat}\,z, \mathbf{box}\,x\}$. But importantly, the reasoning above establishes *in situ* and *ex situ* translations regardless of the constructor $\Sigma$ or the actual meanings of the subject and predicate. In particular, it guarantees a purely combinatorial derivation of inverse-scope readings for non-commutative effects like $\boxed{\mathtt{T}\,\alpha}$ ::= $\mathtt{s} \to \alpha \times \mathtt{s}$ (introduced in Section 3.5).

The monad operators for $\mathtt{T}$ are not much different than the applicative ones. Given a computation $m$ of type $\boxed{\mathtt{T}\,\alpha}$ and a function $k$ of type $\alpha \to \boxed{\mathtt{T}\,\beta}$, the bind operation creates a composite computation that sequences an incoming state $i$ through $m$, and then passes the resulting value and updated state into $k$. The join operation is defined similarly, threading the output of the outer computation in as input to the inner one.

(4.41) $m \gg\!= k \coloneqq \lambda i.\, k\,x\,j$ **where** $\langle x, j\rangle = m\,i$

$\mu M \coloneqq \lambda i.\, m\,j$ **where** $\langle m, j\rangle = M\,i$

Exactly the same sequence of algebraic substitutions as in (4.40) show that (4.42a) is denotationally equivalent to (4.42b).

(4.42a)

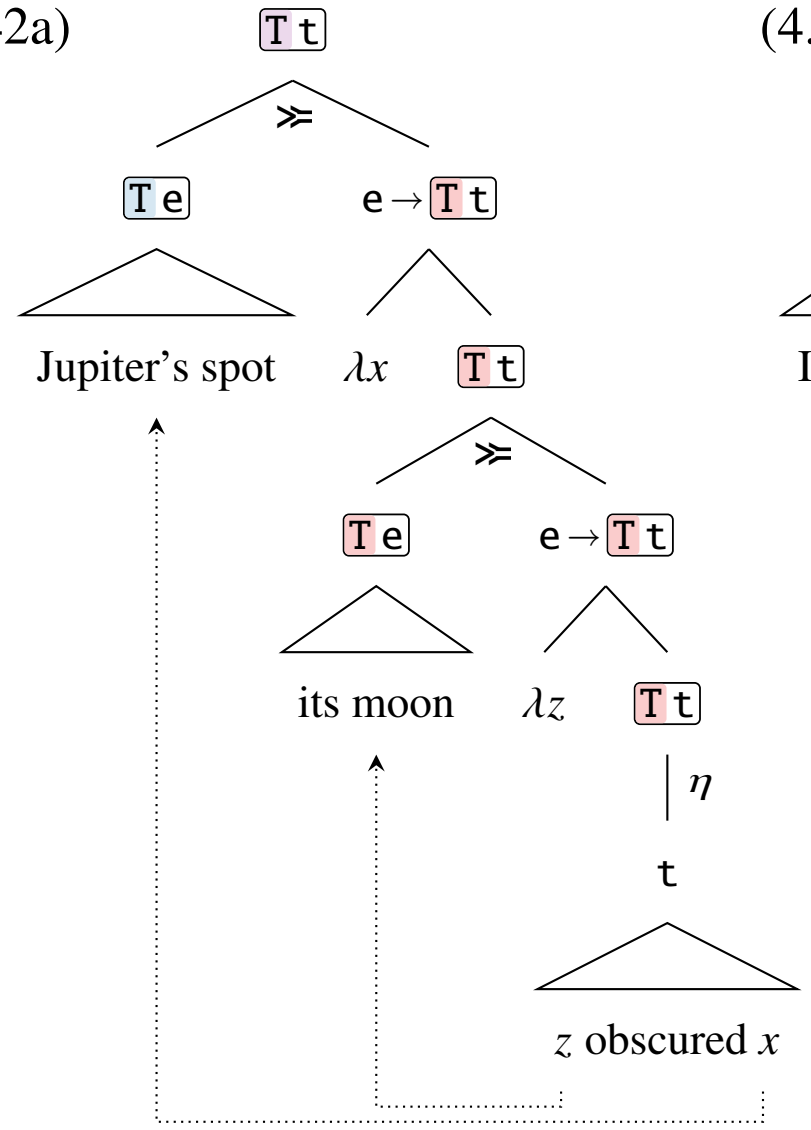

(4.42b)

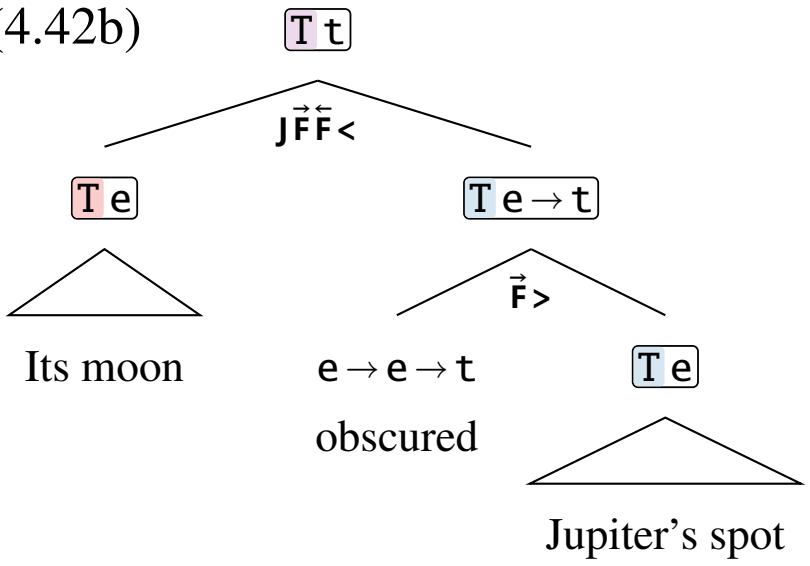

The meaning computed by these derivations is given in (4.42c). Clearly, the effect order has been inverted in a non-trivial way, since the object now binds into the subject!

(4.42c) $[\![\text{Jupiter's spot}]\!] \gg\!= \lambda x.\, ([\![\text{its moon}]\!] \gg\!= \lambda z.\, \eta\, (\mathbf{obs}\, x\, z))$

$= (\lambda i.\, \langle \mathbf{spot\,j}, \mathbf{j} +\!\!+\, i\rangle) \gg\!= \lambda x.\, ((\lambda i.\, \langle \mathbf{moon}\, i_0, i\rangle) \gg\!= \lambda z.\, (\lambda i.\, \langle \mathbf{obs}\, x\, z, i\rangle))$

$= (\lambda i.\, \langle \mathbf{spot\,j}, \mathbf{j} +\!\!+\, i\rangle) \gg\!= \lambda x.\, (\lambda i.\, \langle \mathbf{obs}\, x\, (\mathbf{moon}\, i_0), i\rangle)$

$= \lambda i.\, \langle \mathbf{obs}\, (\mathbf{spot\,j})\, (\mathbf{moon}\, (\mathbf{j} +\!\!+\, i)_0), \mathbf{j} +\!\!+\, i\rangle$

$= \lambda i.\, \langle \mathbf{obs}\, (\mathbf{spot\,j})\, (\mathbf{moon\,j}), \mathbf{j} +\!\!+\, i\rangle$

We discuss an alternative approach to dynamic binding in Section 5.1, which precludes this sort of cataphora in a natural way. For now, let us just make a few remarks. First, T itself is still a quintessentially *dynamic* monad. By design, it shuttles referents asymmetrically from computations to continuations, and the default applicative modes of combination **A>** and **A<** ensure that this happens from left to right regardless of argument structure. A paradigmatic lexical entry for dynamic conjunction might easily be defined in terms of these applicative operations, as in (4.43).[12]

(4.43) $[\![\text{and}]\!] = \lambda R \lambda L.$

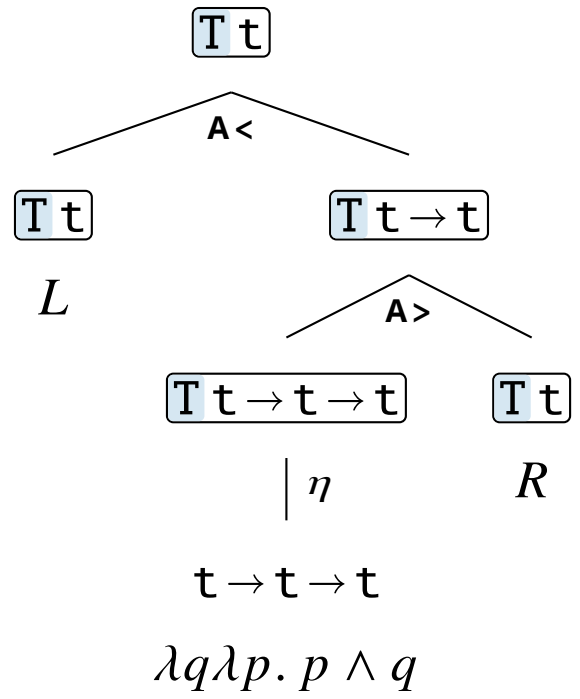

$$= \lambda R \lambda L \lambda i.\ \langle p \wedge q, k \rangle \ \textbf{where}\ \langle p, j \rangle = L\,i$$
$$\langle q, k \rangle = R\,j$$

So the backwards binding on display in (4.42) has a very different source than the binding that would result from the proposal sketched at the end of Section 3.5. There, we noted that constructors like T have secondary, non-canonical applicative instances in which they are effectively re-conceived as computations that by their nature run in reverse, from right to left. Here, it's just orderly Ts all the way down. Instead, the grammatical mechanism for effect inversion emerges organically from T's monadic structure.

As is to be expected at this point, a completely analogous story can be told about quantificational scope in C. As characterized in Chapter 2, Generalized Quantifiers themselves constitute a kind of computation. Any expression of type $(\alpha \to t) \to t$ can appear locally in a position where an ordinary $\alpha$ is expected. But of course the quantifier does not merely contribute a value to that position. Rather, it tests what happens when the rest of the derivation, its **continuation**, is run with different values of type α, and then makes a summary decision based on the results of these experiments. `C` is monadic (and, hence, functorial and applicative; see Section 4.3.2 for further discussion); its $\gg\!=$ and $\mu$ are as follows:

(4.44) $$m \gg\!= k := \lambda c.\, m\,(\lambda x.\, k\,x\,c)$$
$$\mu M := \lambda c.\, M\,(\lambda m.\, m\,c)$$

[12]Though, to be clear, given the applicative modes of combination, there is no good reason to actually define conjunction like this. The grammar will find the derivation inside this lexical entry all by itself, if $[\![\text{and}]\!]$ is just $(\wedge)$. The point is just to show that this effect structure really does instantiate typical notions of dynamic semantics.

Two derivations of the scopally ambiguous 'someone admires everyone' are given below. In (4.45a), **A<** glues the subject and VP together with the attendant linear bias, resulting in a meaning where the quantifiers scope in their surface order. In (4.45b), $\vec{\mathbf{F}}\,\overleftarrow{\mathbf{F}}$**<** first builds a higher-order `C C t` meaning with the object's effects out-scoping the subject's: $\lambda c.\, \forall y.\, c\,(\lambda c'.\, \exists x.\, c'\,(\textbf{admires}\,y\,x))$. Then **J** flattens this higher-order meaning by passing the outer continuation in to the inner layer, resulting in inverse scope. As before, the monad laws guarantee that these *in situ* derivations are equivalent to more familiar transformational ones differing only in how the `C e` quantifiers are raised. (The `C t` meanings that result are not the usual type-`t` ones, but they aren't so far off either. We return to this in Chapter 5.)

(4.45a) (4.45b)

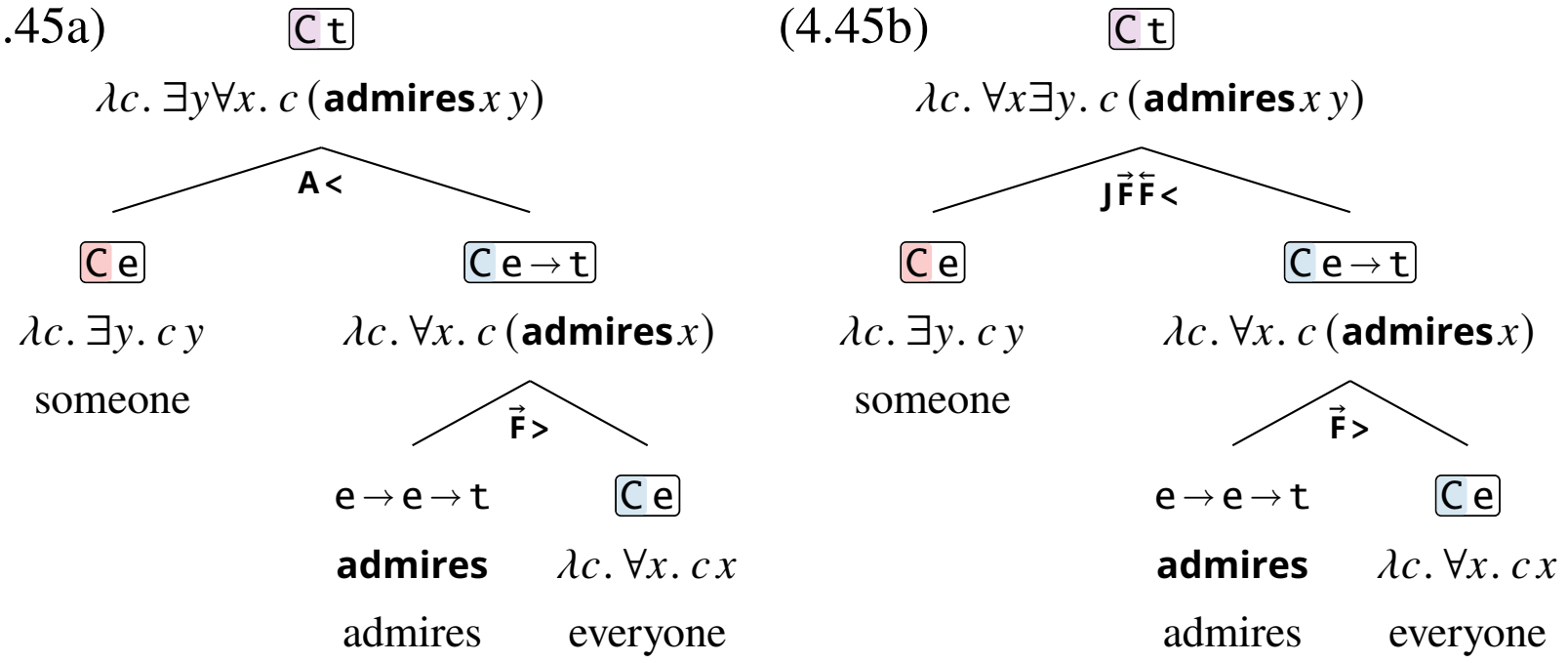

One reaction to these observations is that, as far as generative power goes, monadic effects are only as asymmetric as the syntax-semantics interface they're embedded in. If that interface allows mapping constituents on the right over constituents on the left and joining the results, then it will license full permutations of effect orders. It should be noted, though, that the inverse derivation is necessarily more complex than the surface derivation: **A**$*$ vs. **J**$\,(\vec{\mathbf{F}}\,(\overleftarrow{\mathbf{F}}\,*))$. So even if such inverse readings are grammatical, there is still a place to put a finger on how the underlying non-commutativity might manifest itself empirically. Alternatively, if the grammar really should rigidly enforce that underlying non-commutativity, there is a simple way to do it: just don't allow $\mathbf{J}\,(\vec{\mathbf{F}}\,(\overleftarrow{\mathbf{F}}\,*))$ modes of combination for that effect. This can be decided completely locally as part of the algorithm that determines what modes of combination are possible for two constituents, as we show when we discuss islands in Section 5.4.

Importantly, these *ex situ* and *in situ* equivalences extend to cases of so-called **exceptional scope**. Repeating the process in (4.27)–(4.35), it's easy to see that the derivation in (4.46a) is denotationally identical to the island-violating derivation in (4.46b). This remarkable equivalence opens the door to a uniform treatment of islands as barriers to scope, eliminating the need to make caveats for expressions like indefinites whose semantic force is felt well beyond those boundaries. That is, we are free to deem (4.46b) ungrammatical, based on whatever general principles regulate scope and movement. Such a restriction is toothless in the case of indefinites (and other expressions whose content resides in their effect structure) because the same denotation is *guaranteed* to arise from an island-respecting derivation as well.

(4.46a)

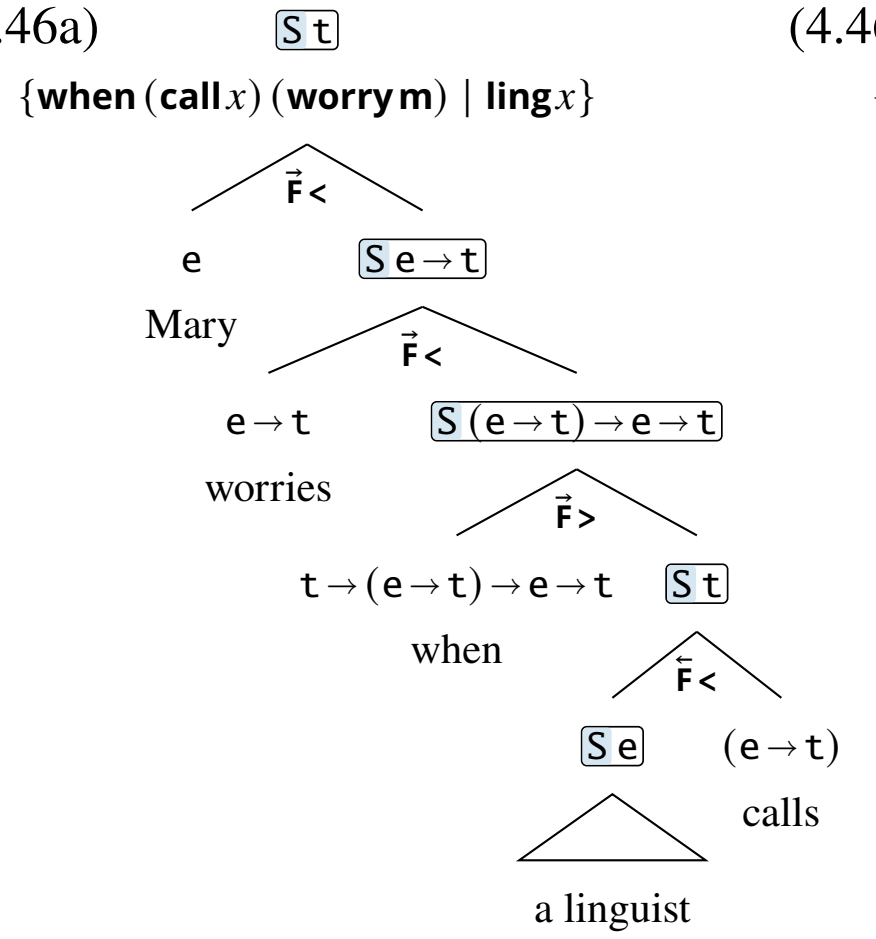

(4.46b)

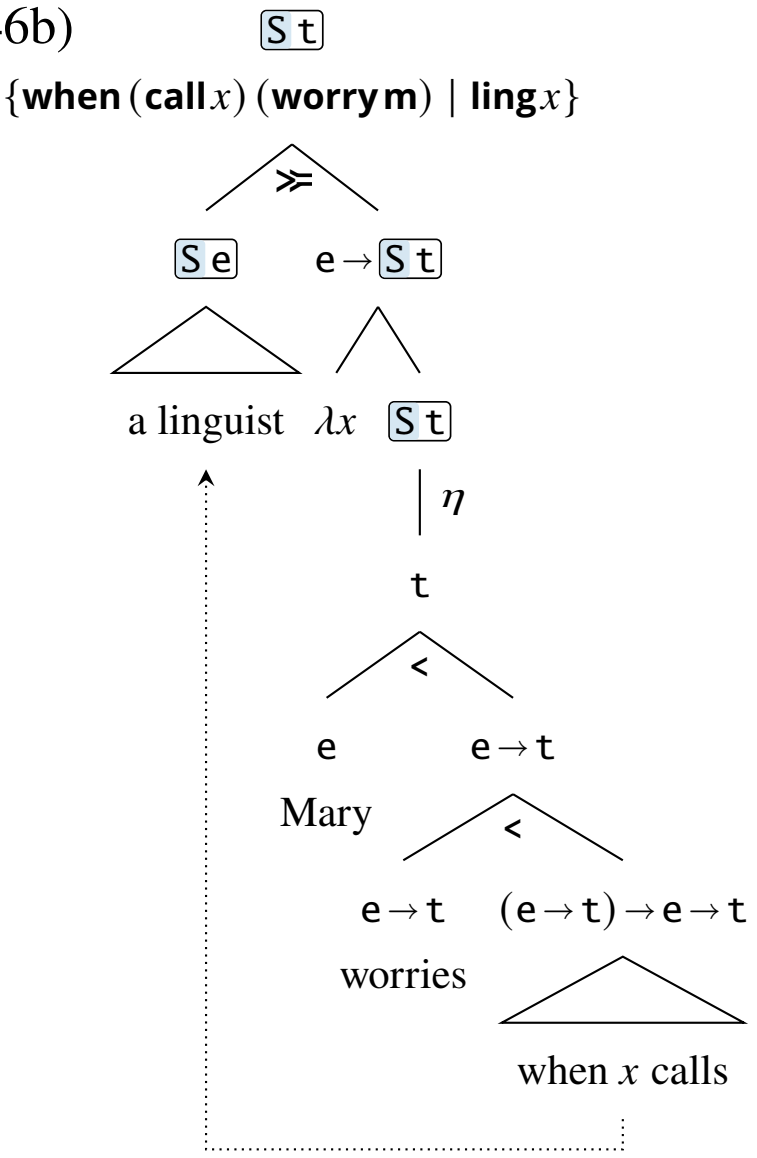

In fact, the exceptional scope of monadic effects can be seen directly in the **Associativity Law** of (4.16). This law is repeated in three different syntactic formats in (4.47). Imagine the constituent denoting $n$ is an island. The derivations on the right-hand sides of (4.47) are island-violating. The $m$ constituent is raised out of the $n$-island to scope over a larger chunk of the sentence, including $o$. But when $\Sigma$ is a monad, these derivations are equivalent to those on the left-hand sides, which are island-respecting. The effectful constituent $m$ takes scope over its enclosing island $n$, but nothing else. It is raised to the specifier of $n$, if you like. From there, the entire island is **pied-piped** to scope over the remainder of expression $o$. But the resulting meaning will be exactly as if $m$ had moved all the way to the top alone. For more on this point, see Charlow 2020.

(4.47a) $(m \mathbin{>\!\!>\!\!=} \lambda x.\, n\, x) \mathbin{>\!\!>\!\!=} \lambda y.\, o\, y \quad = \quad m \mathbin{>\!\!>\!\!=} (\lambda x.\, n\, x \mathbin{>\!\!>\!\!=} \lambda y.\, o\, y)$

(4.47b)

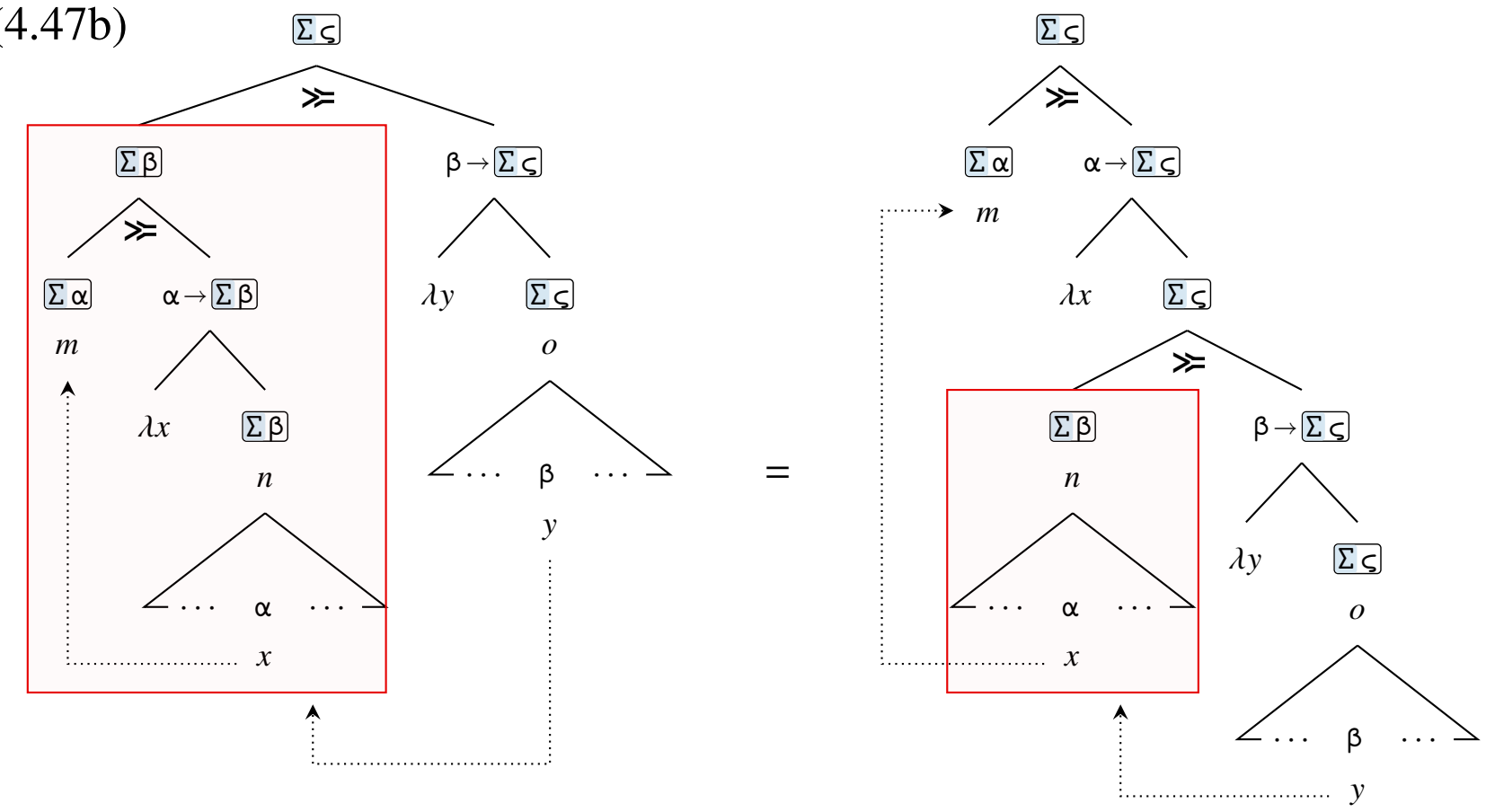

(4.47c)

```
do y <- { do x <- m
             n x      }
   o y
```

=

```
do x <- m
   y <- n x
   o y
```

To recap this section, every monadic computation can be expressed in a `do`-block, structured as a sequence of right-nested actions, concluding with the calculation of an ordinary value. The resemblance to Quantifier Raising and traditional linguistic LFs is uncanny. It is natural then to wonder what monads have to do with Generalized Quantifiers. Following this thread exposes yet a third effect-driven derivational strategy.

### *4.3.2 Transformation into continuations*

To bring out the connection between monads and quantification, let us treat ($\gg\!=$) as a "type-shifting" unary operator, like $\eta$. In this form, it shifts an expression of type $\boxed{\Sigma\,\alpha}$ to one of type $(\alpha \to \boxed{\Sigma\,\beta}) \to \boxed{\Sigma\,\beta}$, as in (4.48).

(4.48)

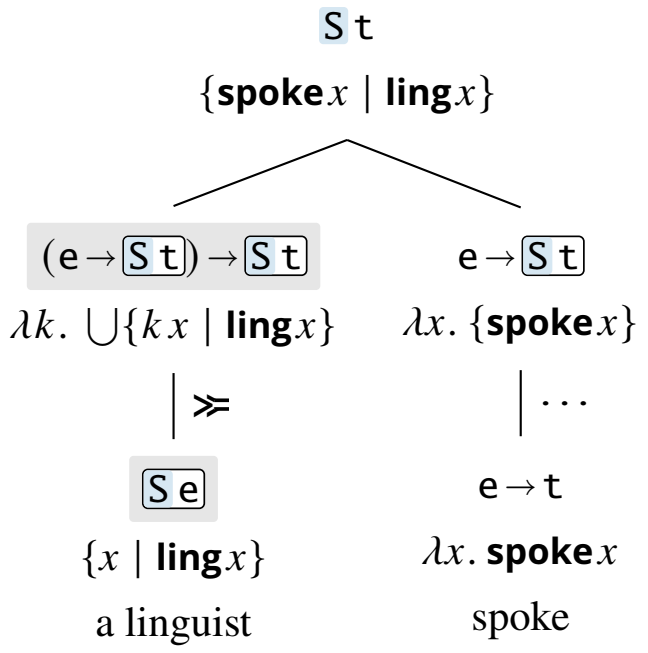

(4.49)

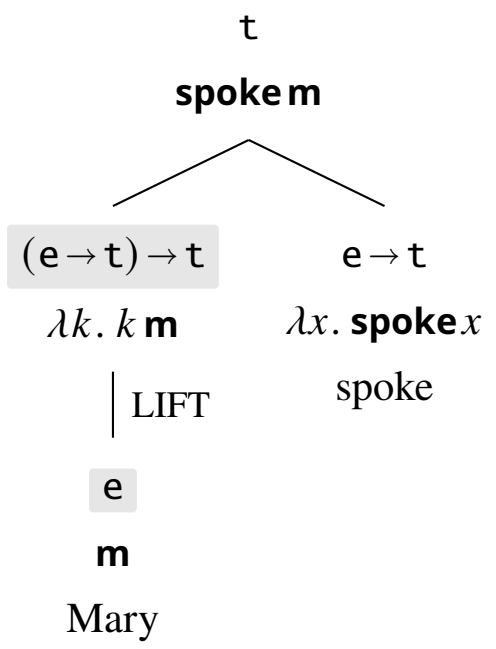

In this guise, ($\gg\!=$) plays a role much like the traditional LIFT operator of Partee (1986). Where LIFT converts an ordinary value of type e into a Generalized Quantifier over properties of individuals, ($\gg\!=$) converts an enriched value of type $\boxed{\Sigma\,e}$ into an enriched quantifier over properties that may have side effects.

In fact, the first monad law ensures that when a computation is trivial (just a value injected into some structure), then ($\gg\!=$) is exactly equivalent to LIFT.

(4.50) **Left Identity:** $\eta x \gg\!= k = k\,x$

This is seen by rewriting the law in (4.50) using the unary version of ($\gg\!=$).

(4.51) **Left Identity:** $(\eta x)^{\gg=} = \lambda k.\, k\,x$
$= \textsc{lift}\, x$

In tree form, this says the following two derivations must be equivalent:

(4.52)

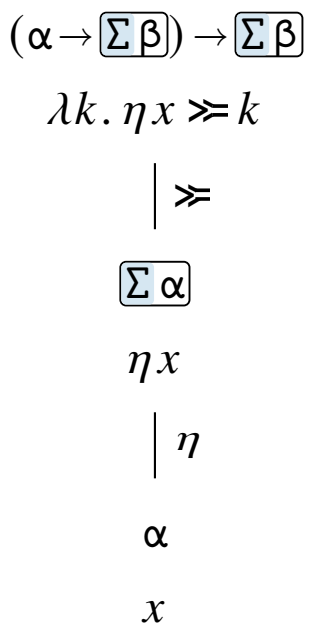

(4.53)

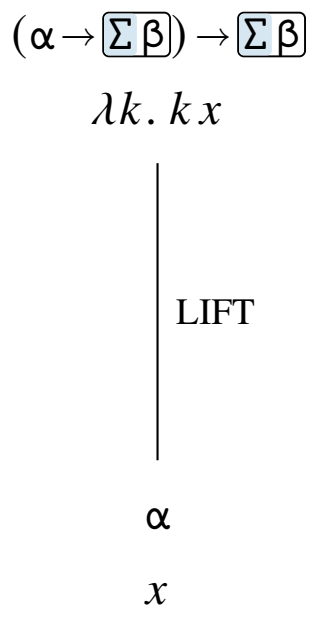

And here's where things get interesting. Recall that Generalized Quantification is a computational effect. Things of type $\boxed{\texttt{C}\,\alpha} = (\alpha \to \texttt{t}) \to \texttt{t}$ appear wherever ordinary $\alpha$'s are expected, and $\texttt{C}$ embodies how a continuation argument of type $\alpha \to \texttt{t}$, representing the quantifier's scope, can be built and passed to such expressions. In Section 4.3.1, we saw how an effect-driven interpreter discovers *in situ* surface and inverse scope derivations for sentences like 'someone admired everyone', underwritten by $\texttt{C}$ being a monadic (and applicative) functor. Here are the functorial and applicative instances for continuized computations.

(4.54a) $k \bullet m := \lambda c.\, m\,(\lambda x.\, c\,(k\,x))$

(4.54b) $F \circledast X := \lambda c.\, F\,(\lambda f.\, X\,(\lambda x.\, c\,(f\,x)))$

From an algebraic perspective, there is nothing special about computations that expect their derivational contexts to compute *truth values*, per se; in other words, nothing special about the type $\texttt{t}$. Much more generally, any function of type $(\alpha \to \texttt{o}) \to \rho$ can be construed as a computation that runs by swallowing some $\texttt{o}$-sized chunk of its context. With this context it tries out different $\alpha$ values to see what $\texttt{o}$ results they lead to. With the collection of these results in hand, it makes a decision on which $\rho$ to return. When $\texttt{o}$ and $\rho$ are $\texttt{t}$, it (often) makes sense to think of these computations as "quantifications", in that the boolean that is returned may depend only on *how many* type-$\alpha$ values tipped the text toward truth. When $\texttt{o}$ and $\rho$ are other types, they may not do anything that we would associate with quantities, but they are still functions of their contexts.

To the point, any higher-order function of type $(\alpha \to \boxed{\Sigma\,\beta}) \to \boxed{\Sigma\,\beta}$ fits this pattern. Such a function may well be construed as a computation purporting to be an $\alpha$, but in reality expecting to see what *computations* $\boxed{\Sigma\,\beta}$ result from filling in its local position with various type-$\alpha$ choices. And every unary application of $(\gg\!=)$ creates a higher-order function of exactly this type.

In light of this, we may choose to conceive of $(\gg\!=)$ not as a mode of combination, but as a way of transforming one kind of computation into another. Formally, this amounts to assigning $(\gg\!=)$ the type in (4.55), where $\texttt{C}$ is understood to be the type of computations that depend on their continuations, in this case continuations of type $(\alpha \to \boxed{\Sigma\,\beta})$.

(4.55) $(\gg\!=) :: \boxed{\Sigma\,\alpha} \to \boxed{\texttt{C}\,\alpha}$

Polymorphic functions like this, that convert values in one functor to values in another functor, are known as **natural transformations**.

Look at what happens when an arbitrary monadic computation $m :: \boxed{\Sigma\,\alpha}$ is so transformed. The resulting computation $m^{\gg=} :: \boxed{\texttt{C}\,\alpha}$ will have functorial and applicative combinators in accordance with the $\texttt{C}$ effect defined in (4.54):

(4.56a) $k \bullet_{\mathsf{C}} m^{\ggcurly} = \lambda c.\, m^{\ggcurly}(\lambda x.\, c\,(k\,x))$

(4.56b) $F^{\ggcurly} \circledast_{\mathsf{C}} X^{\ggcurly} = \lambda c.\, F^{\ggcurly}(\lambda f.\, X^{\ggcurly}(\lambda x.\, c\,(f\,x)))$

Strikingly, these continuized operations are almost identical to those in (4.19), repeated below in (4.57), which are themselves induced directly from the underlying monad $\Sigma$.

(4.57a) $k \bullet_{\Sigma} m = m \ggcurly \lambda x.\, \eta\,(k\,x)$

(4.57b) $F \circledast_{\Sigma} X = F \ggcurly \lambda f.\, X \ggcurly \lambda x.\, \eta\,(f\,x)$

The only difference is that where the original definitions in (4.57) "return" the underlying results with $\eta$, the continuized definitions in (4.56) leave open what will happen to these results by abstracting over the role of $\eta$ with the continuation variable $c$. Naturally, passing $\eta$ in for this function yields the original definitions. That is, for any monad $\Sigma$, we have the equivalences in (4.58). These equations are made slightly more telegraphic by writing $(\cdot)^{\ggcurly}$ as $(\cdot)^{\uparrow}$ and $(\cdot)\,\eta$ as $(\cdot)^{\downarrow}$, to indicate lifting into and lowering out of the $\mathsf{C}$ effect space.

(4.58a) $k \bullet_{\Sigma} m = (k \bullet_{\mathsf{C}} m^{\ggcurly})\,\eta$
$= (k \bullet_{\mathsf{C}} m^{\uparrow})^{\downarrow}$

(4.58b) $F \circledast_{\Sigma} X = (F^{\ggcurly} \circledast_{\mathsf{C}} X^{\ggcurly})\,\eta$
$= (F^{\uparrow} \circledast_{\mathsf{C}} X^{\uparrow})^{\downarrow}$

This means that in principle, every single instance of ($\bullet$) and ($\circledast$), or the corresponding $\vec{\mathbf{F}}/\overleftarrow{\mathbf{F}}$ and **A** meta-combinators, can be simulated with instances of ($\bullet_{\mathsf{C}}$) and ($\circledast_{\mathsf{C}}$), provided access to free applications of the $(\cdot)^{\uparrow}$ and $(\cdot)^{\downarrow}$ operators above. In the framework laid out here, eliminating all effect-specific instances of ($\bullet$) and ($\circledast$) is as simple as restricting the $\vec{\mathbf{F}}$, $\overleftarrow{\mathbf{F}}$, and **A** rules to apply only to $\mathsf{C}$-type computations, rather than arbitrary $\Sigma$ computations. The relevant components of such a grammar are shown in Figure 9.

Going further, we might even replace the free *coercions* of $(\cdot)^{\downarrow}$ and $(\cdot)^{\uparrow}$ with completely free *occurrences* of $\eta$ and $\ggcurly$ in a derivation. That is, we might suppose that not only can $\ggcurly$ be freely applied to an expression (via $(\cdot)^{\uparrow}$), but an expression can be freely applied to $\ggcurly$ as well, and vice versa for $\eta$. This, in effect, lexicalizes the monadic combinators as object-language operators, much as Jacobson's (1999) **g** lexicalizes ($\bullet$) (see Section 2.2.1). Going this route immediately renders the Unit ($\overleftarrow{\mathbf{U}}/\vec{\mathbf{U}}$) and Join (**J**) rules otiose, since they are thin wrappers for applications of $\eta$ and $\ggcurly$, respectively. In fact, even the basic mapping operations ($\bullet$) and their higher-order variants ($\vec{\mathbf{F}}$ and $\overleftarrow{\mathbf{F}}$) become

**Types:**

$\vdots$

**Combinators:**

$\vdots$

**Type-shifters:**

$(\cdot)^{\uparrow} :: (\boxed{\Sigma\,\alpha}) \to \boxed{\mathsf{C}\,\alpha}$ — Up

$m^{\uparrow} \coloneqq \lambda k.\, m \ggg k$

$(\cdot)^{\downarrow} :: (\boxed{\mathsf{C}\,\alpha}) \to \boxed{\Sigma\,\alpha}$ — Down

$m^{\downarrow} \coloneqq m\,\eta$

**Meta-combinators:**

$\overleftarrow{\mathbf{F}} :: (\sigma \to \tau \to \omega) \to \boxed{\mathsf{C}\,\sigma} \to \tau \to \boxed{\mathsf{C}\,\omega}$ — Map Left

$\overleftarrow{\mathbf{F}}\,(*)\,E_1\,E_2 \coloneqq (\lambda a.\, a * E_2) \bullet E_1$

$\overrightarrow{\mathbf{F}} :: (\sigma \to \tau \to \omega) \to \sigma \to \boxed{\mathsf{C}\,\tau} \to \boxed{\mathsf{C}\,\omega}$ — Map Right

$\overrightarrow{\mathbf{F}}\,(*)\,E_1\,E_2 \coloneqq (\lambda b.\, E_1 * b) \bullet E_2$

$\mathbf{A} :: (\sigma \to \tau \to \omega) \to \boxed{\mathsf{C}\,\sigma} \to \boxed{\mathsf{C}\,\tau} \to \boxed{\mathsf{C}\,\omega}$ — Structured App

$\mathbf{A}\,(*)\,E_1\,E_2 \coloneqq (\lambda a \lambda b.\, a * b) \bullet E_1 \circledast E_2$

$\vdots$

**Figure 9** A monadic grammar whose binary rules use continuations

superfluous. Any derivation involving these modes of combination is equivalent to a derivation using just $\gg\!=$ and $\eta$, together with $\mathbf{A}_{\mathsf{C}}$.

For instance, the two derivations in (4.59) generate the same meaning, as may be seen by repeated applications of the monad laws (4.16) and the equivalences in (4.58). Note that in (4.59a), unary applications of $\eta$ and $\gg\!=$ are written as type-shifters, but this is just to save space. Since they both sometimes occur as leaves in the tree, they might just as well always occur as leaves in the tree.

(4.59a)

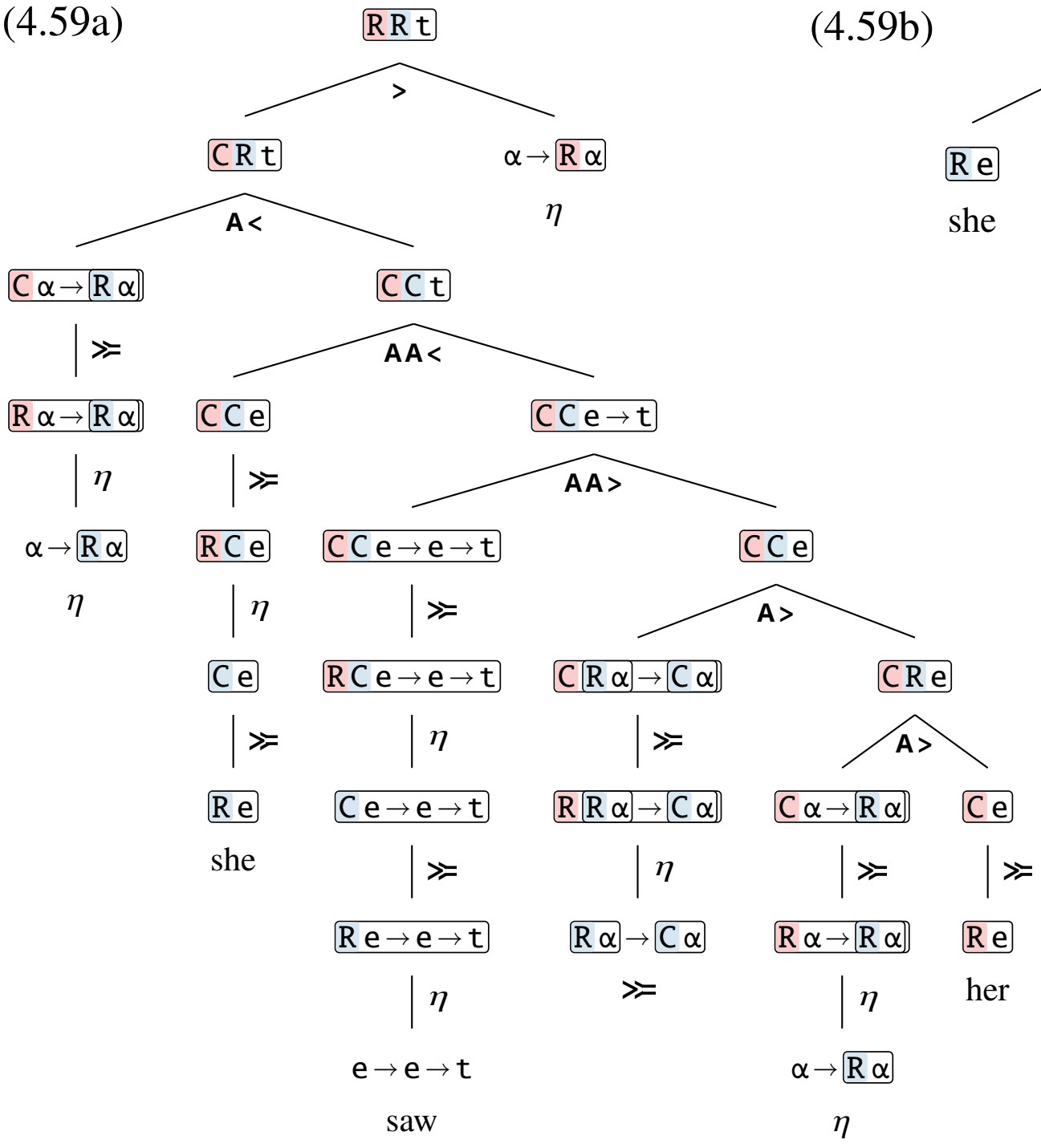

(4.59b)

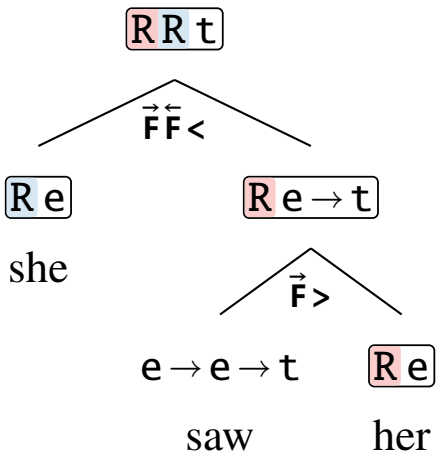

For an at-length presentation of compositional semantics in this style, see Charlow 2014. The resulting grammar is stunningly compact: two (overloaded) lexical operators $\eta$ and $\gg\!=$, and one (parametric) higher-order combinator $\mathbf{A}_{\mathsf{C}}$, together with whatever basic modes of combination (application, modification, restriction, etc.) are desired.[13]

[13] It might seem disingenuous to say the grammar contains just one $\eta$, one $\gg\!=$, and one $\mathbf{A}_{\mathsf{C}}$, since all of these operators are polymorphic. The first two in particular can mean dramatically different things in different contexts. While it's certainly true that the overloading of the symbols hides a lot of semantic complexity, that's, well, that's sort of the point. As far as the syntax-semantics

As usual, this formal economy comes with trade-offs. On the plus side, it is theoretically parsimonious. It isolates all of the *grammatical* action in a very small inventory of operations, arguably just $\mathbf{A}_{\mathrm{C}}$, and leaves the rest to the lexicon, including the language- and effect-specific instances of $\eta$ and $\gg\!=$. On the down side, it requires a bit of expertise, intuition, and practice to actually use. One has to find derivations oneself, by inserting the silent operators where appropriate and/or iteratively applying them to one another. And in principle, infinitely many choices for how to insert the operators will be equivalent, as guaranteed by the various algebraic laws, leading to runaway spurious ambiguity.

In any case, we stick to the type-driven approach in what follows. The guarantees for practical usability are the same as they have been all along: no silent operators or type-shifters to distribute, no covert transformations to sort out, a finite (generally tiny) inventory of modes of combination, and an elementary algorithm to enumerate possible binary combinations. Composition for dummies.

## 4.4 Implementing monadic effects in the type-driven interpreter

Extending the interpreter with a monadic mode of combination requires the same, now familiar modifications as in the last two chapters. We first add a mode `JN` representing the meta-mode **J**.

```
data Mode
  = FA | BA | PM          -- etc
  | MR Mode | ML Mode   -- map right and map left
  | AP Mode             -- structured app
  | UR Mode | UL Mode   -- unit right and unit left
  | JN Mode             -- join
```

To determine the applicability of the `JN` rule, we'll need to declare which effects are monadic. Again, the only constructor to watch out for is `W`, which is only monadic when it is applicative (i.e., when its parameter is a monoid). Other than that, all of the effects under consideration in this Element are monads, so there is nothing else to check.

interface goes, that complexity *should* be hidden. Exactly which instance of $\gg\!=$ is intended in a particular derivation will be completely determined by the type of the expression it is applied to. If the first letter is `S`, then it will be $\gg\!=_{\mathrm{S}}$, etc. All the theory of composition needs to know is that the language's grammar defines a bind operation for `S`, and more generally, for each of its lexical effects `Σ`.

```
functor, applicative, monad :: EffX -> Bool
functor      _       = True
applicative (WX o) = monoid o
applicative f      = functor f && True
monad f            = applicative f && True

monoid :: Ty -> Bool
monoid T = True
monoid _ = False
```

However, the logic determining when to dispatch `JN` must be slightly different than that of the other meta-rules. To see this, compare the types of the $\vec{\mathsf{F}}$ and $\mathsf{J}$ modes, repeated below.

$$\text{(4.60)} \quad \vec{\mathsf{F}}\,(* :: \sigma \to \tau \to \upsilon) :: \sigma \to \boxed{\Sigma\,\tau} \to \boxed{\Sigma\,\upsilon}$$

$$\text{(4.61)} \quad \mathsf{J}\,(* :: \sigma \to \tau \to \boxed{\Sigma\,\Sigma\,\upsilon}) :: \sigma \to \tau \to \boxed{\Sigma\,\upsilon}$$

Provided with a function $(*) :: \sigma \to \tau \to \upsilon$, the $\vec{\mathsf{F}}$ rule returns a mode of combination of type $\sigma \to \boxed{\Sigma\,\tau} \to \boxed{\Sigma\,\upsilon}$. This means that $\vec{\mathsf{F}}\,(*)$ is only ever applicable when the right daughter is of type $\boxed{\Sigma\,\tau}$. Hence the logic of `addMR`:

```
addMR l r = case r of
  Comp f t | functor f
    -> [ (MR op, Comp f u) | (op, u) <- combine l t ]
  _ -> [                                            ]
```

This rule only fires when the right daughter's type has a particular shape: `Comp f t`. Otherwise it immediately returns an empty list. And when that daughter does in fact denote a computation, `combine` is called recursively on the *underlying* type `t`. Because a type is finite, this strategy is guaranteed to terminate. At every step it strips off some effect wrapper and tries again with what is left.

But the $\mathsf{J}$ meta-combinator is different. Provided with a function $(*) :: \sigma \to \tau \to \boxed{\Sigma\,\Sigma\,\upsilon}$, it returns a mode of combination of type $\sigma \to \tau \to \boxed{\Sigma\,\upsilon}$. This means, in principle, that it could apply to *any* two types $\sigma$ and $\tau$, if there happens to be a way to combine them to yield something of type $\boxed{\Sigma\,\Sigma\,\upsilon}$. There is thus no way to know simply by looking at the shapes of the daughters' types `l` and `r` whether the $\mathsf{J}$ mode might apply.

Instead, we need to try and combine the daughters first, and then inspect the *results* to see if we ended up with anything joinable. Thus we split `combine` into two parts. The `binaryCombs` list holds all of the results computed in the preceding chapters; that is, all the up-front binary combinations we can find. Then the `unaryCombs` function sets to work on each result. At this point there are only two things to do with any such result. First, keep it! A good combination is still a good combination. Second, if its type happens to be `Comp f (Comp g a)`, where `f` and `g` are the same monadic effect, then additionally go ahead and join it. Again, these are not exclusive. We hold on to both the layered and the lowered combinations, since they are both valid.

```
combine :: Ty -> Ty -> [(Mode, Ty)]
combine l r = binaryCombs >>= unaryCombs
  where
    binaryCombs =
      modes l r
      ++ addMR l r ++ addML l r
      ++ addAP l r
      ++ addUR l r ++ addUL l r
    unaryCombs e =
      -- keep any result from above
      return e
      -- and if it happens to have a two-layered
      -- monadic type, also join it
      ++ addJN e

addJN e = case e of
  (op, Comp f (Comp g a)) | f == g, monad f
    -> [ (JN op, Comp f a) ]
  _ -> [                   ]
```

Note that `binaryCombs` is a list of modes. And `unaryCombs` is a function from modes to an extended list of modes. Fittingly, the way to take each element of the former, pass it in to the latter, and then flatten out all the results, is just the `>>=` operation on lists. In programs, as in language, this pattern just has a way of showing up.

Once again, we demonstrate the extended `combine` function by training it on a couple key type combinations considered in this chapter. In the first example, with inputs corresponding to S e and a Kleisli arrow e → S t, `combine` finds two derivations, one which maps the Kleisli arrow over S e, yielding a higher-order S S t, and another which subsequently applies **J**, flattening the higher-order meaning to S t. In the second example, with inputs corresponding to C e and C e → t, `combine` finds two higher-order derivations and their flattened counterparts (deriving, respectively, inverse and surface scope), along with a comparatively straightforward applicative derivation resulting in surface scope.

```
ghci> combine (Comp SX E) (E :-> Comp SX (E :-> T))
[(ML BA, Comp SX (Comp SX T)),
 (JN (ML BA), Comp SX T)]
```

```
ghci> combine (Comp (CX T) E) (Comp (CX T) (E :-> T))
[(MR (ML BA), Comp (CX T) (Comp (CX T) T)),
 (JN (MR (ML BA)), Comp (CX T) T),
 (ML (MR BA), Comp (CX T) (Comp (CX T) T)),
 (JN (ML (MR BA)), Comp (CX T) T),
 (AP BA, Comp (CX T) T)]
```

# 5 Eliminating effects

## 5.1 Adjunctions

### *5.1.1 Introducing adjunctions*

Let us return to the issue of anaphora. We would like to analyze a pronoun as a computation that retrieves a salient discourse referent from memory. In its simplest, variable-free form, something like (5.1a) should do. And we would like to analyze its antecedent as a computation that stores a discourse referent in memory. Again with maximal simplification, something like the operator in (5.1b) should suffice to push an antecedent into a secure, second dimension.

(5.1a) it :: $\boxed{\mathtt{R}\,\mathtt{e}}$

$[\![\text{it}]\!] := \lambda x.\, x$

(5.1b) $\triangleright :: \mathtt{e} \to \boxed{\mathtt{W}\,\mathtt{e}}$

$[\![\triangleright]\!] := \lambda x.\, \langle x, x\rangle$

In this construal, the pronoun and its antecedent constitute semantically and type-theoretically distinct effects. This alone is no barrier to composition, given any of the type-driven grammars in the preceding chapters. But in all of those grammars, the result of putting together a sentence with both an antecedent and a pronoun will be a computation with two effects. The antecedent will survive in memory, and the pronoun will continue to await its resolution, as in (5.2).

(5.2)

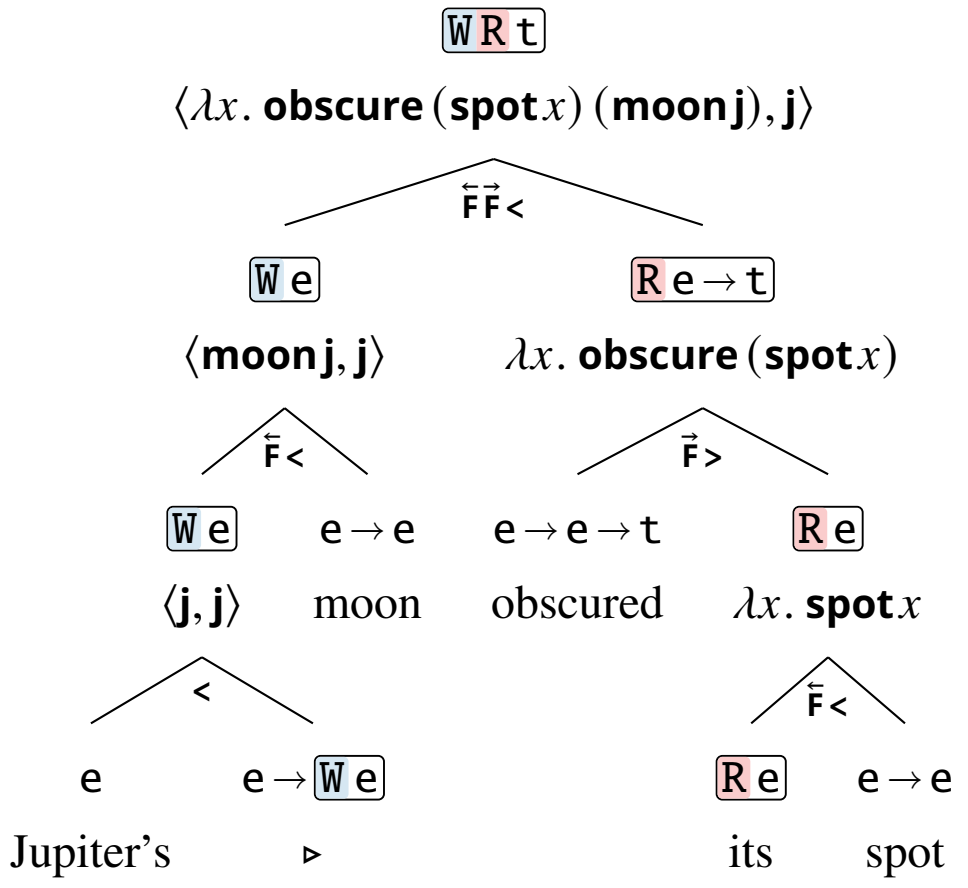

In other words, the composition of $\mathtt{W}$ and $\mathtt{R}$ qua functors gives us writing, and reading, but we don't yet have **binding**. In Chapters 3 and 4, we sketched solutions to this that are typical of dynamic approaches to semantics. These analyses create a new, generalized effect that models any sort of interaction with a discourse state. In general, a sentence with unresolved pronouns and fresh new antecedents will denote a state transition that both reads from and modifies the state. But on their own, pronouns will comprise the special case of programs that read from but do not modify the state, and antecedents the special case of programs that modify but do not inspect the state. In this sense, the dynamic solution is a generalization to the worst case.

Here, we offer a different solution, inspired by Shan (2001b). The essential intuition is that a discourse in which all pronouns are bound should denote an ordinary proposition, a pure value. Once the antecedent has supplied its referent to the pronoun, both the writing and the reading effects should be considered resolved. The two effects are closure operators to one another.

This duality is mathematically manifest in the isomorphism defined by the following functions.

(5.3) $\Phi :: (\boxed{\mathtt{W}\,\alpha} \to \beta) \to \alpha \to \boxed{\mathtt{R}\,\beta}$ $\quad\quad \Psi :: (\alpha \to \boxed{\mathtt{R}\,\beta}) \to \boxed{\mathtt{W}\,\alpha} \to \beta$

$\Phi := \lambda c \lambda a \lambda x.\, c\,\langle a, x\rangle$ $\quad\quad \Psi := \lambda k \lambda \langle a, x\rangle.\, k\, a\, x$

Any co-Kleisli arrow from $\mathtt{W}$ can be converted to a Kleisli arrow into $\mathtt{R}$, and any Kleisli arrow into $\mathtt{R}$ can be converted to a co-Kleisli arrow from $\mathtt{W}$. These transformations proceed without loss of information. They are inverses: $\Phi\,(\Psi\, k) = k$ for any $k :: \alpha \to \boxed{\mathtt{R}\,\beta}$, and $\Psi\,(\Phi\, c) = c$ for any $c :: \boxed{\mathtt{W}\,\alpha} \to \beta$. Setting aside the type constructors, this isomorphism is just the familiar equivalence between curried and uncurried presentations of multi-argument functions.

Whenever two functors $\Omega$ and $\Gamma$ have this dual property, they are said to be **adjoint**. Specifically, $\Omega$ is **left adjoint** to $\Gamma$, written $\Omega \dashv \Gamma$, when functions *from* $\Omega$ are isomorphic to functions *into* $\Gamma$. In the case at hand, we say: $\mathtt{W} \dashv \mathtt{R}$.

Every adjunction $\Omega \dashv \Gamma$ gives rise to a pair of functions, $\eta$ and $\varepsilon$, by applying the components of this isomorphism to identity functions. These functions are known respectively as the **unit** and **co-unit** of the adjunction. For the $\mathtt{W} \dashv \mathtt{R}$ adjunction specified in (5.3), this yields the functions in (5.4).

(5.4) $\eta :: \alpha \to \boxed{\mathtt{R}\,\mathtt{W}\,\alpha}$ $\quad\quad \varepsilon :: \boxed{\mathtt{W}\,\mathtt{R}\,\alpha} \to \alpha$

$\eta := \Phi\,\mathbf{id}$ $\quad\quad \varepsilon := \Psi\,\mathbf{id}$

$= \lambda a \lambda x.\, \langle a, x\rangle$ $\quad\quad = \lambda \langle f, x\rangle.\, f\, x$

It turns out that whenever $\Omega \dashv \Gamma$ is an adjunction, the composite functor $\Gamma\,\Omega$ is a monad. The $\eta$ function determined by (5.4) is in fact its $\eta$, in the sense of Chapters 3 and 4. We will say more about this connection in Section 5.3.4, but for now, just note that the existence of $\eta$ guarantees a way of lifting any value into a trivial computation with the structure of both $\Gamma$ and $\Omega$.

The $\varepsilon$ function, on the other hand, is entirely new. It ensures that any composite $\Gamma\,\Omega$ computation can be *deconstructed*, eliminating all of the $\Gamma$ and $\Omega$ structure. How does this work in the case of $\mathtt{W} \dashv \mathtt{R}$? Well, a computation of type $\boxed{\mathtt{W}\,\mathtt{R}\,\alpha} ::= (\mathtt{e} \to \alpha) \times \mathtt{e}$ is a pair consisting of two things: a function from antecedents to values $f :: \mathtt{e} \to \alpha$, alongside a stored antecedent $x :: \mathtt{e}$. To extract that result of type $\alpha$, we need only pass the remembered referent $x$ into the context-dependent function $f$. And this is exactly what $\varepsilon$ does. For example, applying $\varepsilon$ to (5.2) gives (5.5), a pure truth value in which the role of the pronoun is happily saturated by its antecedent.

(5.5) $\varepsilon\,\langle \lambda x.\, \textbf{obscure}\,(\textbf{spot}\,x)\,(\textbf{moon}\,\textbf{j}), \textbf{j}\rangle = \textbf{obscure}\,(\textbf{spot}\,\textbf{j})\,(\textbf{moon}\,\textbf{j})$

The derivation in (5.2) is just a handful of maps over the basic reading and writing effects, nothing more complicated than what was introduced in Chapter 2. Apparently, dynamic semantics was a mere function application away, and the fact in (5.5) suggests that this application may come from the duality between the effects. Let us then put this exceedingly useful adjunction to work in the effect-driven interpretation scheme.

### *5.1.2 Adjunction as a higher-order mode of combination*

Adjunction is a binary, asymmetric relation between functors. In this respect, constructing a mode of combination that takes advantage of adjunctions is particularly straightforward. After all, a mode of combination is itself a binary, asymmetric relation between denotations. We could, for instance, imagine the forward and backward co-unit combinators in (5.6).

(5.6a) $(\dashv_{>}) :: \boxed{\Omega\,\alpha \to \beta} \to \boxed{\Gamma\,\alpha} \to \beta$

$L \dashv_{>} R \coloneqq \varepsilon\,((\lambda f.\,(\lambda x.\, f\,x) \bullet_{\Gamma} R) \bullet_{\Omega} L)$

(5.6b) $(\dashv_{<}) :: \boxed{\Omega\,\alpha} \to \boxed{\Gamma\,\alpha \to \beta} \to \beta$

$L \dashv_{<} R \coloneqq \varepsilon\,((\lambda x.\,(\lambda f.\, f\,x) \bullet_{\Gamma} R) \bullet_{\Omega} L)$

These might be put to use in a derivation like (5.7), which is identical to (5.2) except for the last step, where this time binding is achieved thanks to the $\varepsilon$ in the $\dashv_{<}$ rule. Notice that this is a kind of dynamic or discourse binding; the referent persists *up* the computation of the left branch, until the subject meets the predicate, at which point it sinks *down* the right branch to value the pronoun.

(5.7)

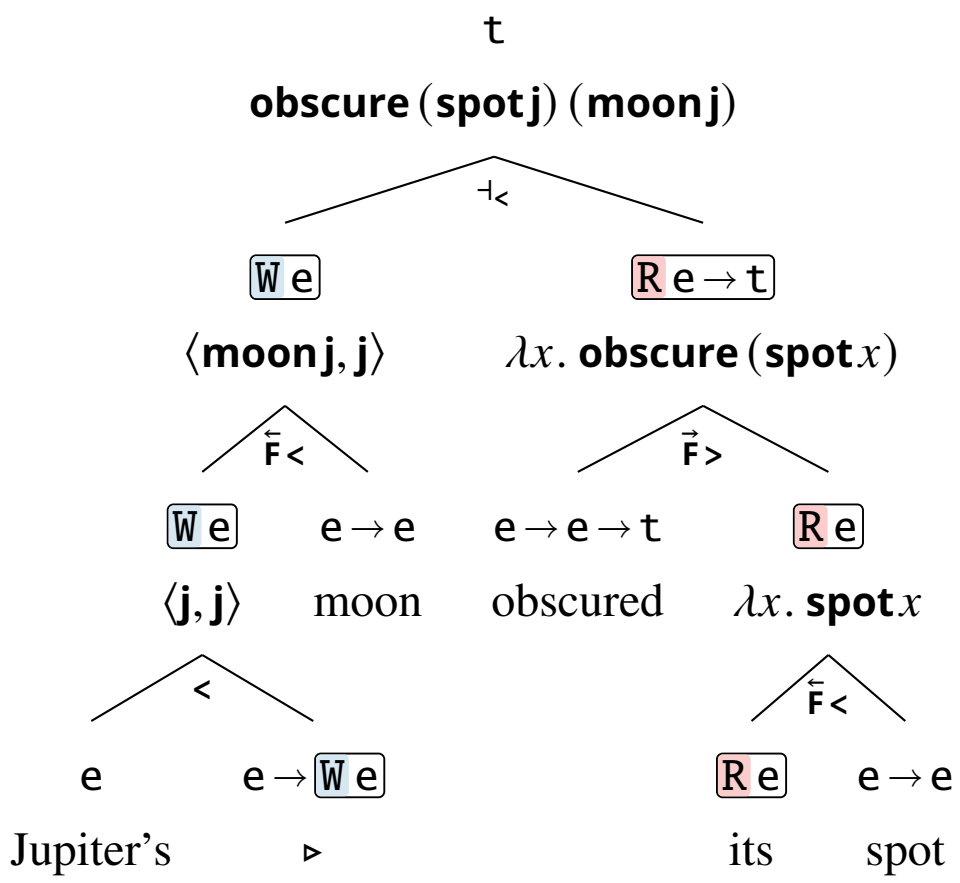

But as usual, this first-order combinatorial approach would need special cases for every basic mode of combination (in addition to **>** and **<**), and then would inevitably fall flat in the presence of other, irrelevant effects. So we generalize in the now standard fashion: whenever there is any way of combining the underlying types of a $\Omega$ constituent and a $\Gamma$ constituent, where $\Omega \dashv \Gamma$, we can combine the constituents by mapping over the two effects, combining the underlying values, and then applying the co-unit $\varepsilon$ to the result.

(5.8)
$$\mathbf{C} :: (\sigma \to \tau \to \omega) \to \boxed{\Omega\,\sigma} \to \boxed{\Gamma\,\tau} \to \omega$$
$$\mathbf{C}\,(*)\,E_1\,E_2 := \varepsilon\,((\lambda l.\,(\lambda r.\,l * r) \bullet E_2) \bullet E_1)$$

The simple combinators $(\dashv_>)$ and $(\dashv_<)$ are thus reproduced as **C >** and **C <**. And the derivation in (5.7) is equivalent to the one in (5.9), yielding the value (5.5).

(5.9)

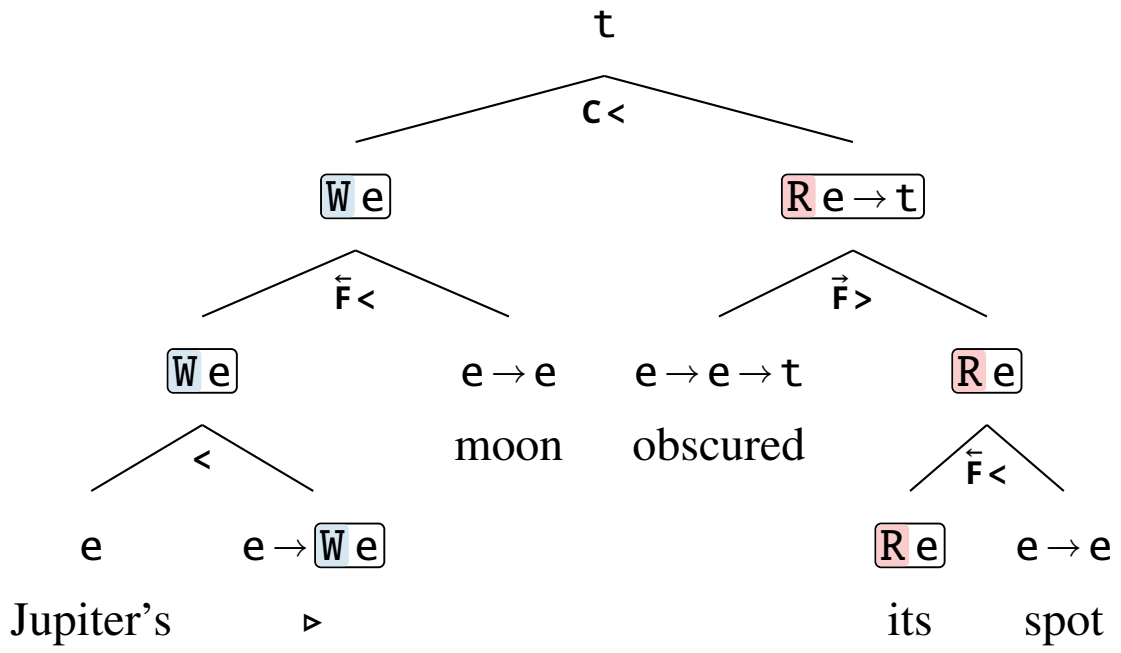

What's more, incidental effects above and below the adjoint ones no longer interfere with the adjunction, as they shouldn't. For instance, adding some indeterminacy in one of the constituents of (5.9), as in (5.10), does not preclude binding:

(5.10)

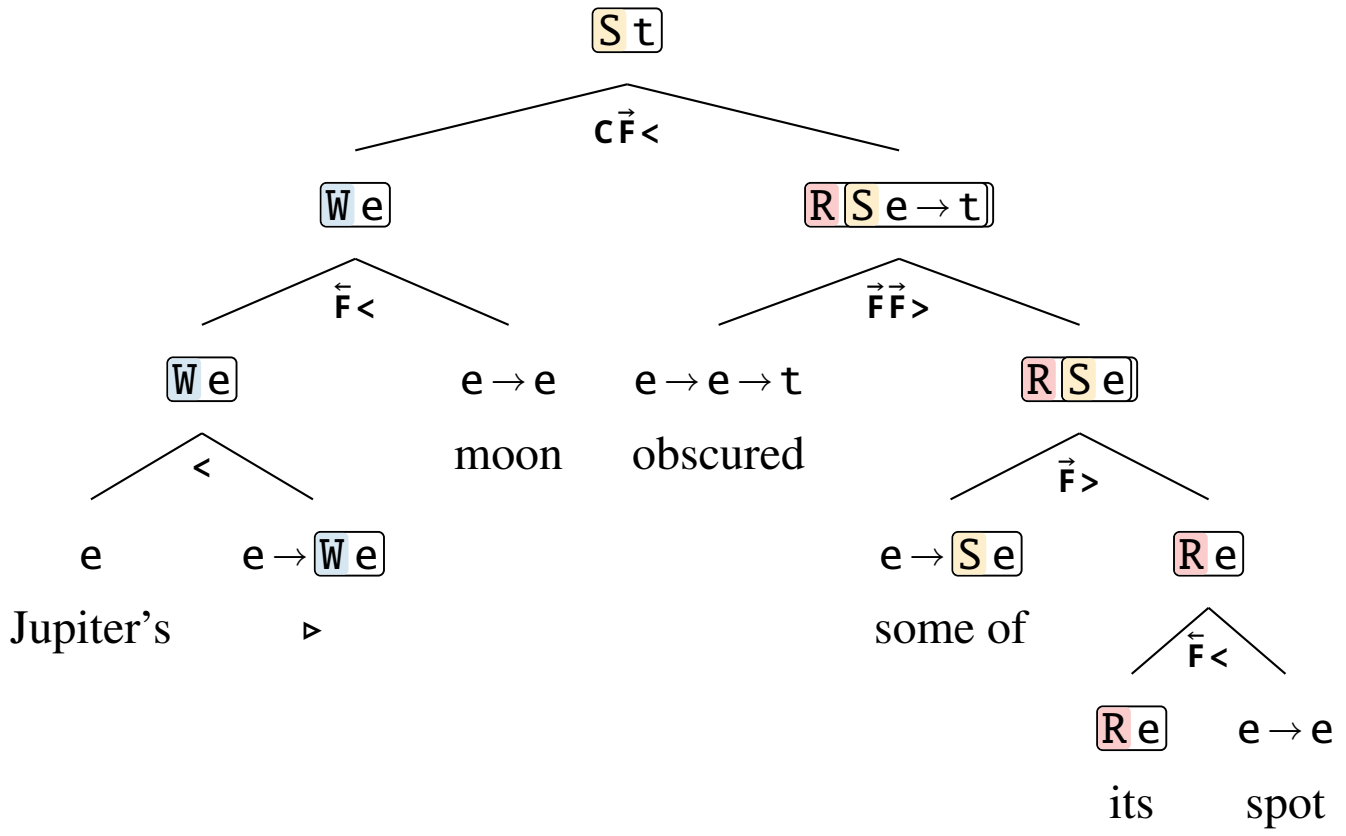

The naive combinators $(\dashv_{>})$ and $(\dashv_{<})$ would not have known what to do in this circumstance, as the underlying types — `e` and `S e → t` — cannot be combined by any basic mode of combination. But with **C**, these underlying types are combined in the obvious way — with $\vec{\mathbf{F}}$**<** — and then the `W` and `R` effects cancel each other out. The referent coming from the left is passed into the function requesting a referent on the right, and only the indeterminacy remains.

The example in (5.10) demonstrates how binding into an indefinite emerges. The inverse is also possible:

(5.11)

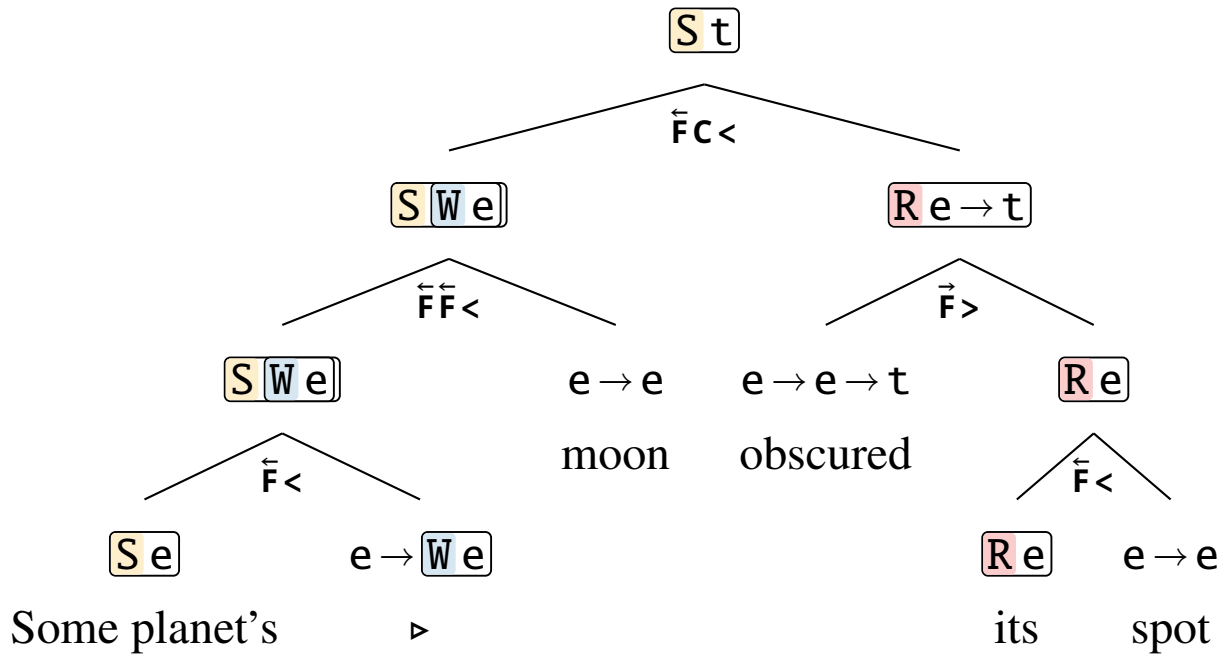

And for the *coup de grâce*: indefinites can also bind into other indefinites. This is shown in (5.12), which we discuss below.

(5.12)

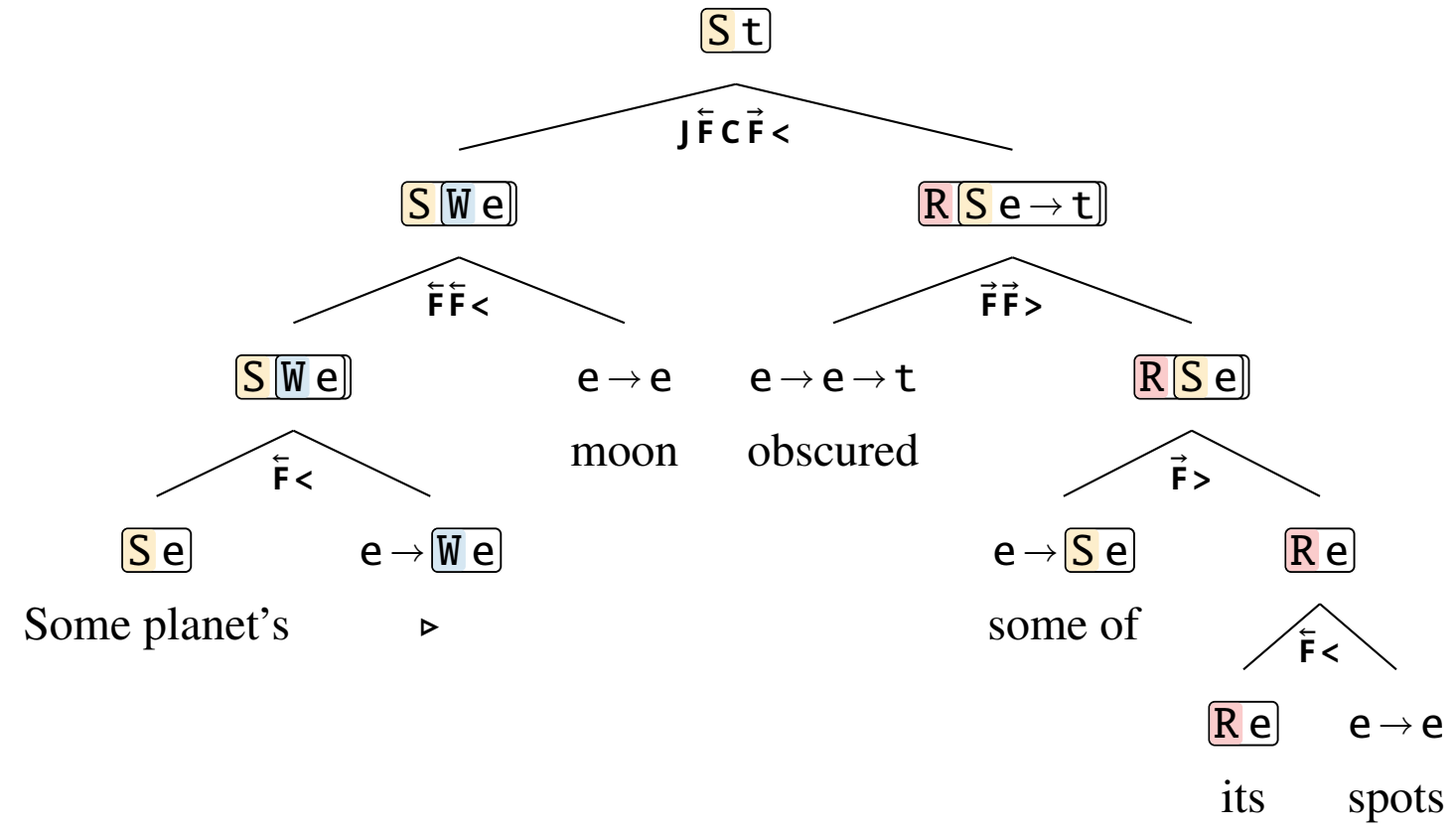

Let us walk through the final step. The monadic meta-combinator **J** combines the two daughters via $\overleftarrow{\mathbf{F}}\,\mathbf{C}\,\overrightarrow{\mathbf{F}}\,\mathbf{<}$ before flattening the result with $\mu$. That inner complex combinator is a combination of the two top-level combinators in (5.11) and (5.10). As in (5.11), the outer $\overleftarrow{\mathbf{F}}$ means we begin by skipping over the left daughter's top effect, S. This leaves us with W e on the left and R S e → t on the right. These are combined via $\mathbf{C}\,\overrightarrow{\mathbf{F}}\,\mathbf{<}$, exactly as in (5.10). The result of these combined combinations, just before the $\mu$ imposed by the **J**, is of type S S e. The outer S corresponds to the left indefinite, which was mapped over first, and the inner S to the right indefinite, which is mapped over while executing the $\varepsilon$ of **C**. Finally, this doubly-layered set is unioned by the outermost **J**, delivering a single indeterminate proposition that varies both with planets and their spots.

## 5.2 Crossover

One fortuitous consequence of this formulation of binding is that it inherits the non-commutativity of adjunction. The **C** rule expects the left adjoint to come from the left daughter. For the W ⊣ R adjunction, this means that discourse antecedents must precede the pronouns they bind. Expressions in which this order is reversed are still composable, and even composable in such a way that the would-be binder outscopes its would-be bindee, but the two effects will never cancel out. Nothing in the grammar will ever pass the remembered referent coming from the right into the request for a referent coming from the left.

(5.13)

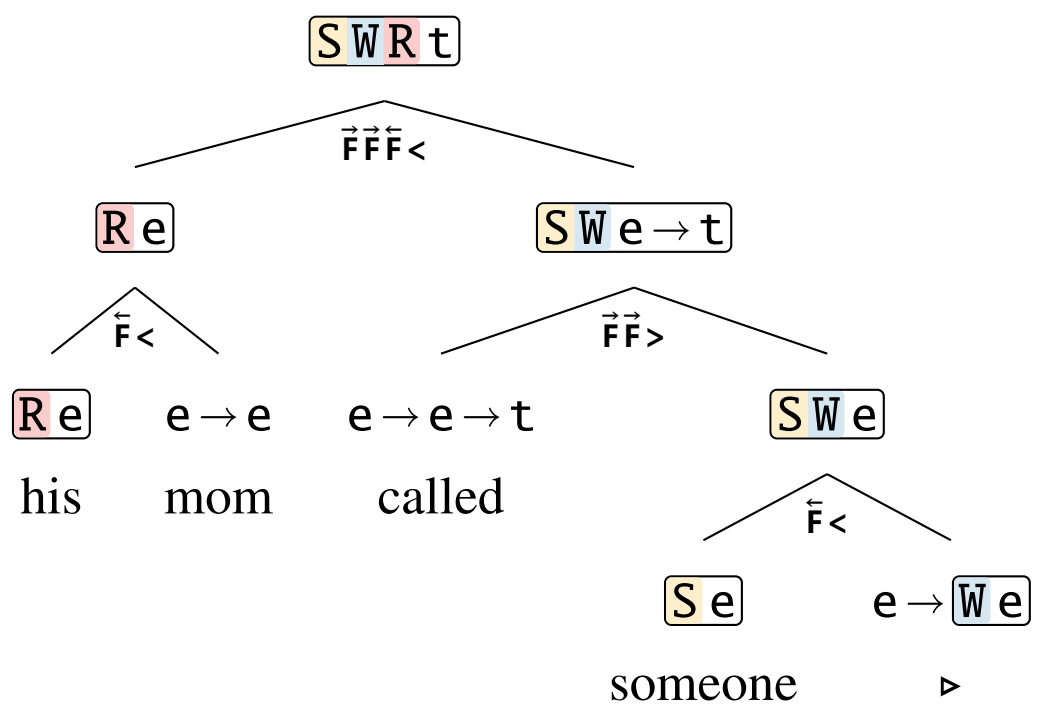

In this manner, the mode of combination derives the scope and binding pattern long known to linguists as **crossover**. Namely, discourse antecedence and retrieval proceeds from left to right, even while general semantic scope may be arbitrarily inverted. To put a point on this, consider the derivation in (5.14). The quantificational object takes logical scope over the indefinite subject, so that teachers vary with students. But since the computation still includes a `R` constructor, the pronoun's request for an antecedent remains open.

(5.14)

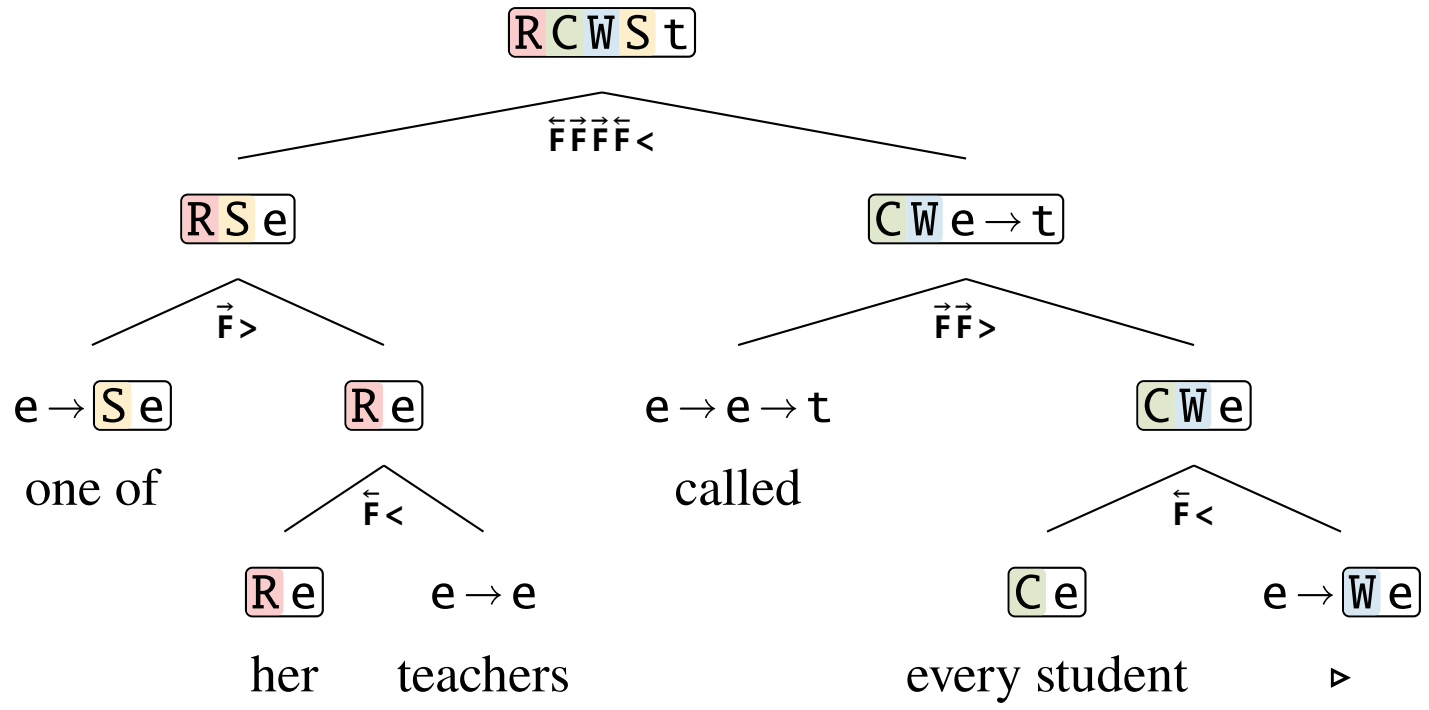

To be sure, there are many ways to combine the subject and predicate of (5.14), which determine different layerings of the various effects. But none of them will fill the pronoun's request for an antecedent. The best that one can do is scope the object's referent `W` over the subject's request `R`, yielding the layering $\boxed{\mathtt{C\,W\,R\,S\,t}}$. But as there is no independent $\varepsilon$-mechanism beside the **C** rule, they will just carry on like this, anaphoric ships passing in the night.[14]

## 5.3 From adjunctions to monads

### 5.3.1 Fused effects

Despite their prominent practical and theoretical roles in functional programming, monadic techniques for handling multiple effects face a well-known obstacle: in contrast with functors and applicatives, which are closed under composition, the composition of two monads is not necessarily monadic (Jones and Duponcheel 1993, King and Wadler 1993). That is, there is no general way to define a law-abiding function $\mu_{\Theta\mathtt{K}} :: \boxed{\mathtt{\Sigma\,K}\,\boxed{\mathtt{\Sigma\,K\,\alpha}}} \to \boxed{\mathtt{\Sigma\,K\,\alpha}}$, even when $\Sigma$ is a monad with an associated $\mu_\Sigma :: \boxed{\mathtt{\Sigma\,\Sigma\,\alpha}} \to \boxed{\mathtt{\Sigma\,\alpha}}$, and $\mathtt{K}$ is a monad with $\mu_\mathtt{K} :: \boxed{\mathtt{K\,K\,\alpha}} \to \boxed{\mathtt{K\,\alpha}}$. The reason, intuitively, is that $\Sigma$ and $\mathtt{K}$ are interleaved in $\mu_{\Theta\mathtt{K}}$, which means there's no way to use the respective $\mu_\Sigma$ and $\mu_\mathtt{K}$ to help flatten out the layers of effects.

To see a situation where this fact might rear its head, consider the description in (5.17).

(5.15) another picture of a book in her library

For the purposes of the demonstration, imagine that 'another' is a determiner with both anaphoric and indeterminate effects. It requires an antecedent — *other than what?* — and generates a set of possible referents. Which referents it computes depends on the antecedent: roughly, those left in its restrictor once the antecedent is removed. Also for simplicity, let's assume the input type for anaphora is now an assignment function, or list, so that resolution is a matter of choosing an index rather than choosing a scope. With these assumptions, the type of 'another' is plausibly rendered as $(\mathtt{e} \to \mathtt{t}) \to \boxed{\mathtt{R\,S\,e}}$, with the lexical semantics in (5.16).

(5.16) $\text{another}_n :: (\mathtt{e} \to \mathtt{t}) \to \boxed{\mathtt{R\,S\,e}}$

$[\![\text{another}_n]\!] := \lambda P \lambda i.\ \{x \mid Px,\ x \neq i_n\}$

Putting this determiner together with a restrictor that itself contains both anaphoric and indeterminate effects looks like (5.17).

[14] See Barker and Shan (2014) for a crossover solution with a similar character. Barker and Shan manage all effect combinations using layers of continuations, as sketched in Section 4.3.1. Rightward referent introductions may outscope leftward pronouns, but the two continuation layers can never be merged. In contrast, leftward referents and rightward pronouns may simply meet on the same level, where they neutralize each other.

(5.17)

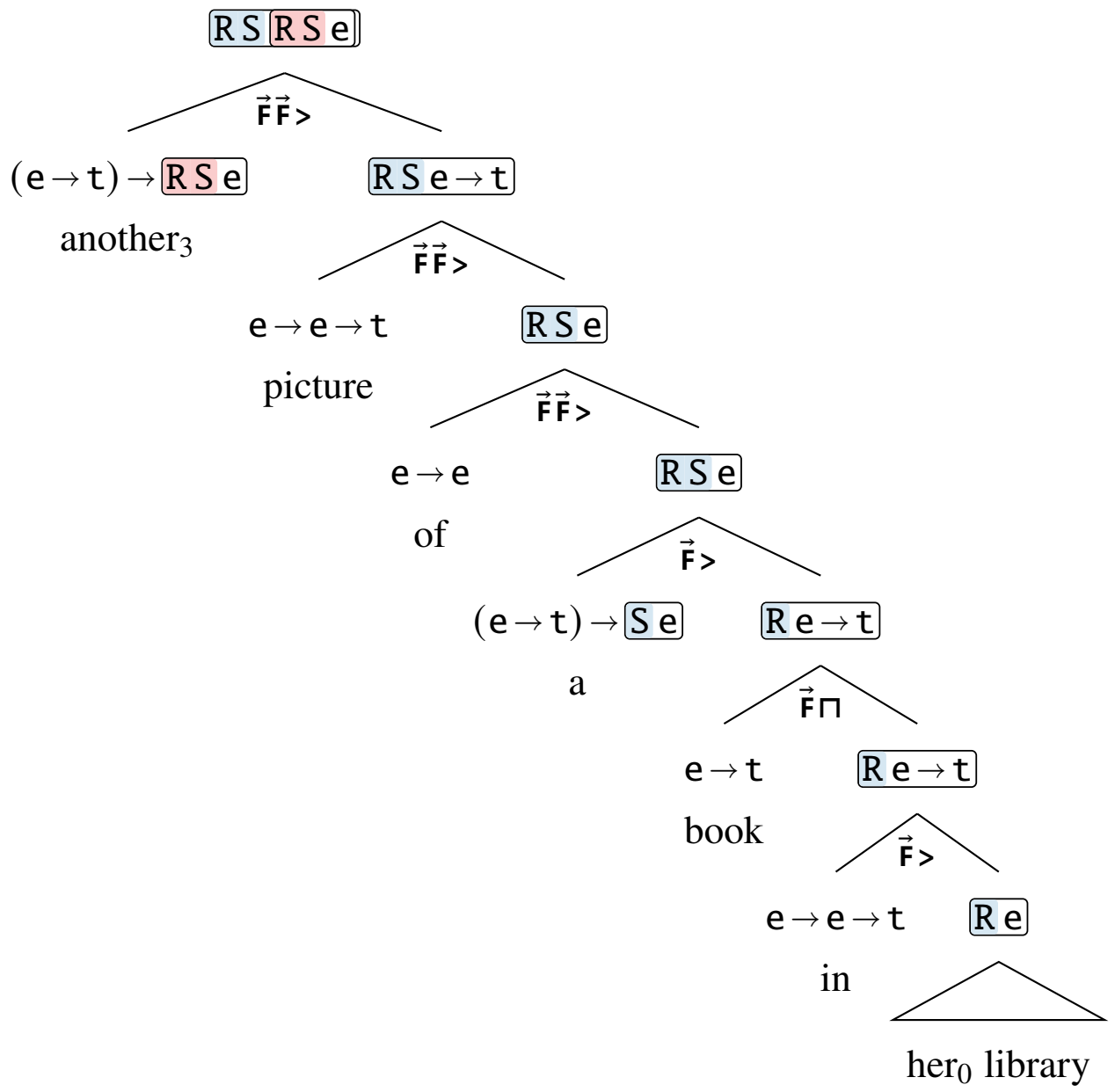

There is no other way to combine these pieces, given the grammar in Figure 8. This is a shame, since there's no *semantic* reason why the assignment-dependencies induced by 'another' and 'her library' cannot be collapsed into a single R-layer. Likewise the alternatives generated by 'a' and those by 'another' could certainly in principle be represented in a single set determined by the cross product of the two restrictors. Indeed, the meaning given in (5.18) is a perfectly plausible denotation for the phrase, in line with the sorts of denotations otherwise assigned to nested indefinite expressions here. The trouble is just that there's no way to compose the determiner and restrictor using the **J** rules for R and S that would bring it about.

(5.18) (5.17) :: $\boxed{\mathtt{R\,S}\ \mathtt{e}}$

$$[\![(5.17)]\!] = \lambda i.\ \{x \mid (\textbf{book} \sqcap \textbf{in}\,(\textbf{lib}\, i_0))\, y,\ \textbf{pic}\, y x,\ x \neq i_3\}$$

And this is as it must be. The **J**, $\overleftarrow{\mathbf{F}}$, and $\overrightarrow{\mathbf{F}}$ modes operate without regard to the particular effects being combined. So if there were somehow a derivation for (5.17) resulting in the flattened type $\boxed{\mathtt{R\,S}\ \mathtt{e}}$, it would work just as well for any composite effects $\Sigma\mathtt{K}$, providing a combinatorial template for marrying an arbitrary pair of monads. But since it is known that some compositions of monads are not themselves monadic, there can't possibly be any such completely general derivation.

Here's the kicker, though: the composition of $\mathtt{R}$ and $\mathtt{S}$ *is a monad*! Let $\boxed{\mathtt{H}\,\alpha} ::= \mathtt{i} \to \{\alpha\}$ represent computations that are inherently environment-sensitive *and* parallel. This constructor is a monad. Its bind operation is as follows:

(5.19) $$(\ll\!\!=_{\mathtt{H}}) :: (\alpha \to \boxed{\mathtt{H}\,\beta}) \to \boxed{\mathtt{H}\,\alpha} \to \boxed{\mathtt{H}\,\beta}$$
$$m \ll\!\!=_{\mathtt{H}} k := \lambda i.\ \bigcup\{k\,a\,i \mid a \in m\,i\}$$

It would appear that the lack of any completely general strategy for composing monads forces us into an obligatorily higher-order representation even for those effects that do happen to fit together in a monadic way. This raises the same sorts of empirical red flags as in Section 4.1. Without a way to derive (5.17) at type $\boxed{\mathtt{R}\,\mathtt{S}\,\mathtt{e}}$, the embedded indefinite should obligatorily outscope any otherwise unselective closure operator that captures its host. And now, dubiously, the prediction is that this unusual selectivity should arise only when the nested phrase contains anaphoric elements gumming up the works.

Since English does not work this way, it looks as if the appropriate combinatorial apparatus, e.g., (5.19), will on occasion have to be hand-engineered and manually deployed. If this were true, it would threaten to undermine the modular approach we have championed in this Element. Fortunately, in Section 5.3.2, we show that there is, after all, a principled way to iron out the derivation in (5.17) without introducing the ad-hoc combinator in (5.19). Surprisingly, the solution is a consequence of the $\mathtt{W} \dashv \mathtt{R}$ adjunction.

### 5.3.2 Ejection from Adjunctions

Look again at the final step in (5.17). The problem is that there's no way to combine these two constituents to produce a meaning in the domain of $\boxed{\mathtt{R}\,\mathtt{S}\,\mathtt{e}}$.

(5.20)

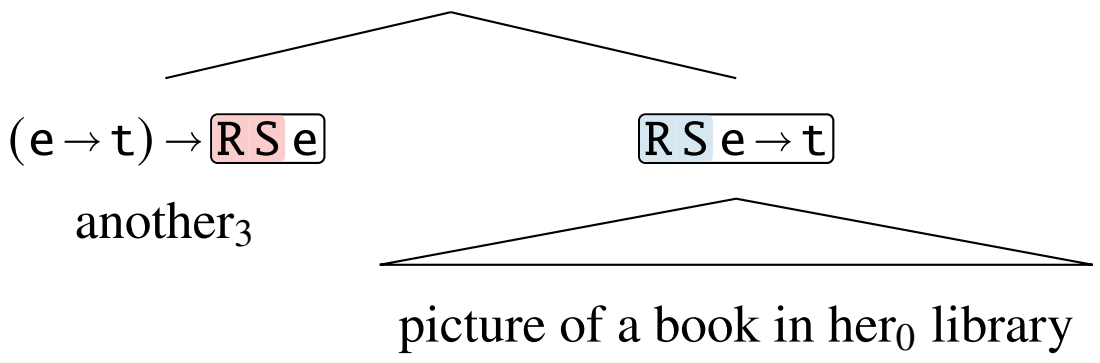

Even more frustratingly, notice that if the $\mathtt{R}$ constructor were outside the entire determiner type, the problem would disappear. The grammar through Chapter 4 would immediately derive the desired semantic object.

(5.21)

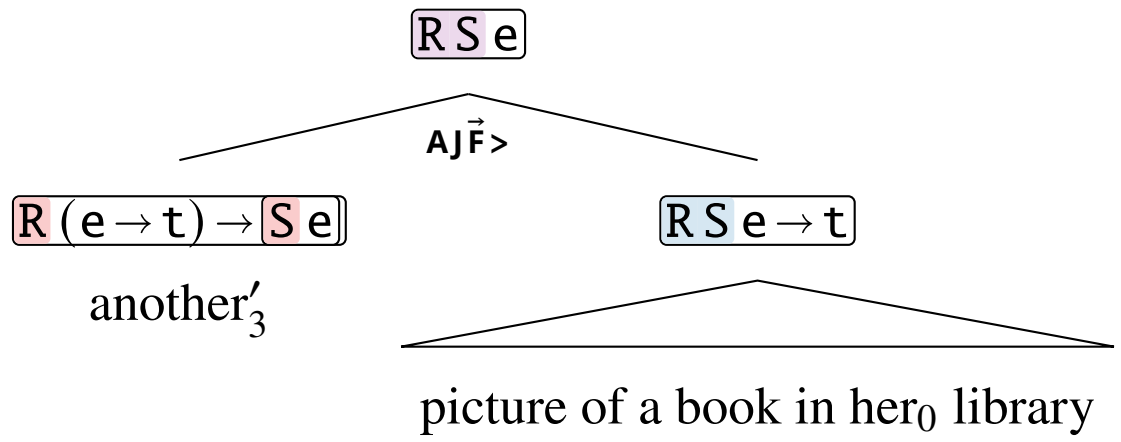

What's more, a moment's reflection will reveal that $\alpha \to \boxed{\mathtt{R}\,\beta}$ and $\boxed{\mathtt{R}\,\alpha \to \beta}$ are actually isomorphic:

(5.22) $\Upsilon_{\mathtt{R}} :: (\alpha \to \boxed{\mathtt{R}\,\beta}) \to \boxed{\mathtt{R}\,\alpha \to \beta}$ $\quad \Upsilon_{\mathtt{R}}^{-1} :: \boxed{\mathtt{R}\,\alpha \to \beta} \to \alpha \to \boxed{\mathtt{R}\,\beta}$

$\Upsilon_{\mathtt{R}} \coloneqq \lambda k.\, \lambda i \lambda a.\, k\, a\, i$ $\quad \Upsilon_{\mathtt{R}}^{-1} \coloneqq \lambda m.\, \lambda a \lambda i.\, m\, i\, a$

This is certainly not true for all effects. Consider $\mathtt{M}$, for instance. Something of type $(\alpha \to \boxed{\mathtt{M}\,\beta})$ is a partial function, but something of type $\boxed{\mathtt{M}\,\alpha \to \beta}$ is either a total function or nothing at all. The only parametric way to convert the former to the latter is to return # if *any* input leads to #. That is, if the function is undefined anywhere, its image under $\Upsilon$ would have to be undefined full stop. But then, since all properly partial functions would be mapped to #, $\Upsilon$ would not be injective, and there would thus be no way to recover the original partial function.

Squinting at the types in (5.22), one may detect an eerie resemblance to the isomorphism that defines an adjunction, repeated in (5.23).

(5.23) $\Psi :: (\alpha \to \boxed{\Gamma\,\beta}) \to \boxed{\Omega\,\alpha} \to \beta$ $\quad \Phi :: (\boxed{\Omega\,\alpha} \to \beta) \to \alpha \to \boxed{\Gamma\,\beta}$

Indeed, it turns out that for *any right adjoint* $\Gamma$, the types $(\alpha \to \boxed{\Gamma\,\beta})$ and $\boxed{\Gamma\,\alpha \to \beta}$ are isomorphic. The components of the isomorphism are given by the functions in (5.24), where # is any value of any type.[15]

(5.24) $\Upsilon_{\Gamma} :: (\alpha \to \boxed{\Gamma\,\beta}) \to \boxed{\Gamma\,\alpha \to \beta}$ $\quad \Upsilon_{\Gamma}^{-1} :: \boxed{\Gamma\,\alpha \to \beta} \to \alpha \to \boxed{\Gamma\,\beta}$

$\Upsilon_{\Gamma} \coloneqq \lambda k.\, \Phi\, (\lambda \omega \lambda a.\, \Psi\, (\lambda_.\, k\, a)\, \omega)\, \#$ $\quad \Upsilon_{\Gamma}^{-1} \coloneqq \lambda m \lambda a.\, (\lambda f.\, f\, a) \bullet m$

As the reader may verify, substituting in the relevant $\mathtt{R}$-effect $\Phi$, $\Psi$, and $(\bullet)$ operators reduces these equations to those of (5.22).

This suggests that a very simple fix to the problem at the root of (5.20) is to first transform it into (5.21), via $\Upsilon$. Because the Kleisli arrow $\alpha \to \boxed{\Gamma\,\beta}$ may appear in either the left or right daughter, we need two variants of the rule,

[15] This is ultimately a consequence of the fact that all right adjoints in the category of sets are **distributive**, in the sense of J. Beck (1969). See Asudeh and Giorgolo 2020: Ch. 7 for more direct applications of J. Beck's distributive laws in natural language composition.

which we call **eject**.

(5.25a) $\overleftarrow{\blacktriangle} :: (\boxed{\Gamma\,\sigma \to \sigma'} \to \tau \to \upsilon) \to (\sigma \to \boxed{\Gamma\,\sigma'}) \to \tau \to \upsilon$

$\overleftarrow{\blacktriangle}\,(*)\, E_1\, E_2 \coloneqq \Upsilon\, E_1 * E_2$

(5.25b) $\overrightarrow{\blacktriangle} :: (\sigma \to \boxed{\Gamma\,\tau \to \tau'} \to \upsilon) \to \sigma \to (\tau \to \boxed{\Gamma\,\tau'}) \to \upsilon$

$\overrightarrow{\blacktriangle}\,(*)\, E_1\, E_2 \coloneqq E_1 * \Upsilon\, E_2$

With these eject modes in place, at last, the relevant derivation of (5.17) is given in (5.26). For completeness, Figure 10 displays the full type-driven grammar with adjunction rules, extending the grammar of the previous chapter.

(5.26)

$\boxed{\mathtt{R\,S\,e}}$

$\lambda i.\, \{x \mid (\textbf{book} \sqcap \textbf{in}\,(\textbf{lib}\, i_0))\, y,\ \textbf{pic}\, y\, x,\ x \neq i_3\}$

$\overleftarrow{\blacktriangle}\mathbf{AJ}\overrightarrow{\mathbf{F}}\mathbf{>}$

$(\mathtt{e} \to \mathtt{t}) \to \boxed{\mathtt{R\,S\,e}}$ — $\lambda P \lambda i.\, \{x \mid P\, x,\ x \neq i_3\}$ — another$_3$

$\boxed{\mathtt{R\,S\,e \to t}}$ — $\lambda i.\, \{x \mid (\textbf{book} \sqcap \textbf{in}\,(\textbf{lib}\, i_0))\, y,\ \textbf{pic}\, y\, x\}$ — picture of a book in her$_0$ library

The context-dependence in the return type of the determiner parachutes out over the rest of the word's meaning, and from there combination procedes exactly as in (5.21).

### *5.3.3 Monad Transformers*

Notice that the underlying indeterminacy of the meanings in (5.26) played no particular role in the solution. The entire story would unfold just the same for *any* inner monad in place $\mathtt{S}$. Take the definite article, for instance, on an anaphoric view of definiteness (e.g., Heim 1982).[16]

(5.27)

$\boxed{\mathtt{R\,M\,e}}$

$\lambda i{:}\, |\textbf{friend}\, i_0| = 2 \wedge \textbf{book}\, i_3 \wedge \textbf{friend}\, i_0 \subseteq \textbf{rec}\, i_3.\, i_3$

$\overleftarrow{\blacktriangle}\mathbf{AJ}\overrightarrow{\mathbf{F}}\mathbf{>}$

$(\mathtt{e} \to \mathtt{t}) \to \boxed{\mathtt{R\,M\,e}}$ — $\lambda P \lambda i{:}\, P\, i_3.\, i_3$ — the$_3$

$\boxed{\mathtt{R\,M\,e \to t}}$ — $\lambda i{:}\, |\textbf{friend}\, i_0| = 2.\, \lambda x.\, \textbf{book}\, x \wedge \textbf{friend}\, i_0 \subseteq \textbf{rec}\, x$ — book recommended by both her$_0$ friends

[16] $\lambda i{:}\, E_1.\, E_2$ represents a partial function defined only for those $i$ making $E_1$ true, and mapping all other $i$ to #.

**Types:**

$$\begin{aligned} \tau ::= {} & \mathtt{e} \mid \mathtt{t} \mid \cdots && \text{Base types} \\ \mid {} & \tau \to \tau && \text{Function types} \\ \mid {} & \ldots && \ldots \\ \mid {} & \boxed{\Sigma\,\tau} && \text{Computation types} \end{aligned}$$

**Effects:**

$$\begin{aligned} \Sigma ::= {} & \mathtt{R} && \text{Input} \\ \mid {} & \mathtt{W} && \text{Output} \\ \mid {} & \mathtt{S} && \text{Indeterminacy} \\ \mid {} & \ldots && \ldots \end{aligned}$$

**Basic Combinators:**

$$(\mathbf{>}) :: (\alpha \to \beta) \to \alpha \to \beta \qquad \text{Forward Application}$$
$$f \mathbf{>} x \coloneqq f\,x$$

$$(\mathbf{<}) :: \alpha \to (\alpha \to \beta) \to \beta \qquad \text{Backward Application}$$
$$x \mathbf{<} f \coloneqq f\,x$$

$$\ldots \qquad \ldots$$

**Meta-combinators:**

$$\overleftarrow{\mathsf{F}} :: (\sigma \to \tau \to \upsilon) \to \boxed{\Sigma\,\sigma} \to \tau \to \boxed{\Sigma\,\upsilon} \qquad \text{Map Left}$$
$$\overleftarrow{\mathsf{F}}\,(*)\,E_1\,E_2 \coloneqq (\lambda a.\, a * E_2) \bullet E_1$$

$$\overrightarrow{\mathsf{F}} :: (\sigma \to \tau \to \upsilon) \to \sigma \to \boxed{\Sigma\,\tau} \to \boxed{\Sigma\,\upsilon} \qquad \text{Map Right}$$
$$\overrightarrow{\mathsf{F}}\,(*)\,E_1\,E_2 \coloneqq (\lambda b.\, E_1 * b) \bullet E_2$$

$$\mathsf{A} :: (\sigma \to \tau \to \omega) \to \boxed{\Sigma\,\sigma} \to \boxed{\Sigma\,\tau} \to \boxed{\Sigma\,\omega} \qquad \text{Structured App}$$
$$\mathsf{A}\,(*)\,E_1\,E_2 \coloneqq (\lambda a \lambda b.\, a * b) \bullet E_1 \circledast E_2$$

$$\overleftarrow{\mathsf{U}} :: (\sigma \to (\tau \to \tau') \to \upsilon) \to \sigma \to (\boxed{\Sigma\,\tau} \to \tau') \to \upsilon \qquad \text{Unit Left}$$
$$\overleftarrow{\mathsf{U}}\,(*)\,E_1\,E_2 \coloneqq E_1 * (\lambda b.\, E_2\,(\eta\, b))$$

$$\overrightarrow{\mathsf{U}} :: ((\sigma \to \sigma') \to \tau \to \upsilon) \to (\boxed{\Sigma\,\sigma} \to \sigma') \to \tau \to \upsilon \qquad \text{Unit Right}$$
$$\overrightarrow{\mathsf{U}}\,(*)\,E_1\,E_2 \coloneqq (\lambda a.\, E_1\,(\eta\, a)) * E_2$$

$$\mathsf{J} :: (\sigma \to \tau \to \boxed{\Sigma\,\Sigma\,\omega}) \to \sigma \to \tau \to \boxed{\Sigma\,\omega} \qquad \text{Join}$$
$$\mathsf{J}\,(*)\,E_1\,E_2 \coloneqq \mu\,(E_1 * E_2)$$

$$\mathsf{C} :: (\sigma \to \tau \to \omega) \to \boxed{\Omega\,\sigma} \to \boxed{\Gamma\,\tau} \to \omega \qquad \text{Co-unit}$$
$$\mathsf{C}\,(*)\,E_1\,E_2 \coloneqq \varepsilon\,((\lambda l.\, (\lambda r.\, l * r) \bullet E_2) \bullet E_1)$$

$$\overleftarrow{\underline{\blacktriangle}} :: (\boxed{\Gamma\,\sigma \to \sigma'} \to \tau \to \upsilon) \to (\sigma \to \boxed{\Gamma\,\sigma'}) \to \tau \to \upsilon \qquad \text{Eject Left}$$
$$\overleftarrow{\underline{\blacktriangle}}\,(*)\,E_1\,E_2 \coloneqq \Upsilon\,E_1 * E_2$$

$$\overrightarrow{\underline{\blacktriangle}} :: (\sigma \to \boxed{\Gamma\,\tau \to \tau'} \to \upsilon) \to \sigma \to (\tau \to \boxed{\Gamma\,\tau'}) \to \upsilon \qquad \text{Eject Right}$$
$$\overrightarrow{\underline{\blacktriangle}}\,(*)\,E_1\,E_2 \coloneqq E_1 * \Upsilon\,E_2$$

**Figure 10** A type-driven grammar with adjunctions

This means that the combinator defined by $\overleftarrow{\blacktriangle}\,\mathbf{A}\,\mathbf{J}\,\vec{\mathbf{F}}\,\mathbf{>}$ acts as a bind operator ($=\!\!\ll_{\mathtt{R}\Sigma}$) for any composite effect $\mathtt{R}\,\Sigma$, so long as $\Sigma$ is a monad. What is this combinator? Well, for starters, note that $\mathbf{J}\,\vec{\mathbf{F}}\,\mathbf{>}$ is just the ($=\!\!\ll_{\Sigma}$) of the inner effect. The rest unfolds pretty easily:

(5.28)
$$
\begin{aligned}
k =\!\!\ll_{\mathtt{R}\Sigma} m &= \overleftarrow{\blacktriangle}\,\mathbf{A}\,\mathbf{J}\,\vec{\mathbf{F}}\,\mathbf{>}\,k\,m \\
&= \overleftarrow{\blacktriangle}\,(\mathbf{A}\,(=\!\!\ll_{\Sigma}))\,k\,m \\
&= \mathbf{A}\,(=\!\!\ll_{\Sigma})\,(\Upsilon\,k)\,m \\
&= \lambda g.\,\Upsilon\,k\,g =\!\!\ll_{\Sigma} m\,g \\
&= \lambda g.\,(\lambda a.\,k\,a\,g) =\!\!\ll_{\Sigma} m\,g
\end{aligned}
$$

That is, for any monad $\Sigma$, the derived operator in (5.28) serves as a well-typed bind for the composite effect signature $\mathtt{R}\,\Sigma$. Given this, we might define a **higher-order constructor** $\mathtt{R}^{+}$ as in (5.29), parameterized by an inner constructor $\Sigma$ as well as a concrete type $\alpha$.

(5.29) $\boxed{\mathtt{R}^{+}\,\Sigma\,\alpha} ::= \mathtt{i} \to \boxed{\Sigma\,\alpha}$

Where an ordinary constructor maps types to types, a higher-order constructor like the one in (5.29) maps constructors to constructors. They take kinds of computations as arguments and produce enhanced computations as results. From this perspective, $\mathtt{R}^{+}$ *adds* the ability to respond to an environment to an existing computation $\Sigma$.

And it's not hard to prove that whenever the inner computation $\Sigma$ is monadic, this enhanced environment-sensitive computation $\mathtt{R}^{+}\Sigma$ will also satisfy the monad laws. Higher-order constructors like this, that produce new monads from old ones, are sometimes known as **monad transformers** (Liang, Hudak, and Jones 1995).

Notice also that the basic $\mathtt{R}$ monad is exactly the $\mathtt{R}^{+}$ transformer applied to the **identity monad**, defined in (5.30).

(5.30)
$$
\begin{aligned}
&\mathtt{I}\,\alpha ::= \alpha \\
&k \bullet_{\mathtt{I}} m := k\,m \\
&\mu_{\mathtt{I}}\,M := M
\end{aligned}
$$

The $\mathtt{I}$ constructor represents a computation that doesn't do anything but hold a value. Mapping a function over an $\mathtt{I}$ computation is just function application, since an $\mathtt{I}$ computation is nothing but an argument. Using the definitions in (5.30), we see that ($\lll_{\mathtt{I}}$) is also just function application. So the equation in (5.28) with ($\lll_{\mathtt{I}}$) in place of ($\lll_{\Sigma}$) reduces to $\lambda g.\, k\,(m\,g)\,g$, i.e., ($\lll_{\mathtt{R}}$).

All of the effects in Table 2 can be seen as applications of some transformer to the identity monad. That is, they are all special cases of the act of *enhancing* an existing computation with a new, particular effect; namely, the special case when that enhancement is performed on the trivial computation $\mathtt{I}$ that merely holds a value.

We will not dwell on the applications of this technique (see Shan 2001a, Charlow 2014), except to draw attention to one interesting case, that of $\mathtt{T}$. The transformer that gives rise to $\mathtt{T}$ when applied to the identity monad is given in (5.31). This constructor enhances a computational type $\Sigma$ by adding the ability to read and write to a common state $\mathsf{s}$ that is carried throughout the computation.

(5.31) $\boxed{\mathtt{T}^{+}\Sigma\,\alpha} ::= \mathsf{s} \to \boxed{\Sigma\,\alpha \times \mathsf{s}}$

$\eta\, a \coloneqq \lambda i.\, \eta_{\Sigma}\,\langle a, i\rangle$

$k \bullet_{\mathtt{T}^{+}\Sigma}\, m \coloneqq \lambda i.\, m\, i \ggg_{\Sigma} \lambda\langle a, j\rangle.\, \eta_{\Sigma}\,\langle k\, a, j\rangle$

$\mu_{\mathtt{T}^{+}\Sigma}\, M \coloneqq \lambda i.\, M\, i \ggg_{\Sigma} \lambda\langle m, j\rangle.\, m\, j$

When we apply this transformer to $\mathtt{S}$, we get $\mathtt{T}^{+}\mathtt{S} \cong \mathtt{D}$, the type representing computations in the style of dynamic semantics. In typical presentations of dynamic semantics, pronouns, discourse referents, and indefinites interact in a single pervasive framework encompassing the interleaved effects of reading, writing, and nondeterminism. For instance, we might find lexical entries like those in (5.32), where the ▹ operator adds its argument to the discourse state that is carried along to subsequent pronouns.

(5.32a) she$_n$ :: $\boxed{\mathtt{D}\,\mathsf{e}}$

$[\![\text{she}_n]\!] \coloneqq \lambda i.\, \{\langle i_n, i\rangle\}$

(5.32b) ▹ :: $\mathsf{e} \to \boxed{\mathtt{D}\,\mathsf{e}}$

$[\![\triangleright]\!] \coloneqq \lambda x \lambda i.\, \{\langle x, x \mathbin{+\!\!+} i\rangle\}$

(5.32c) someone :: $\boxed{\mathtt{D}\,\mathsf{e}}$

$[\![\text{someone}]\!] \coloneqq \lambda i.\, \{\langle x, i\rangle \mid \textbf{person}\, x\}$

(5.32d) if :: $\boxed{\mathtt{D}\,\mathsf{t}} \to \boxed{\mathtt{D}\,\mathsf{t}} \to \boxed{\mathtt{D}\,\mathsf{t}}$

$[\![\text{if}]\!] \coloneqq \lambda m \lambda n \lambda i.\, \{\langle r, i\rangle\}$

**where** $r = \forall j.\, \langle \textbf{true}, j\rangle \in m\, i \Rightarrow \exists k.\, \langle \textbf{true}, k\rangle \in n\, j$

And with these, we might put together derivations of such classic dynamic phenomena as cross-sentential and donkey binding, as in (5.33) and (5.34).

(5.33)

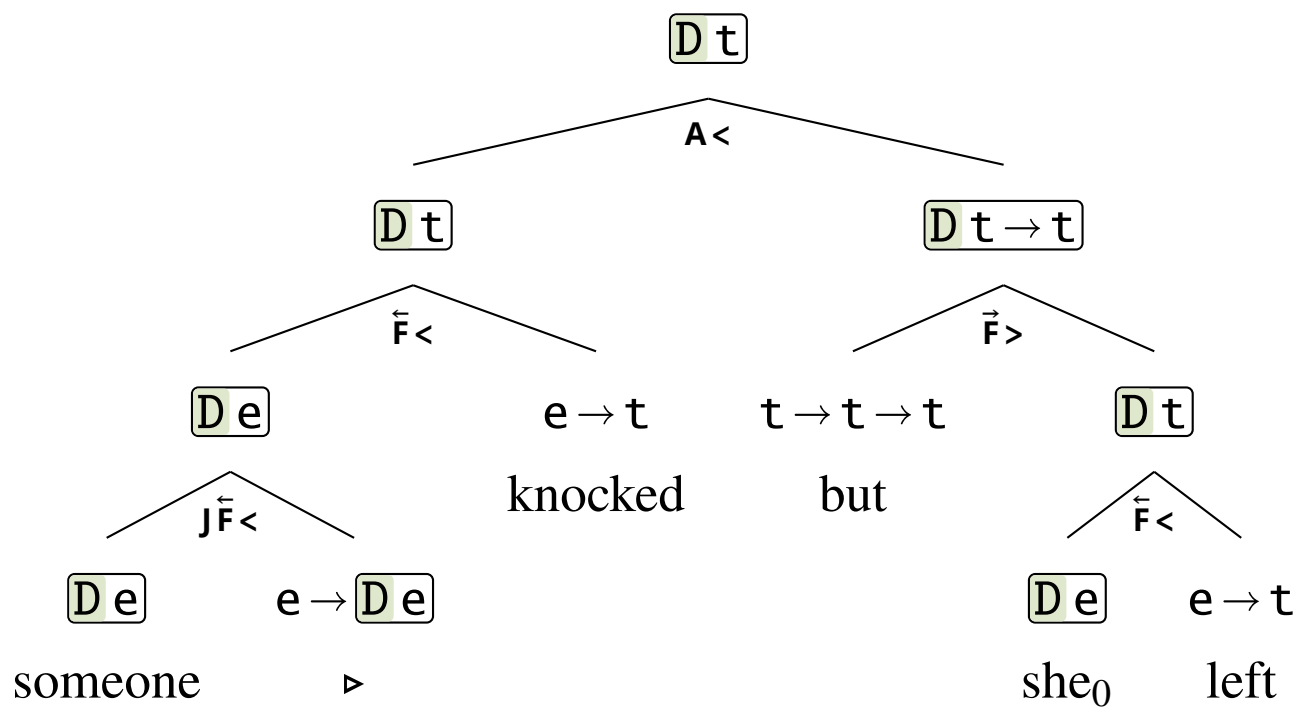

(5.34)

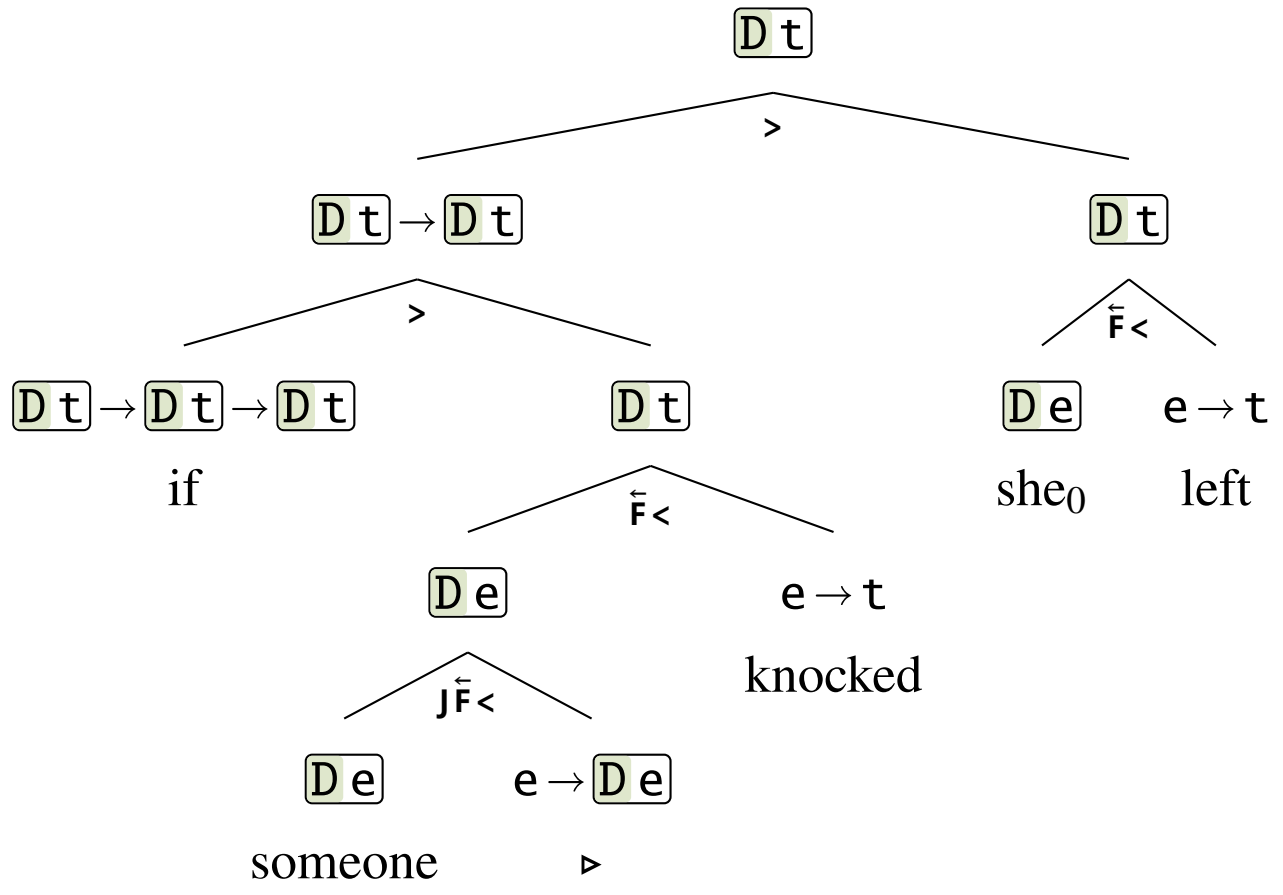

As elegant and well-studied as this system is (see, e.g., Charlow 2014), one can hardly shake the feeling that it is overcooked. Nothing brings this out more than the canonical lexical entries in (5.32a)–(5.32c). The pronoun doesn't do anything interesting except read from the input *i*; it is deterministic and doesn't change the state *i* at all. The referent-pushing operator ▹ adds its prejacent to the state, but is deterministic and doesn't read from its input. And the indefinite ramifies the semantic computation, adding a new alternative for each witness, but does not on its own interact with the state in any way. Yet the three are typed uniformly as `D e`. This is exactly the sort of generalization to the worst case that we have so far avoided.

It is conceivable that some combinations of intuitively distinguishable effects act together in a self-contained computational system deserving of its own encapsulated type. But recall that we already derived sentences similar to the ones just above way back in Section 5.1.2, without recourse to the conglomerate effect `D`. The derivations in (5.11) and (5.12), for example, evince the same nondeterministic and linear antecedence patterns as (5.33) and (5.34). This is no accident; it is, in fact, another surprising corollary of the `W` $\dashv$ `R` adjunction.

### *5.3.4 Monads from Adjunctions*

Consider the abstract configuration in (5.35). The left daughter reads from the input and stores an output for later, while computing an entity. The right daughter, given an entity, also reads from the input and stores an output, while computing a truth value.

(5.35)

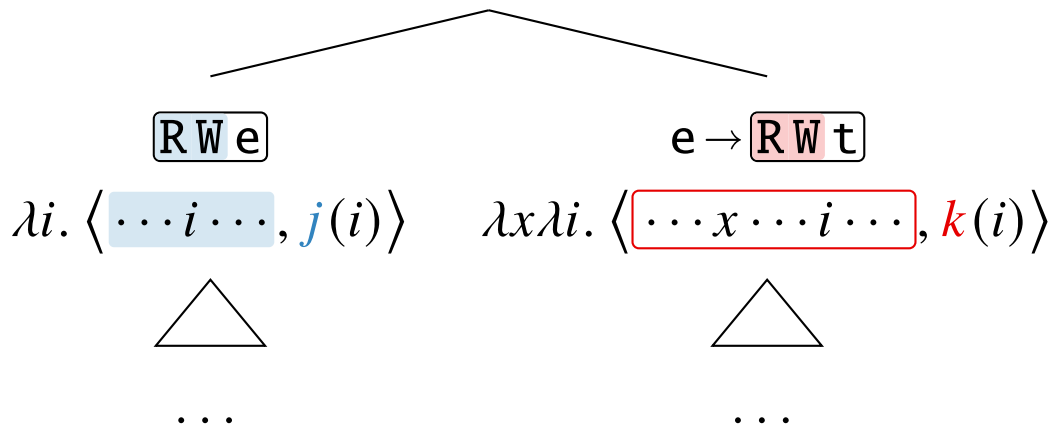

If `W` is monadic, then all of the reasoning from Section 5.3.2 kicks in and we derive a meaning of type `R W` via the $\mathtt{R}^{+}$ transformer. The bind of this composite monad $\mathtt{R}^{+}\mathtt{W}$ is: $m \mathbin{=\!\!\ll} f \;=\; \lambda i.\,\langle p, j +\!\!\!+\, k \rangle$, **where** $\langle x, j \rangle \;=\; m\,i, \langle p, k \rangle \;=\; f\,x\,i$. (Here, $+\!\!\!+$ is the accumulator operation for the monoid underlying any monadic `W`.) Schematically, the input is given to both daughters (and thus, neither's output is passed in as the other's input), and their respective outputs are aggregated:

(5.36)

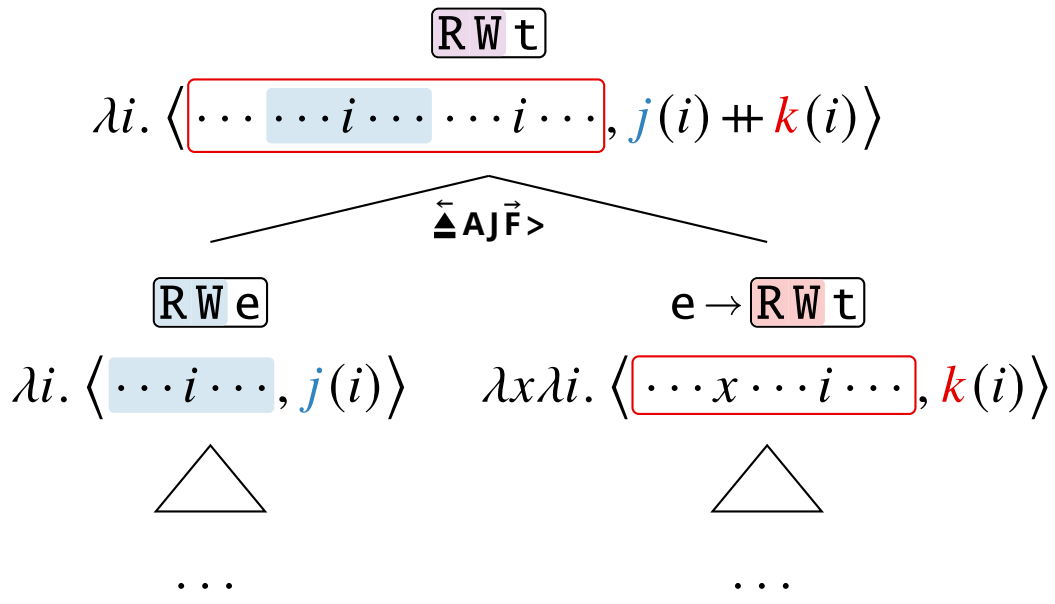

But when `R` and `W` are variable-free, they read and write a single entity at a time. There is no way to accumulate or merge entities in the output and so `W` is not a monad. Are we then stuck with the sort of `RWRW` higher-order meaning that plagued Section 5.3.2? Let us put the question to `ghci` (after implementing the adjunction modes described above; see Section 5.5).

```
ghci> let lt = Comp (RX E) (Comp (WX E) E)
ghci> let rt = E :-> Comp (RX E) (Comp (WX E) T)
ghci> combine lt rt
[...
(ML (ER (CU BA)), Comp (RX E) (Comp (WX E) T))
...]
```

Among the various ways to combine these daughters, there is apparently guaranteed to be a meaning of type `RWt` after all, via $\overleftarrow{\mathsf{F}}\overrightarrow{\blacktriangle}\mathsf{C}{<}$. This mode of combination skips over the `R` from the left daughter, and ejects the `R` from the right daughter. It then combines the `Wt` underlying the left with the (post-ejection) `Re→Wt` on the right. Since the `W` and `R` are adjoint, their effects cancel out and only the underlying `Wt` from the right remains. This, finally, is tucked back under the left's `R` that was skipped.

What does this combination amount to, semantically?

(5.37) $\overleftarrow{\mathsf{F}}\overrightarrow{\blacktriangle}\mathsf{C}{<}\, m\, k = (\lambda x.\, \varepsilon\, ((\lambda a.\, (\lambda f.\, f\, a) \bullet_{\mathtt{R}} \Upsilon\, k) \bullet_{\mathtt{W}} x)) \bullet_{\mathtt{R}} m$

by def of $(\bullet_{\mathtt{R}})$ $= \lambda i.\, \varepsilon\, ((\lambda a.\, \lambda j.\, \Upsilon\, k\, j\, a) \bullet_{\mathtt{W}} m\, i)$

by def of $\Upsilon$ $= \lambda i.\, \varepsilon\, ((\lambda a.\, \lambda j.\, k\, a\, j) \bullet_{\mathtt{W}} m\, i)$

by def of $(\bullet_{\mathtt{W}})$ $= \lambda i.\, \varepsilon\, (\lambda j.\, k\, \langle (m\, i) \rangle_1\, j, (m\, i)_2)$

by def of $\varepsilon$ $= \lambda i.\, k\, (m\, i)_1\, (m\, i)_2$

It evaluates the left daughter at the input, and then sends the output of the left daughter in as input to the right daughter (along with the value it computes). Schematically:

(5.38)

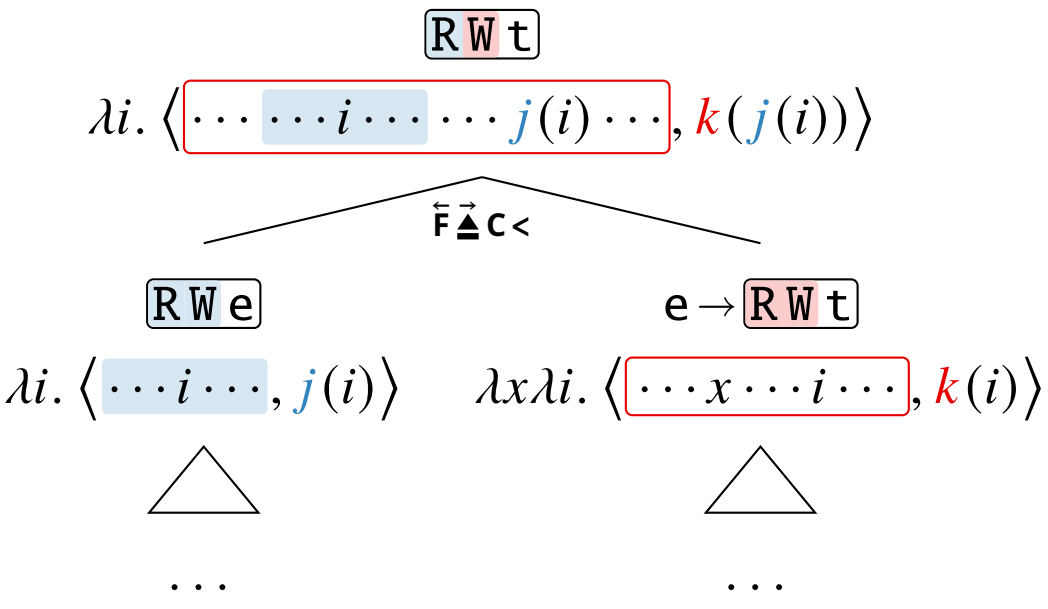

Of course, this is just the bind operator of the state monad $\mathtt{T}$! What we have discovered is that it emerges completely organically from the $\mathtt{W} \dashv \mathtt{R}$ adjunction. Once the modes of combination for $\varepsilon$ and $\Upsilon$ are in place, there is absolutely no need to fuse $\mathtt{R}$ and $\mathtt{W}$ into a single synthetic type $\mathtt{T}$. Any functionality of the latter is perfectly mimicked by the former.

In fact, every adjunction gives rise to a monad via the derivation in (5.38). That is, for any adjunction $\Omega \dashv \Gamma$, the combinator $\overleftarrow{\mathtt{F}}\overrightarrow{\blacktriangle}\mathtt{C}{<}$ with the relevant operators from $\Omega$ and $\Gamma$ is a law-abiding ($\ggg\!=$) for the composite effect $\Gamma\Omega$.

What then of the dynamic type $\mathtt{D} \cong \mathtt{T}^{+}\mathtt{S}$? Let us repeat the experiment, this time with a bit of indeterminacy, `combine`-ing $\mathtt{R}\,\mathtt{S}\,\mathtt{W}\,\mathtt{e}$ and $\mathtt{e} \to \mathtt{R}\,\mathtt{S}\,\mathtt{W}\,\mathtt{t}$.

```
ghci> let lt = Comp (RX E) (Comp SX (Comp (WX E) E))
ghci> let rt = E :-> Comp (RX E) (Comp SX (Comp (WX E) T))
ghci> combine lt rt
[...
(ML (JN (ML (ER (CU BA)))), Comp (RX E) (Comp SX (Comp (WX E) T)))
...]
```

Here too we see that there is already a way to combine these elements to produce a single-layered composition $\mathtt{R}\,\mathtt{S}\,\mathtt{W}\,\mathtt{t}$, namely $\overleftarrow{\mathtt{F}}\mathtt{J}\overleftarrow{\mathtt{F}}\overrightarrow{\blacktriangle}\mathtt{C}{<}$. What then does this fancy combinator do?

(5.39) $\overleftarrow{\mathtt{F}}\mathtt{J}\overleftarrow{\mathtt{F}}\overrightarrow{\blacktriangle}\mathtt{C}{<}\, m\, k = (\lambda s.\, \mu_{\mathrm{S}}\,((\lambda p.\, \varepsilon\,((\lambda a.\, (\lambda f.\, f a) \bullet_{\mathrm{R}} \Upsilon k) \bullet_{\mathrm{W}} p)) \bullet_{\mathrm{S}} s)) \bullet_{\mathrm{R}} m$

by def of $(\bullet_{\mathrm{R}})$ $\quad = \lambda i.\, \mu_{\mathrm{S}}\,((\lambda p.\, \varepsilon\,((\lambda a.\, \lambda i'.\, \Upsilon\, k\, i'\, a) \bullet_{\mathrm{W}} p)) \bullet_{\mathrm{S}} m\, i)$

by def of $\Upsilon$ $\quad = \lambda i.\, \mu_{\mathrm{S}}\,((\lambda p.\, \varepsilon\,((\lambda a.\, \lambda i'.\, k\, a\, i') \bullet_{\mathrm{W}} p)) \bullet_{\mathrm{S}} m\, i)$

by def of $\mu_{\mathrm{S}}$, $(\bullet_{\mathrm{S}})$ $\quad = \lambda i.\, \bigcup\{\varepsilon\,((\lambda a.\, \lambda i'.\, k\, a\, j) \bullet_{\mathrm{W}} \langle x, j\rangle) \mid \langle x, j\rangle \in m\, i\}$

by def of $(\bullet_{\mathrm{W}})$ $\quad = \lambda i.\, \bigcup\{\varepsilon\,\langle \lambda i'.\, k\, x\, i', j\rangle \mid \langle x, j\rangle \in m\, i\}$

by def of $\varepsilon$ $\quad = \lambda i.\, \bigcup\{k\, x\, j \mid \langle x, j\rangle \in m\, i\}$

As might be expected by now, this is exactly the ($\ggg\!=$) of $\mathtt{D}$, according to the definitions in (5.31). Once again, there is simply nothing to be gained by bottling up the three ingredients and re-packaging them as a new kind of computation.

Also once again, the $\overleftarrow{\mathtt{F}}\mathtt{J}\overleftarrow{\mathtt{F}}\overrightarrow{\blacktriangle}\mathtt{C}{<}$ combinator is entirely abstract with respect to the adjunction *as well as* the medial effect $\mathtt{S}$. That sandwiched type could have been any monad whatsoever. In other words, the adjunction precipitates not only the $\mathtt{T}$ monad, but also the $\mathtt{T}^{+}$ monad transformer. Whenever $\Omega \dashv \Gamma$ and $\Sigma$ is a monad, the entire mode of combination will define a law-abiding ($\ggg\!=$) for the composite effect $\Gamma\Sigma\Omega$.

## 5.4 Islands

Some effects may be delimited not by any particular closure operator, but rather by a certain syntactic configuration. Loosely following common parlance in linguistic literature, we will refer to such encapsulating domains as **islands**. For example, the scopes of distributive quantifiers like ‘every’ and ‘no’ are almost always bounded by their enclosing finite clauses, regardless of what other expressions they co-occur with.

It is tempting to associate such scope-delimiting nodes with the presence of syntactically obligatory closure operators. But a moment’s reflection on the discussion in Section 3.3 will show that this does not in general suffice. There, it is already demonstrated how nondeterminism can spill right over the edge of an existential closure operator, permitting the sort of exceptional scope associated with indefinites. With distributive quantifiers, the story is no different. The troublesome derivation would look as in (5.40).

(5.40)

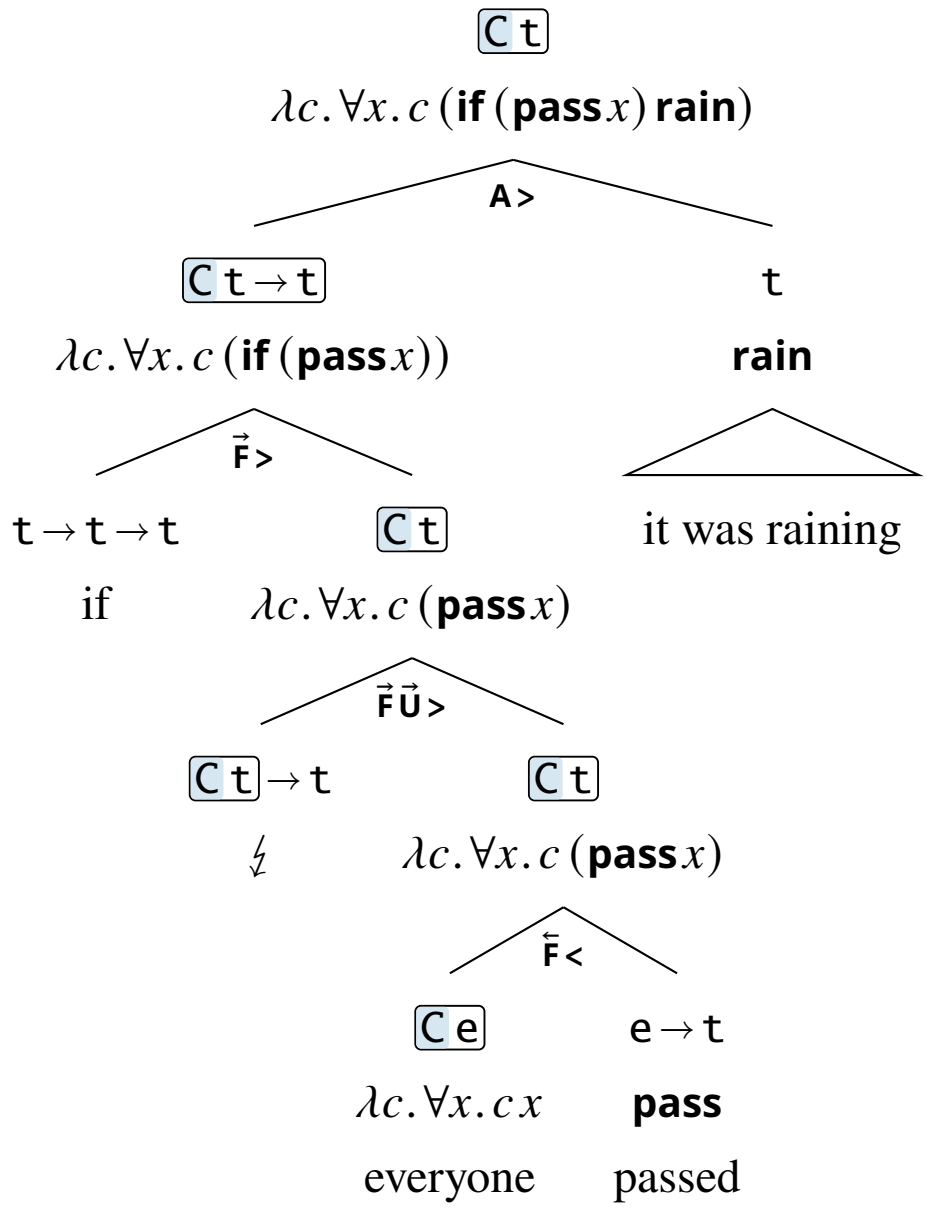

Let’s say the closure operator ϟ has the Lowering semantics assigned by Barker and Shan (2014) in (5.41).

(5.41) $\Downarrow :: \boxed{\mathtt{C}\ \mathtt{t}} \rightarrow \mathtt{t}$

$\Downarrow := \lambda m.\, m\, \textbf{id}$

If it were to apply directly to the conditional antecedent, it would result in the ordinary proposition that is true if everyone passed, false if anyone didn't.

(5.42) $$\begin{aligned} \Downarrow [\![\text{everyone passed}]\!] &= [\![\text{everyone passed}]\!]\,\mathbf{id} \\ &= (\lambda c.\, \forall x.\, c\,(\mathbf{pass}\,x))\,\mathbf{id} \\ &= \forall x.\, \mathbf{pass}\,x \end{aligned}$$

But in (5.40), the operator does not apply directly to the antecedent. Instead, it is mapped over the antecedent by $\vec{\mathbf{F}}$, and then applied to a $\eta$-ified version of the underlying proposition. That underlying proposition is everything in the scope of the continuation, roughly $\mathbf{pass}\,x$. Applying $\eta$ to this yields $\lambda k.\, k\,(\mathbf{pass}\,x)$. And then applying $\Downarrow$ to this LIFTed proposition brings us right back where we started: $\mathbf{pass}\,x$. The net result is a no-op; the antecedent remains a quantificationally-charged computation, type $\boxed{\mathtt{C\,t}}$. The conditional operator is then mapped over this higher-order meaning, and the island is escaped.

How then can narrow scope be enforced? One of the benefits of the effect-theoretic approach we have taken is that nodes' types are revealing of their contents. So far, we have used this information only to determine the ways that two constituents may be combined. But we might just as easily use the information to curtail, or alternatively boost, certain kinds of interpretations. Looking again at the tree in (5.40), we can tell at a glance that something has gone empirically awry because the conditional antecedent has unclosed continuations in it. That is, its type includes the letter $\mathtt{C}$. That's enough to know that some quantifier has not yet closed its scope, and if nothing is done, interpretations will be generated in which that quantifier continues to gobble up, and quantify over, more of its syntactic context.

So one way to implement quantificational islands is simply to filter out any derivations with $\mathtt{C}$-effects in their types. In other words, islandhood may be type-driven. Assuming tensed clauses constitute such an island for quantifiers, the derivation in (5.40) would never arise because the conditional antecedent does not have a valid type for a tensed clause. Fortunately, the types do allow for other derivations, like the obvious one in (5.43).

(5.43)

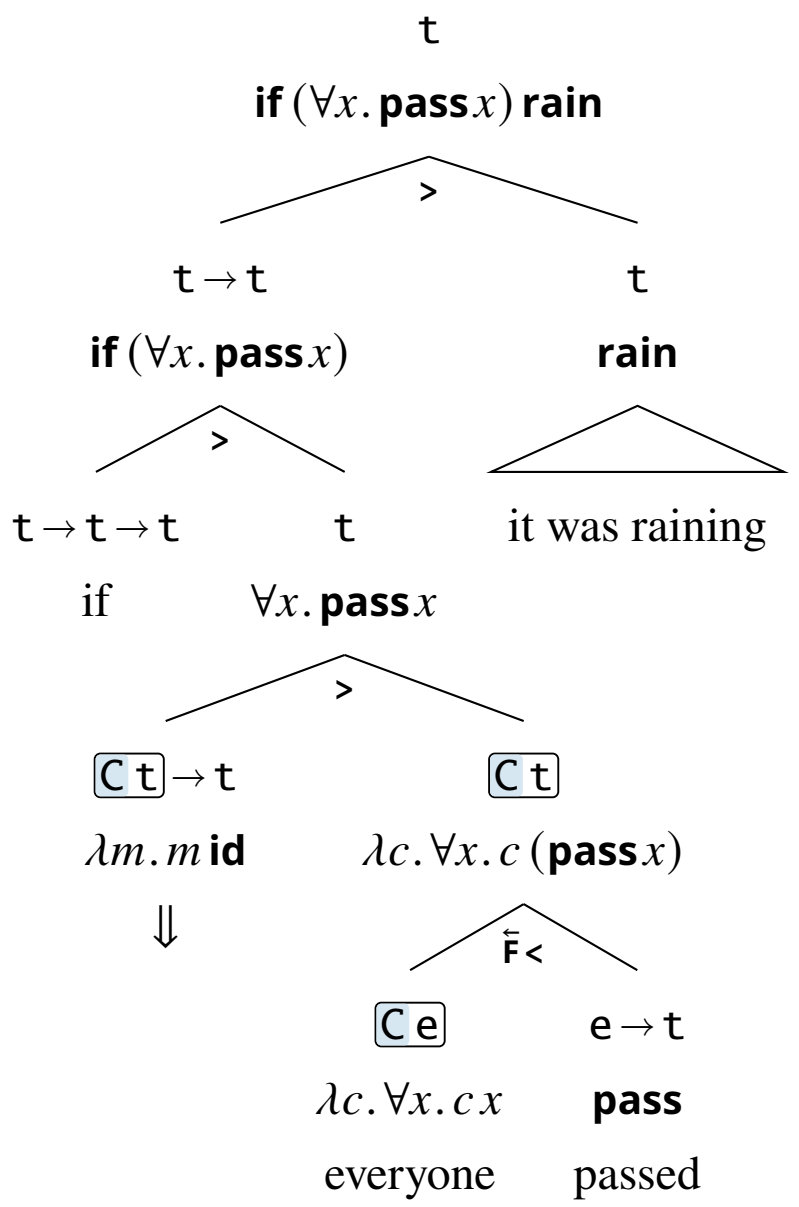

In fact, given the syntactic restriction on the acceptable types of clausal interpretations, the explicit lowering operator is unnecessary. If desired, closure, too, may be type-driven, at least for any operator $\lightning :: \boxed{\Sigma\,\upsilon} \to \xi$ where $\xi$ is effect-free. The reason is that such operators reduce the complexity of their prejacents. They only apply to denotations with particular computational signatures, and in so doing, they strip off the types that trigger them. This guarantees that $\lightning$ can only apply to its own output finitely many times. Incorporating this into the type-driven framework of Figure 10, we might add meta-combinators like (5.44) for any appropriately typed closure operators $\lightning :: \boxed{\Sigma\,\upsilon} \to \xi$.

(5.44) $\mathbf{D}_{\lightning} :: (\sigma \to \tau \to \boxed{\Sigma\,\upsilon}) \to \sigma \to \tau \to \xi$

$\mathbf{D}_{\lightning}\,(*)\,E_1\,E_2 := \lightning\,(E_1 * E_2)$

Suppose that two constituents $E_1 :: \sigma$ and $E_2 :: \tau$ can be combined via $(*)$ to make something of type $\boxed{\Sigma\,\upsilon}$. And further suppose that $\boxed{\Sigma\,\upsilon}$ is a type that can be closed, a complete thought, so to speak. For instance, if $\Sigma = \mathtt{C}$ and $\upsilon = \mathtt{t}$, then the computation is ripe for lowering via the $\Downarrow$ operator defined in (5.41), producing an ordinary proposition of type $\mathtt{t}$. In that case, the rule in (5.44) combines and lowers the two constituents $E_1$ and $E_2$ in one fell swoop, and the de-continuized composition continues.

To be clear, even if a grammar includes type-driven binary closure rules like $\mathbf{D}_{\lightning}$, these do not by themselves enforce island constraints any more than obligatory occurrences of $\lightning$ at island boundaries do. Such rules merely add to the set of possible meanings for the combined constituents. Clearly any two constituents that can be combined via $\mathbf{D}_{\lightning}$ can also be combined via the underlying combinator $(*)$, since all $\mathbf{D}_{\lightning}$ does is apply $\lightning$ after applying $(*)$. For instance, with $\mathbf{D}_{\Downarrow}$ in the grammar, the constituent 'everyone passed' will in principle be ambiguous between a closed type-`t` meaning (composed via $\mathbf{D}(\overleftarrow{\mathbf{F}}<)$) and an unclosed `C t`-meaning (composed via $\overleftarrow{\mathbf{F}}<$). The force of the island comes from the syntax-semantics interface, which upon seeing that this constituent has a designated syntactic status, filters out any interpretations with unclosed types. If this still seems nebulous, see the next section for a concrete implementation of this strategy.

## 5.5 Implementing adjoint effects in the type-driven interpreter

All the scaffolding for adding adjoint and closure operators has already been laid. As ever, we first add variants to the `Mode` representing the **C**, ▲, and **D** operations.

```
data Mode
  = FA | BA | PM         -- etc
  | MR Mode | ML Mode    -- map right and map left
  | AP Mode              -- structured app
  | UR Mode | UL Mode    -- unit right and unit left
  | JN Mode              -- join
  | CU Mode              -- co-unit
  | ER Mode | EL Mode    -- eject right and eject left
  | DN Mode              -- closure
  deriving (Show)
```

Then we add a predicate `adjoint` characterizing the relation between adjoint effects. The only adjunction we treat here is that between `W` and `R`, which are adjoint so long as they read and write the same kind of data. Thus we check that the parameters to the `WX` and `RX` effects are the same.

```
functor, applicative, monad :: EffX -> Bool
functor _              = True
applicative (WX s) = monoid s
applicative f       = functor f && True
monad f             = applicative f && True

monoid :: Ty -> Bool
monoid T = True
monoid _ = False

leftAdj :: EffX -> [EffX]
leftAdj (RX i) = [(WX i)]
leftAdj _      = [ ]

adjoint :: EffX -> EffX -> Bool
adjoint w g = w `elem` leftAdj g
```

The `combine` rule is extended with binary cases for **C** and **▲**, and a unary case for **D**. The binary rule `addCU` looks for a pair of daughters `Comp f s` and `Comp g t` that have adjoint effects. When it finds them, it attempts to combine their underlying types `s` and `t`. For every way `(op, u)` that these types can be combined, a top-level combinator `CU op` is returned with result type `u`.

The ejection rules `ER` and `EL` fire whenever a daughter is a Kleisli arrow into a right adjoint. They simply eject the adjoint effect out of the arrow and try combining again. Note that these are the only recursive rules that do not make at least one type smaller before recursing. In principle, this could lead to non-termination, but clearly neither rule feeds itself (or the other), and since every other rule that changes a type in fact shrinks that type, no cycles can arise.

The unary rule `addDN` looks at the results of a binary combination to see whether it can be closed. For illustrative purposes, we include only the closure rule for continuation effects, $\mathbf{D}_{\Downarrow}$, defined in (5.41). Actually we slightly generalize this to allow for lowering a quantifier at any location where it would be well-typed to do so. That is, we allow $\Downarrow$ to apply whenever we have reached a quantificational type $(\alpha \to \alpha) \to \mathrm{o}$ that might be applied to the identity function, and thus closed. Checking that this is possible is just a matter of checking that the underlying type and intermediate parameter of the `C` effect are the same.

```
combine :: Ty -> Ty -> [(Mode, Ty)]
combine l r = binaryCombs >>= unaryCombs
  where
    binaryCombs =
```

```
      modes l r
      ++ addMR l r ++ addML l r
      ++ addAP l r
      ++ addUR l r ++ addUL l r
      -- if the left and right daughters are adjoint, try
      -- to cancel them out with their co-unit
      ++ addCU l r
      -- if the left or right daughter is a Kleisli arrow
      -- into a right adjoint, try ejecting it
      ++ addER l r ++ addEL l r
    unaryCombs e =
      return e
      ++ addJN e
      -- and if the result type is close-able, close it
      ++ addDN e

addCU l r = case (l, r) of
  (Comp f s, Comp g t) | adjoint f g
    -> [ (CU op, u) | (op, u) <- combine s t ]
  _ -> [                                    ]

addER l r = case r of
  s :-> Comp g t | leftAdj g /= []
    -> [ (ER op, u) | (op, u) <- combine l (Comp g (s :-> t)) ]
  _ -> [                                                      ]

addEL l r = case l of
  s :-> Comp g t | leftAdj g /= []
    -> [ (EL op, u) | (op, u) <- combine (Comp g (s :-> t)) r ]
  _ -> [                                                      ]

addDN e = case e of
  (op, Comp (CX o) a) | o == a
    -> [ (DN op, o) ]
  _ -> [            ]
```

The extended functionality of `combine`, which we've already previewed, is now demonstrated on some key type combinations. First, `W e` and `R e → t` (in that order) give the expected higher-order results, together with a new value that exploits the adjunction `W` ⊣ `R` to derive a pure value of type `t` with (dynamic) binding via **C**.

```
ghci> combine (Comp (WX E) E) (Comp (RX E) (E :-> T))
[(MR (ML BA), Comp (RX E) (Comp (WX E) T)),
 (ML (MR BA), Comp (WX E) (Comp (RX E) T)),
 (CU BA, T)]
```

Symmetrically, we observe that combining `R e` and `W e→t` (in that order) does not result in binding due to the linearly biased formulation of `addCU`.

```
ghci> combine (Comp (RX E) E) (Comp (WX E) (E :-> T))
[(MR (ML BA), Comp (WX E) (Comp (RX E) T)),
 (ML (MR BA), Comp (RX E) (Comp (WX E) T))]
```

Next, we verify that `S W e` and `R S e→t` (in that order) derive type `S t`. The combinator discovered by `combine` recapitulates the last step of (5.12).

```
ghci> let lt = Comp SX (Comp (WX E) E)
ghci> let rt = Comp (RX E) (Comp SX (E :-> T))
ghci> combine lt rt
[ ... -- all possible functorial interleavings of SX, WX, RX, SX
  ... -- plus applicative and monadic combinations of SX, SX
 (JN (ML (CU (MR BA))), Comp SX T) ]
```

Finally, we `combine` `t→t` and `C t` to demonstrate lowering. We observe two possible outputs: an effectful (unlowered) result and a pure (lowered) result of type `t` resulting from a further application of **D**.

```
ghci> combine (T :-> T) (Comp (CX T) T)
[(MR FA, Comp (CX T) T),
 (DN (MR FA), T)]
```

To complete the picture on closure, we implement the discussion of island enforcement from Section 5.4. Again the goal is just to give a proof of concept, showing how effect types can be used to do syntactic work. So we assume that islands correspond to particular syntactic nodes, and that the parser will one way or another have done the work of identifying such boundaries before the type-driven interpreter sets to work. We thus extend the `Syn` type to include a branching node that has been identified as an island for quantifier scopes.

```
data Syn
  = Leaf String
  | Branch Syn Syn
  | Island Syn Syn
```

Next we have to specify which types the island seeks to trap. We do this with the predicate `evaluated`. `C`-type computations are obviously out. These are denotations with quantifiers that are still up in the air, consuming their contexts, so they do not count as `evaluated`. Other effects are passed over, but we recurse into their underlying types to make sure that they are not hiding any embedded un-lowered quantifiers. Likewise with function types, we check that they are not going to spring into life as newly escaped quantifiers as soon as they are saturated with an argument down the road. That leaves only atomic types, which all count as `evaluated`.

```
evaluated :: Ty -> Bool
evaluated t = case t of
  Comp (CX _) _ -> False
  Comp _ a      -> evaluated a
  _ :-> a       -> evaluated a
  _             -> True
```

We put this predicate to work when extending the interpreter `synsem`. The first two cases are the same as before. All that remains is to say how `Island` nodes are interpreted. And here, all we do is interpret them as if they were ordinary branching nodes, and then discard any unevaluated results. That is, any `e` among the results of `synsem (Branch lsyn rsyn)` must have a properly `evaluated` type to escape the island. And that's it!

```
synsem :: Lexicon -> Syn -> [Sem]
synsem lex syn = case syn of
  (Leaf w)           -> [Lex t w | t <- lex w]
  (Branch lsyn rsyn) ->
    [ Comb ty op lsem rsem
      | lsem     <- synsem lex lsyn
      , rsem     <- synsem lex rsyn
      , (op, ty) <- combine (getType lsem) (getType rsem) ]
  (Island lsyn rsyn) ->
    [ e | e <- synsem lex (Branch lsyn rsyn)
        , evaluated (getType e) ]
  where
    getType (Lex ty _) = ty
    getType (Comb ty _ _ _) = ty
```

# 6 Conclusion

## 6.1 Recap

Let us summarize the key ingredients of the approach. Not all expressions of the same syntactic category have the same sort of denotation. In particular, some classes of expressions seem systematically enriched relative to their referential or predicative counterparts. For instance, where a syntactic name identifies a single, determinate entity, a 'wh'-phrase picks out a class of such entities, a definite description defines an entity only in certain situations and defines nothing at all in others, a pronoun must be told who it designates every time it's used, and a quantificational phrase doesn't single out anyone at all, but rather sets up a procedure to count the entities that it could have.

Loosely speaking, we refer to those semantic elements of expressions that exceed the expectations of their compositional contexts as those expressions' computational effects. Mathematically, we model effects as type constructors that endow ordinary types with additional algebraic structure. In this Element, we encapsulated and named the following semantic effects. Note that in the interest of generality, effect constructors are here annotated with parameters specifying the type of environment they expect to interact with. So $\texttt{R}_\iota$ is the type of computations that read from a type-$\iota$ context; $\texttt{W}_\circ$ is the type of computations that store a supplemental type-$\circ$ value; $\texttt{C}_\rho$ the type of computations whose continuations return a value of type $\rho$; and so on.

| Effects: | Expression | Type | Denotation |
|---|---|---|---|
| $\texttt{R}_\iota\,\alpha ::= \iota \to \alpha$ | it | $\texttt{R}_\mathsf{e}\,\mathsf{e}$ | $\lambda x.\, x$ |
| $\texttt{M}\;\alpha ::= \alpha + \bot$ | the planet | $\texttt{M}\;\mathsf{e}$ | $x$ if $\textbf{planet} = \{x\}$ else $\#$ |
| $\texttt{W}_\circ\,\alpha ::= \alpha \times \circ$ | Jupiter, a planet | $\texttt{W}_\mathsf{t}\,\mathsf{e}$ | $\langle \textbf{j}, \textbf{planet}\,\textbf{j} \rangle$ |
| $\texttt{S}\;\alpha ::= \{\alpha\}$ | which planet | $\texttt{S}\;\mathsf{e}$ | $\{x \mid \textbf{planet}\,x\}$ |
| $\texttt{C}_\rho\,\alpha ::= (\alpha \to \rho) \to \rho$ | no planet | $\texttt{C}_\mathsf{t}\,\mathsf{e}$ | $\lambda c.\, \neg\exists x.\, \textbf{planet}\,x \wedge c\,x$ |
| $\texttt{F}\;\alpha ::= \alpha \times \{\alpha\}$ | JUPITER | $\texttt{F}\;\mathsf{e}$ | $\langle \textbf{j}, \{x \mid x \sim \textbf{j}\} \rangle$ |
| $\texttt{T}_\sigma\,\alpha ::= \sigma \to (\alpha \times \sigma)$ | as for Jupiter | $\texttt{T}_{\vec{\mathsf{e}}}\,\mathsf{e}$ | $\lambda s.\, \langle \textbf{j}, \textbf{j} \mathbin{+\!\!+} s \rangle$ |
| $\texttt{D}_\sigma\,\alpha ::= \sigma \to \{\alpha \times \sigma\}$ | a planet | $\texttt{D}_{\vec{\mathsf{e}}}\,\mathsf{e}$ | $\lambda s.\, \{\langle x, x \mathbin{+\!\!+} s \rangle \mid \textbf{planet}\,x\}$ |

For each of these constructors, there are functions $(\bullet)$, $(\circledast)$, $\eta$, $\mu$, and $(\gg\!=)$ that satisfy the functor, applicative, and monad laws (with the usual proviso that the $\circ$ parameterizing $\texttt{W}_\circ$ computations must be monoidal). For $\texttt{W}_\iota$ and $\texttt{R}_\iota$ with matching parameters, there are additionally functions $\Phi$ and $\Psi$ satisfying the adjunction laws, and from these, $\varepsilon$ canceling the two effects, and $\Upsilon$ distributing an effect over an argument. These various operations are collected with their type classes and signatures in Figure 11.

| Class | Operators and types | |
|---|---|---|
| **Functor** $\Sigma$ | $(\bullet) :: (\alpha \to \beta) \to \boxed{\Sigma\,\alpha} \to \boxed{\Sigma\,\beta}$ | |
| **Applicative** $\Sigma$ | $\eta :: \alpha \to \boxed{\Sigma\,\alpha}$ | |
| | $(\circledast) :: \boxed{\Sigma\,\alpha \to \beta} \to \boxed{\Sigma\,\alpha} \to \boxed{\Sigma\,\beta}$ | |
| **Monad** $\Sigma$ | $(\gg\!=) :: \boxed{\Sigma\,\alpha} \to (\alpha \to \boxed{\Sigma\,\beta}) \to \boxed{\Sigma\,\beta}$ | $(=\!\ll) :: (\alpha \to \boxed{\Sigma\,\beta}) \to \boxed{\Sigma\,\alpha} \to \boxed{\Sigma\,\beta}$ |
| | $\mu :: \boxed{\Sigma\,\boxed{\Sigma\,\alpha}} \to \boxed{\Sigma\,\alpha}$ | |
| **Adjoint** $\Omega\,\Gamma$ | $\Phi :: (\boxed{\Omega\,\alpha} \to \beta) \to \alpha \to \boxed{\Gamma\,\beta}$ | $\Psi :: (\alpha \to \boxed{\Gamma\,\beta}) \to \boxed{\Omega\,\alpha} \to \beta$ |
| | $\varepsilon :: \boxed{\Omega\,\boxed{\Gamma\,\alpha}} \to \alpha$ | $\eta :: \alpha \to \boxed{\Gamma\,\boxed{\Omega\,\alpha}}$ |
| | $\Upsilon :: (\alpha \to \boxed{\Gamma\,\beta}) \to \boxed{\Gamma\,\alpha \to \beta}$ | |

**Figure 11** Algebraic operations

Built around these various functions, we defined an inventory of semantic combinators to compose effectful expressions in a principled and modular way. The complete architecture is laid out in Figure 12. Included in this grammar is a single **D** rule based on the Lowering semantics in (5.41), but really it is a template for whatever closure operators a particular theory calls for.

We should emphasize here that, while we would be delighted for you to take on Figure 12 wholesale, this need not be a package deal. Certainly, various subsets of these meta-combinators make for coherent grammars (as already discussed in Section 3.4), and some may be explanatorily tailored to various phenomena, empirical domains, and theoretical priors. If you're working in an area where Kleisli arrows never come up, you can safely bracket monads and ejection. If you're happy to assume an unbounded number of silent closure operators or effect handlers in the left periphery, you may be able to get by with mapping alone. Nevertheless, because Kleisli arrows do have a way of showing up (which speaks in favor of monads, and hence of applicatives, functors, and ejection), and because there are strong arguments that binding in natural language is inherently dynamic (which speaks in favor of adjunctions), we think that the full grammar of Figure 12 is a natural starting point for effects-oriented linguistic theorizing.

**Types:**

| | |
|---|---|
| $\tau ::= \mathtt{e} \mid \mathtt{t} \mid \cdots$ | Base types |
| $\mid\ \tau \to \tau \mid \tau \times \tau \mid \tau + \tau \mid \{\tau\} \mid \cdots$ | Compound types |
| $\mid\ \boxed{\Sigma\,\tau}$ | Computation types |

**Effects:**

$\Sigma ::= \mathtt{S} \mid \mathtt{R}_\iota \mid \mathtt{W}_o \mid \mathtt{C}_\rho \mid \cdots$

Indeterminacy, Input, Output, Quantification, . . .

**Basic Combinators:**

| | |
|---|---|
| $(>) :: (\alpha \to \beta) \to \alpha \to \beta$<br>$f > x \coloneqq f\,x$ | Forward Application |
| $(<) :: \alpha \to (\alpha \to \beta) \to \beta$<br>$x < f \coloneqq f\,x$ | Backward Application |
| $(\sqcap) :: (\alpha \to \mathtt{t}) \to (\alpha \to \mathtt{t}) \to (\alpha \to \mathtt{t})$<br>$f \sqcap g \coloneqq \lambda x.\ f\,x \wedge g\,x$ | Predicate Modification |
| ... | ... |

**Meta-combinators:**

| | |
|---|---|
| $\overleftarrow{\mathsf{F}} :: (\sigma \to \tau \to \upsilon) \to \boxed{\Sigma\,\sigma} \to \tau \to \boxed{\Sigma\,\upsilon}$<br>$\overleftarrow{\mathsf{F}}\,(*)\,E_1\,E_2 \coloneqq (\lambda a.\ a * E_2) \bullet E_1$ | Map Left |
| $\overrightarrow{\mathsf{F}} :: (\sigma \to \tau \to \upsilon) \to \sigma \to \boxed{\Sigma\,\tau} \to \boxed{\Sigma\,\upsilon}$<br>$\overrightarrow{\mathsf{F}}\,(*)\,E_1\,E_2 \coloneqq (\lambda b.\ E_1 * b) \bullet E_2$ | Map Right |
| $\mathsf{A} :: (\sigma \to \tau \to \omega) \to \boxed{\Sigma\,\sigma} \to \boxed{\Sigma\,\tau} \to \boxed{\Sigma\,\omega}$<br>$\mathsf{A}\,(*)\,E_1\,E_2 \coloneqq (\lambda a \lambda b.\ a * b) \bullet E_1 \circledast E_2$ | Structured App |
| $\overleftarrow{\mathsf{U}} :: (\sigma \to (\tau \to \tau') \to \upsilon) \to \sigma \to (\boxed{\Sigma\,\tau} \to \tau') \to \upsilon$<br>$\overleftarrow{\mathsf{U}}\,(*)\,E_1\,E_2 \coloneqq E_1 * (\lambda b.\ E_2\,(\eta\,b))$ | Unit Left |
| $\overrightarrow{\mathsf{U}} :: ((\sigma \to \sigma') \to \tau \to \upsilon) \to (\boxed{\Sigma\,\sigma} \to \sigma') \to \tau \to \upsilon$<br>$\overrightarrow{\mathsf{U}}\,(*)\,E_1\,E_2 \coloneqq (\lambda a.\ E_1\,(\eta\,a)) * E_2$ | Unit Right |
| $\mathsf{J} :: (\sigma \to \tau \to \boxed{\Sigma\,\Sigma\,\omega}) \to \sigma \to \tau \to \boxed{\Sigma\,\omega}$<br>$\mathsf{J}\,(*)\,E_1\,E_2 \coloneqq \mu\,(E_1 * E_2)$ | Join |
| $\mathsf{C} :: (\sigma \to \tau \to \omega) \to \boxed{\Omega\,\sigma} \to \boxed{\Gamma\,\tau} \to \omega$<br>$\mathsf{C}\,(*)\,E_1\,E_2 \coloneqq \varepsilon\,((\lambda l.\ (\lambda r.\ l * r) \bullet E_2) \bullet E_1)$ | Co-unit |
| $\overleftarrow{\blacktriangle} :: (\boxed{\Gamma\,\sigma \to \sigma'} \to \tau \to \upsilon) \to (\sigma \to \boxed{\Gamma\,\sigma'}) \to \tau \to \upsilon$<br>$\overleftarrow{\blacktriangle}\,(*)\,E_1\,E_2 \coloneqq \Upsilon\,E_1 * E_2$ | Eject Left |
| $\overrightarrow{\blacktriangle} :: (\sigma \to \boxed{\Gamma\,\tau \to \tau'} \to \upsilon) \to \sigma \to (\tau \to \boxed{\Gamma\,\tau'}) \to \upsilon$<br>$\overrightarrow{\blacktriangle}\,(*)\,E_1\,E_2 \coloneqq E_1 * \Upsilon\,E_2$ | Eject Right |
| $\mathsf{D}_{\Downarrow} :: (\sigma \to \tau \to \mathtt{C}_\upsilon\,\upsilon) \to \sigma \to \tau \to \upsilon$<br>$\mathsf{D}_{\Downarrow}\,(*)\,E_1\,E_2 \coloneqq \Downarrow(E_1 * E_2)$ | Lower |

**Figure 12** Complete effect-driven grammar

In tandem with these theoretical proposals, we offered a simple computational implementation in the form of a type-driven semantic parser coded in Haskell. This parser interprets syntactic structures by recursively interpreting their parts (this part is standard), and also by recursively peeling back and combining layers of functorial, applicative, monadic, and adjoint effects (this part is new). This algorithm is guaranteed to find every possible interpretation that a provided sentence can have according to our theory (and none that it doesn't).

This semantic parser powers the interactive application that can be accessed at `https://dylanbumford.com/effects.html`. We have not, however, discussed many of the bells and whistles that make this application practical, useful, and (we hope) fun: a reasonable lexicon associating words with syntactic categories and semantic types; a grammar governing syntactic formation, alongside a syntactic parser with standard optimizations for efficiency; optimizations of `combine` that avoid repeated computation on recursive calls (via memoization) and spurious ambiguity (via a notion of normal-form derivation); and routines for evaluating and normalizing the objects produced by the semantic parser. Readers interested in these aspects of a computational implementation may view the complete source code at `https://github.com/schar/TDParse`.

## 6.2 Outlook

Why bother abstracting out these algebraic patterns? What is actually gained over, say, a bunch of independent grammars tailored to their specific phenomena of interest? For instance, the semantic literature includes many proposals for composing rich denotations that mimic almost exactly the applicative modes of combination for various effects (e.g., for sets of values (Hamblin 1973), assignment-dependent values (Cooper 1975), values that quantify over their contexts (Barker 2002), values with supplemental content (Portner 2007), values embedded within focus alternatives (Rooth 1985), values of dubious definedness (Heim and Kratzer 1998), context-transforming values (Bumford 2015), etc.). So what exactly is the point of introducing the **A>** and **A<** combinators that are polymorphic over these various compositional schemes?

For starters, the higher-order applicative operators do recapitulate those various effect-wise modes of combination, so we certainly don't lose anything relative to the isolated grammars. And even if all we gained was a recognition that there are algebraic patterns common to all of these grammars, that is still illuminating. Equational reasoning can be very useful in practice, as seen in many places throughout this Element, e.g., (4.40).

But what's more, the recursive formulation of the combinatory inventory empowers the grammar to handle meanings of arbitrary computational depth. On the low end, this flexibility means we are not forced to assign the word 'cat' a dozen or so different interpretations — an ordinary denotation, a focus denotation, an expressive denotation, a continuized denotation, etc. In other words, composition does not require us to pretend that 'cat' denotes a fancy computation of any sort, nor are we required to coerce it into one in order to put it together with its surroundings. It just denotes the property **cat** in any compositional context.

In the other direction, the functorial combinators allow computations to nest within and slide over one another. This alone predicts the exceptional syntactic distance over which these effects may be felt (Section 2.4). These layered computations then enable expressions to associate with effects selectively, as when two focus-sensitive operators target two different foci, or when two indefinites take differentiated scopes with respect to some island-external operator (Section 3.3). Obversely, monadic combinators allow layered computations to be flattened, thereby synthesizing their effects. This makes it possible to provide categorematic definitions of Kleisli-typed expressions like determiners (Section 4.2.1), and additionally opens the door to both transformational and non-transformational mechanisms of effect inversion (Section 4.3).

Finally, and importantly, the effect-driven system developed here is inherently modular and fundamentally extensible. First, the effect-driven interpreter is agnostic as to the basic modes of composition that operate on effect-free types. And second, the lexical semantics of irrelevant vocabulary items nor the methods of composition need to be re-defined, or worse, re-imagined, every time the grammar is extended to accommodate some novel phenomenon. In the words of Cartwright and Felleisen (1994), an important antecedent for the approach to interpretation advocated here, this modularity allows us to accommodate "orthogonal extensions of a language without changing the denotations of existing phrases" and to "construct interpreters for complete languages by composing interpreters for language fragments". On the view we've set out here, the grammar itself is compositionally constituted — the sum of maximally simple parts, and sometimes (as with adjunctions) more. From a practical point of view, this modularity also allows independent researchers to work on independent problems and rest assured that the semantic constructs they develop are interoperable. This is perhaps an underappreciated methodological virtue in a nascent, exploratory field like natural language semantics where new mathematical structures pop up all the time.

# Appendix A

## *Implementations of combinatoric operations*

### A1 Types

Here we give Haskell encodings of the effect types used in this Element. As indicated in the main text, many of these mimic data types that are defined in Haskell's prelude and standard libraries. We spoof them here to match the notation in the Element and to simplify their presentation.

```
data R i a = R (i -> a)             -- Haskell's `Reader`
data W o a = W (a, o)               -- Haskell's `Writer`
data M a   = Valid a | Failure      -- Haskell's `Maybe`
data T s a = T (s -> (a, s))        -- Haskell's `State`
data D s a = D (s -> [(a, s)])      -- Haskell's `StateT []`
data C r a = C ((a -> r) -> r)      -- Haskell's `Cont`
data S a   = S [a]                  -- Haskell's `[ ]`
data F a   = F (a, [a])             -- Haskell's `Product Identity []`
```

### A2 Algebraic classes

The following type classes define the signatures for the algebraic operations in Figure 11. The `Functor`, `Applicative`, and `Monad` classes, and the attendent operators, are brought in scope by default in any Haskell program.

```
class Functor f where
  fmap :: (a -> b) -> f a -> f b

class Functor f => Applicative f where
  pure :: a -> f a
  (<*>) :: f (a -> b) -> f a -> f b

class Applicative f => Monad f where
  (>>=) :: f a -> (a -> f b) -> f b
  return :: a -> f a
  return = pure

join :: Monad f => f (f a) -> f a
join m = m >>= id

(=<<) :: (a -> f b) -> f a -> f b
k =<< m = m >>= k
```

The `Adjoint` class is similar to that in the `adjunctions` package. Note that this class, unlike the preceding ones, defines a relation between constructors rather than a predicate of constructors (in fact, it defines a function as adjoints are unique). Because this makes type inference harder, it requires a "pragma" extending the accepted language.

```
{-# LANGUAGE FunctionalDependencies #-}
class (Functor f, Functor g) => Adjoint f g
  | f -> g, g -> f -- f and g uniquely determine each other
                   -- this helps with type inference
  where
  phi    :: (f a -> b) -> a -> g b
  psi    :: (a -> g b) -> f a -> b
  unit   :: a -> g (f a)
  counit :: f (g a) -> a

  -- unit/counit and phi/psi are interdefinable; an instance
  -- of Adjoint need only declare one or the other pair:
  unit   = phi id
  counit = psi id
  phi c  = fmap c . unit
  psi k  = counit . fmap k

eject :: Adjoint f g => (a -> g b) -> g (a -> b)
eject k = phi (\w a -> psi (\_ -> k a) w) ()
```

## A3 Functor instances

For each effect, we define a law-abiding `fmap` implementing the (•) operation assumed in Chapter 2.

```
instance Functor (R i) where
  fmap k (R m) =
    R (\i -> k (m i))

instance Functor (W o) where
  fmap k (W m) =
    W (k (fst m), snd m)

instance Functor M where
  fmap k m =
    case m of
      Valid a -> Valid (k a)
      Failure -> Failure

instance Functor (T s) where
  fmap k (T m) =
    T (\s -> let (a, t) = m s in (k a, t))

instance Functor (D s) where
  fmap k (D m) =
    D (\s -> let outs = m s in [(k a, t) | (a, t) <- outs])

instance Functor (C r) where
  fmap k (C m) =
    C (\c -> m (\a -> c (k a)))

instance Functor S where
  fmap k (S m) =
    S [k a | a <- m]

instance Functor F where
  fmap k (F m) =
    F (k (fst m), [k a | a <- snd m])
```

## A4 Applicative instances

For each effect, we define a law-abiding `<*>`, implementing the (⊛) operation of Chapter 3.

```
instance Applicative (R i) where
  pure x = R (\i -> x)
  (R ff) <*> (R xx) =
    R (\i -> ff i (xx i))

instance Monoid o => Applicative (W o) where
  pure x = W (x, mempty)
  (W ff) <*> (W xx) =
    W (fst ff (fst xx), snd ff <> snd xx)

instance Applicative M where
  pure x = Valid x
  ff <*> xx =
    case (ff, xx) of
      (Valid f, Valid x) -> Valid (f x)
      (_      , _      ) -> Failure

instance Applicative (T s) where
  pure x = T (\s -> (x, s))
  (T ff) <*> (T xx) =
    T (\s -> let (f, t) = ff s
                 (x, u) = xx t
             in (f x, u))

instance Applicative (D s) where
  pure x = D (\s -> [(x, s)])
  (D ff) <*> (D xx) =
    D (\s -> [(f x, u) | (f, t) <- ff s, (x, u) <- xx t])

instance Applicative (C r) where
  pure x = C (\c -> c x)
  (C ff) <*> (C xx) =
    C (\c -> ff (\f -> xx (\x -> c (f x))))

instance Applicative S where
  pure x = S (pure x)
  (S ff) <*> (S xx) =
    S [f x | f <- ff, x <- xx]

instance Applicative F where
  pure x = F (x, [x])
  (F ff) <*> (F xx) =
```

```
  F (fst ff (fst xx), [f x | f <- snd ff, x <- snd xx])
```

## A5 Monad instances

For each effect, we define a law-abiding `>>=`, implementing the (≫=) operation of Chapter 4.

```
instance Monad (R i) where
  (R m) >>= k =
    R (\i -> let R n = k (m i) in n i)

instance Monoid o => Monad (W o) where
  (W m) >>= k =
    W (let W n = k (fst m) in (fst n, snd m <> snd n))

instance Monad M where
  m >>= k =
    case m of
      Valid a -> k a
      Failure -> Failure

instance Monad (T s) where
  (T m) >>= k =
    T (\s -> let (a, t) = m s
                 T n = k a
             in n t)

instance Monad (D s) where
  (D m) >>= k =
    D (\s -> [(b, u) | (a, t) <- m s
                     , let D n = k a
                     , (b, u) <- n t])

instance Monad (C r) where
  (C m) >>= k =
    C (\c -> m (\x -> let C n = k x in n c))

instance Monad S where
  (S m) >>= k =
    S [b | x <- m, let S n = k x, b <- n]

instance Monad F where
  (F m) >>= k =
    F ( let F n = k (fst m) in fst n
      , [b | a <- snd m, let F n = k a, b <- snd n])
```

## A6 Adjunction instances

Finally, the adjunction between `W` and `R`, relied upon in Chapter 5.

```
instance Adjoint (W io) (R io) where
  unit a = R (\io -> W (a, io))
  counit (W (R m, io)) = m io
```

# Appendix B

## *The complete type-driven interpreter*

Finally, we implement the grammar of Figure 12 as a type-driven semantic parser. The code below merely collects the snippets defined throughout the text.

```
-- representations of types
data Ty
  = TyE | TyT | TyV -- base types
  | Ty :-> Ty       -- function types
  -- other compound types, as desired
  | Comp EffX Ty    -- computation types
  deriving (Eq, Show)

-- representations of effect constructors
data EffX
  = SX        -- computations with indeterminate results
  | RX Ty     -- computations that query an environment of type Ty
  | WX Ty     -- computations that store information of type Ty
  | CX Ty     -- computations that quantify over Ty contexts
  -- and so on for other effects, as desired
  deriving (Eq, Show)

-- predicates characterizing the algebraic properties of the EffX
functor, applicative, monad :: EffX -> Bool
functor _            = True
applicative (WX o) = monoid o
applicative f       = functor f && True
monad f              = applicative f && True

monoid :: Ty -> Bool
monoid TyT = True
monoid _   = False

leftAdj :: EffX -> [EffX]
leftAdj (RX i) = [(WX i)]
leftAdj _      = [ ]

adjoint :: EffX -> EffX -> Bool
adjoint w g = w `elem` leftAdj g
```

```
-- an inventory of combinatory modes
data Mode
  = FA | BA | PM        -- basic modes
  | MR Mode | ML Mode  -- map right and map left
  | AP Mode            -- structured app
  | UR Mode | UL Mode  -- unit right and unit left
  | JN Mode            -- join
  | CU Mode            -- co-unit
  | ER Mode | EL Mode  -- eject right and eject left
  | DN Mode            -- continuation closure
  deriving (Show)

-- a lexicon is a dictionary of types
type Lexicon = String -> [Ty]

-- syntactic objects to be interpreted
data Syn
  = Leaf String
  | Branch Syn Syn
  | Island Syn Syn

-- semantic objects describing an interpretation
data Sem
  = Lex Ty String
  | Comb Ty Mode Sem Sem

-- the recursive interpreter
synsem :: Lexicon -> Syn -> [Sem]
synsem lex syn = case syn of
  (Leaf w)           -> [Lex t w | t <- lex w]
  (Branch lsyn rsyn) ->
    [ Comb ty op lsem rsem
      | lsem     <- synsem lex lsyn
      , rsem     <- synsem lex rsyn
      , (op, ty) <- combine (getType lsem) (getType rsem) ]
  (Island lsyn rsyn) ->
    [ e | e <- synsem lex (Branch lsyn rsyn)
        , evaluated (getType e) ]
  where
    getType (Lex ty _) = ty
    getType (Comb ty _ _ _) = ty
    evaluated t = case t of
      Comp (CX _) _ -> False
```

```
      Comp _ a      -> evaluated a
      _ :-> a       -> evaluated a
      _             -> True

-- basic modes of combination
modes :: Ty -> Ty -> [(Mode, Ty)]
modes l r = case (l, r) of
  (a :-> b  , _      )   | r == a -> [(FA, b)]
  (_        , a :-> b)   | l == a -> [(BA, b)]
  (a :-> TyT, b :-> TyT) | a == b -> [(PM, a :-> TyT)]
  _                               -> [ ]

-- the logic of applying higher-order modes to types
combine :: Ty -> Ty -> [(Mode, Ty)]
combine l r = binaryCombs >>= unaryCombs
  where
    binaryCombs =
      modes l r
      ++ addMR l r ++ addML l r
      ++ addAP l r
      ++ addUR l r ++ addUL l r
      ++ addCU l r
      ++ addER l r ++ addEL l r
    unaryCombs e =
      return e
      ++ addJN e
      ++ addDN e

-- if the right daughter is functorial, try to map over it
addMR l r = case r of
  Comp f t | functor f
    -> [ (MR op, Comp f u) | (op, u) <- combine l t ]
  _ -> [                                            ]

-- if the left daughter is functorial, try to map over it
addML l r = case l of
  Comp f s | functor f
    -> [ (ML op, Comp f u) | (op, u) <- combine s r ]
  _ -> [                                            ]

-- if both daughters are applicative, try structured application
addAP l r = case (l, r) of
  (Comp f s, Comp g t) | f == g, applicative f
```

```
    -> [ (AP op, Comp f u) | (op, u) <- combine s t ]
  _ -> [                                            ]

-- if the left daughter closes an applicative effect,
-- try to purify the right daughter
addUR l r = case l of
  Comp f s :-> s' | applicative f
    -> [ (UR op, u) | (op, u) <- combine (s :-> s') r ]
  _ -> [                                              ]

-- if the right daughter closes an applicative effect,
-- try to purify the left daughter
addUL l r = case r of
  Comp f t :-> t' | applicative f
    -> [ (UL op, u) | (op, u) <- combine l (t :-> t') ]
  _ -> [                                              ]

-- if the left and right daughters are adjoint, try co-unit
addCU l r = case (l, r) of
  (Comp f s, Comp g t) | adjoint f g
    -> [ (CU op, u) | (op, u) <- combine s t ]
  _ -> [                                     ]

-- if the right daughter is a kleisli arrow into a distributive
-- effect, try ejecting the effect
addER l r = case r of
  s :-> Comp g t | leftAdj g /= [ ]
    -> [ (ER op, u) | (op, u) <- combine l (Comp g (s :-> t)) ]
  _ -> [                                                      ]

-- if the left daughter is a kleisli arrow into a distributive
-- try ejecting the effect
addEL l r = case l of
  s :-> Comp g t | leftAdj g /= [ ]
    -> [ (EL op, u) | (op, u) <- combine (Comp g (s :-> t)) r ]
  _ -> [                                                      ]

-- if a result of combination has a layered structure, try
-- joining it
addJN e = case e of
  (op, Comp f (Comp g a)) | f == g, monad f
    -> [ (JN op, Comp f a) ]
  _ -> [                   ]
```

```
-- if a result of combination can be closed, try closing it
addDN e = case e of
  (op, Comp (CX r) a) | r == a
    -> [ (DN op, r) ]
  _ -> [             ]
```

# Acknowledgements

# References

Ades, A. E., & Steedman, M. J. (1982). On the order of words. *Linguistics and philosophy*, *4*(4), 517–558.

Aloni, M. (2007). Free choice, modals, and imperatives. *Natural Language Semantics*, *15*(1), 65–94.

Alonso-Ovalle, L. (2009). Counterfactuals, correlatives, and disjunction. *Linguistics and Philosophy*, *32*(2), 207–244.

Asudeh, A., & Giorgolo, G. (2020). *Enriched meanings: Natural language semantics with Category Theory*. Oxford University Press.

Barker, C. (2002). Continuations and the nature of quantification. *Natural Language Semantics*, *10*(3), 211–242.

Barker, C., & Shan, C.-c. (2014). *Continuations and natural language*. Oxford University Press.

Barwise, J., & Cooper, R. (1981). Generalized quantifiers and natural language. *Linguistics and Philosophy*, *4*(2), 159-219.

Beaver, D., & Krahmer, E. (2001). A partial account of presupposition projection. *Journal of Logic, Language and Information*, *10*, 147–82.

Beck, J. (1969). Distributive laws. In *Seminar on triples and categorical homology theory* (pp. 119–140).

Beck, S. (2006). Intervention effects follow from focus interpretation. *Natural Language Semantics*, *14*(1), 1–56. doi: 10.1007/s11050-005-4532-y

Bittner, M. (2001). Topical referents for individuals and possibilities. In *Semantics and linguistic theory (SALT) 11* (pp. 36–55).

Bumford, D. (2015). Incremental quantification and the dynamics of pair-list phenomena. *Semantics and Pragmatics*, *8*(9), 1–70.

Büring, D. (2005). *Binding theory*. Cambridge University Press.

Cartwright, R., & Felleisen, M. (1994). Extensible denotational language specifications. In M. Hagiya & J. C. Mitchell (Eds.), *Theoretical aspects of computer software* (pp. 244–272). Berlin, Heidelberg: Springer Berlin Heidelberg.

Charlow, S. (2014). *On the semantics of exceptional scope* (PhD Dissertation). New York University, New York, NY.

Charlow, S. (2018). A modular theory of pronouns and binding. In *Logic and engineering of natural language semantics (LENLS) 14*.

Charlow, S. (2020). The scope of alternatives: Indefiniteness and islands. *Linguistics and Philosophy*, *43*, 427–472.

Charlow, S. (2022). On Jacobson's "Towards a variable-free semantics". In L. McNally & Z. G. Szabó (Eds.), *A reader's guide to classic papers* *in formal semantics* (Vol. 100, pp. 171–196). Springer International Publishing.

Chierchia, G., & McConnell-Ginet, S. (1990). *Meaning and grammar: An introduction to semantics*. Cambridge, MA: MIT Press.

Chung, S., & Ladusaw, W. (2003). *Restriction and saturation*. Cambridge, MA: MIT press.

Church, A. (1940). A formulation of the simple theory of types. *The journal of symbolic logic*, *5*(2), 56–68.

Cooper, R. (1975). *Montague's semantic theory and Transformational Grammar* (PhD Dissertation). University of Massachusetts, Amherst.

de Groote, P. (2001). Type raising, continuations, and classical logic. In R. van Rooij & M. Stokhof (Eds.), *The thirteenth Amsterdam Colloquium* (p. 97-101).

Dekker, P. (1994). Predicate logic with anaphora. In M. Harvey & L. Santelmann (Eds.), *Semantics and linguistic theory (SALT) 4* (pp. 79–95).

Dowty, D. (1988). Type raising, functional composition, and non-constituent conjunction. In *Categorial grammars and natural language structures* (pp. 153–197). Dordrecht, The Netherlands: Springer.

Fodor, J. D. (1982). The mental representation of quantifiers. In *Processes, beliefs, and questions: Essays on formal semantics of natural language and natural language processing* (pp. 129–164). Springer.

Geach, P. (1972). *Logic matters*. University of California Press.

Goldstein, S. (2019). Free choice and homogeneity. *Semantics and Pragmatics*, *12*(23), 1–53.

Groenendijk, J., & Stokhof, M. (1984). *Studies on the semantics of questions and the pragmatics of answers* (PhD Dissertation). Universiteit van Amsterdam.

Groenendijk, J., & Stokhof, M. (1988). Context and information in dynamic semantics. In H. Bouma & B. A. G. Elsendoorm (Eds.), *Working models of human perception* (pp. 457–486). London: Academic Press.

Groenendijk, J., & Stokhof, M. (1991). Dynamic predicate logic. *Linguistics and Philosophy*, *14*(1), 39–100.

Grosz, B., Joshi, A., & Weinstein, S. (1995). Centering: A framework for modelling the local coherence of discourse. *Computational Linguistics*, *21*(2), 203–225.

Hagstrom, P. (1998). *Decomposing questions* (PhD Dissertation). Massachusetts Institute of Technology, Cambridge, MA.

Hamblin, C. L. (1973). Questions in Montague English. *Foundations of Language*, *10*(1), 41-53.

Heim, I. (1982). *The semantics of definite and indefinite noun phrases* (PhD

Dissertation). University of Massachusetts, Amherst.

Heim, I. (1983). On the projection problem for presuppositions. In M. Barlow, D. P. Flickinger, & M. T. Wescoat (Eds.), *Proceedings of the Second West Coast Conference on Formal Linguistics* (p. 114-125). Stanford: Stanford University Press.

Heim, I., & Kratzer, A. (1998). *Semantics in generative grammar.* Oxford: Blackwell.

Hutton, G. (2016). *Programming in Haskell.* Cambridge University Press.

Jacobson, P. (1999). Towards a variable-free semantics. *Linguistics and Philosophy*, *22*(2), 117-184. doi: 10.1023/A:1005464228727

Jacobson, P. (2014). *Compositional semantics: An introduction to the syntax/semantics interface.* Oxford Textbooks in Linguistics.

Jones, M. P., & Duponcheel, L. (1993). *Composing monads* (Technical Report No. YALEU/DCS/RR-1004). New Haven: Yale University, Department of Computer Science.

Kamp, H. (1975). Two theories about adjectives. In E. Keenan (Ed.), *Formal semantics of natural language* (pp. 123–155). Cambridge University Press.

King, D. J., & Wadler, P. (1993). Combining monads. In J. Launchbury & P. Sansom (Eds.), *Functional programming, Glasgow 1992* (pp. 134–143). London: Springer London.

Kiselyov, O. (2015). Applicative abstract categorial grammars in full swing. In *Logic and engineering of natural language semantics (LENLS) 11* (pp. 66–78).

Kiselyov, O., & Shan, C.-c. (2014). Continuation hierarchy and quantifier scope. In *Formal approaches to semantics and pragmatics* (pp. 105–134). Springer.

Klein, E., & Sag, I. A. (1985). Type-driven translation. *Linguistics and Philosophy*, *8*(2), 163–201.

Kratzer, A. (1996). Severing the external argument from its verb. In *Phrase structure and the lexicon* (pp. 109–137). Springer.

Kratzer, A., & Shimoyama, J. (2002). Indeterminate pronouns: The view from Japanese. In Y. Otsu (Ed.), *Third Tokyo conference on psycholinguistics* (p. 1-25). Tokyo.

Krifka, M. (1992). A compositional semantics for multiple focus constructions. In *Informationsstruktur und grammatik. linguistische berichte* (pp. 17–53). Springer.

Krifka, M. (1995). The semantics and pragmatics of polarity items. *Linguistic Analysis*, *25*(3-4), 209–257.

Kroch, A. S. (1974). *The semantics of scope in English* (PhD Dissertation).

Massachusetts Institute of Technology, Cambridge, MA.
Lewis, D. (1975). Adverbs of quantification. In E. Keenan (Ed.), *Formal semantics of natural language* (p. 3-15). Cambridge, MA: Cambridge University Press.
Liang, S., Hudak, P., & Jones, M. (1995). Monad transformers and modular interpreters. In *The 22nd principles of programming languages (POPL)* (pp. 333–343).
May, R. (1985). *Logical form: Its structure and derivation*. Cambridge, MA: MIT Press.
McBride, C., & Paterson, R. (2008). Applicative programming with effects. *Journal of functional programming*, *18*(1), 1–13.
McCready, E. (2010). Varieties of conventional implicature. *Semantics and Pragmatics*, *3*(8), 1–57.
Montague, R. (1973). The proper treatment of quantification in ordinary English. In *Approaches to natural language* (pp. 221–242). Dordrecht: D. Reidel Publishing Company.
Muskens, R. (1990). Anaphora and the logic of change. In *European workshop on logics in artificial intelligence* (pp. 412–427).
Muskens, R. (1996). Combining Montague semantics and Discourse Representation. *Linguistics and Philosophy*, *19*(2), 143–186.
Partee, B. (1986). Noun phrase interpretation and type-shifting principles. In J. Groenendijk, D. de Jongh, & M. Stokhof (Eds.), *Studies in Discourse Representation Theory and the theory of Generalized Quantifiers* (p. 115-144). Dordrecht: Foris Publications.
Poesio, M. (1996). Semantic ambiguity and perceived ambiguity. In K. van Deemter & S. Peters (Eds.), *Semantic ambiguity and underspecification* (Vol. 55, pp. 159–201). Stanford: CSLI Publications.
Portner, P. (2007). Instructions for interpretation as separate performatives. In K. Schwabe & S. Winkler (Eds.), *On information structure, meaning and form* (pp. 407–425). John Benjamins Publishing Co.
Potts, C. (2005). *The logic of conventional implicatures*. Oxford: Oxford University Press.
Reinhart, T. (1983). *Anaphora and semantic interpretation*. London: Croom Helm.
Reynolds, J. (1983). Types, abstraction and parametric polymorphism. In *Information processing 83: Proceedings of the IFIP 9th world computer congress* (pp. 513–523). Amsterdam.
Romero, M., & Novel, M. (2013). Variable binding and sets of alternatives. In *Alternatives in semantics* (pp. 174–208). Springer.
Rooth, M. (1985). *Association with focus* (PhD Dissertation). University of Massachusetts, Amherst.

Rothschild, D., & Yalcin, S. (2016). Three notions of dynamicness in language. *Linguistics and Philosophy*, *39*, 333–355.

Shan, C.-c. (2001a). Monads for natural language semantics. In K. Striegnitz (Ed.), *ESSLLI 2001 student session* (pp. 285–298).

Shan, C.-c. (2001b). A variable-free dynamic semantics. In R. van Rooy & M. Stokhof (Eds.), *The thirteenth Amsterdam Colloquium* (pp. 204–209).

Shan, C.-c. (2005). *Linguistic side effects* (PhD Dissertation). Harvard University, Cambridge, MA.

Shan, C.-c., & Barker, C. (2006). Explaining crossover and superiority as left-to-right evaluation. *Linguistics and Philosophy*, *29*(1), 91-134.

Shimoyama, J. (2006). Indeterminate phrase quantification in Japanese. *Natural Language Semantics*, *14*, 139-173.

Siegel, M. E. A. (1976). *Capturing the adjective* (PhD Dissertation). University of Massachusetts, Amherst.

Strawson, P. (1950). On referring. *Mind*, *59*, 320–44.

Szabolcsi, A. (1989). Bound variables in syntax (are there any?). In R. Bartsch, J. van Benthem, & P. van Emde Boas (Eds.), *Semantics and contextual expressions* (p. 295-318). Dordrecht: Foris.

Tarski, A. (1956). The concept of truth in formalized languages. In *Logic, semantics, and metamathematics.* Hackett Publishing.

van Eijck, J. (2001). Incremental dynamics. *Journal of Logic, Language and Information*, *10*(3), 319–351.

van Eijck, J., & Unger, C. (2010). *Computational semantics with functional programming*. Cambridge University Press.

Vermeulen, C. F. M. (1993). Sequence semantics for Dynamic Predicate Logic. *Journal of Logic, Language and Information*, *2*(3), 217–254.

Wadler, P. (1989). Theorems for free! In *Proceedings of the fourth international conference on functional programming languages and computer architecture* (pp. 347–359).

Winter, Y. (2016). *Elements of formal semantics: An introduction to the mathematical theory of meaning in natural language*. Edinburgh University Press.